\documentclass[10pt,twocolumn,letterpaper]{article}
\usepackage{cvpr}
\usepackage{times}
\usepackage{epsfig}
\usepackage{graphicx}
\usepackage{amsmath}
\usepackage{amssymb}

\usepackage{color}
\usepackage{multirow}
\usepackage{rotating}
\usepackage{gensymb}
\usepackage{wrapfig}
\usepackage{soul}
\usepackage{epstopdf}
\usepackage{subfig}
\usepackage{bm}
\usepackage{verbatim}
\usepackage{multirow}
\usepackage{svg}
\usepackage[ruled]{algorithm2e}

\newcommand{\figLabel}{Figure\xspace}

\newcommand{\tblLabel}{Table\xspace}
\newcommand{\mysection}[1]{\vspace{3pt}\noindent\textbf{#1}}

\definecolor{turquoise}{cmyk}{0.65,0,0.1,0.1}
\definecolor{purple}{rgb}{0.65,0,0.65}
\definecolor{darkgreen}{rgb}{0.0, 0.5, 0.0}
\definecolor{darkred}{rgb}{0.5, 0.0, 0.0}
\definecolor{darkblue}{rgb}{0.0, 0.0, 0.5}
\definecolor{blue}{rgb}{0.0, 0.0, 1.0}
\definecolor{orange}{rgb}{1.0, 0.5, 0.0}

\usepackage[pagebackref=true,breaklinks=true,letterpaper=true,colorlinks,bookmarks=false]{hyperref}

\cvprfinalcopy

\begin{document}
\title{Learning a Controller Fusion Network by Online Trajectory Filtering for Vision-based UAV Racing}

\author{Matthias M\"uller\thanks{equal contribution}~~ Guohao Li\footnotemark[1]~~ Vincent Casser~~ Neil Smith~~ Dominik L. Michels~~ Bernard Ghanem\\
Visual Computing Center, King Abdullah University of Science and Technology, Saudi Arabia\\
{\{\tt\small matthias.mueller.2, guohao.li, bernard.ghanem\}@kaust.edu.sa}
}
	
\maketitle

\begin{abstract}
 Autonomous UAV racing has recently emerged as an interesting research problem. The dream is to beat humans in this new fast-paced sport. A common approach is to learn an end-to-end policy that directly predicts controls from raw images by imitating an expert. However, such a policy is limited by the expert it imitates and scaling to other environments and vehicle dynamics is difficult. One approach to overcome the drawbacks of an end-to-end policy is to train a network only on the perception task and handle control with a PID or MPC controller. However, a single controller must be extensively tuned and cannot usually cover the whole state space. In this paper, we propose learning an optimized controller using a DNN that fuses multiple controllers. The network learns a robust controller with online trajectory filtering, which suppresses noisy trajectories and imperfections of individual controllers. The result is a network that is able to learn a good fusion of filtered trajectories from different controllers leading to significant improvements in overall performance. We compare our trained network to controllers it has learned from, end-to-end baselines and human pilots in a realistic simulation; our network beats all baselines in extensive experiments and approaches the performance of a professional human pilot.
\end{abstract}

\section{Introduction} \label{sec: intro}
Recent advances in UAV technology by industry leaders such as DJI, Amazon and Intel make UAV design and point-to-point stabilized flight navigation appear to be a well-solved problem. However, autonomous navigation of UAVs in more complex and real-world scenarios, such as in unknown congested environments, GPS-denied areas, and narrow spaces, is still far from being solved. 
This is a complex problem, since it requires \emph{both} the sensing of the environment and the execution of appropriate control policies for interaction at low latencies, running on on-board hardware with typically very limited computational resources. The emerging sport of UAV racing displays a lot of these real-world navigation challenges, and is one of the areas where the performance gap between human pilots and machine-driven navigation approaches is most evident. UAV racing requires human pilots or agents to sequentially control the UAV to fly through a race track based on the feedback (visual information, physical measurements, or both) of previous actions. It requires control over six degrees of freedom (6-DoF) at high speeds while traversing tight spaces, and passing consistently through racing gates. These complex sense-and-understand tasks are conducted at extreme speeds reaching over 100\,km/h, and thus can serve as a controlled and challenging benchmark for machine-driven agile navigation approaches.

\begin{figure}[!tb]
  \centering
  \includegraphics[width=\columnwidth]{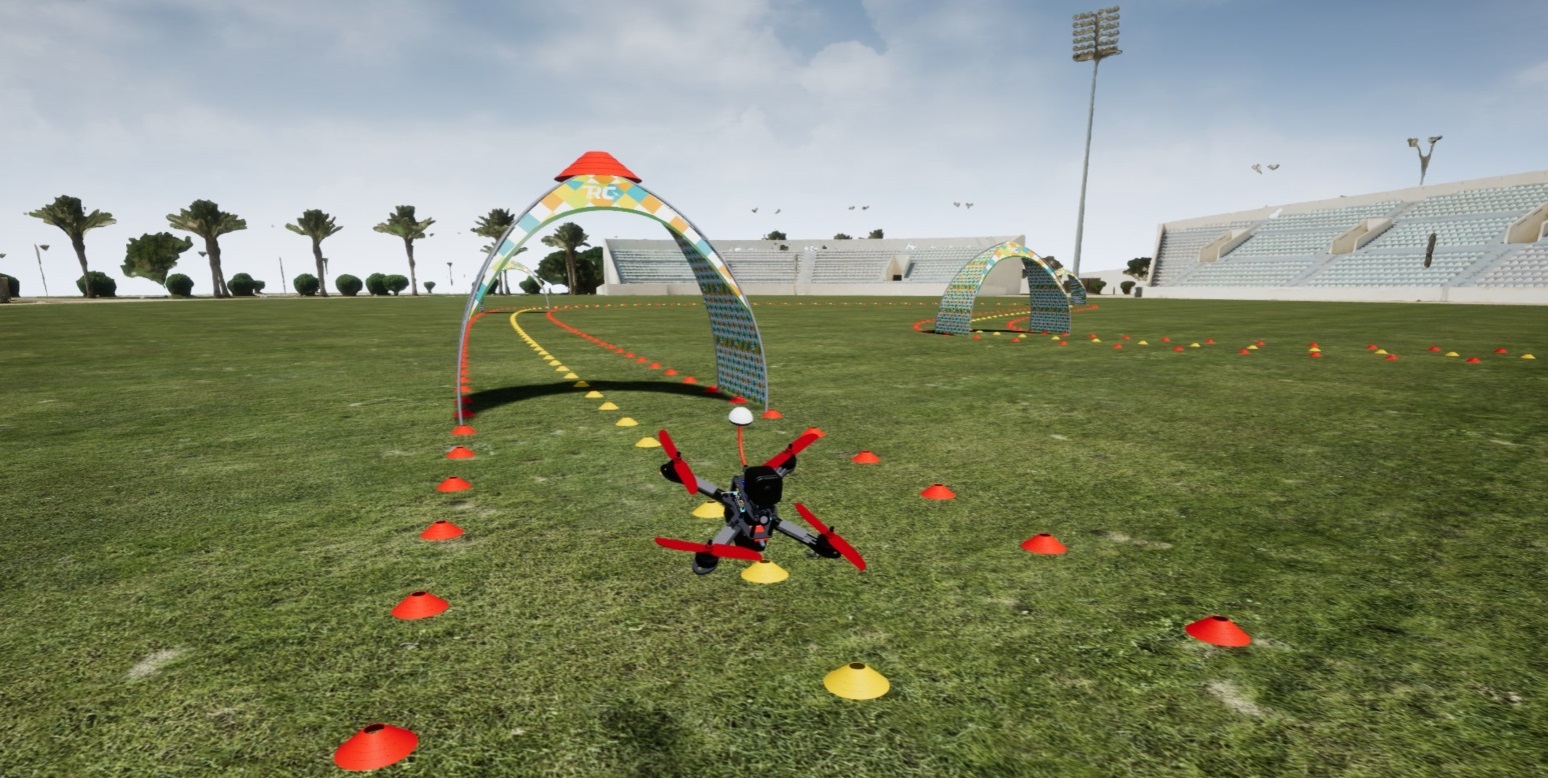}
  \caption{UAV flying in Sim4CV while being controlled by our controller fusion network.}
\label{fig:teaser}
\end{figure}

One of the more difficult tasks in UAV racing is the prediction of the proper trajectory in order to traverse a course while maintaining high speeds. In earlier work, either a PID controller or model predictive controller with Kalman filters has been used. However, in practice a single controller cannot usually cover the whole state space. In certain states the controller may not perform as expected or even fail. In this paper, we propose the fusion of multiple controllers using a DNN to cover a much larger state space and to outperform any single controller. 

\begin{figure*}
\includegraphics[trim = 15mm 20mm 0mm 10mm, clip, width=18cm]{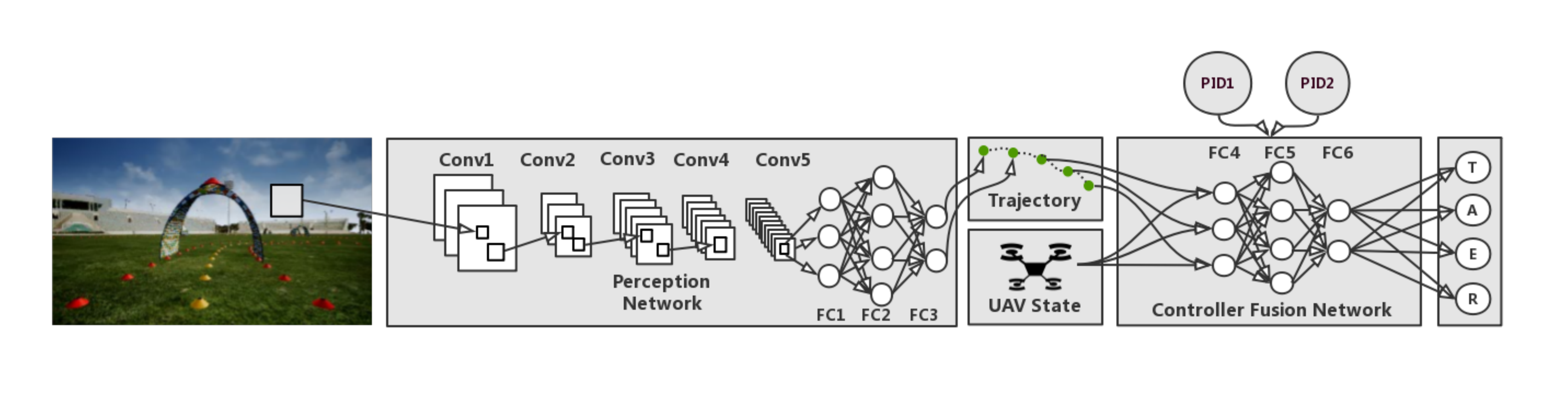}
	\caption{Our complete system consists of a perception network $\phi$ and a controller fusion network (CFN). The perception network predicts local trajectories from a monocular RGB image. The CFN $\varphi$ takes the predicted trajectory and UAV state as input and outputs the low-level controls:  throttle (T), roll/aileron(A), pitch/elevator (E), yaw/rudder (R). The CFN is trained by fusing filtered trajectories from multiple classical controllers.}
\label{fig:pipeline}
\end{figure*}

However, fusing different controller's trajectories naively will result in incorporating both their good and bad trajectories due to their high variability and limitations in specific scenarios. We overcome this problem by storing trajectories in an online buffer that can be accessed by our deep neural network (DNN) during learning. By filling this buffer with only good trajectories, we are able to trim out failing trajectories. The network learns an optimized controller by sampling from the buffer containing these pre-filtered trajectories. The result is a network that is able to learn a good fusion of filtered trajectories from different controllers allowing a much larger coverage of the whole state space. We show that this leads to significant improvements in overall performance.

We use Sim4CV \cite{sim4cv} to simulate UAV racing with accurate physics using the Unreal Engine 4 (UE4). \cite{TeachingUAVstoRace}. 
We also develop a customizable racing area in the form of a stadium based on a 3D scan of a real-world location. Race tracks are generated within a GUI interface and can be loaded at game-time to allow training on multiple tracks automatically. Inspired by recent work on self-driving cars \cite{NvidiaCar}, we are trying to imitate UAV racing at a professional level. The key difference is that we train a perception network that predicts desired trajectories rather than controls. We then train a separate neural network to produce the low-level controls. We call this network Controller Fusion Network (CFN) and train it by fusing filtered trajectories from multiple controllers. Through extensive experiments in simulation we demonstrate that CFN outperforms the controllers it learned from, end-to-end baselines, and even human pilots flying via a remote control used for real UAVs.

\vspace{3pt}\noindent\textbf{Contributions.} 
\textbf{(1)} We propose to learn the fusion of trajectories by a neural network. This allows for combining trajectories from different controllers in a principled way, and separates the control from the perception task. 
\textbf{(2)} By implementing online trajectory filtering we are able to learn from multiple noisy trajectories without incorporating their imperfections. While the control task is learned online, a buffer/memory also allows for semi-offline training. Our approach leads to a robust network outperforming several state-of-the-art approaches and human pilots.

\section{Related Work} \label{sec: related work}
The use of deep neural networks (DNNs) to control UAVs dates back to work on learning acrobatic helicopter flight \cite{Pieter2007}. More recent work has studied training UAV controller networks with SL,  RL or combined methods but with a focus on indoor flight, collision avoidance, and trajectory optimization \cite{Plato_KahnZLA16, guidedpolicysearch,SadeghiL16,ForestTrail,Andersson2017,Kim2015DeepNN,Shah:2016,pmlr-v87-kaufmann18a,TeachingUAVstoRace,BlukisBBKA18,SpicaFCMSS18}.
An important insight from \cite{guidedpolicysearch,Plato_KahnZLA16,Levine:2016,SadeghiL16} is that a trajectory optimizer such as a Model Predictive Controller (MPC) can function similar to traditional SL to help regress an agent's sub-optimal policy towards the optimal one with much fewer iterations. 
By jointly learning  perception and control with the self-supervision of MPC, full end-to-end navigation and collision avoidance can be learned. Recently, \cite{pmlr-v87-kaufmann18a} implement such an approach where a DNN is used to predict a trajectory and a MPC is used to properly output the motor control of the UAV. Although not functioning at racing speeds, initial experiments demonstrate the advantages of such a setup. A major limitation of this approach is that the DNN is only used for perception, while the MPC requires extensive setup, tuning, and a full knowledge of the UAV dynamics. We propose in this paper that a DNN can also be applied to control allowing the UAV dynamics to be inherently learned.

Our network is able to learn from imperfect proportional-integral-derivative (PID) controllers allowing both self-supervision and extensive exploration. The network training with extensive exploration is most similar to DAGGER \cite{Dagger} and its variants \cite{BlukisBBKA18,zhang2016query,Pan_RSS2018,chang2015learning}. Our approach differs in that our control network does not learn strictly from the controller. Our network learns the appropriate actions of the PID controllers at each time step and selects the best predictions based on the filtered trajectories. The buffering strategy is similar to DQN's \cite{AtariNature} "experience replay mechanism" in that it stores a limited set of experiences in a buffer and then selects samples randomly. However, our buffer is dynamic being updated during online training and continually filters the buffer with only good samples. By design, our RL motivated training process enables extensive exploration by only observing the best behavior from various controllers. It differs from \cite{Levine:2016} in that the controllers never have to deviate from their optimal control to induce exploration.  Also, unlike trajectory optimization \cite{Hehn2015}, the trajectory in our setup is without known global 3D position. This enables the prediction of local waypoints without needing  precise knowledge of the UAV's current state and dynamics, which are rarely available in the real-world. Similar to adaptive trajectory optimization, our predicted waypoints are updated every time step allowing for adaptation to environment changes.

\section{Methodology} \label{sec: methodology}
In this section, we introduce the setup of the Controller Fusion Network (CFN) which enables automatic removal of bad trajectory segments from imperfect controllers by filtering bad samples in an online manner before the CFN is updated. We build a modular system for UAV racing (see \figLabel \ref{fig:pipeline}) that separates the task into a high dimensional perception module and a low dimensional control module (CFN). The perception network predicts trajectories which are used by the the CFN along with the UAV state to produce low-level controls.

\subsection{Controller Fusion Network (CFN)}
We learn a \emph{Controller Fusion Network} agent by a learning strategy that integrates knowledge from multiple controllers and the dynamic environment into the learning process (refer to Algorithm \ref{alg: CFN} for a detailed description during training). At each time step $t$, the agent receives a state (or partial state) $s_t$  from the environment and executes an action $a_t$. Thus, the trajectory of the agent behaviors are denoted as $\tau = (s_1, a_1, s_2, a_2, ..., s_n)$. The CFN policy is a parameterized function $\pi(s|\theta)$ mapping the state to a deterministic action that can be continuous or discrete. In our case, the action is a 4D continuous control signal for a UAV. The PID controllers' policy $\mu(s)$ also maps the state to an action. In practice, the controller can be either an automated controller or a human, and demonstration can be performed online or offline from a recording. In this paper, we use two PID (Proportional-Integral-Derivative) controllers, thus,  avoiding the need for hours of human recorded control. 

\begin{algorithm}[t] 
\caption{Controller Fusion Network (CFN) during training.}\label{alg: CFN}
\small
\SetAlgoLined
 Initialize controllers $\{\mu_i(s)\}_{i=1}^n$ and CFN $\pi(s|\theta)$ with random weights $\theta$\; 
 Initialize CFN training database $D \leftarrow \emptyset$ and CFN temporary buffers $\{B_i \leftarrow \emptyset\}_{i=1}^n$ corresponding to the controllers\;
 // \emph{for each controller $\mu_i$}\\
 \For{episode $\leftarrow 1$ \KwTo $M$}{
   Initial state $s_1$ provided by the environment\;
   \For{$t\leftarrow 1$ \KwTo $T$}{
    Controller $\mu_i$ demonstrates $a_{\mu_i t}=\mu_i(s_t)$\;
    Execute controller action $a_{\mu_i t}$; observe new state $s_{t+1}$ and feedback\;
    Update $B_i$ $\leftarrow B_i$ $\bigcup \{(s_t, a_{\mu_i t})\}$\;
    Discard unwanted demonstrations from $B_i$ (according to \emph{buffering strategy})\;
    Add desirable demonstrations from $B_i$ to $D$;\\
    Sample mini-batch $(s, a)$ from $D$ and perform SGD to minimize $\mathbb{L}(\pi(s|\theta),a)$\;
    Break if $s_{t+1}$ is terminal state\;
   }
 }
\end{algorithm}

\begin{figure*}[!h]
  \centering
  \includegraphics[page=1, trim = 1mm 20mm 1mm 20mm, clip, width=18cm]{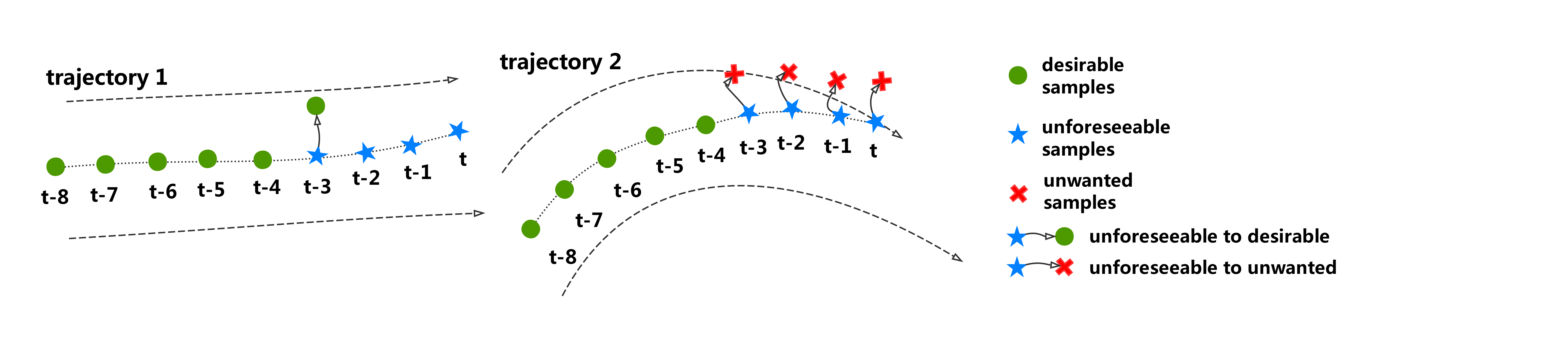}
	  \caption{Illustration of the Buffering Strategy. In Trajectory 1 the PID controller remains on track: the sample $(s_{t-k}, a_{t-k})$ will become a desirable sample and will be added to the ground truth database $D$. In Trajectory 2 the PID controller leaves the track: the samples $\{(s_{t}, a_{t})\}_{t-k}^{t}$ will be discarded.}
		\label{fig:trajectory}\vspace{-15pt}
\end{figure*}

To clarify which demonstrations we should learn from, we refer to the demonstrations that lead the CFN agent to good behavior as \emph{desirable} demonstrations, the ones that lead to bad behavior as \emph{unwanted} demonstrations, and the uncertain ones as \emph{unforeseeable} demonstrations. We introduce a \emph{temporary buffer} $B$ to store the demonstrations of the active controller. At each time step $t$, by observing the feedback of the interactions between controller and environment, the CFN agent can determine whether to discard some \emph{unwanted} demonstrations from $B$, retain \emph{unforeseeable} demonstrations in $B$, or augment its training data $D$ by adding \emph{desirable} demonstrations from $B$. We call this operation \emph{buffer strategy}. Note that the CFN agent maintains a temporary buffer for each controller. See \figLabel\ref{fig:trajectory} for details.

\begin{figure}[!htb]
  \centering
  \includegraphics[page=1, trim = 0mm 10mm 0mm 15mm, clip, width=8 cm]{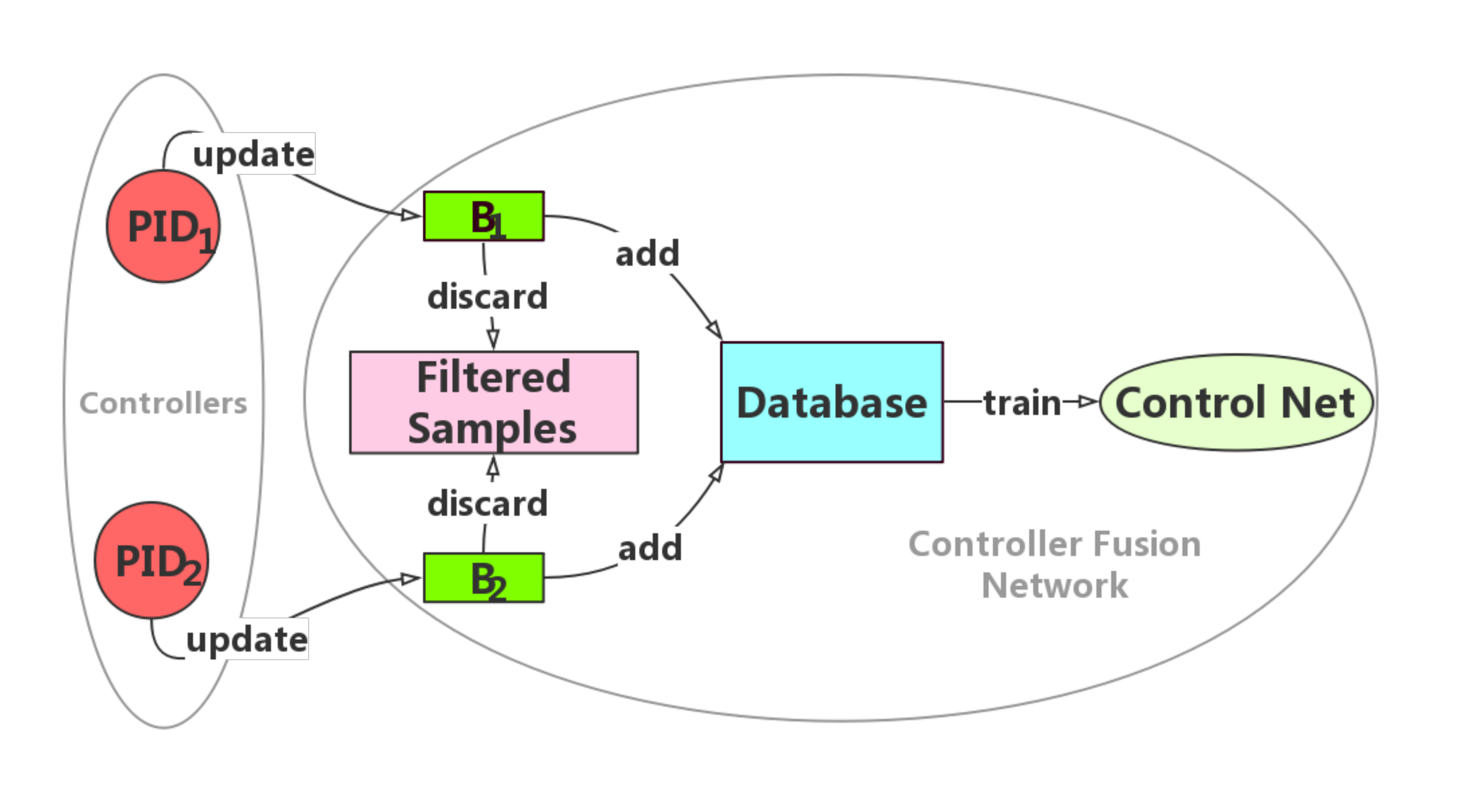}
	  \caption{Visualization of the CFN buffer strategy.}
		\label{fig:buffer}
\end{figure}

\mysection{CFN buffer strategy.~}
We use two simple PID controllers. The CFN agent maintains two \emph{temporary buffers} B corresponding to each controller. At each time step, the state action pair $(s_t, a_t)$ of each controller will be buffered into B. Then the CFN agent will decide which state action pairs to add to the training database, and discard others based on the \emph{buffer strategy}. Finally, the CFN agent will perform an SGD update on the training database. A schematic diagram of the buffer used during training is shown in \figLabel \ref{fig:buffer}.

The training database $D$ can be viewed as a distilled set of demonstrations collected by applying the buffering strategy on $B$ at each time step. Our goal is to train a CFN policy $\pi^*$ to minimize the loss function $\mathbb{L}(\pi(s|\theta),a^{*})$:
\begin{equation} \label{eq1}
\begin{split}
	\pi^*(s|\theta) &= \underset{\pi}{\arg\min} ~~\mathbb{E}_{(s,a)\thicksim D}[\mathbb{L}(\pi(s|\theta),a^{*})]
\end{split}
\end{equation}
The total loss consists of the perception loss $\mathbb{L}_p$ and a control loss $\mathbb{L}_c$:
\begin{align} 
&\mathbb{L} = \mathbb{L}_p(\phi(s_I|\theta_p),z^{*})  + \lambda \mathbb{L}_c(\pi(s|\theta),a^{*})
\label{eq3}\\
&\mathbb{L}_c(\pi(s|\theta),a^{*}) = \mathbb{L}_c(\varphi(\phi(s_I|\theta_p),s_M|\theta_c),a^{*})\label{eq4}
\end{align}
Here, $\theta_p$ and $\theta_c$ are the learnable parameters of the perception and controller fusion network, respectively; $\lambda$ scales the control loss $\mathbb{L}_c$; $z^{*}$ represents the groundtruth waypoints; $s_I$ and $s_M$ are the input image state and UAV state.

\section{Network Architecture and Training Details}
Our overall network architecture is depicted in \figLabel \ref{fig:pipeline}. It consists of two networks, one for perception and one for control. The perception network outputs a local trajectory and the control network produces low-level controls given this trajectory and the UAV state as input. 

\subsection{Perception} \label{sec: perception}

\mysection{Network Architecture.~} For our perception network, we use a network architecture that is inspired by the one used by Bojarski \etal\cite{NvidiaCar} for autonomous driving. However, we make changes to accommodate the complexity of the task at hand and to improve robustness in training. Our network architecture is shown in \figLabel \ref{fig:pipeline} as the perception module. It consists of eight layers: five convolutional with $\{20,24,28,30,32\}$ filters and three fully-connected layers with $\{1800,800,100\}$ hidden units, respectively. Instead of pooling operations, we introduce convolutional strides of two to downsample the input consecutively, inspired by more recent architecture designs such as MobileNet \cite{howard2017mobilenets}.

\mysection{Training Details.~}
In contrast to the network by Bojarski \etal\cite{NvidiaCar}, our network regresses local trajectories rather than raw controls. Predicting raw controls has several shortcomings; raw controls are specific to a vehicle and controller. In addition, different sequences of controls can lead to the same trajectory which makes data augmentation very difficult. As a result, we are able to train a very robust perception network and can validate it with ground-truth trajectories which are well defined. Our perception network takes images from a monocular RGB camera as input and ouputs a trajectory relative to the current position of the UAV. We represent the trajectory with five uniformly sampled points. We regress these points to the groundtruth (waypoints along the center line of the track) applying a L2-loss and dropout of $0.5$ in the fully-connected layers. We implement our model in TensorFlow and train it with a learning rate of $5\mathrm{e}{-4}$ using the Adam optimizer. 
Given the dynamic nature of the racing task we use the maximum frame rate of 60\,fps unlike other works that downsample \cite{deepDriving,NvidiaCar,ForestTrail}. Note that when predicting controls directly from the input image, a high frame rate can be problematic, since fast transitions in control can lead to similar images with different labels, causing a regression towards averages. Since the image to trajectory correspondence is well defined and more stable, our approach is not affected by sampling rate variations.

\subsection{Control} \label{sec: controller}
Here, we present the details of our controller fusion network, including network architecture and training strategy. The network takes the predicted trajectory from the perception network and the UAV state (orientation and velocity) as input and predicts the four UAV controls: throttle (T), roll/aileron (A), pitch/elevator (E), yaw/rudder (R).

\mysection {Learning the Controller Fusion Network.~} 
In our experiments, we use two naive PID controllers and denote them as $\mu_1$ and $\mu_2$. 
We briefly tune two PID controllers on the training tracks. It is not necessary to tune controllers to be optimal on all the training tracks since CFN is robust to learn from sub-optimal controllers.
The first PID controller $\mu_1$ is conservative, as it accurately follows the center of the track and flies through gates precisely but at relatively low speeds. Its output control values are a function of the first predicted waypoint $wp_1$ and UAV state. The second PID controller $\mu_2$ is more dynamic, as it flies at maximum speed but can often overshoot gates on sharp turns due to inertia and limitations of the UAV. Its output control values are a function of the fourth predicted waypoint $wp_4$ and the UAV state. We use a three-layer fully connected network to approximate the policy $\varphi$ of the CFN agent. The state of the CFN agent is a vector concatenation of the predicted waypoints and the UAV state (physical measurements):  $s = [s_{wp_{1-n}}, s_M] = [\phi(s_I|\theta_p), s_M]$.

As such, the states of the CFN agent $\varphi$, the conservative PID $\mu_1$, and the aggressive PID $\mu_2$, are $[s_{wp_{1-n}}, s_M]$, $[s_{wp_1}, s_M]$ and $[s_{wp_4}, s_M]$ respectively. At each time step, the state-action pairs $(s, a)$ in the \emph{temporary buffers} $B_1$ and $B_2$ are $([s_{wp_{1-n}}, s_M], \mu_1(s_{wp_1}))$ and  $([s_{wp_{1-n}}, s_M], \mu_2(s_{wp_4}))$.

An illustration of the \emph{buffering strategy} is shown in \figLabel\ref{fig:trajectory} with buffer size $k=3$. At time step $t$, there are $k$ unforeseeable samples $\{(s_{t}, a_{t})\}_{t-k}^{t-1}$ in $B$. The next sample is $(s_t, a_t)$. The samples ahead of $(s_{t-k}, a_{t-k})$ are stored in the ground truth training database $D$. \figLabel\ref{fig:trajectory} illustrates two cases of buffering operations.

\mysection {Network Architecture and Training Details.~} The goal of the control network is to find a control policy $\varphi$ that minimizes the control loss $\mathbb{L}_c$:
\begin{align} 
\varphi^*(s|\theta_c) &= \underset{\varphi}{\arg\min}~~ \mathbb{E}_{(s,a)\thicksim D}[\mathbb{L}_c(\varphi(s|\theta_c),a^{*})]\label{eq9}
\end{align}
Here, $\theta_c$ represents the learnable parameters of the controller fusion network $\varphi$. We use Tensorflow to implement our CFN. 
A three-layer fully-connected network with $\{64, 32, 16\}$ hidden units is used to represent CFN. To regularize the network, we apply dropout to the second layer with a 0.5 ratio. A weighted $L1$-loss 
is used for our loss function $\mathbb{L}_c$. We use an Adam Optimizer with a learning rate of $1\mathrm{e}{-3}$ to train CFN in an online fashion, while the UAV flies through a training track.
A \emph{temporary buffer} with size $k$ is used to temporally store the last $k$ state-action pairs from each controller ($k_1=1$ for $B_1$ and $k_2=50$ for $B_2$). The buffer size mainly depends on the controller's trajectory quality. For example, if an aggressive PID controller overshoots or crashes after $100$ steps at a straight and $50$ steps at a bend with a high probability, $40$--$60$ would be considered a good buffer size; a buffer size of $0$--$20$ would lead the CFN agent to learn dangerous behaviours, while a buffer size of over $100$ would make the CFN agent unable to benefit from the speed advantages of the aggressive PID controller on straightaways. In our case, we chose $k_1=1$ for the conservative PID controller to benefit from its accuracy and $k_2=50$ for the aggressive PID controller to filter out its dangerous behaviors in curves. To improve learning, we add an Ornstein Uhlenbeck process \cite{deepReinforcementSimulator} noise to the output of the controller fusion network to allow for exploration at the beginning and move the UAV by the agent's control predictions, but the action is labeled by the conservative PID. After the initial exploration, the controllers take over control. At each time step, if a controller leaves the track, its temporary buffer is flushed and no state-action pairs are added to $D$ for training. If the controller keeps the UAV within the track boundaries, the oldest state-action pair in its temporary buffer is added to $D$, as shown in \figLabel\ref{fig:trajectory}. At each time step, the network parameters $\theta_c$ are updated by back propagation to minimize the difference between the controller's UAV policy $\varphi$ and the demonstrations (state-action pairs) in $D$, which are considered to be believable after discarding unwanted behaviors. 

\begin{figure}[!b]
	\includegraphics[width=\linewidth]{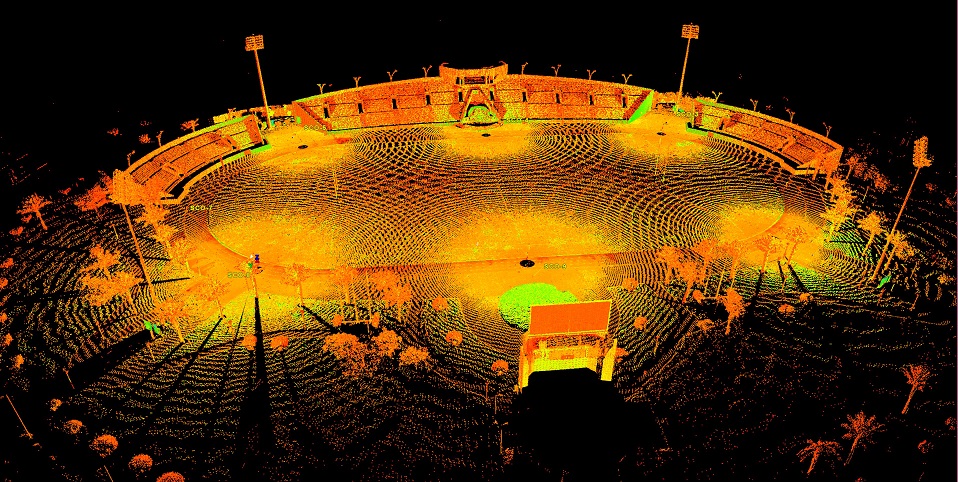}
	\caption{The point cloud from the LiDAR scan of the stadium collected from six different locations.}
	\label{fig:stadium_lidar}
\end{figure}

\begin{figure*}[!htb]
    \begin{tabular}{cc}
	\includegraphics[width=0.49\linewidth]{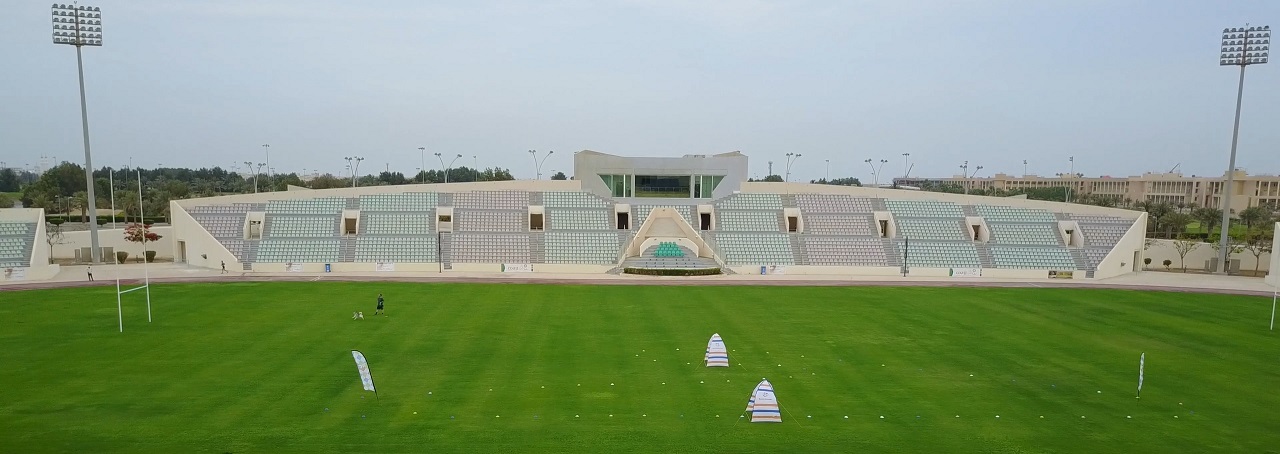} & \includegraphics[width=0.49\linewidth]{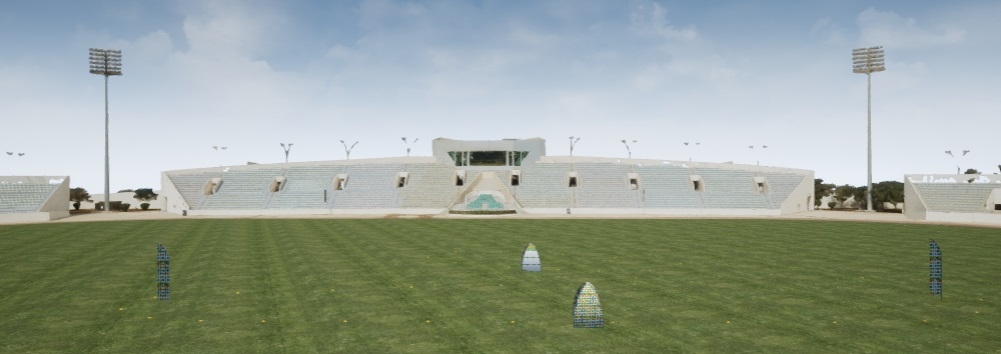}
    \end{tabular}
	\caption{\emph{Left:} Aerial image captured from an UAV hovering above the stadium racing track. \emph{Right:} Rendering of the reconstructed stadium generated at a similar altitude and viewing angle within the simulator.}
	\label{fig:stadium_racetrack}
\end{figure*}

\begin{figure*}[!tb]
  \centering
  \includegraphics[width=\linewidth]{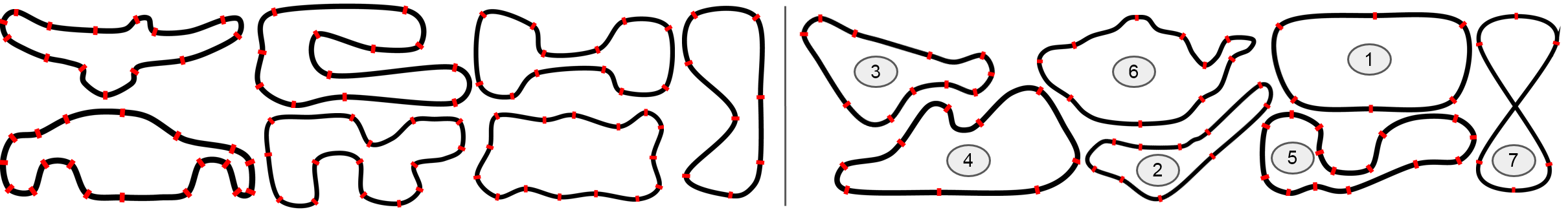}
	  \caption{The seven training tracks (left) and the seven testing tracks (right). Gates are marked in red.}
		\label{fig:tracks}
\end{figure*}

\section{Experiments} \label{sec: results}
\mysection{Creation of the UAV Racing Simulation.}
Many professional pilots compete in time trials on well-known tracks such as those posted by the MultiGP Drone Racing League. Following this paradigm, our simulator race course is modeled after a football stadium, where local professional pilots regularly setup MultiGP tracks. Using a combination of LiDAR scanning and aerial photogrammetry, we captured the stadium with an accuracy of 0.5\,cm (see \figLabel \ref{fig:stadium_lidar}). A team of architects used the dense point cloud and textured mesh to create an accurate solid model with physics based rendering (PBR) textures in 3DS Max for export to Unreal. This resulted in a geometrically accurate and photo-realistic race course that remains low in poly count, so as to run within Sim4CV in real-time, in which all training and testing experiments are conducted. We refer to \figLabel \ref{fig:stadium_racetrack} for a side-by-side comparison of the real and virtual stadiums. 

\mysection{Experimental Setup.~} \label{sec: experimental_setup}
We use our UAV racing environment in Sim4CV \cite{sim4cv} (see Figure \ref{fig:teaser}); we design seven racing tracks for training and seven tracks for testing. To avoid user bias, we collect online images and trace their contours to create uniquely stylized tracks. We select the tracks with two aspects in mind. (1) The tracks should be similar to what racing professionals are accustomed to, and (2) they should offer enough diversity for network generalization on unseen tracks (see \figLabel\ref{fig:tracks}). 
For all of the following evaluations, both the trained networks and human pilots are tasked with flying two laps on each of the test tracks. The score comprises three components: the percentage of successfully passed gates, the time to complete both laps, and the number of required resets. The UAV is reset at the next gate, if it does not reach it within 10 seconds after passing through the previous gate. This occurs if the UAV crashes beyond recovery or drifts off the track. Visualizations of the UAV's trajectory for all models on each track are provided in the appendix \ref{sec: supp}. 

\begin{figure*}
\centering
\begin{tabular}{@{}c@{\hspace{5mm}}c@{\hspace{5mm}}c@{}}
		\includegraphics[height=2.5cm]{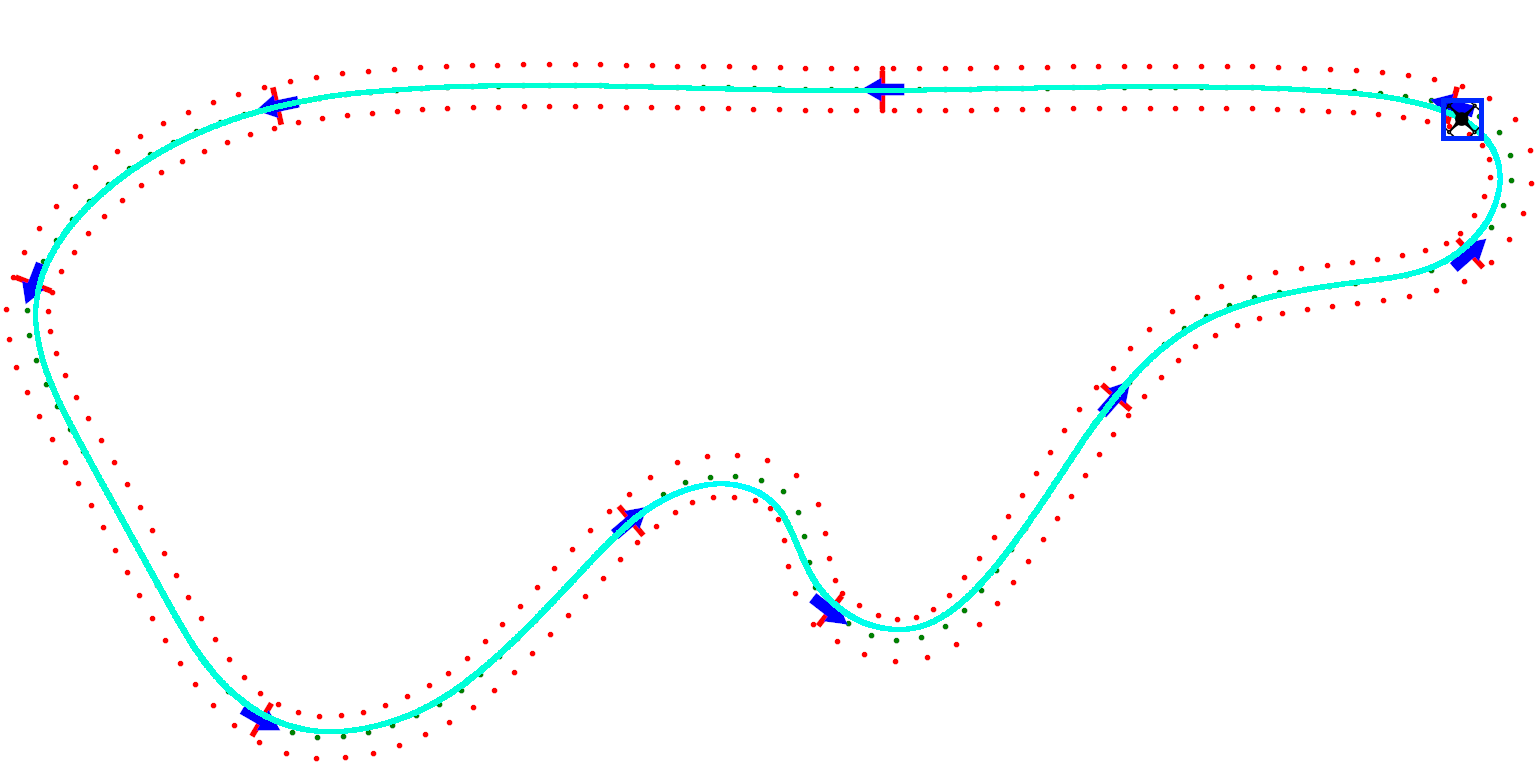}&
		\includegraphics[height=2.5cm]{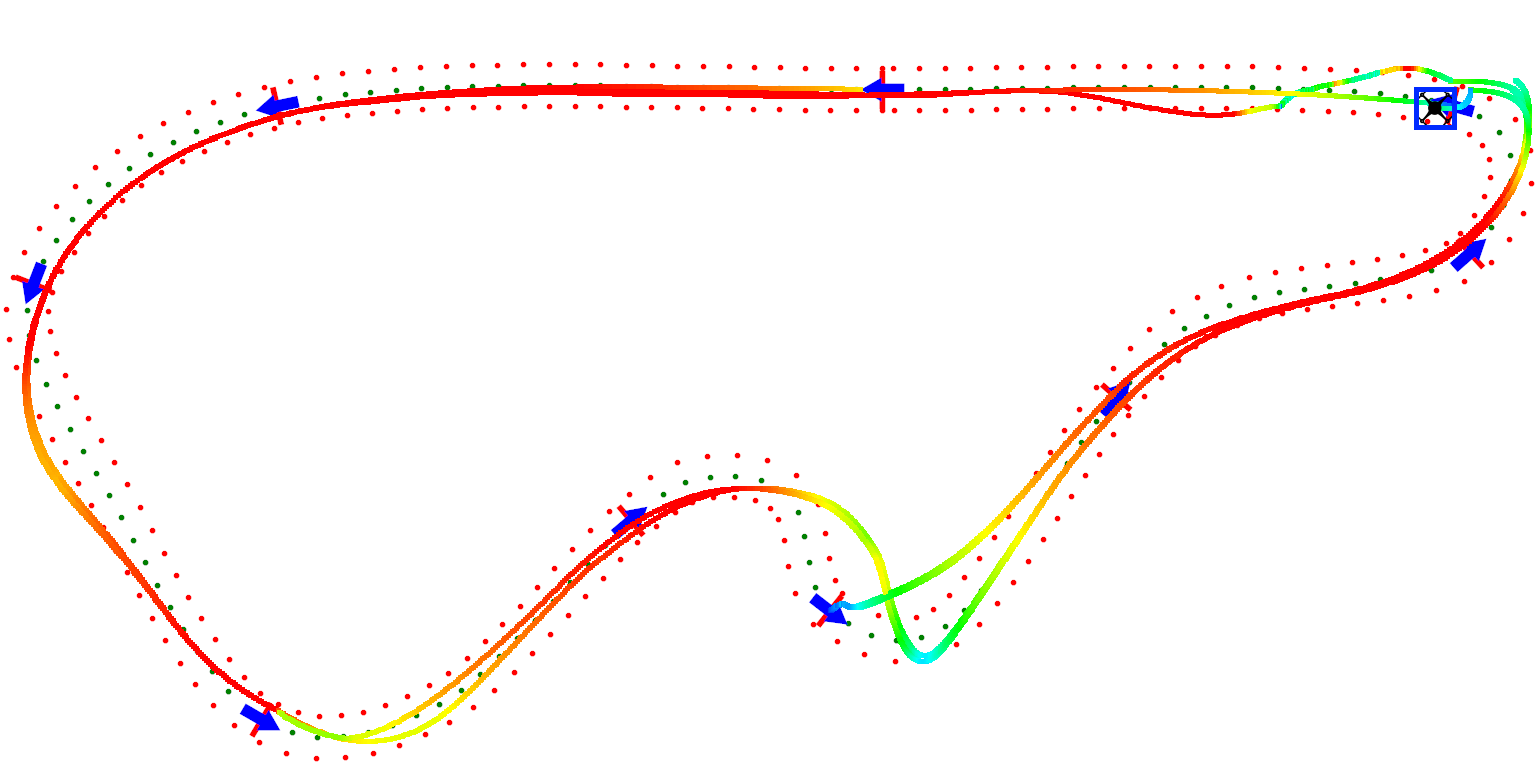}&
		\includegraphics[height=2.5cm]{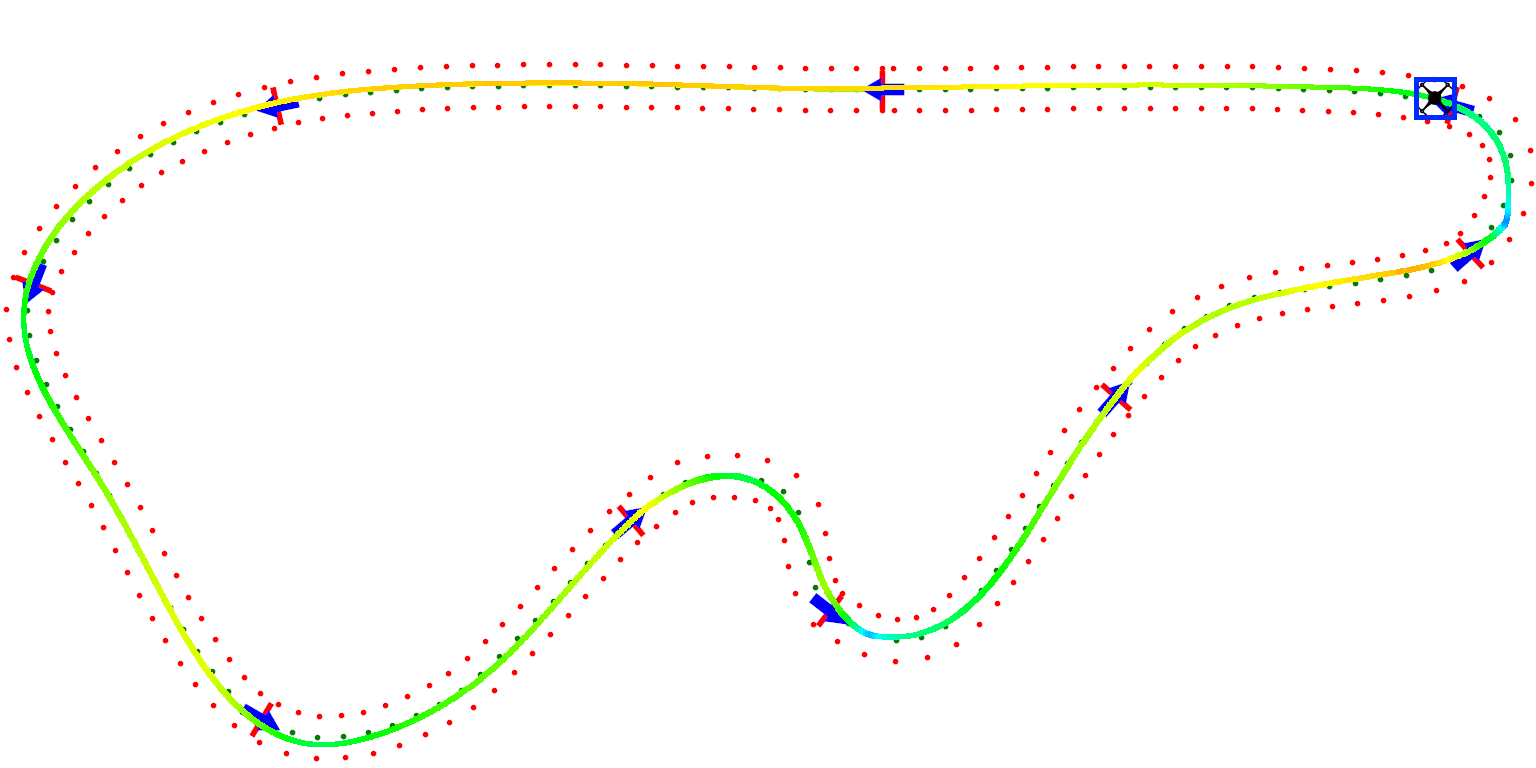} \\
		\small (a) PID1 (Conservative) &
		\small (b) PID2 (Aggressive) &
		\small (c) Ours \\
		\includegraphics[height=2.5cm]{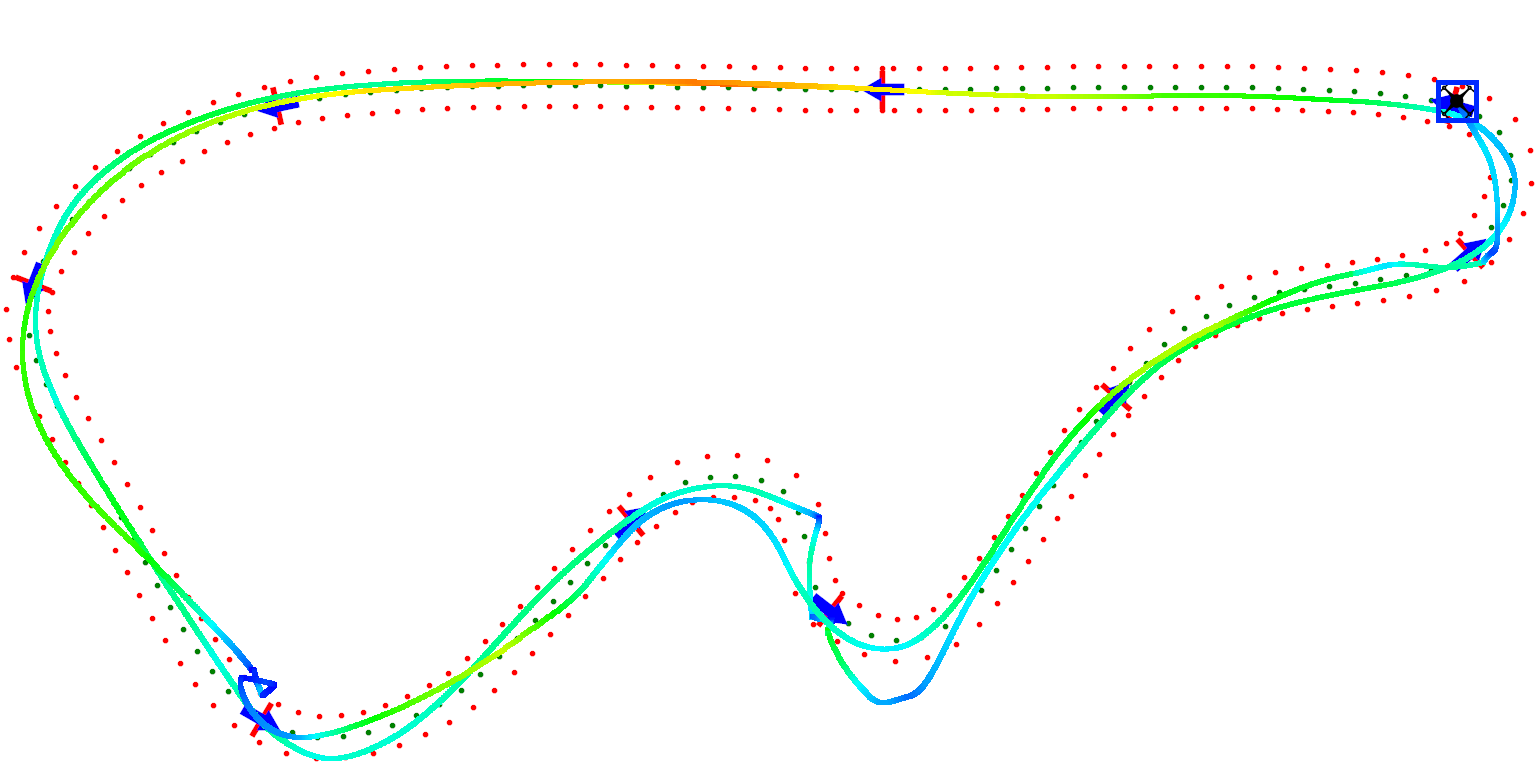}&
		\includegraphics[height=2.5cm]{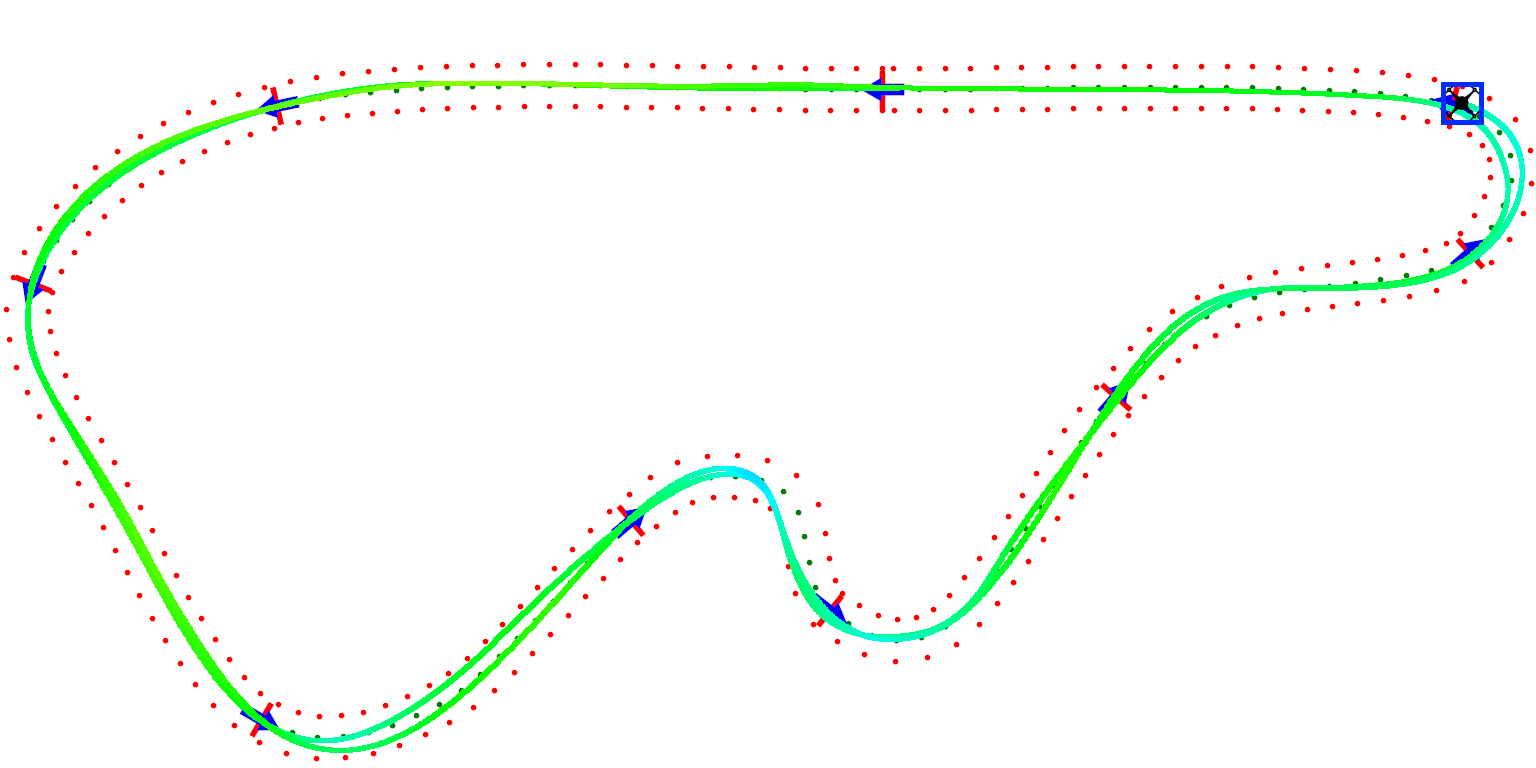}&
		\includegraphics[height=2.5cm]{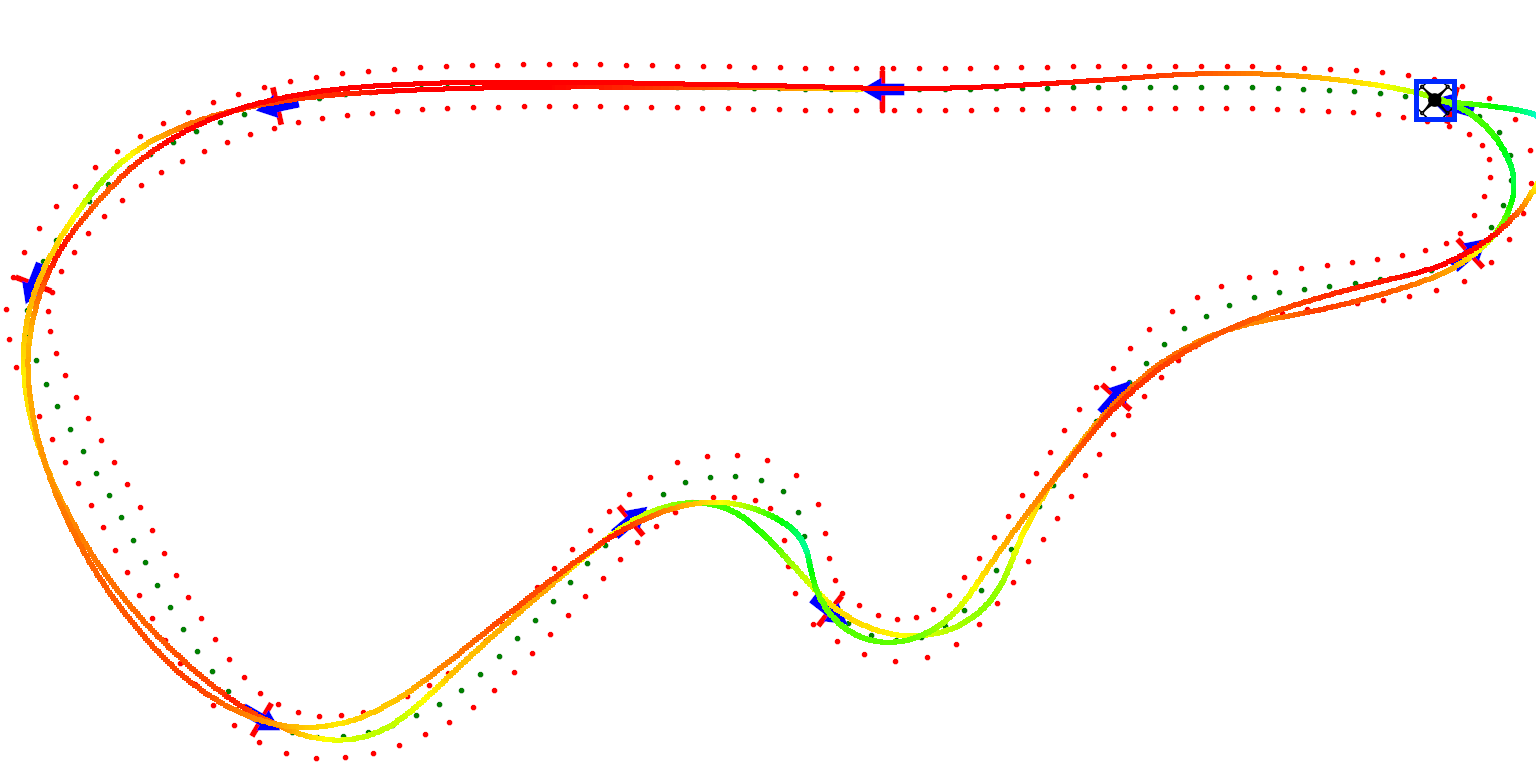}\\
		\small (d) Novice &
		\small (e) Intermediate &
		\small (f) Professional \\
        \multicolumn{3}{c}{\includegraphics[height=1cm]{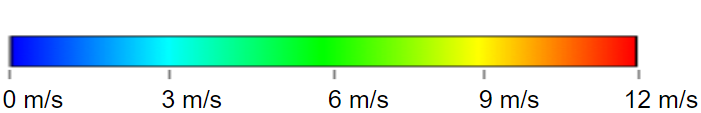}} 
\end{tabular}
\captionof{figure}{Comparison between our learned CFN policy and PID controllers (\emph{row1}) and human pilots (\emph{row2}), on a test track. Color encodes speed as a heatmap, where blue is the minimum speed and red is the maximum speed.}
\label{fig:qualitive_results}
\end{figure*}

\begin{table*}[!htb]
\centering
\footnotesize
\caption{Comparison of CFN to baselines}
\label{tbl: baselines}
\begin{tabular}{lcccccccccccc}
\hline
              & \multicolumn{3}{c}{End2End (MAV)} & \multicolumn{3}{c}{End2End (Nvidia)} & \multicolumn{3}{c}{End2End (Ours)} & \multicolumn{3}{c}{Ours (WP + CFN)}           \\
              & Score & Time   & Resets & Score  & Time    & Resets  & Score       & Time       & Resets     & Score          & Time           & Resets     \\ \hline
Track1        & 6/12  & 98.60  & 6      & 7/12   & 101.15  & 5      &   12/12       &  80.11       &   0         & 12/12          & 52.20          & 0          \\
Track2        & 16/20 & 113.55 & 4      & 15/20  & 140.92   & 5       & 20/20        &  91.88      &   0         & 20/20          & 64.75          & 0          \\
Track3        & 11/22 & 161.85 & 11      & 19/22  & 110.94  & 3       & 22/22         & 81.26       &   0         & 22/22          & 62.00          & 0          \\
Track4        & 10/18 & 152.27 & 8      & 15/18  & 121.07  & 3       &  18/18        &  97.10    &    0       & 18/18          & 71.93          & 0          \\
Track5        & 18/30 & 207.07 & 12      & 21/30  & 197.11  & 9       &  30/30          & 100.47     &  0          & 30/30          & 71.16          & 0          \\
Track6        & 15/20 & 136.69 & 5      & 20/20  & 108.42  & 0       &   16/20        &137.14        &   4         & 20/20          & 81.66          & 0          \\
Track7        & 8/12  & 115.05 & 4      & 10/12  & 105.37  & 2       &   10/12        &104.97        &   2         & 12/12          & 64.86          & 0          \\ \hline
\textbf{Avg.} & 62.69\%  & 140.72 & 7.14 & 79.85\%   & 126.43  & 3.86  & \textbf{95.52\%}   & \textbf{98.99}  & \textbf{0.86}  & \textbf{100\%} & \textbf{66.94} & \textbf{0} \\ \hline
\end{tabular}
\end{table*}

\begin{table*}[!htb]
\centering
\footnotesize
\caption{Ablation study}
\label{tbl: ablation}
\begin{tabular}{lcccccccccccc}
\hline
              & \multicolumn{3}{c}{PID1 (Conservative)} & \multicolumn{3}{c}{PID2 (Aggressive)} & \multicolumn{3}{c}{Ours (No Buffer)}                     & \multicolumn{3}{c}{Ours (WP + CFN)}                     \\
              & Score       & Time        & Resets      & Score      & Time        & Resets     & Score          & Time           & Resets     & Score          & Time           & Resets     \\ \hline
Track1        & 12/12       & 130.76      & 0           & 12/12      & 40.04       & 0          & 10/12         & 70.35           & 2          & 12/12          & 52.20          & 0          \\
Track2        & 20/20       & 136.19      & 0           & 17/20      & 77.41       & 3          & 18/20         & 75.58           & 2          & 20/20          & 64.75          & 0          \\
Track3        & 22/22       & 121.54      & 0           & 11/22      & 149.45      & 11         & 17/22         & 102.72          & 5          & 22/22          & 62.00          & 0          \\
Track4        & 18/18       & 139.09      & 0           & 15/18      & 81.08       & 3          & 14/18         & 102.27          & 4          & 18/18          & 71.93          & 0          \\
Track5        & 30/30       & 144.49      & 0           & 12/30      & 212.79      & 18         & 28/30         & 89.93           & 2          & 30/30          & 71.16          & 0          \\
Track6        & 20/20       & 151.95      & 0           & 12/20      & 118.69      & 8          & 13/20         & 126.77          & 7          & 20/20          & 81.66          & 0          \\
Track7        & 10/12       & 139.28      & 2           & 9/12       & 72.90       & 3          & 7/12          & 86.53           & 5          & 12/12          & 64.86          & 0          \\ \hline
\textbf{Avg.} & 98.51\%     & 137.61      & 0.29        & 65.67\%    & 107.48      & 6.57       & 79.85\%       & 93.45           & 3.86       & \textbf{100\%} & \textbf{66.94} & \textbf{0} \\ \hline
\end{tabular}
\end{table*}

\begin{table*}[!htb]
\centering
\footnotesize
\caption{Comparison of CFN to humans}
\label{tbl: human}
\begin{tabular}{lcccccccccccc}
\hline
              & \multicolumn{3}{c}{Human (Novice)} & \multicolumn{3}{c}{Human (Intermediate)} & \multicolumn{3}{c}{Human (Professional)}             & \multicolumn{3}{c}{Ours (WP + CFN)} \\
              & Score  & Time    & Resets  & Score     & Time     & Resets    & Score          & Time           & Resets     & Score  & Time   & Resets \\ \hline
Track1        & 12/12  & 87.44   & 0       & 12/12     & 62.80    & 0         & 12/12          & 40.50          & 0          & 12/12  & 52.20  & 0      \\
Track2        & 20/20  & 166.11  & 0       & 20/20     & 88.21    & 0         & 20/20          & 49.23          & 0          & 20/20  & 64.75  & 0      \\
Track3        & 21/22  & 118.41  & 1       & 22/22     & 82.17    & 0         & 22/22          & 47.67          & 0          & 22/22  & 62.00  & 0      \\
Track4        & 17/18  & 126.47  & 1       & 18/18     & 91.53    & 0         & 18/18          & 50.10          & 0          & 18/18  & 71.93  & 0      \\
Track5        & 30/30  & 129.49  & 0       & 30/30     & 87.62   & 0         & 30/30          & 55.72          & 0          & 30/30  & 71.16  & 0      \\
Track6        & 20/20  & 196.16  & 0       & 20/20     & 95.99    & 0         & 20/20          & 57.90          & 0          & 20/20  & 81.66  & 0      \\
Track7        & 12/12  & 113.14  & 0       & 12/12     & 74.91   & 0         & 12/12          & 46.98          & 0          & 12/12  & 64.86  & 0      \\ \hline
\textbf{Avg.} & 98.51\%& 133.89  & 0.29    & 100\%     & 83.32    & 0         & \textbf{100\%} & \textbf{49.73} & \textbf{0} & 100\%  & 66.94  & 0      \\ \hline
\end{tabular}
\end{table*}

\mysection{Comparison to State-of-the-Art Baselines.~} \label{sec: baselines}
We compare our system for UAV racing to the two most related and recent network architectures, the first denoted as Nvidia (for self-driving cars \cite{NvidiaCar}) and the second as MAV (for forest path navigating UAVs \cite{ForestTrail}). Both the Nvidia and MAV networks use data augmentation from an additional left and right camera. For the Nvidia network, the exact offset choices for training are not publicly known, so we use a rotational offset of $\{-30^{\circ},30^{\circ}\}$. For the MAV network, we use the same augmentation parameters in the paper, \ie a rotational offset of $\{-30^{\circ},30^{\circ}\}$. We modify the MAV network to allow for a regression output instead of its original classification (left, center and right controls). This is necessary, since our task requires fine-grained control, and discrete controls would be insufficient. We assign corrective controls to the augmentation views using a fairly simple but effective strategy. 
Depending on the camera view, we apply the following offset parameters: one that acts as a horizontal offset (roll-offset) and one that acts as a rotational offset (yaw-offset). For rotational offsets, we couple the yaw correction with a proportional roll correction because the UAV is in motion while rotating, causing it to drift outwards due to its inertia.
 
While the domains of these methods are similar, it should be noted that flying a high-speed racing UAV is a particularly challenging task, since the effect of inertia is more significant and there are more degrees of freedom than for ground vehicles. To ensure a strong end-to-end baseline, we build an end-to-end network that takes the state of the UAV (exactly like our control module) as input along with the image. We also augment the data with 18 additional camera views (exactly like our perception module) and assign the best corrective controls after cross-validation search \cite{TeachingUAVstoRace}. 

\tblLabel \ref{tbl: baselines} compares the performance of these baselines against our method. The MAV reference network needs more than $7$ resets and only completes about $60\%$ of gates on average, while taking more than twice the time. The Nvidia-inspired architecture performs slightly better, but still needs about $4$ resets and only completes $80\%$ of gates on average. While the end-to-end trained version of our network achieves better performance than MAV and Nvidia, our modular network with CFN clearly outperforms it without the need for supervision or approximate corrective controls. 
In fact, CFN outperforms all baselines by a considerable margin in all three evaluation metrics; it completes all seven race tracks with 100\% accuracy compared to 62.69\%, 79.85\% and 95.52\% in about half the time on average.

\begin{figure}
\centering
\begin{tabular}{@{}c@{\hspace{1mm}}c@{\hspace{1mm}}c@{\hspace{1mm}}c@{}}
		\includegraphics[height=2.3cm]{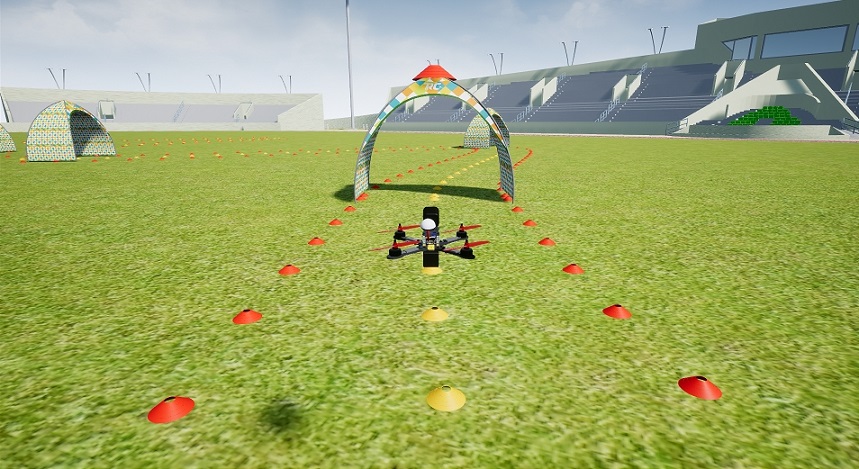} &
		\includegraphics[height=2.3cm]{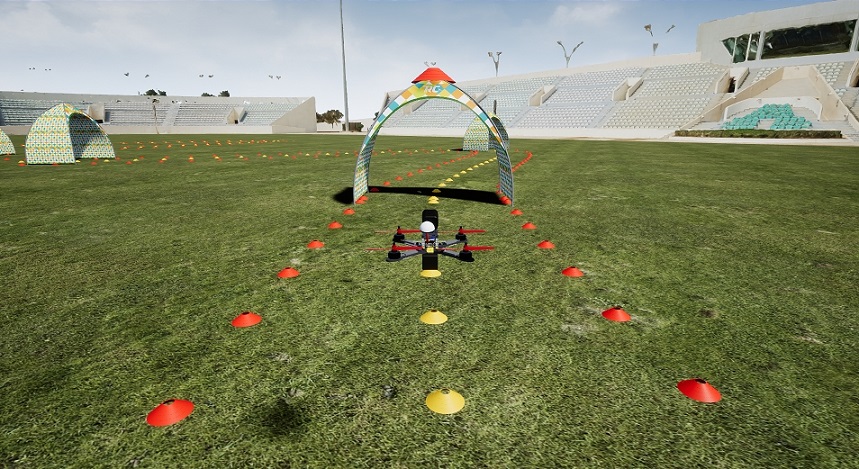} \\
		\small (a) Grass &
		\small (b) HD Grass \\
		\includegraphics[height=2.3cm]{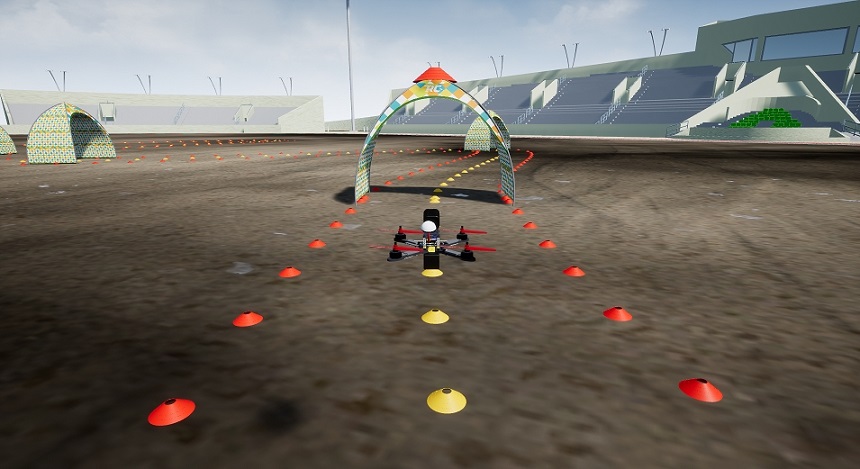} &
		\includegraphics[height=2.3cm]{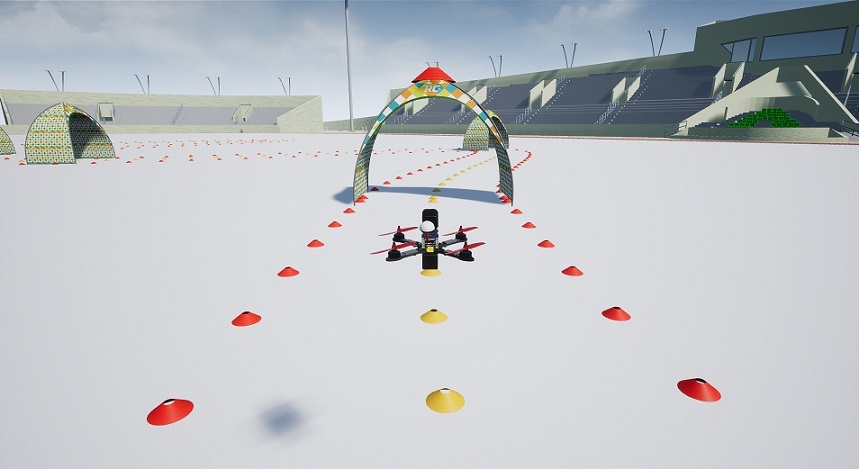}\\
		\small (c) Mud &
		\small (d) Snow \\
		\includegraphics[height=2.3cm]{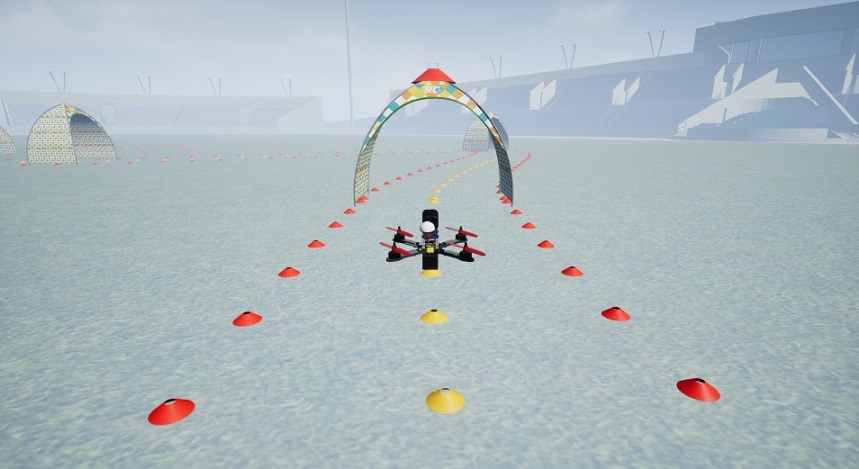} &
		\includegraphics[height=2.3cm]{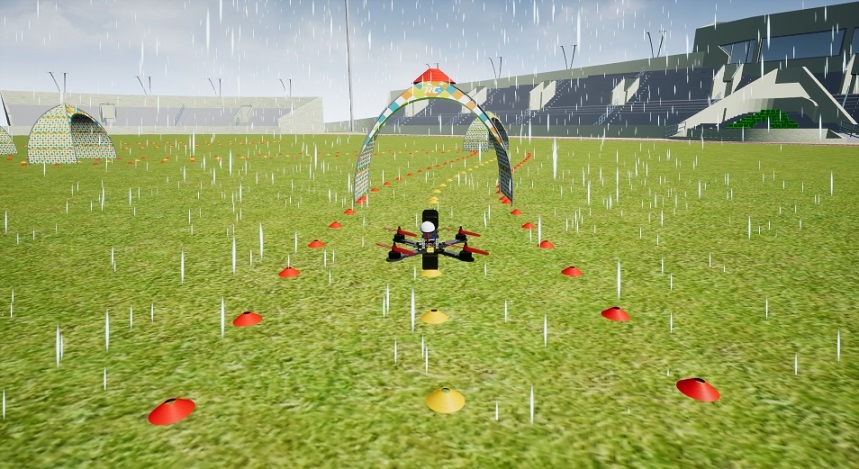} \\
		\small (e) Fog &
		\small (f) Rain \\
		\includegraphics[height=2.3cm]{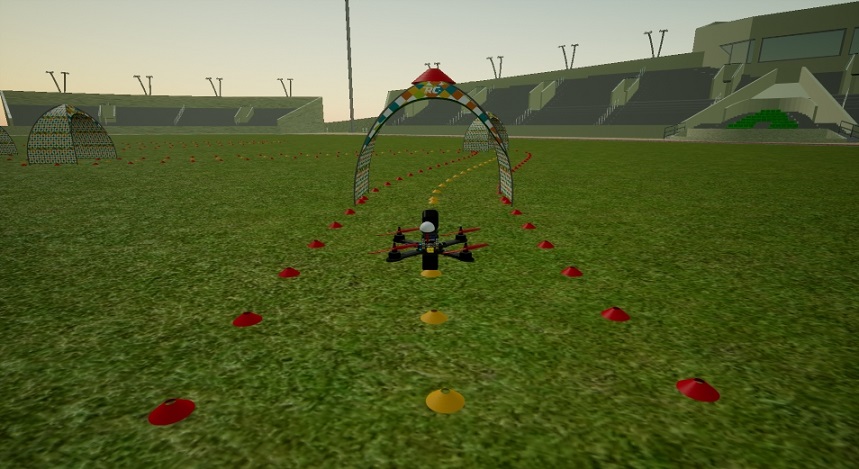} &
    	\includegraphics[height=2.3cm]{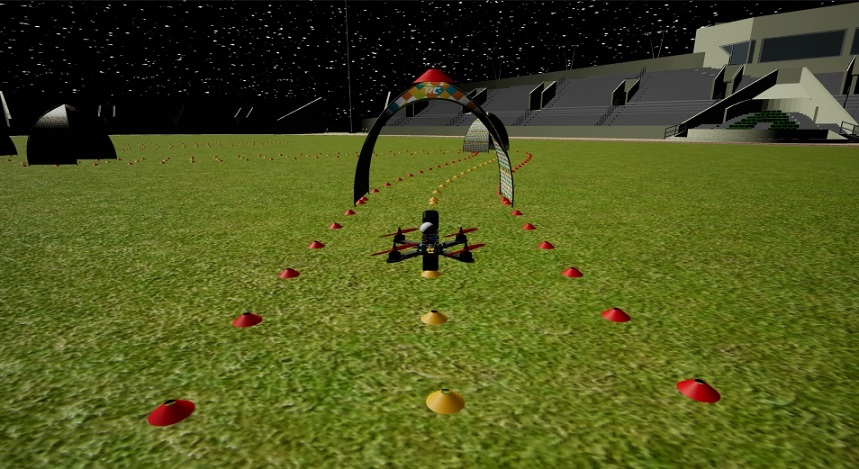} \\
		\small (g) Sunrise &
		\small (h) Night \\

\end{tabular}
\captionof{figure}{Simulated UAV racing stadium with low and high quality grass textures (\emph{a,b}), different ground materials (\emph{c,d}), different weather conditions (\emph{e,f}) and different lighting conditions (\emph{g,h}).}
		\label{fig:generalization}
\end{figure}

\begin{table}[!tb]
\centering
\footnotesize
\caption{Adaptation through modularity. The reported results reflect average performance across all seven testing tracks. Please refer to the appendix \ref{sec: supp} for detailed results per track.}
\setlength\tabcolsep{2.0pt}
\begin{tabular}{rcccccccc}
\hline
\multicolumn{1}{l}{} & \multicolumn{2}{c}{Baseline} & \multicolumn{2}{c}{Weather} & \multicolumn{2}{c}{Lighting} & \multicolumn{2}{c}{Texture} \\
\multicolumn{1}{l}{} & Grass       & GrassHD       & Fog           & Rain        & Sunrise       & Night        & Snow         & Mud          \\ \hline
Score                 & 100\%       & 100\%          & 95.5\%       & 100\%       & 99.2\%       & 96.2\%      & 100\%        & 100\%        \\
Time                  & 66.94       & 69.75          & 85.48         & 68.68       & 73.77         & 86.01        & 68.26        & 67.54        \\
Reset                & 0           & 0              & 0.86          & 0           & 0.14          & 0.71         & 0            & 0            \\ \hline
\end{tabular}
\vspace{7pt}
\label{tbl: adaptation}
\vspace{-12pt}
\end{table}

\mysection{Comparison to PID Controllers and Human Performance.~} We compare our CFN trained control network to the PID controllers it learned from. The perception network stays the same. The summary of this experiment is given in \tblLabel \ref{tbl: ablation}, where we find that our learned policy is able to outperform both PID controllers (PID1 and PID2). Further, as \figLabel \ref{fig:qualitive_results} shows, both PID controllers are imperfect. PID1 completes most gates while flying very slowly and PID2 misses many gates while flying at very high speeds. However, since our control module is designed to learn only from the best behaviour of both PID controllers, it completes all gates at a high speed. We also want to highlight the importance of the \emph{temporary buffer}. Removing it results in a significant drop in performance (score: $79.85\%$, time: $93.45$, resets: $3.86$) since the CFN agent is forced to learn from all the controller demonstrations including the undesirable ones.

We also compare our system to three pilots with different skill levels: novice (has never flown before), intermediate (a moderately experienced pilot), and a professional (a competitive racing pilot with many years of experience). The pilots are given the opportunity to fly the seven training tracks as many times as needed until they successfully complete the tracks at their best time while passing through all gates. The pilots are then scored on the test tracks in the same fashion as the trained networks. The results are summarized in Table \ref{tbl: human}. CFN achieves at least the same accuracy as human pilots but is about two times faster than a novice, 20\% faster than an intermediate pilot, and within 25\% of a professional pilot.
Although slower, our network flies more consistently than even the professional racing pilot while remaining reliably on the track (see \figLabel \ref{fig:qualitive_results}).

\mysection{Adaptation through Modularity.~}
We replace the low-quality grass textures with high-quality grass and show that our perception network generalizes without any modification (see \figLabel \ref{fig:generalization} and \tblLabel \ref{tbl: adaptation}). When changing the weather and lighting conditions, the performance of the perception network starts degrading. The network is not affected by rain, slightly degrades with sunrise lighting and degrades noticeably with fog and night-time lighting. If our perception network was trained on diverse environments/textures, it would learn even more invariance to the background, as demonstrated in \cite{SadeghiL16}. However, generalization only works up to some extent and usually requires heavy data augmentation. In some applications, it might not even be desirable to generalize too broadly as the performance in the target domain often suffers as a result. For such cases, our modular approach allows to simply swap out the perception module to adapt to any environment. To demonstrate this, we train the perception network on different textured environments while keeping the controller fusion network fixed and show successful transfer of the control policy.

\section{Conclusions and Future Work}\label{sec: conclusion}
In this work, we present a controller fusion network (CFN) that allows fusing multiple classical controllers. Extensive experiments demonstrate that a CFN based network outperforms state-of-the-art methods and flies more consistently than human pilots. This a product of both the ability for the network to fuse multiple controller's trajectories and at the same time filter out controller actions leading to poor performance. We expect the framework can be adapted for other robotic and controller based dynamic tasks such as visual grasping tasks or visual placing tasks by making minor changes in buffer strategy. Instead of relying on extensive fine-tuning of a controller or defining an explicit model of a system, a CFN is able to produce an optimized predictive control of dynamic systems. 

\mysection{Acknowledgments} This work was supported by the King Abdullah University of Science and Technology (KAUST) Office of Sponsored Research.

{\small
\bibliographystyle{ieee_fullname}
\bibliography{references}
}

\clearpage
\appendix

\section{Supplementary Material} \label{sec: supp}
Here we provide additional results and the recorded paths of the UAV networks and human pilots evaluations in the paper. All results were recorded as logs during testing inside Sim4CV \cite{sim4cv} allowing plotting on the GUI track interface developed for the paper. The logs record stick input, position, orientation and velocity. These allow visualization of the performance of the pilot/network on the different tracks.  
Tables  \ref{tbl: environment} and \ref{tbl: textures} show the detailed results for adaptation to different textures, lighting and environment conditions. 

\begin{figure*}
\centering
\begin{tabular}{@{}c@{\hspace{1mm}}c@{\hspace{1mm}}c@{\hspace{8mm}}c@{}}
		\includegraphics[height=3cm]{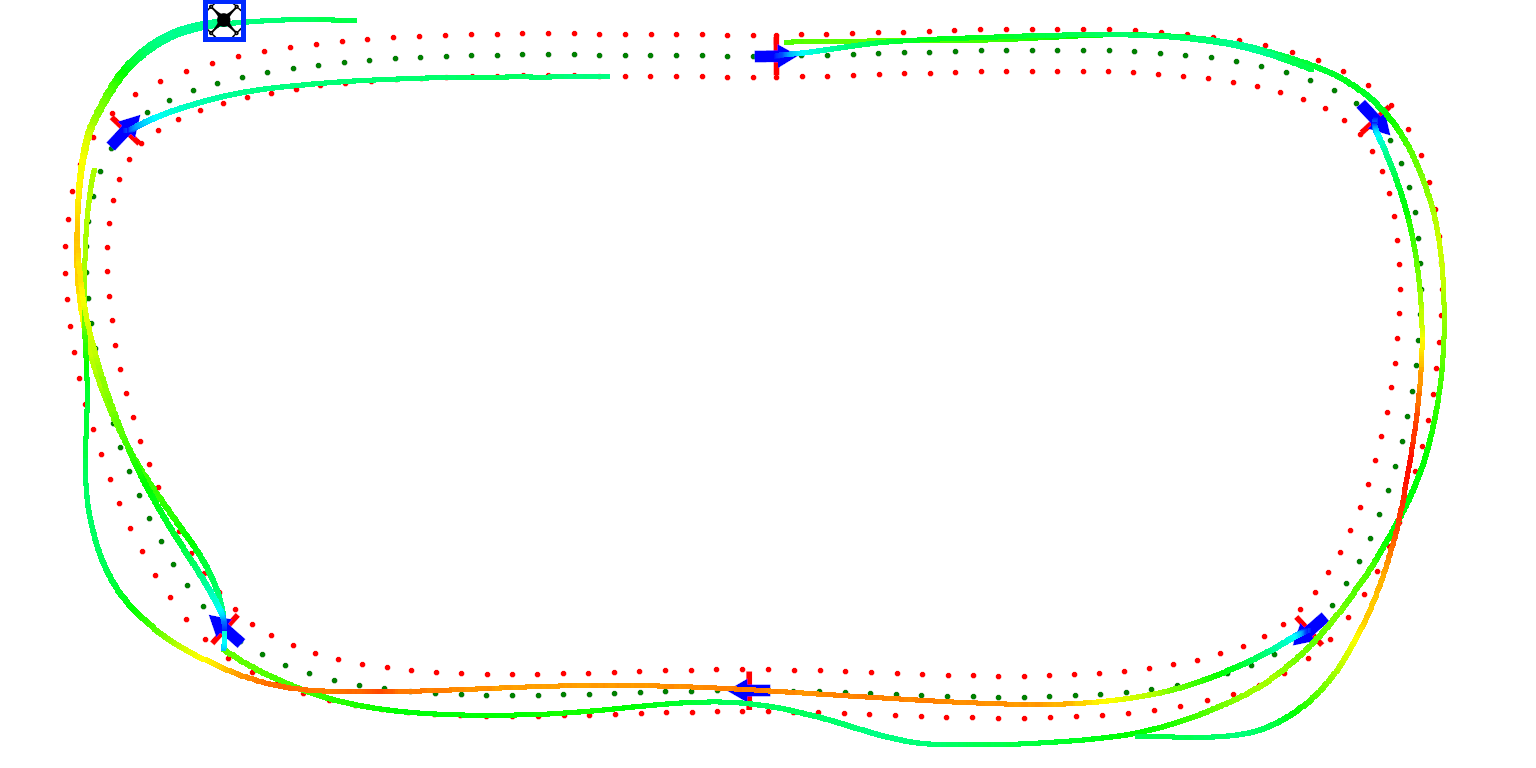} &
		\includegraphics[height=3cm]{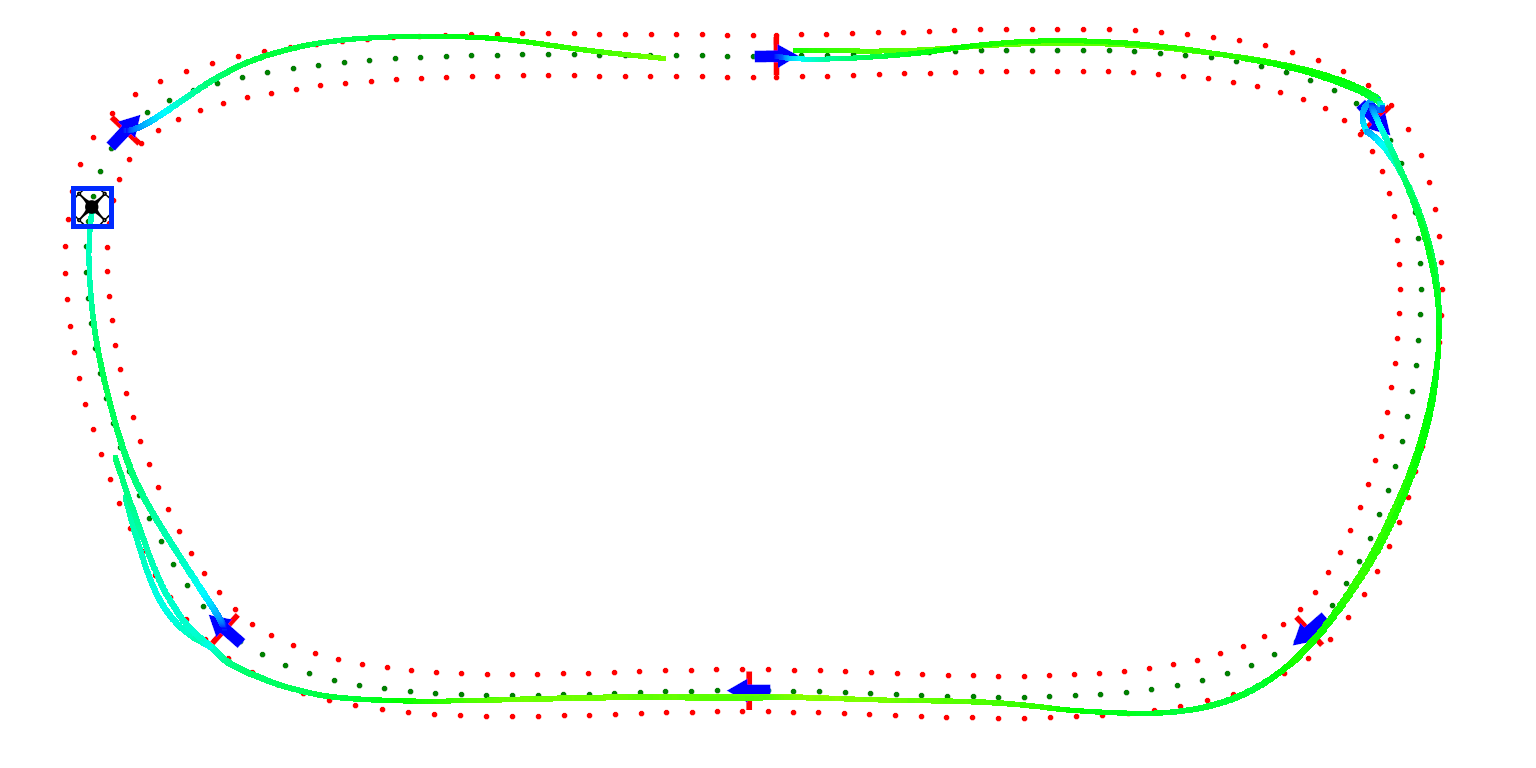} &
		\includegraphics[height=3cm]{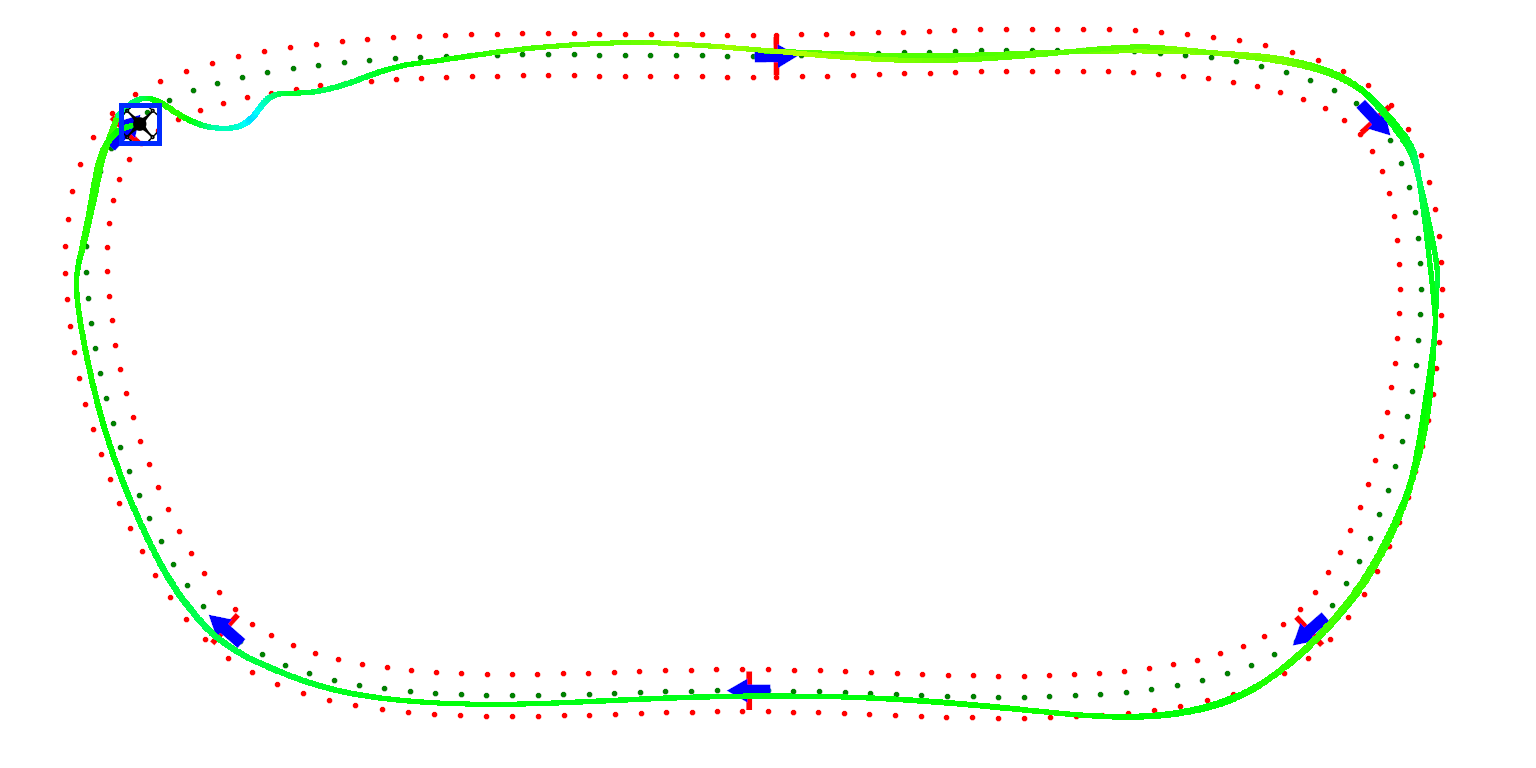} \\
		\small (a) End2End (MAV) & \small (b) End2End (Nvidia) & \small (c) End2End (Ours) \\
		\includegraphics[height=3cm]{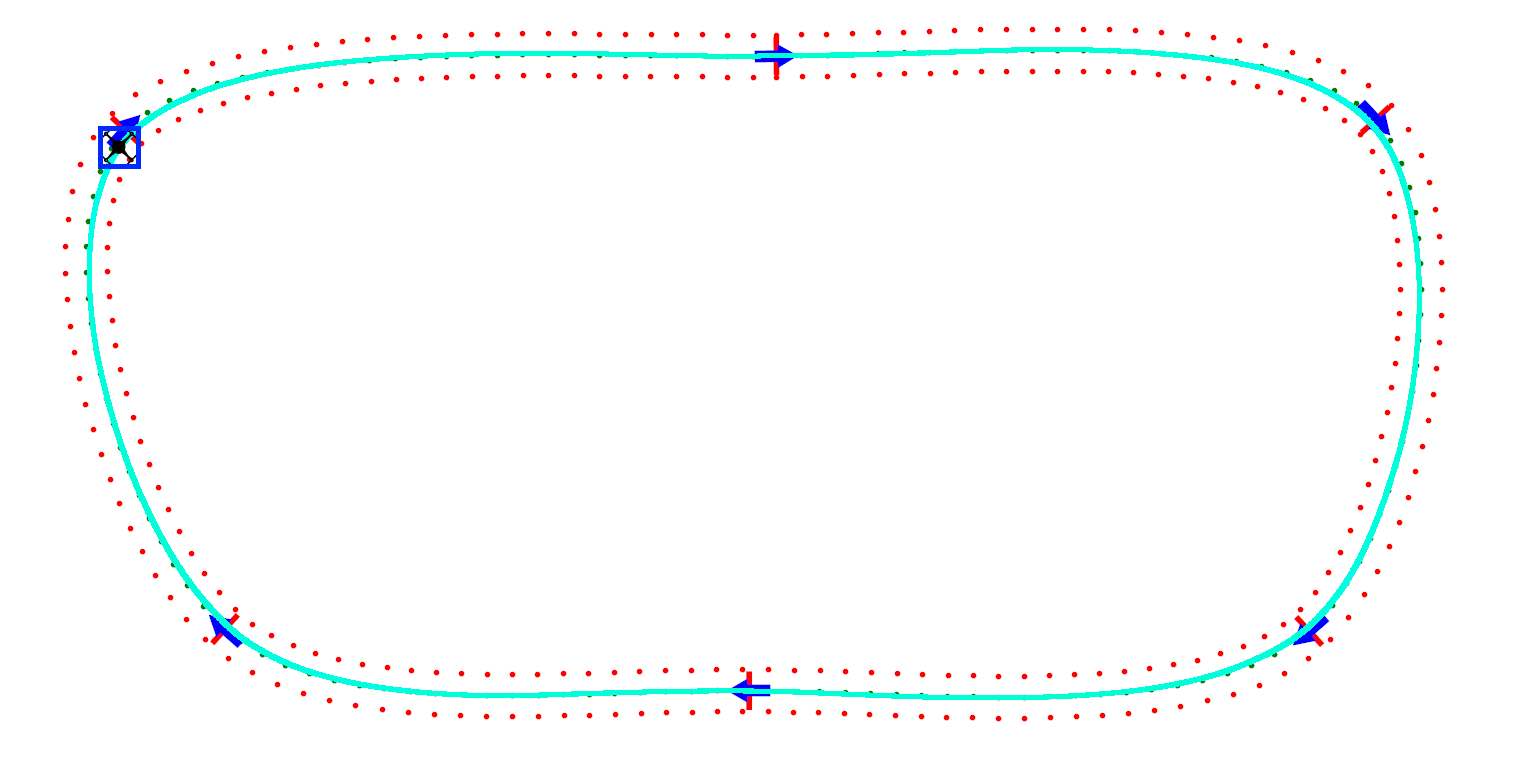} &
		\includegraphics[height=3cm]{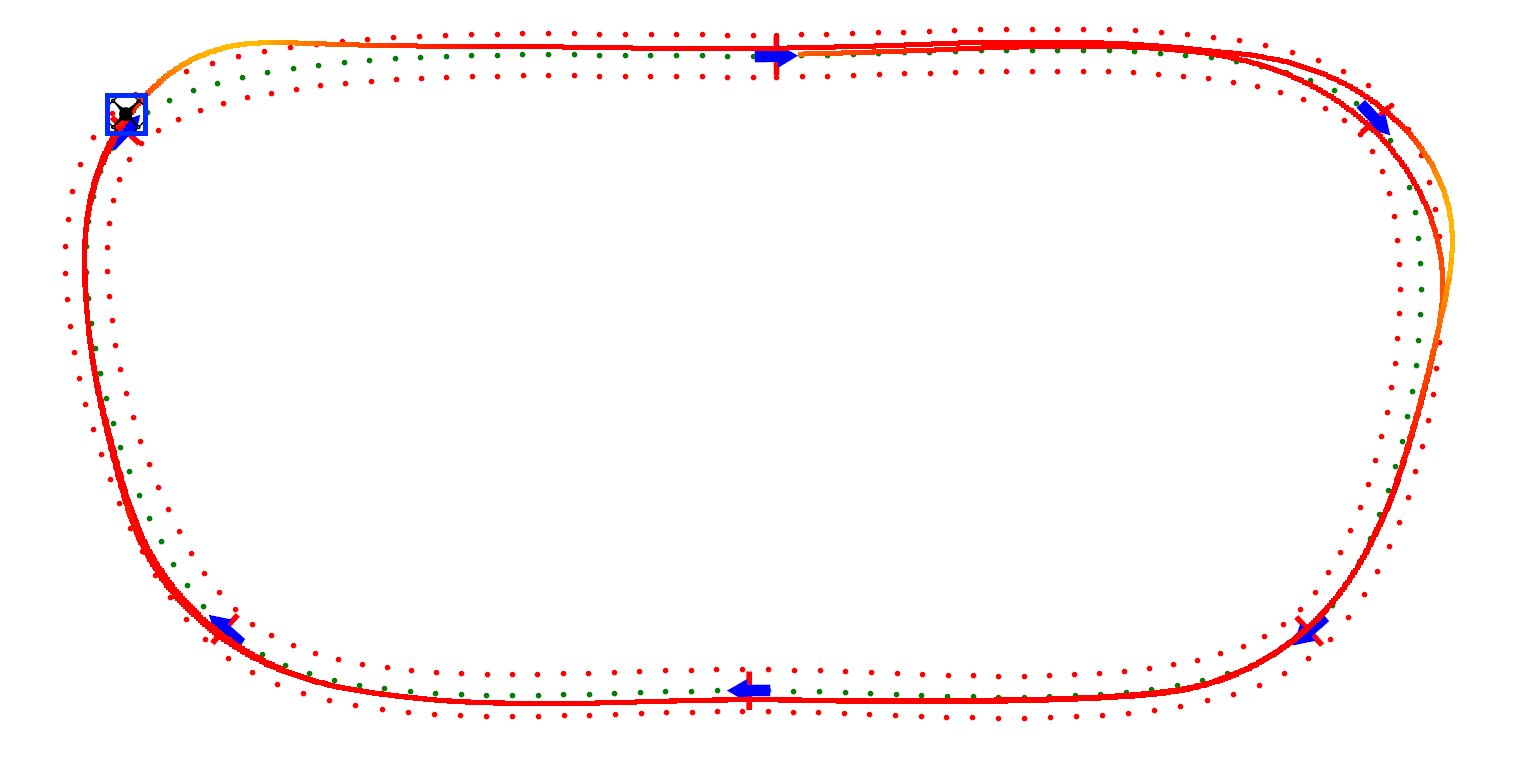} &
		\includegraphics[height=3cm]{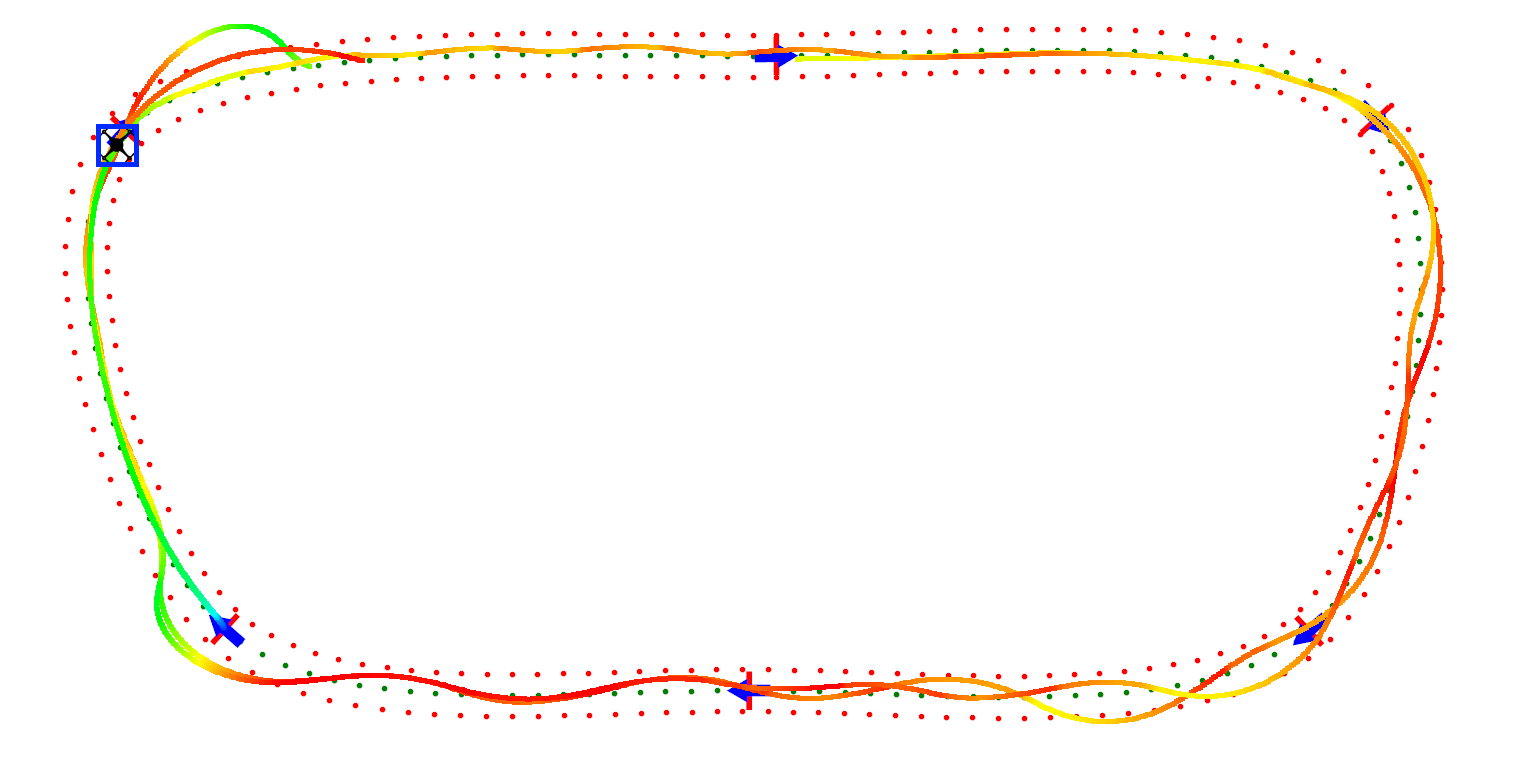} \\
		\small (c) PID1 (Conservative) & \small (d) PID2 (Aggressive) & \small (e) Ours (No Buffer) \\
		\includegraphics[height=3cm]{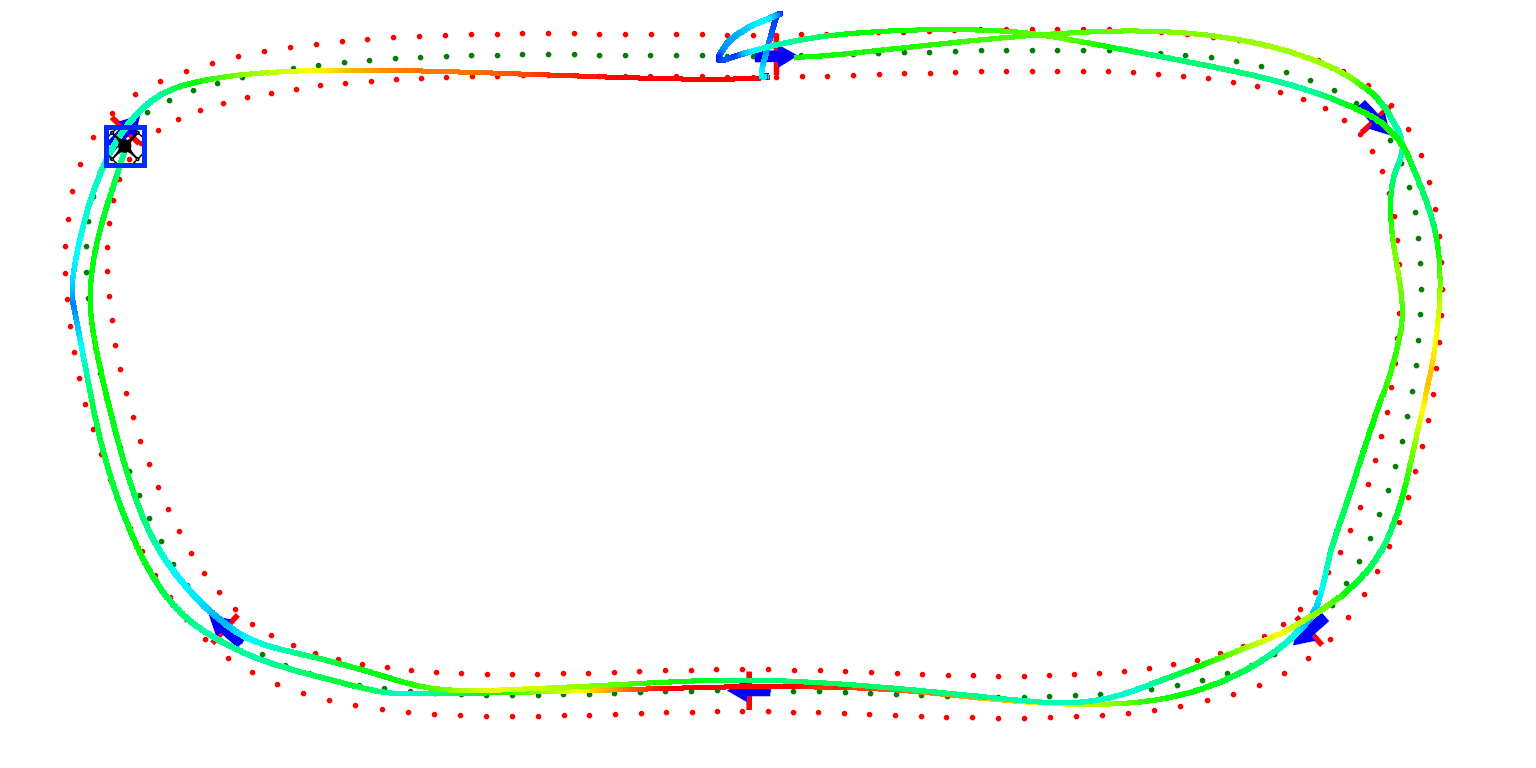} &
		\includegraphics[height=3cm]{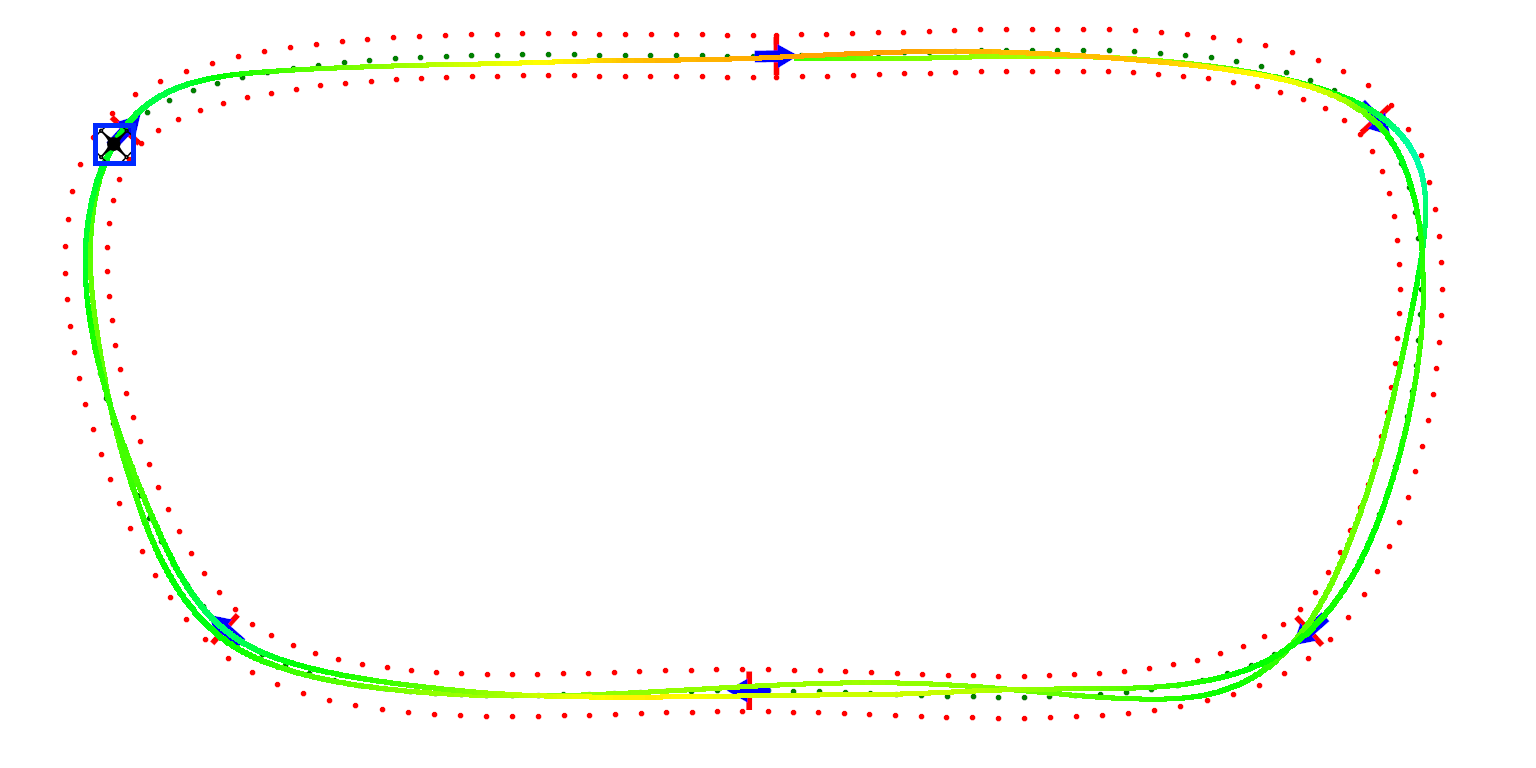} &
		\includegraphics[height=3cm]{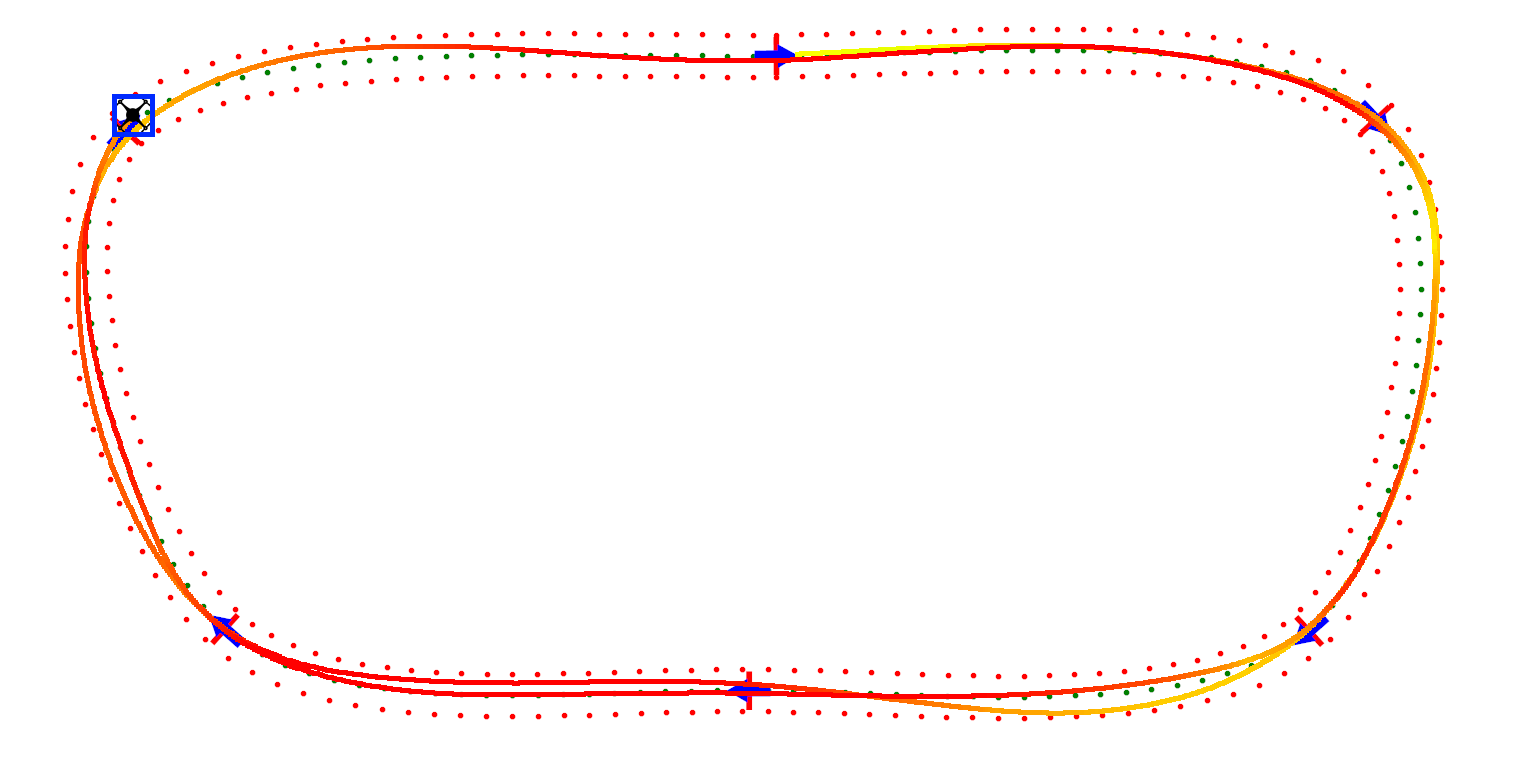} \\
        \small (f) Human (Novice) & \small (g) Human (Intermediate) & \small (h) Human (Professional)\\
		\includegraphics[height=3cm]{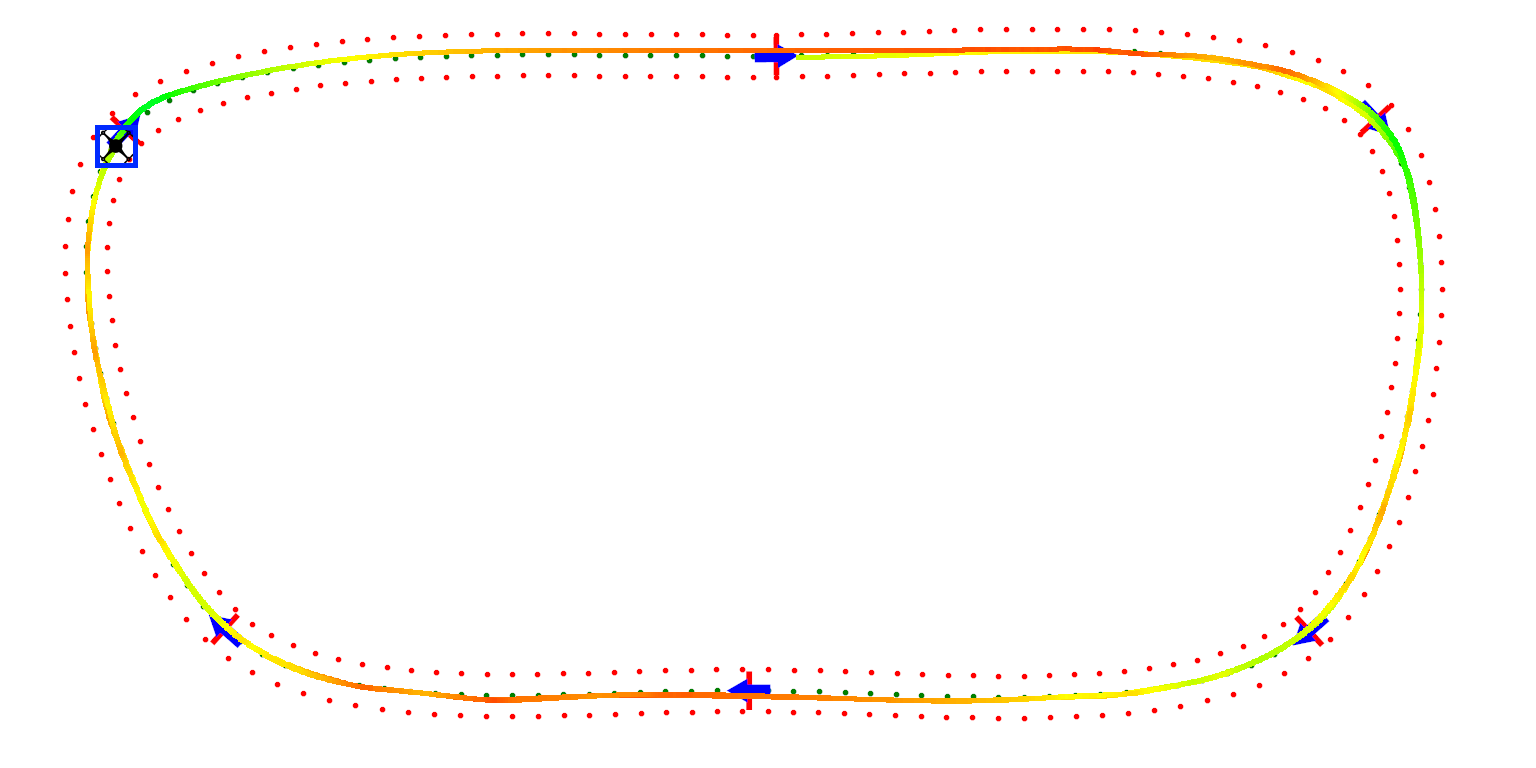} &
		\includegraphics[height=3cm]{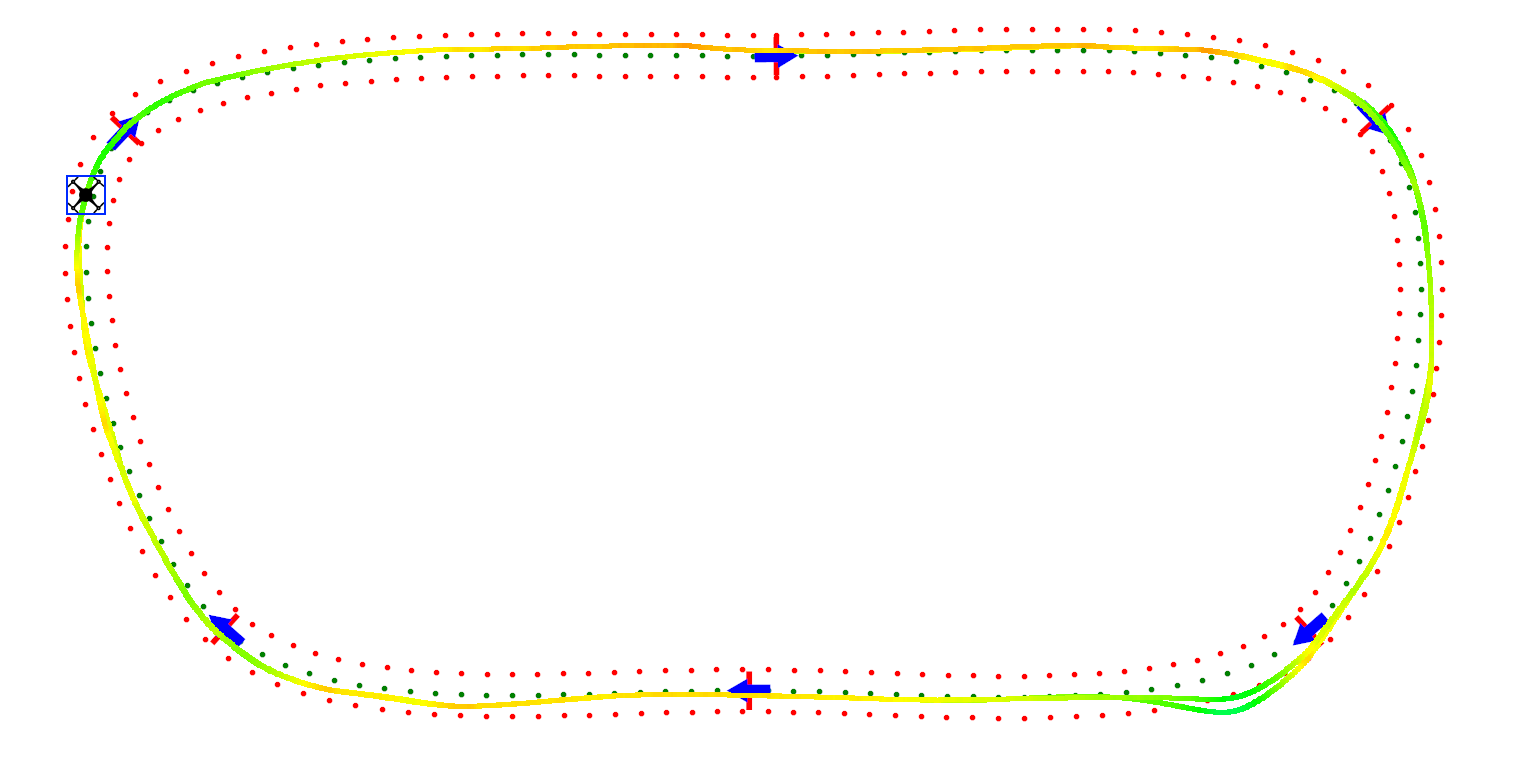} &
		\includegraphics[height=3cm]{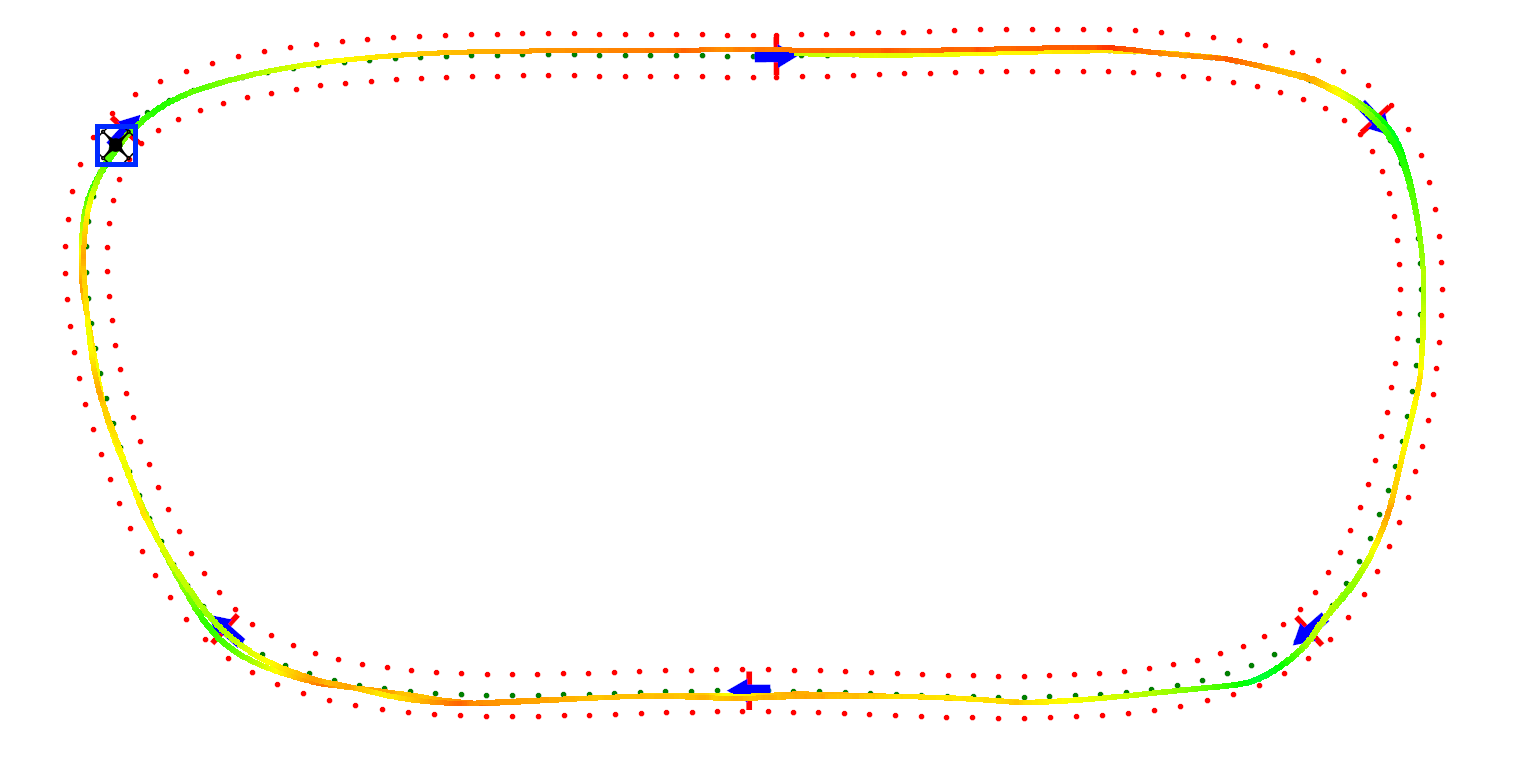} \\
		\small (i) Ours (Reference) & \small (j) Ours (Night) & \small (k)  Ours (Sunrise) \\
		\includegraphics[height=3cm]{sup_figures/track1_ours_grass.png} &
		\includegraphics[height=3cm]{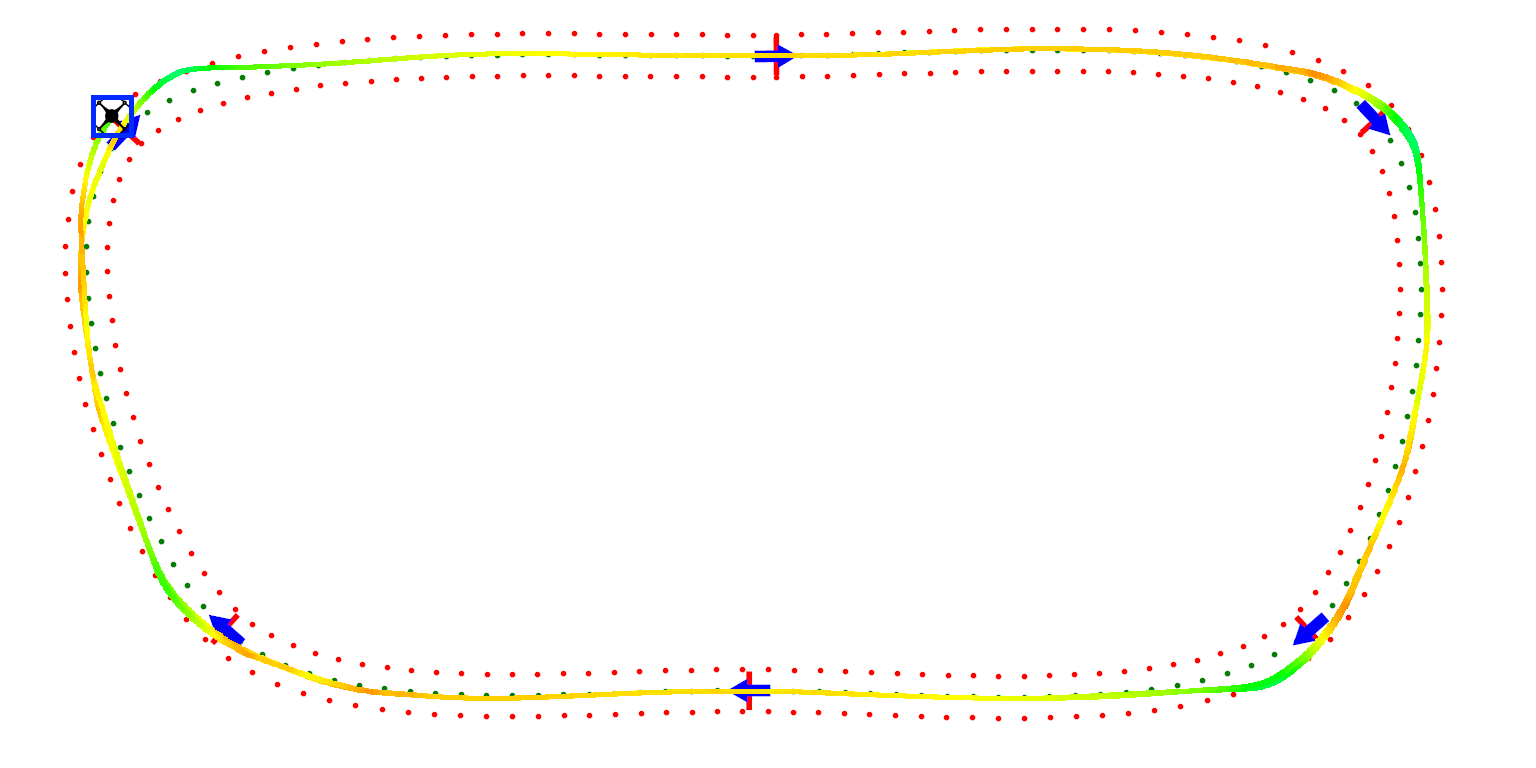} &
		\includegraphics[height=3cm]{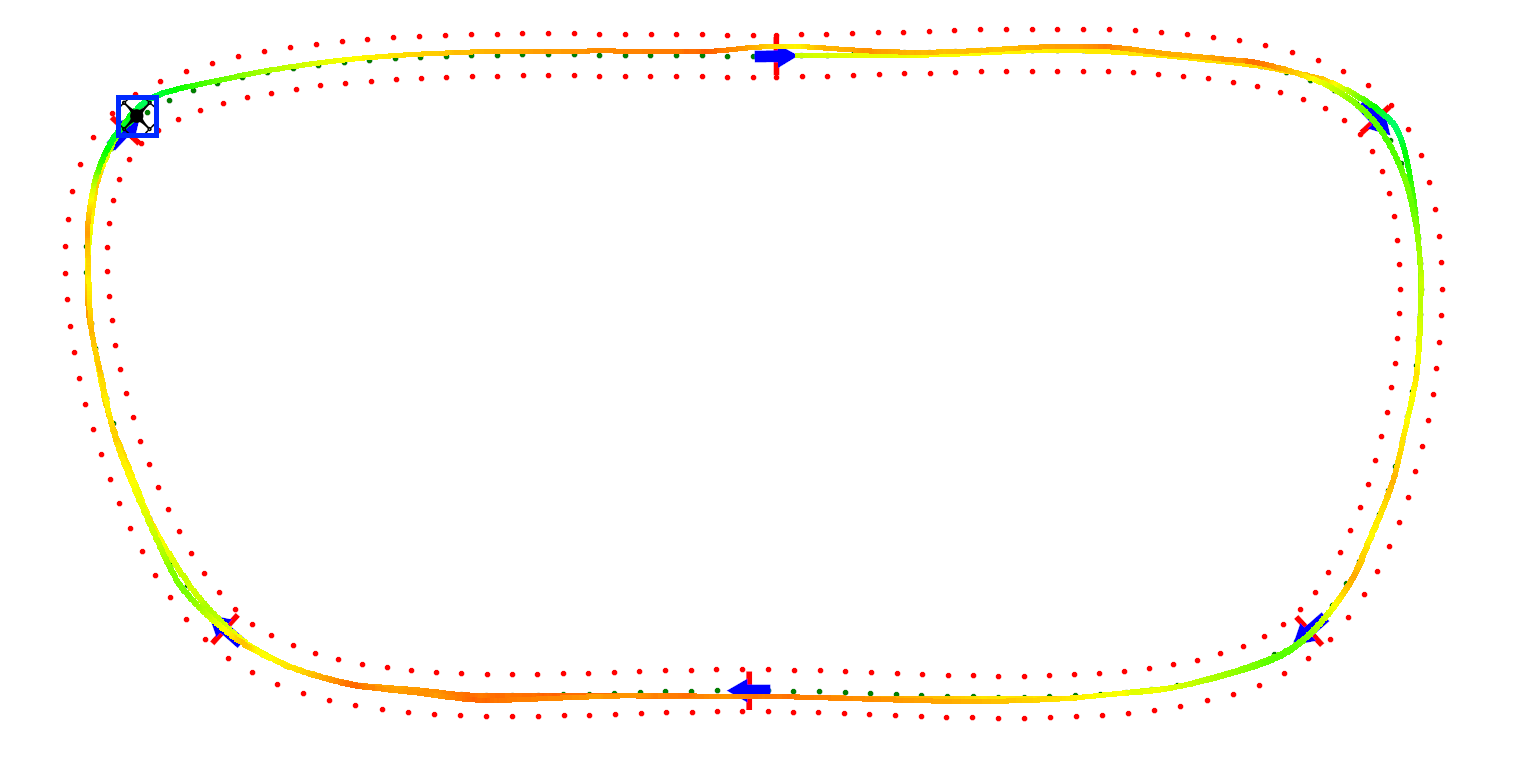} \\
		\small (l) Ours (Reference) & \small (m) Ours (Fog) & \small (o)  Ours (Rain) \\
		\includegraphics[height=3cm]{sup_figures/track1_ours_grass.png} &
		\includegraphics[height=3cm]{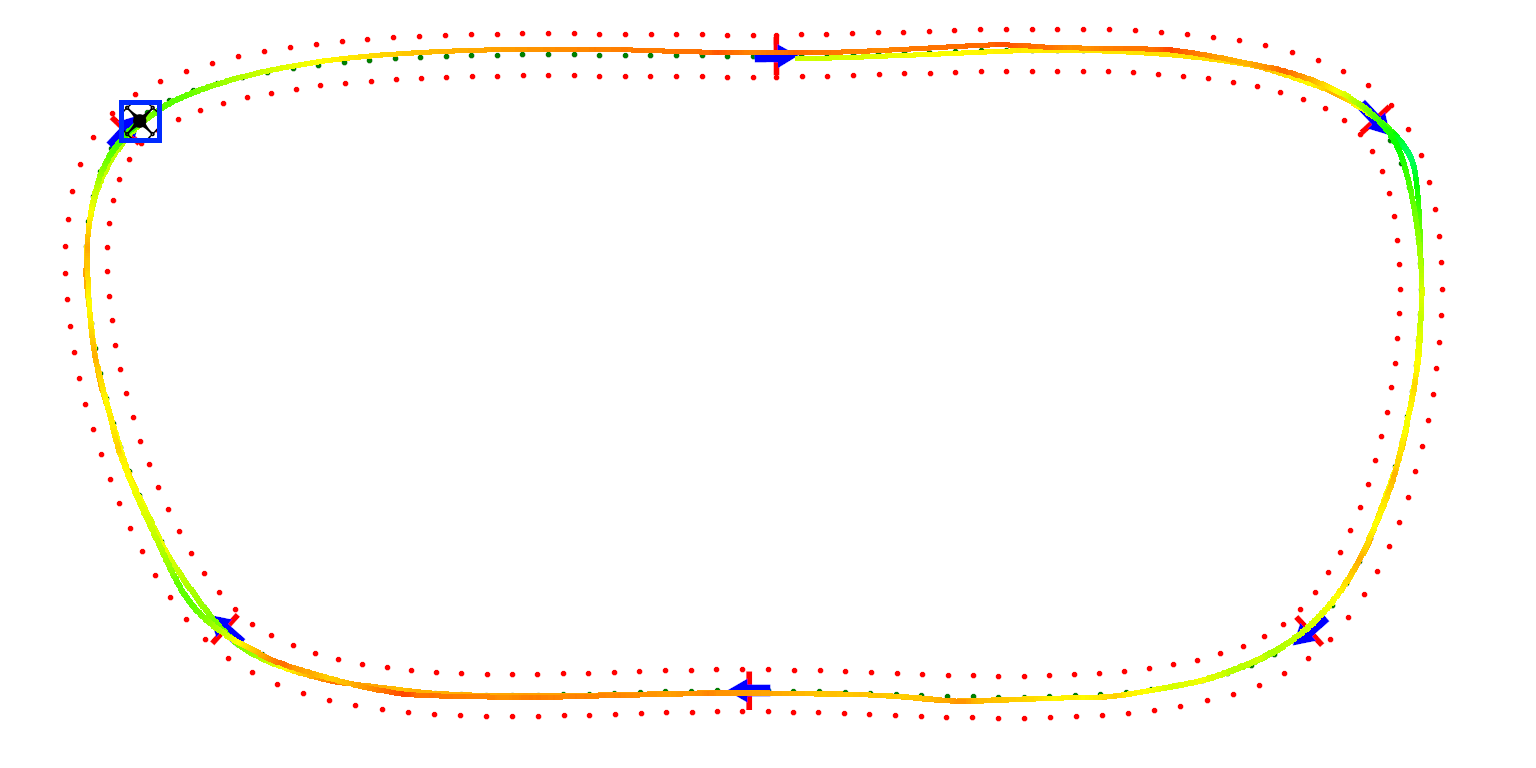} &
		\includegraphics[height=3cm]{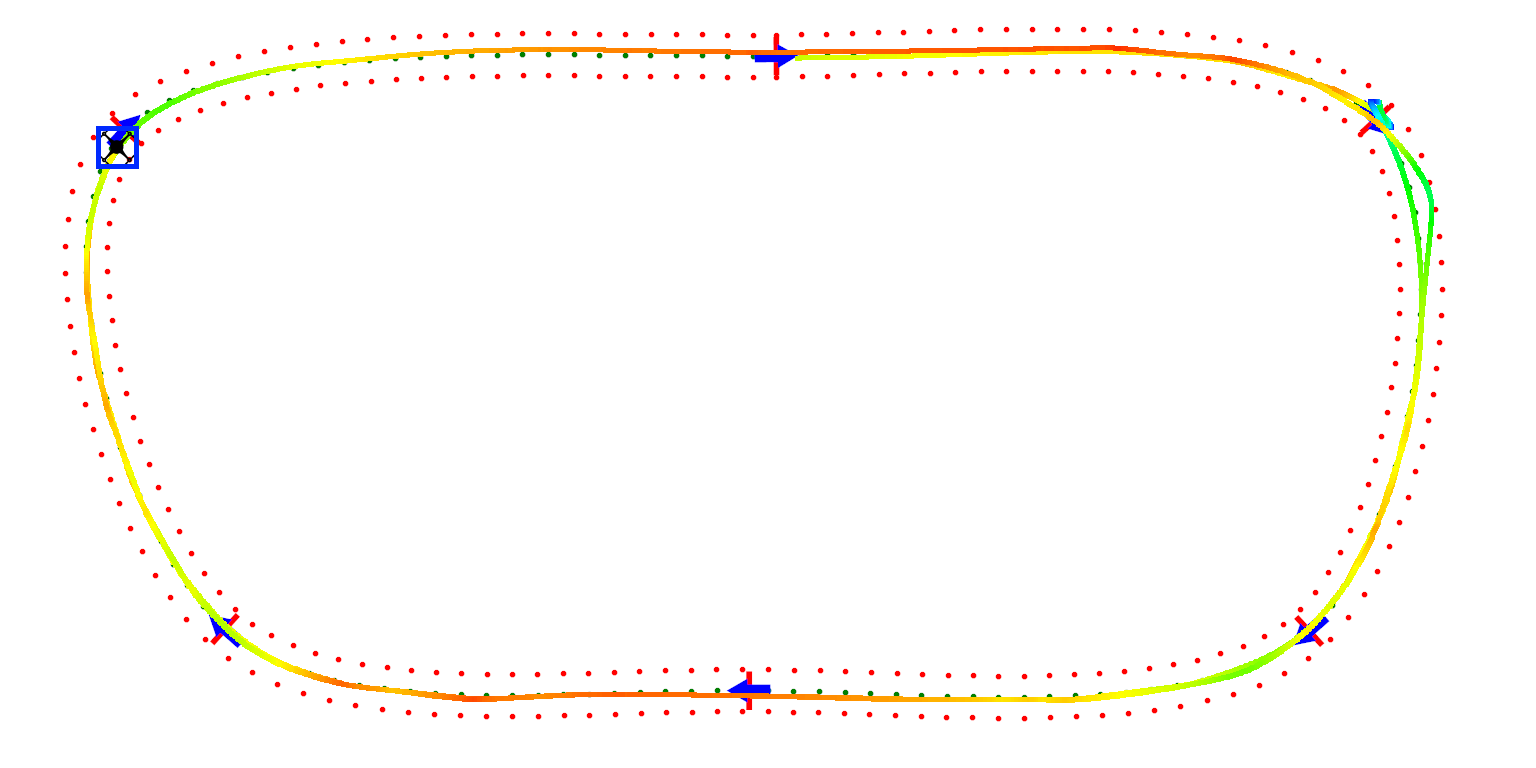} \\
		\small (p) Ours (Grass) & \small (q) Ours (Mud) & \small (r)  Ours (Snow) \\
        \multicolumn{3}{c}{\includegraphics[height=1.2cm]{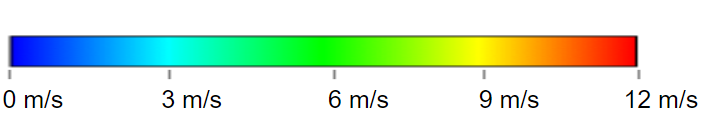}}
\end{tabular}
\captionof{figure}{Qualitative results on track1. The color encodes speed as a heatmap, where blue corresponds to the minimum speed and red to the maximum speed.}
\label{fig:qualitive_results_track1}
\end{figure*}

\begin{figure*}
\centering
\begin{tabular}{@{}c@{\hspace{1mm}}c@{\hspace{1mm}}c@{\hspace{8mm}}c@{}}
		\includegraphics[height=3cm]{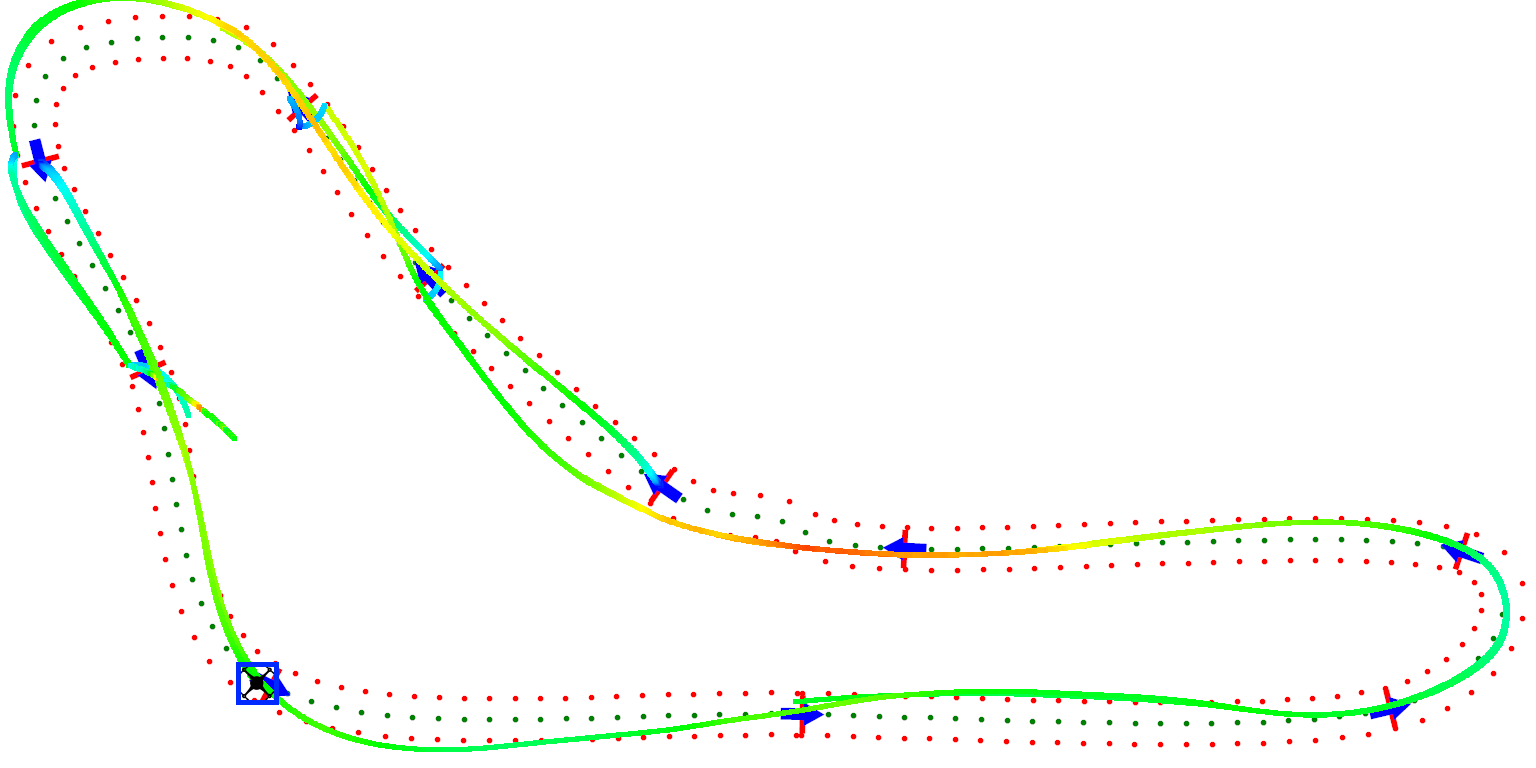} &
		\includegraphics[height=3cm]{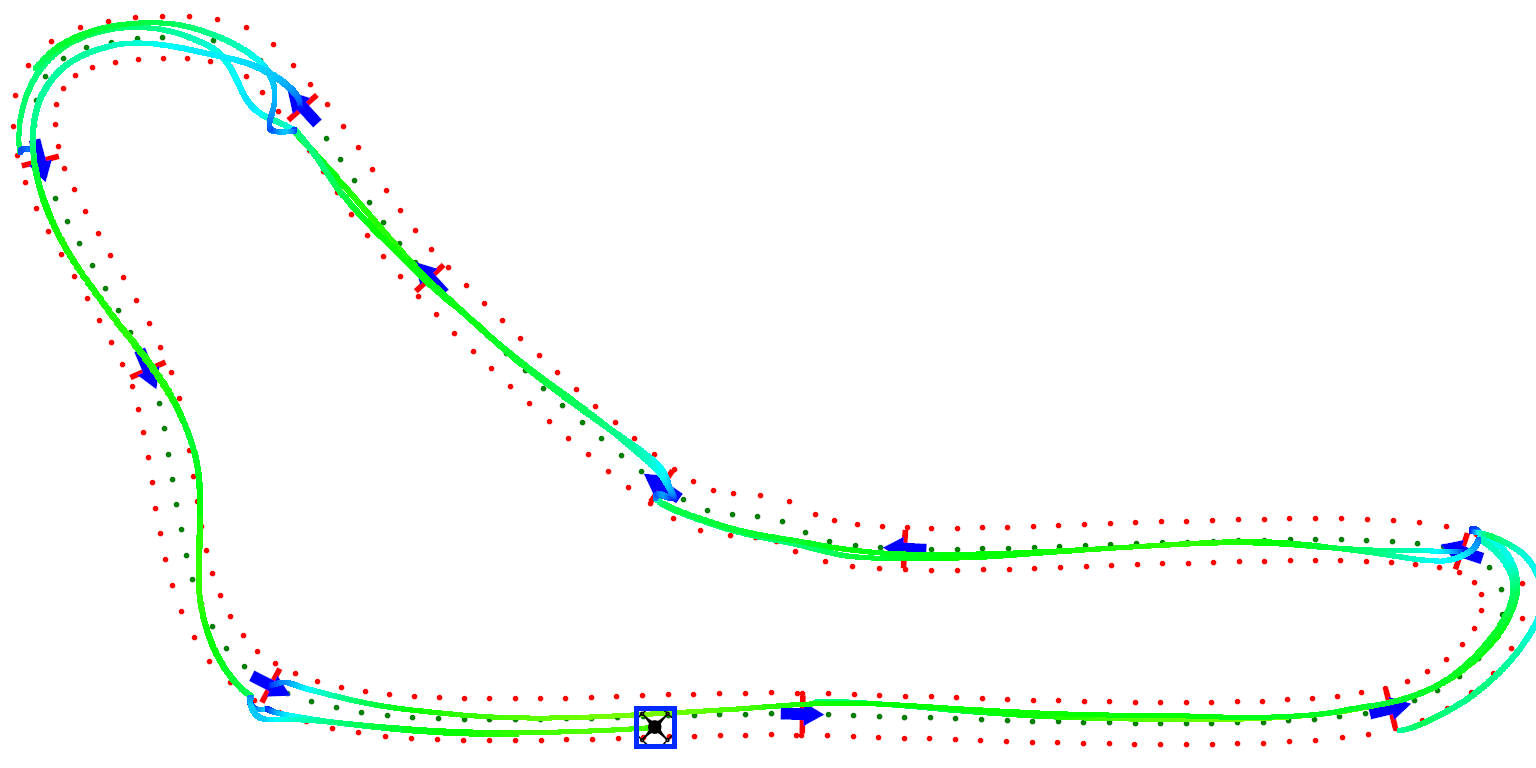} &
		\includegraphics[height=3cm]{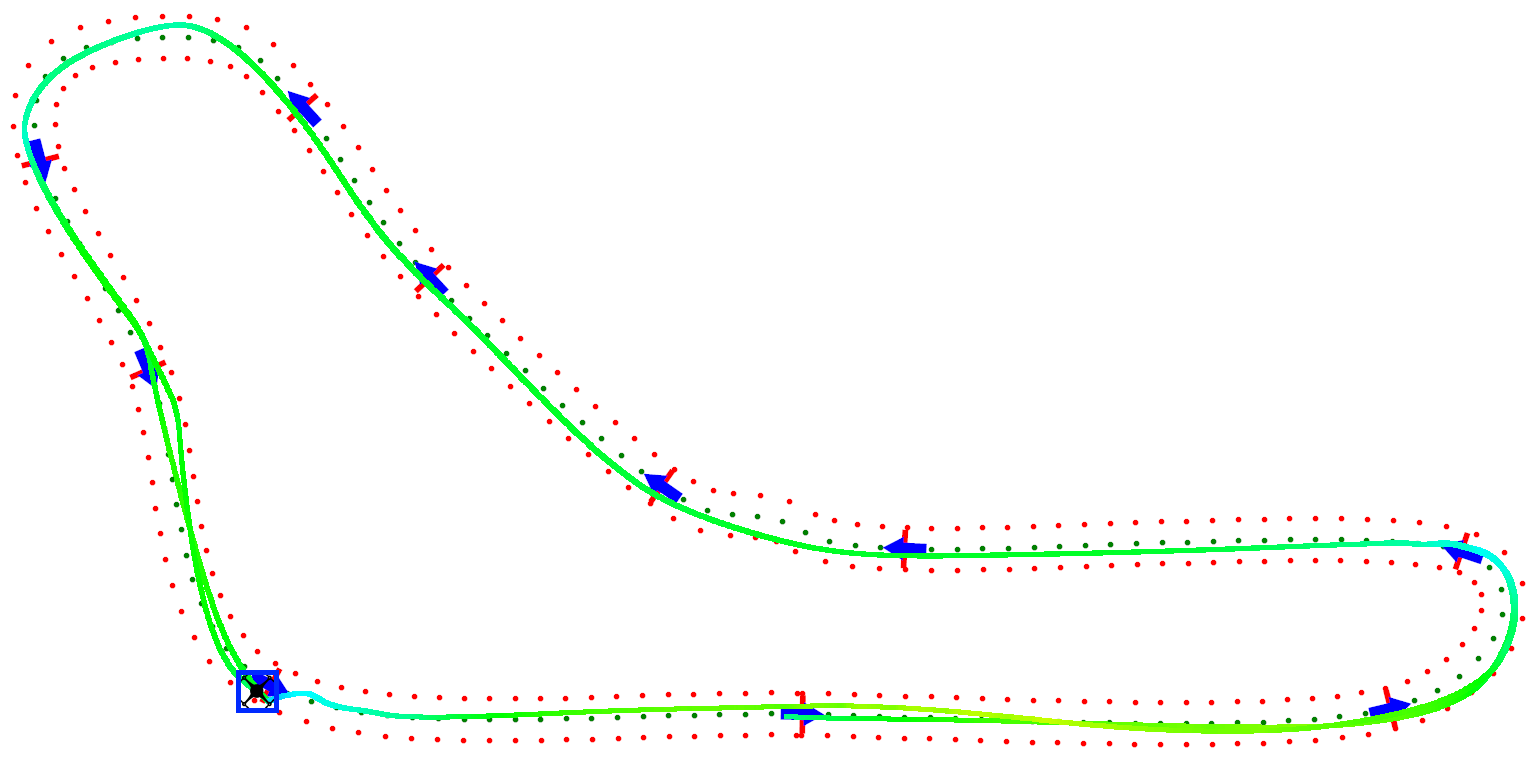} \\
		\small (a) End2End (MAV) & \small (b) End2End (Nvidia) & \small (c) End2End (Ours) \\
		\includegraphics[height=3cm]{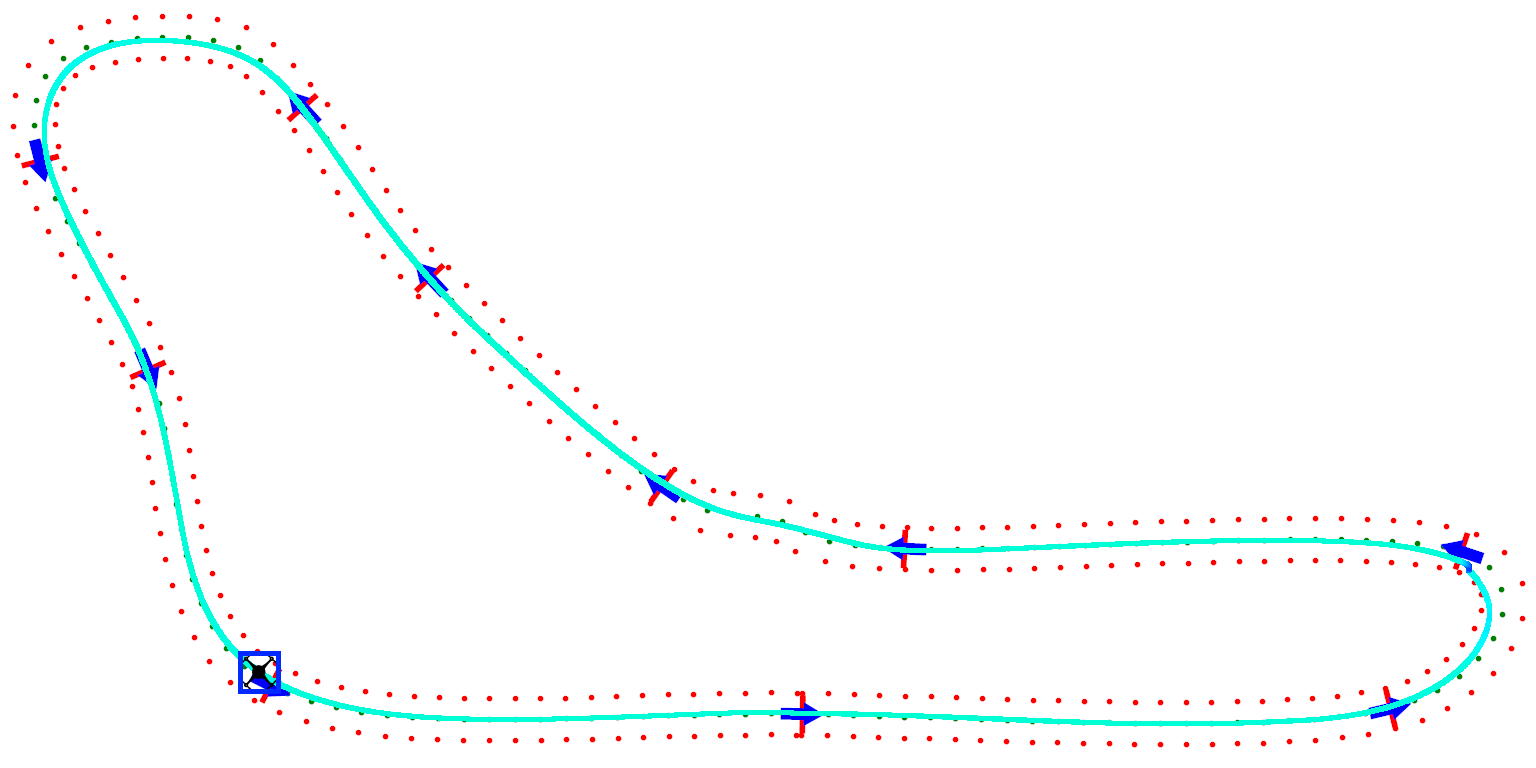} &
		\includegraphics[height=3cm]{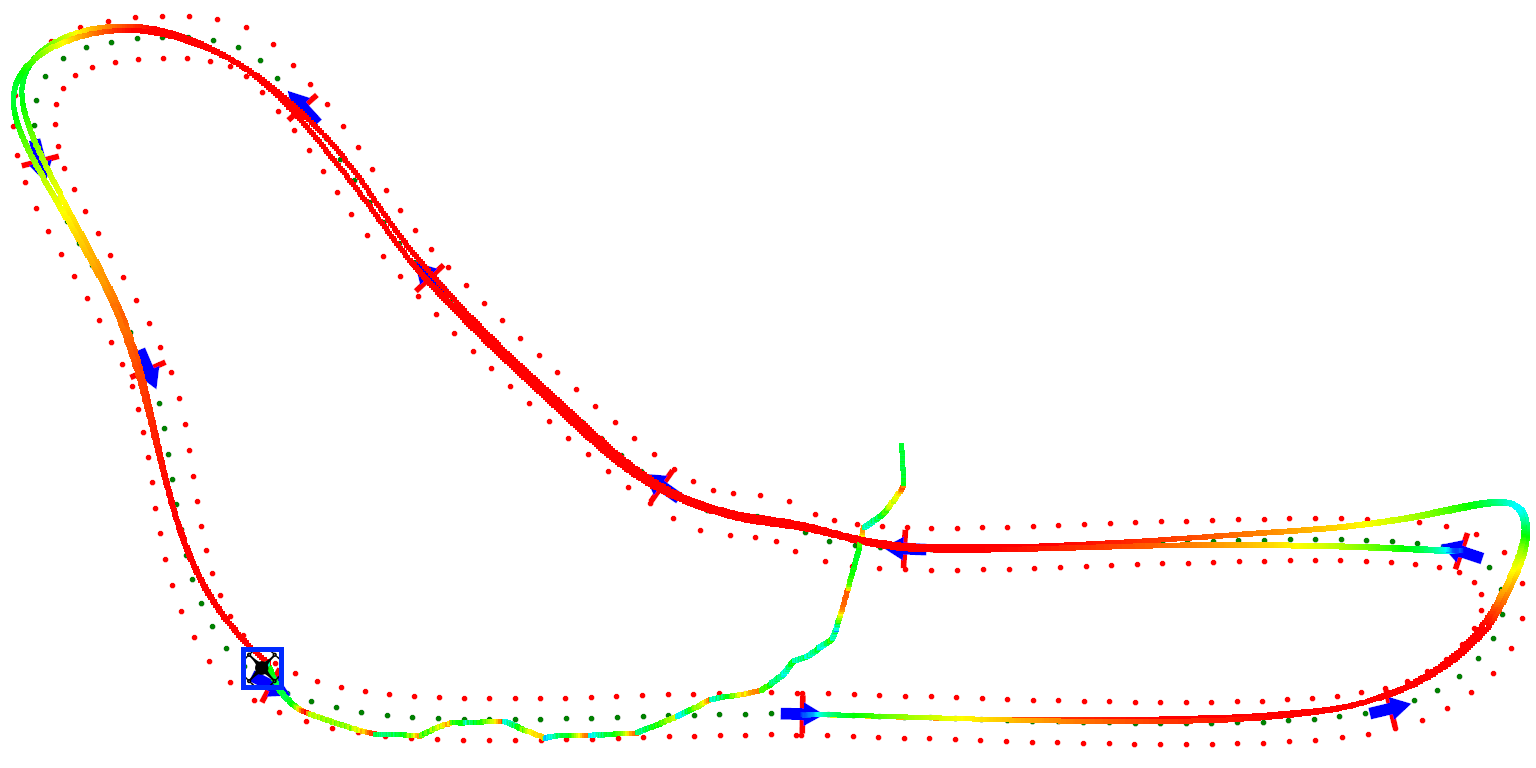} &
		\includegraphics[height=3cm]{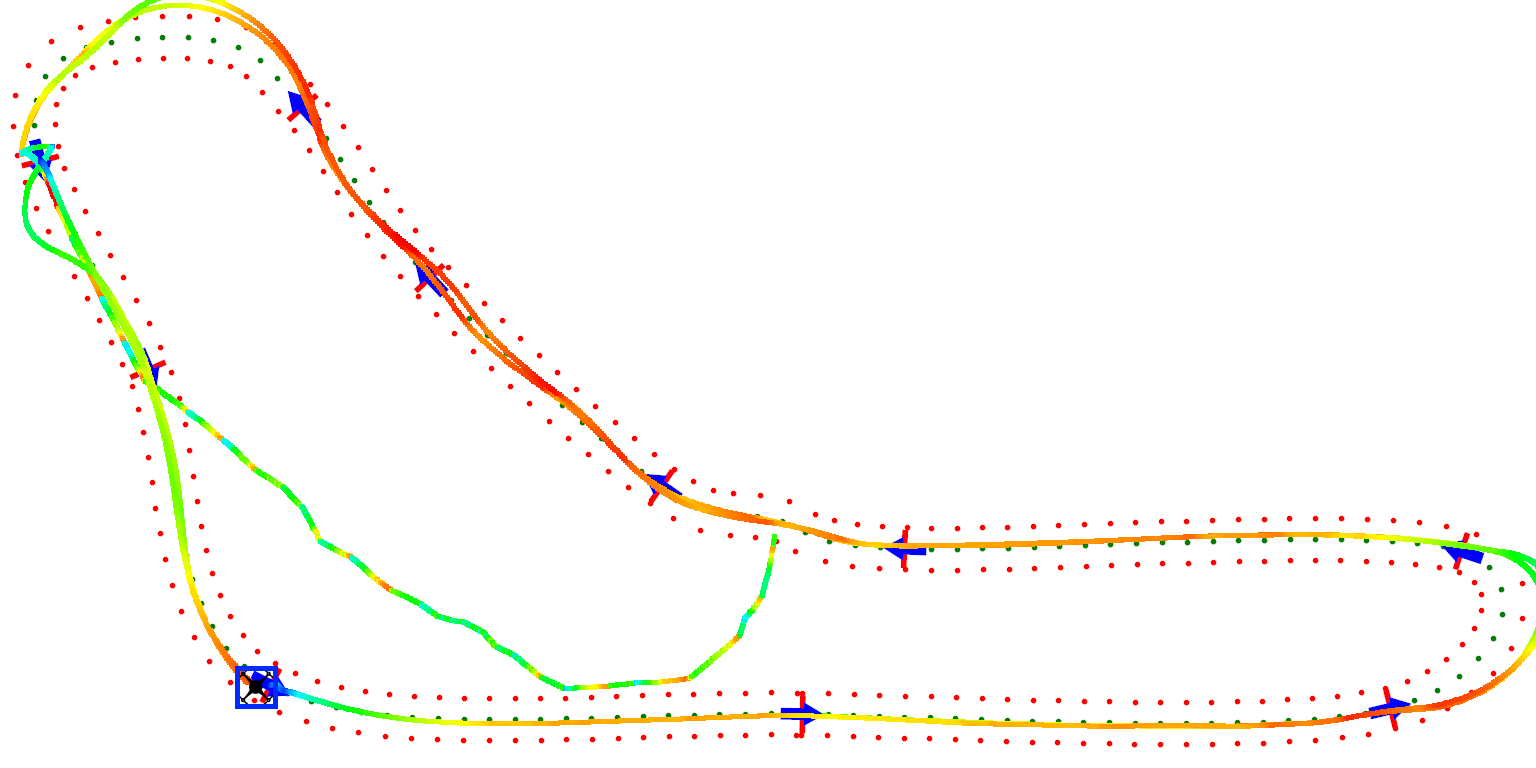} \\
		\small (c) PID1 (Conservative) & \small (d) PID2 (Aggressive) & \small (e) Ours (No Buffer) \\
		\includegraphics[height=3cm]{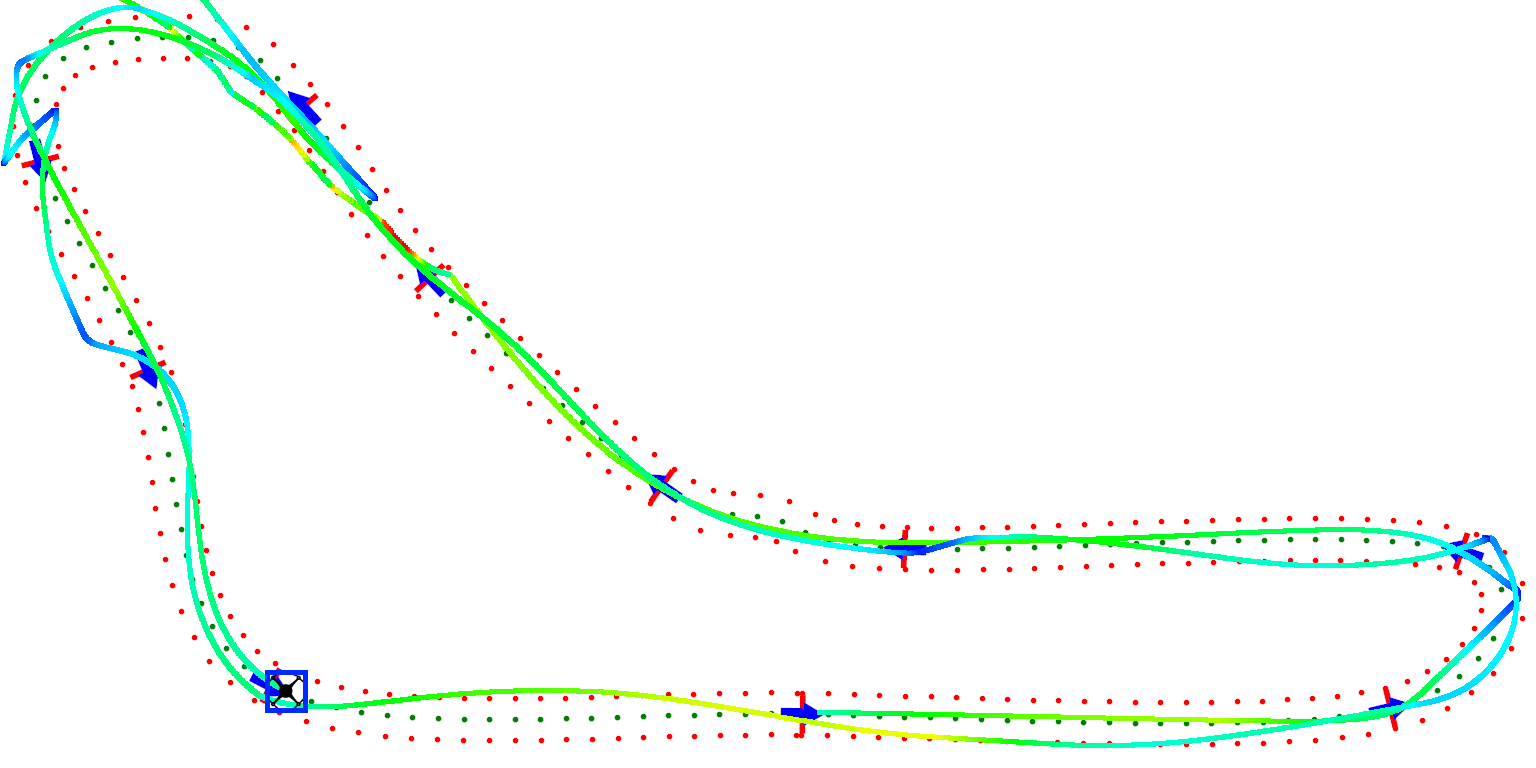} &
		\includegraphics[height=3cm]{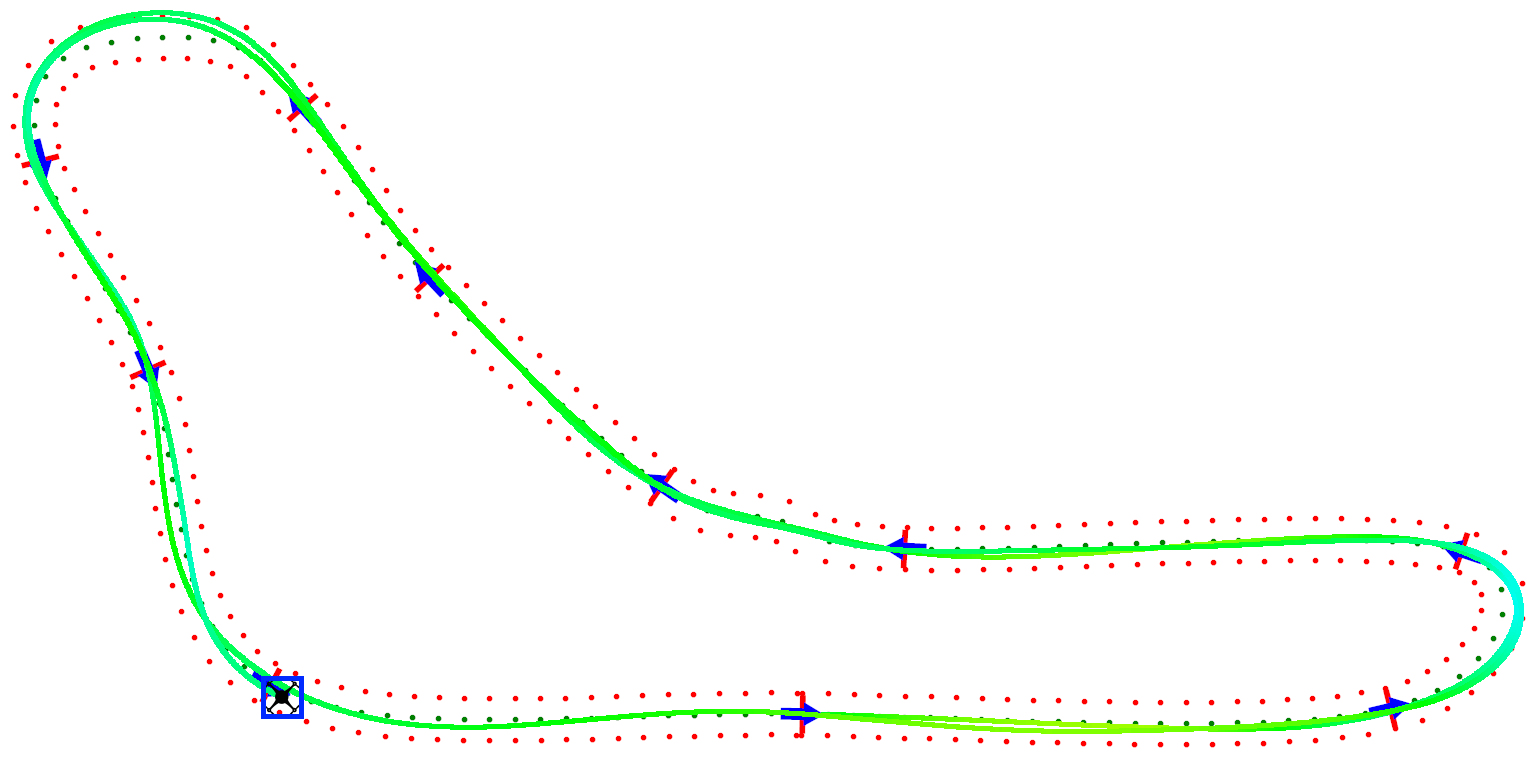} &
		\includegraphics[height=3cm]{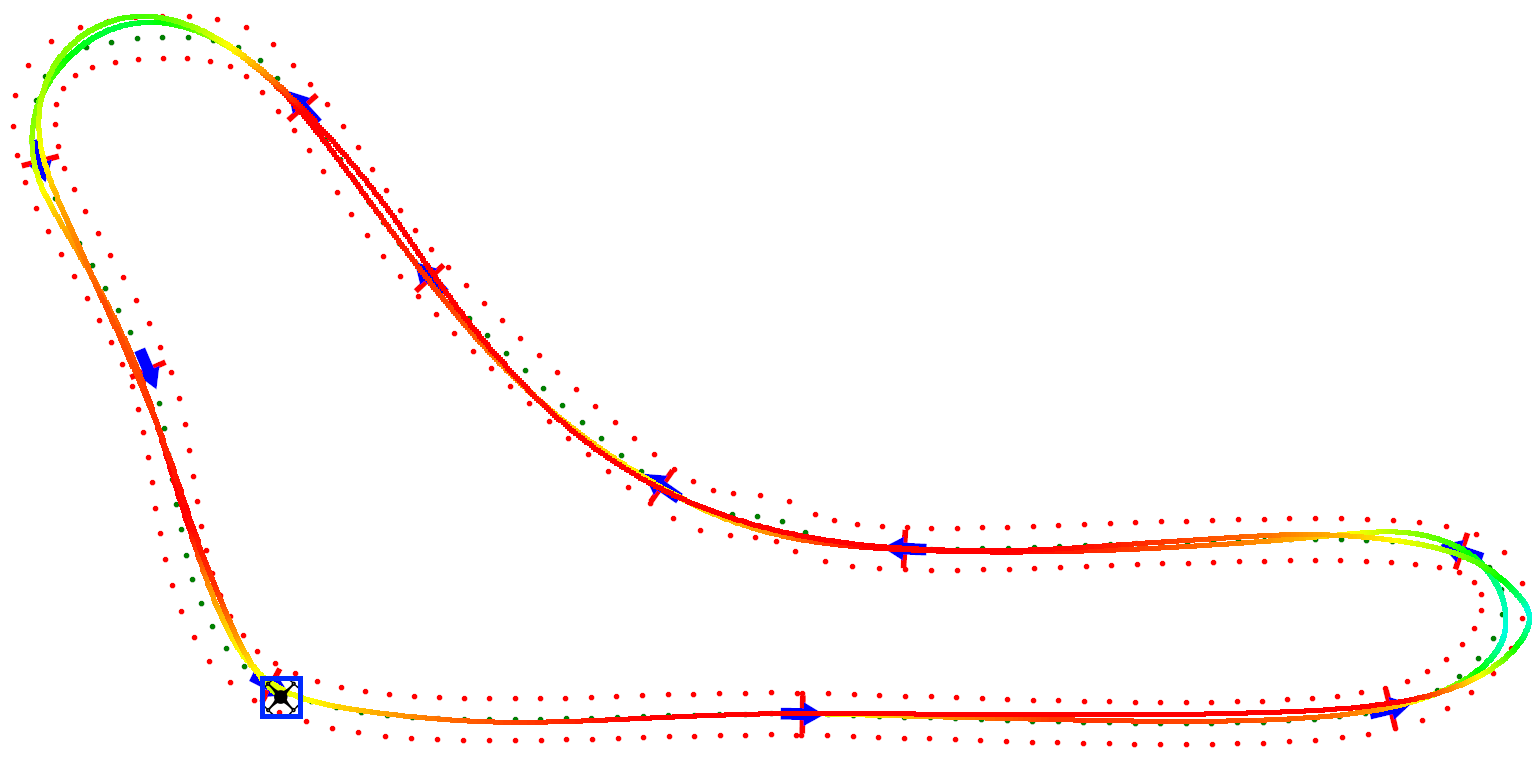} \\
        \small (f) Human (Novice) & \small (g) Human (Intermediate) & \small (h) Human (Professional)\\
		\includegraphics[height=3cm]{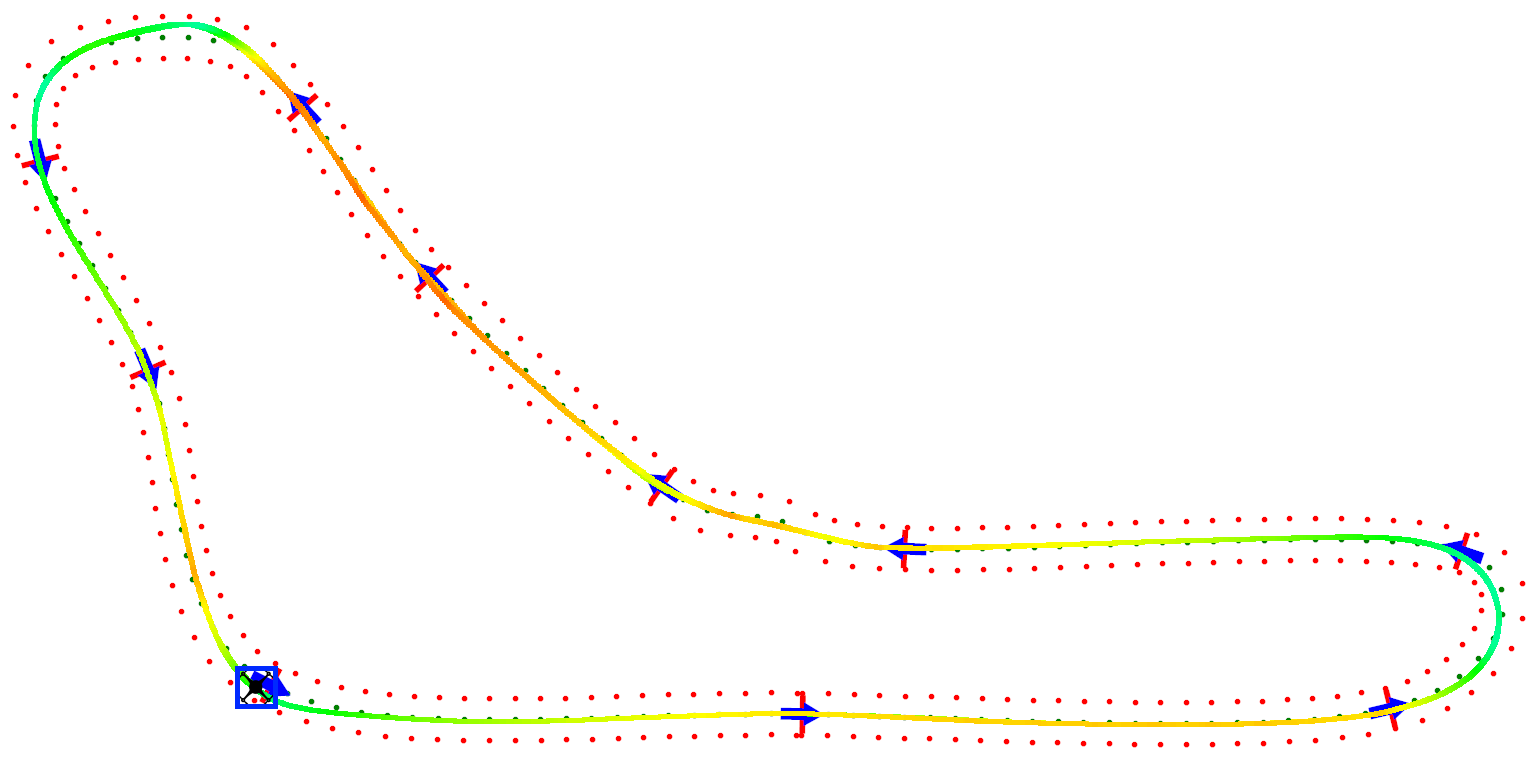} &
		\includegraphics[height=3cm]{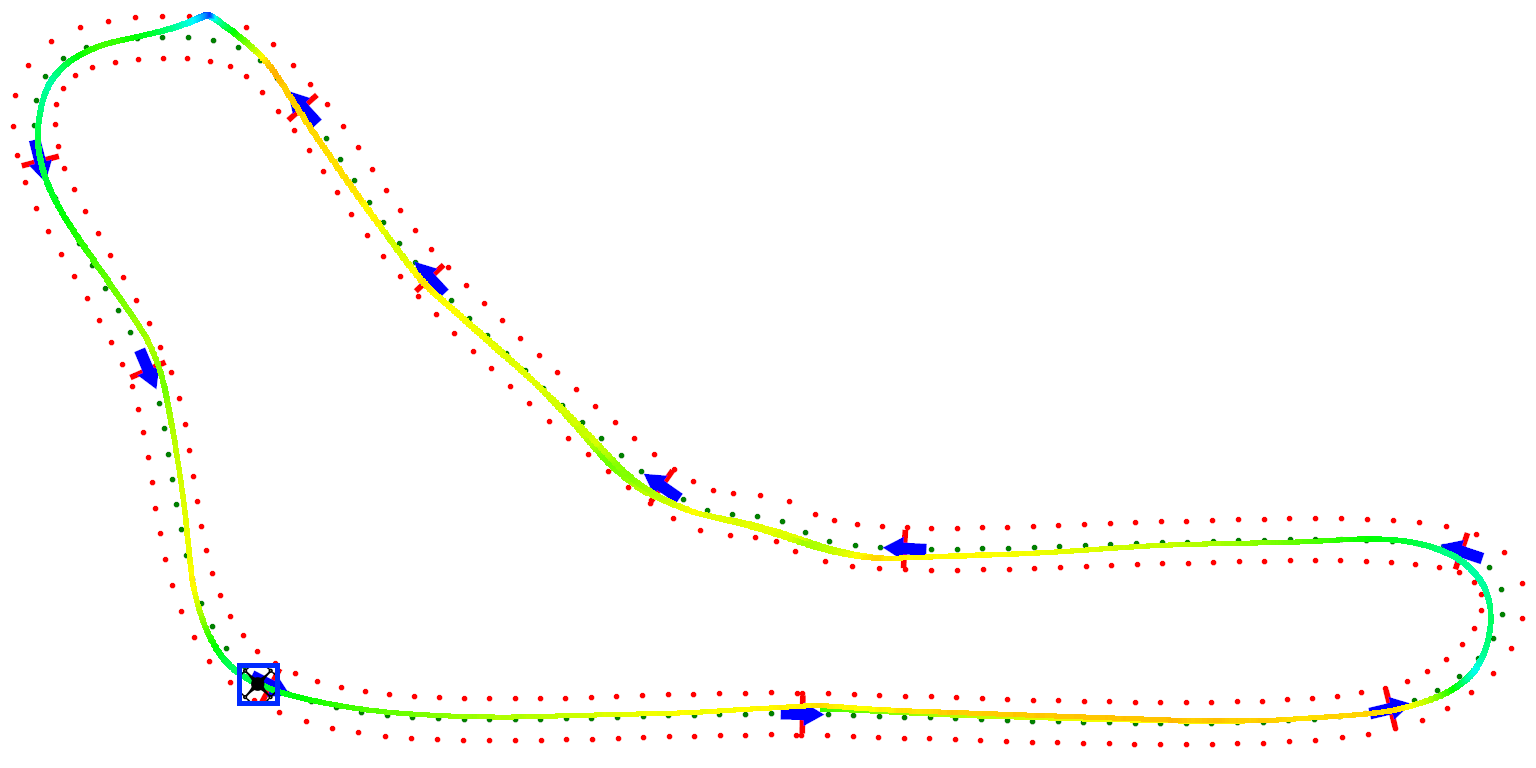} &
		\includegraphics[height=3cm]{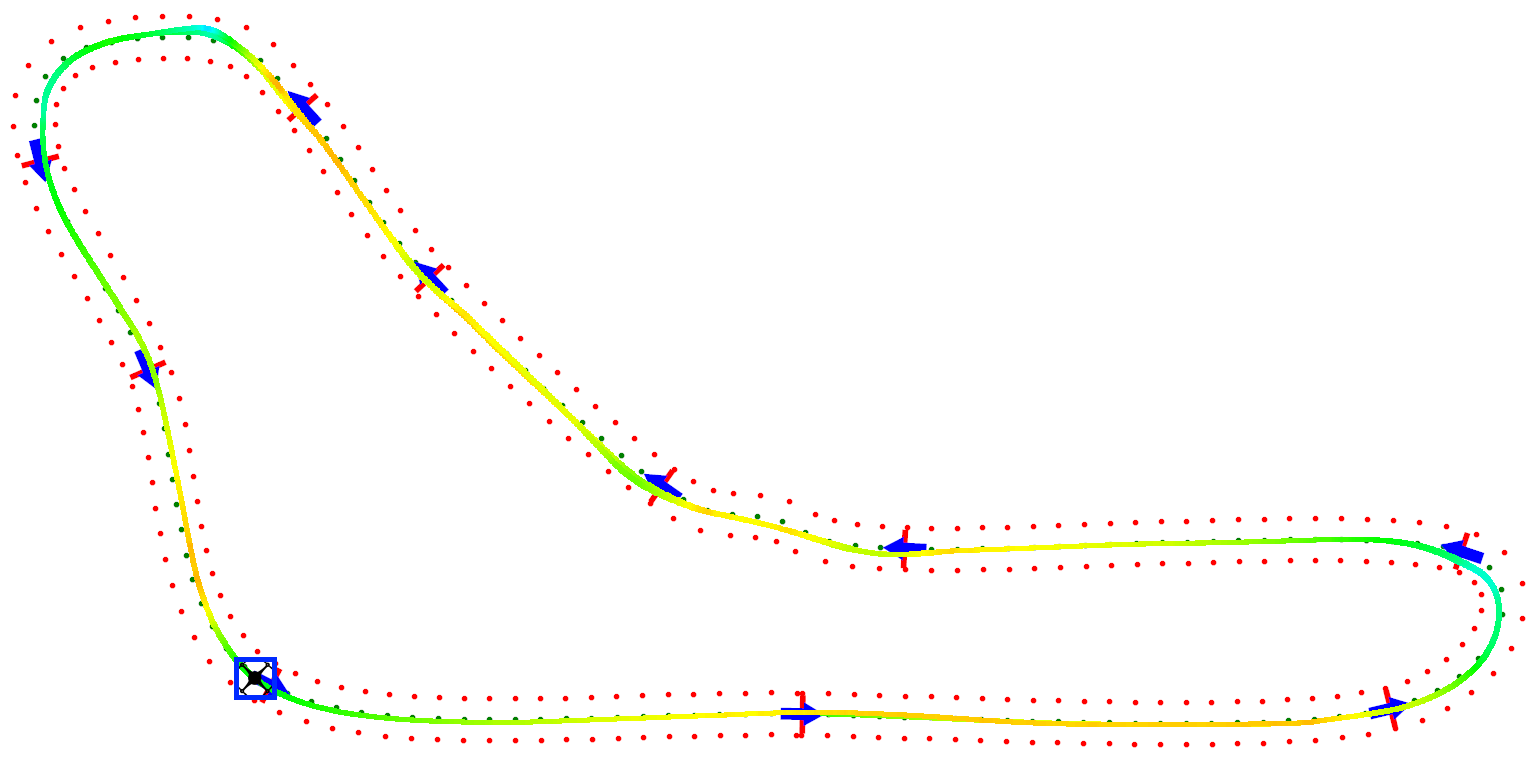} \\
		\small (i) Ours (Reference) & \small (j) Ours (Night) & \small (k)  Ours (Sunrise) \\
		\includegraphics[height=3cm]{sup_figures/track2_ours_grass.png} &
		\includegraphics[height=3cm]{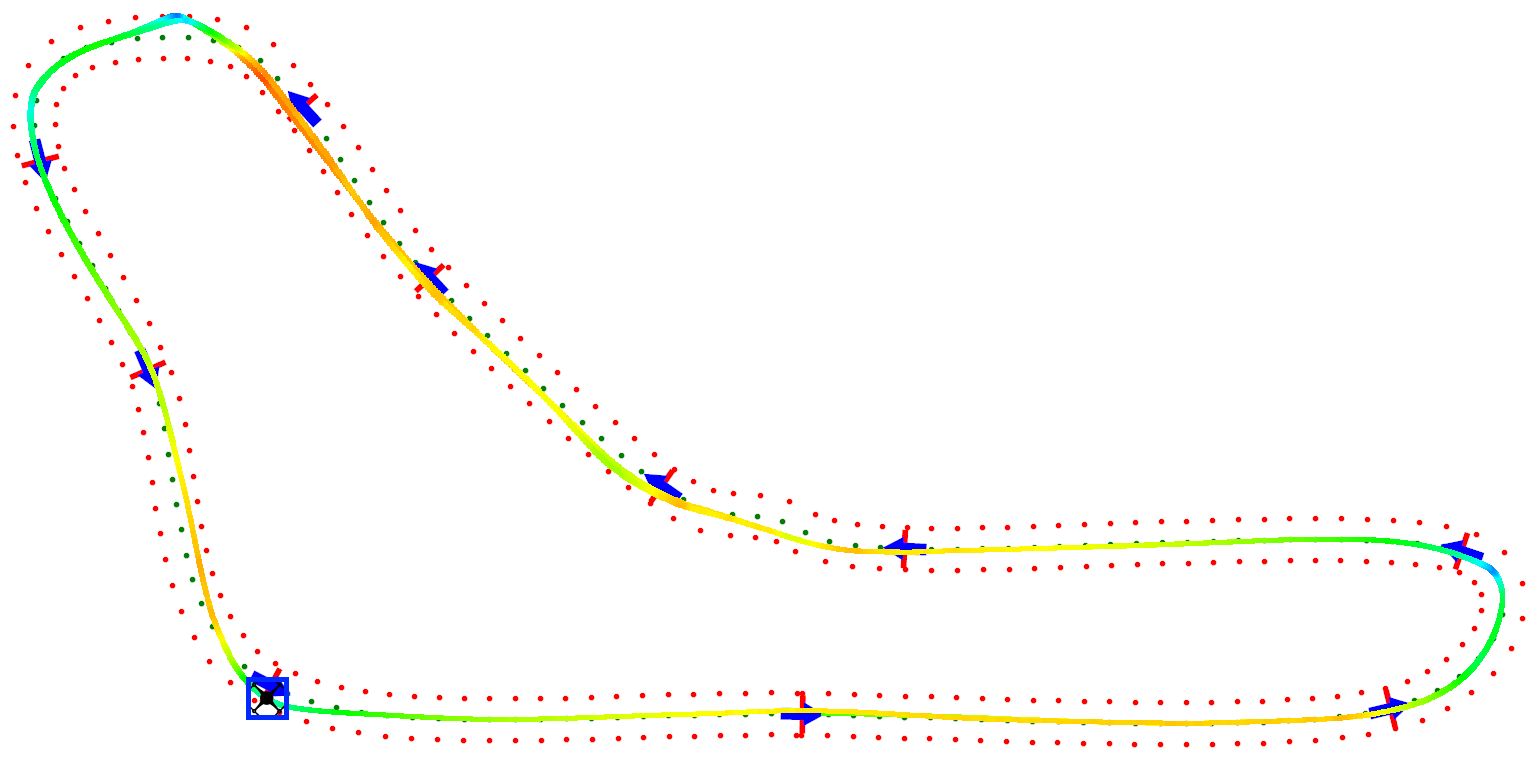} &
		\includegraphics[height=3cm]{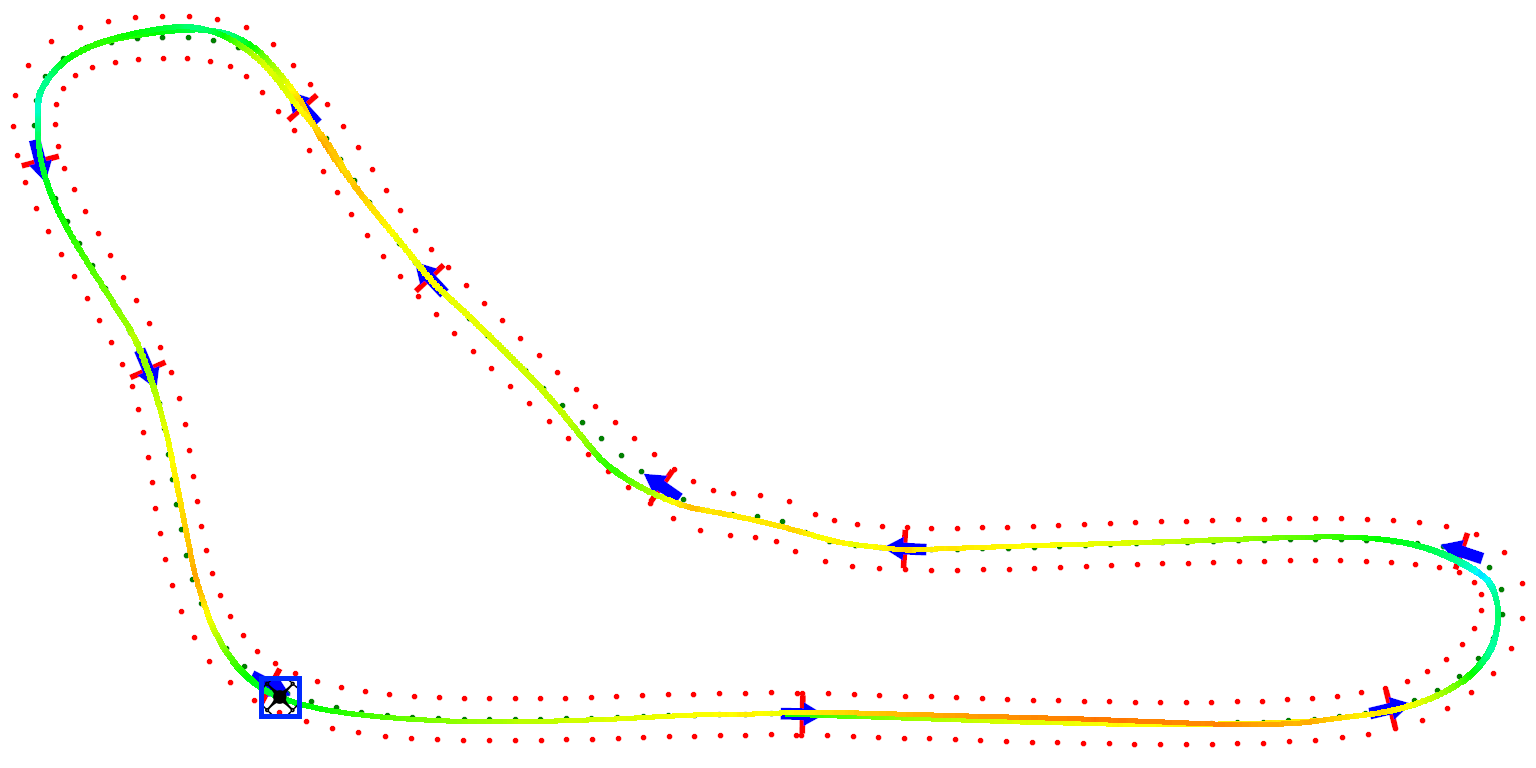} \\
		\small (l) Ours (Reference) & \small (m) Ours (Fog) & \small (o)  Ours (Rain) \\
		\includegraphics[height=3cm]{sup_figures/track2_ours_grass.png} &
		\includegraphics[height=3cm]{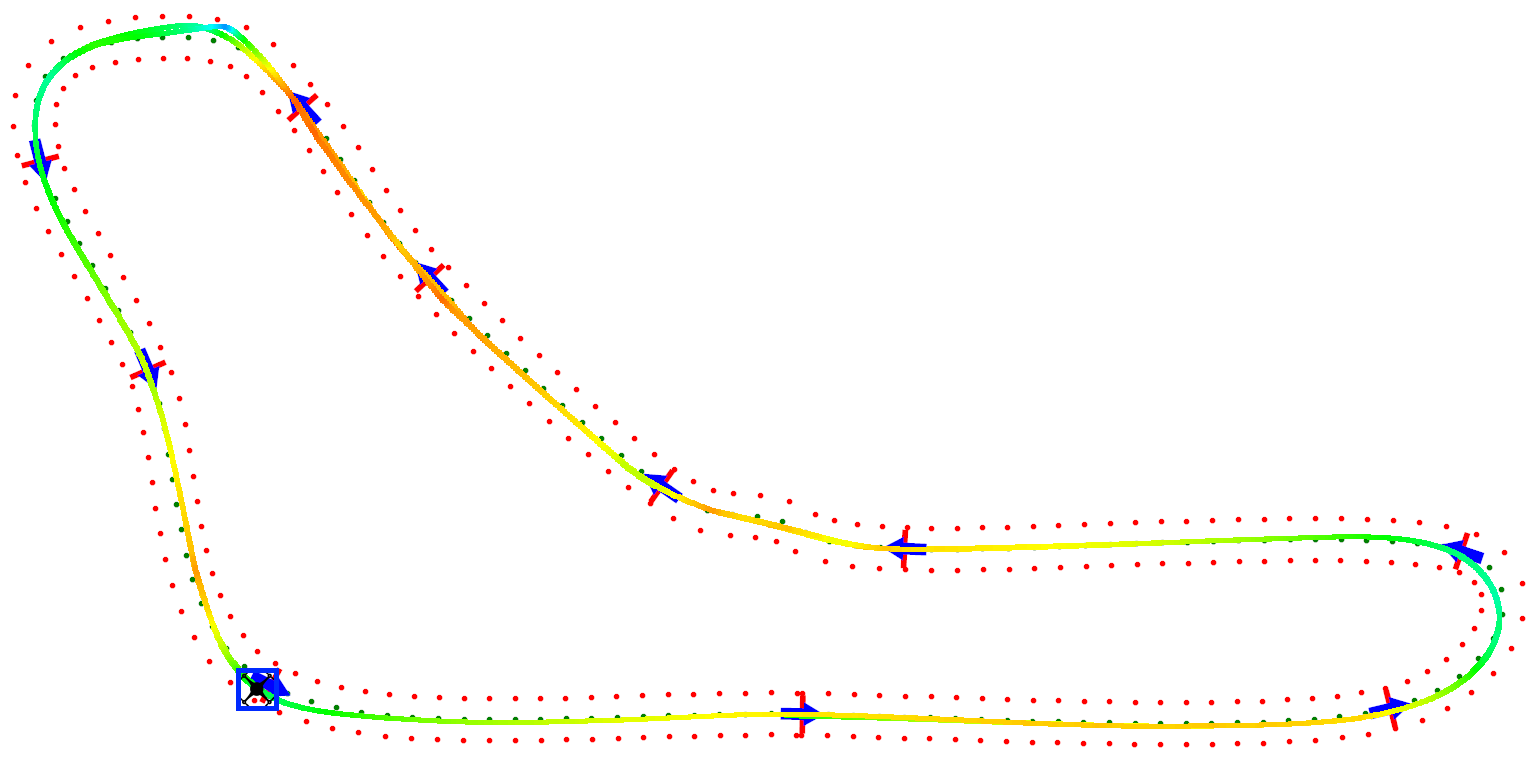} &
		\includegraphics[height=3cm]{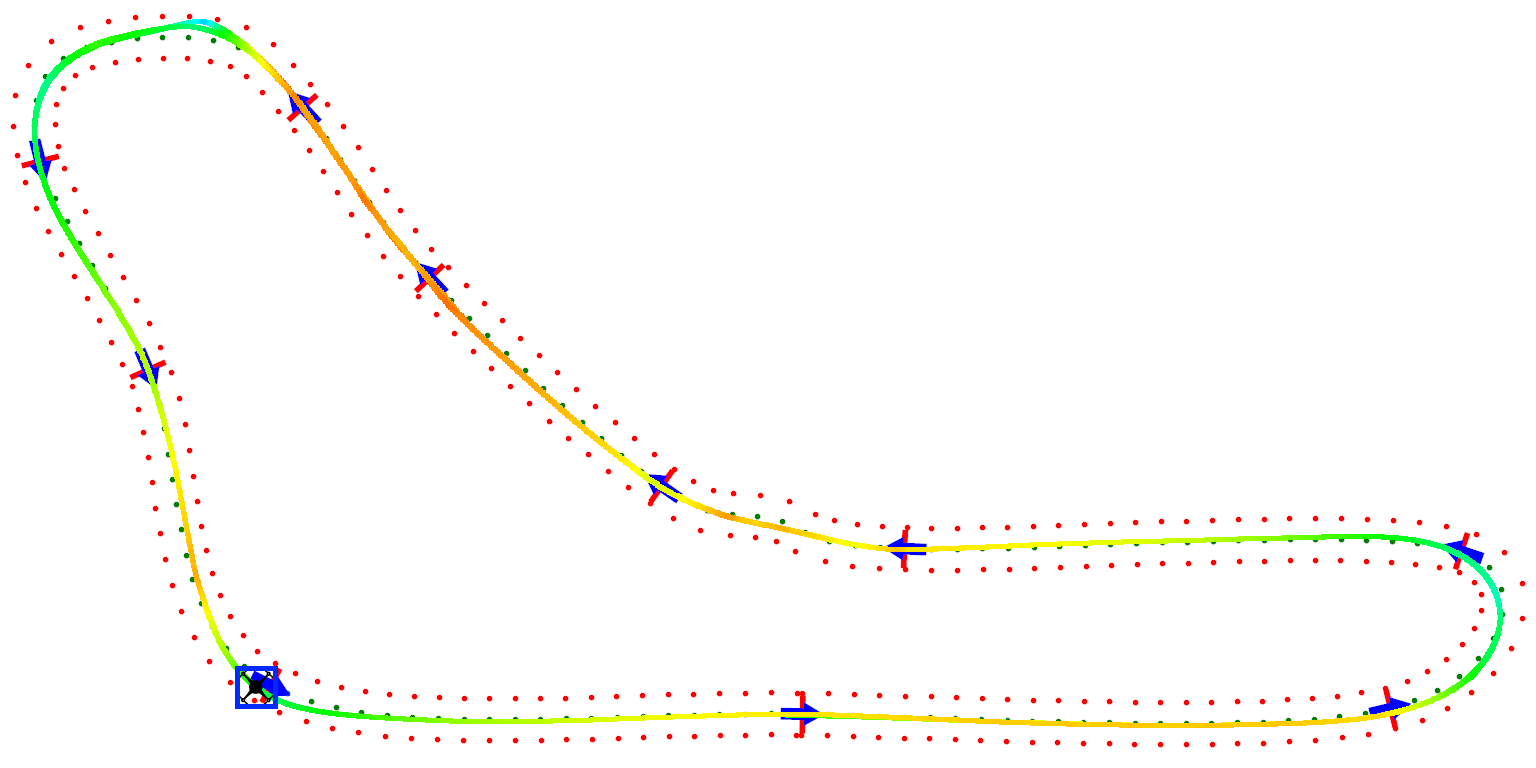} \\
		\small (p) Ours (Grass) & \small (q) Ours (Mud) & \small (r)  Ours (Snow) \\
        \multicolumn{3}{c}{\includegraphics[height=1.2cm]{sup_figures/ColorScaleHorizontal.png}}
\end{tabular}
\captionof{figure}{Qualitative results on track2. The color encodes speed as a heatmap, where blue corresponds to the minimum speed and red to the maximum speed.}
\label{fig:qualitive_results_track2}
\end{figure*}

\begin{figure*}
\centering
\begin{tabular}{@{}c@{\hspace{1mm}}c@{\hspace{1mm}}c@{\hspace{8mm}}c@{}}
		\includegraphics[height=3cm]{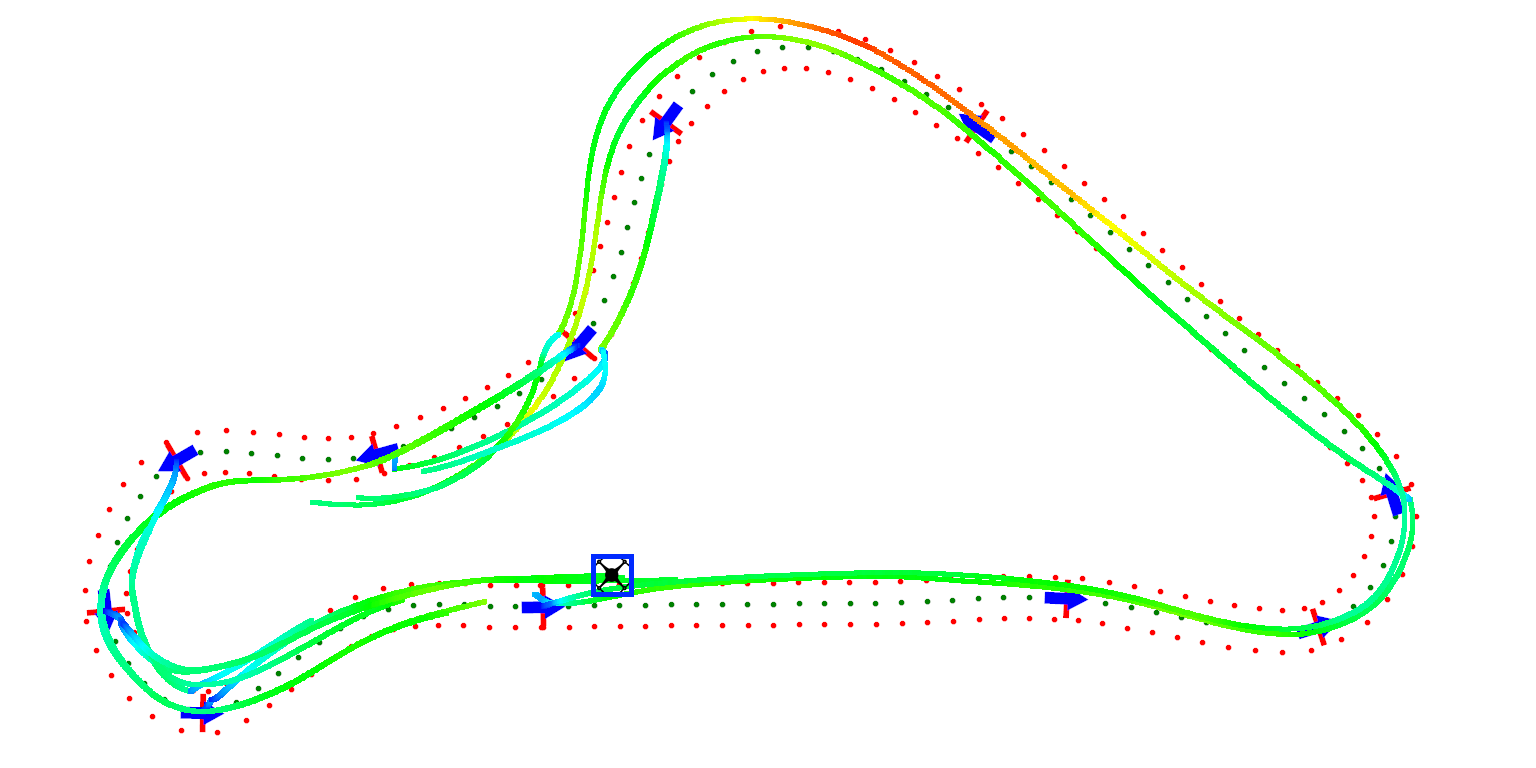} &
		\includegraphics[height=3cm]{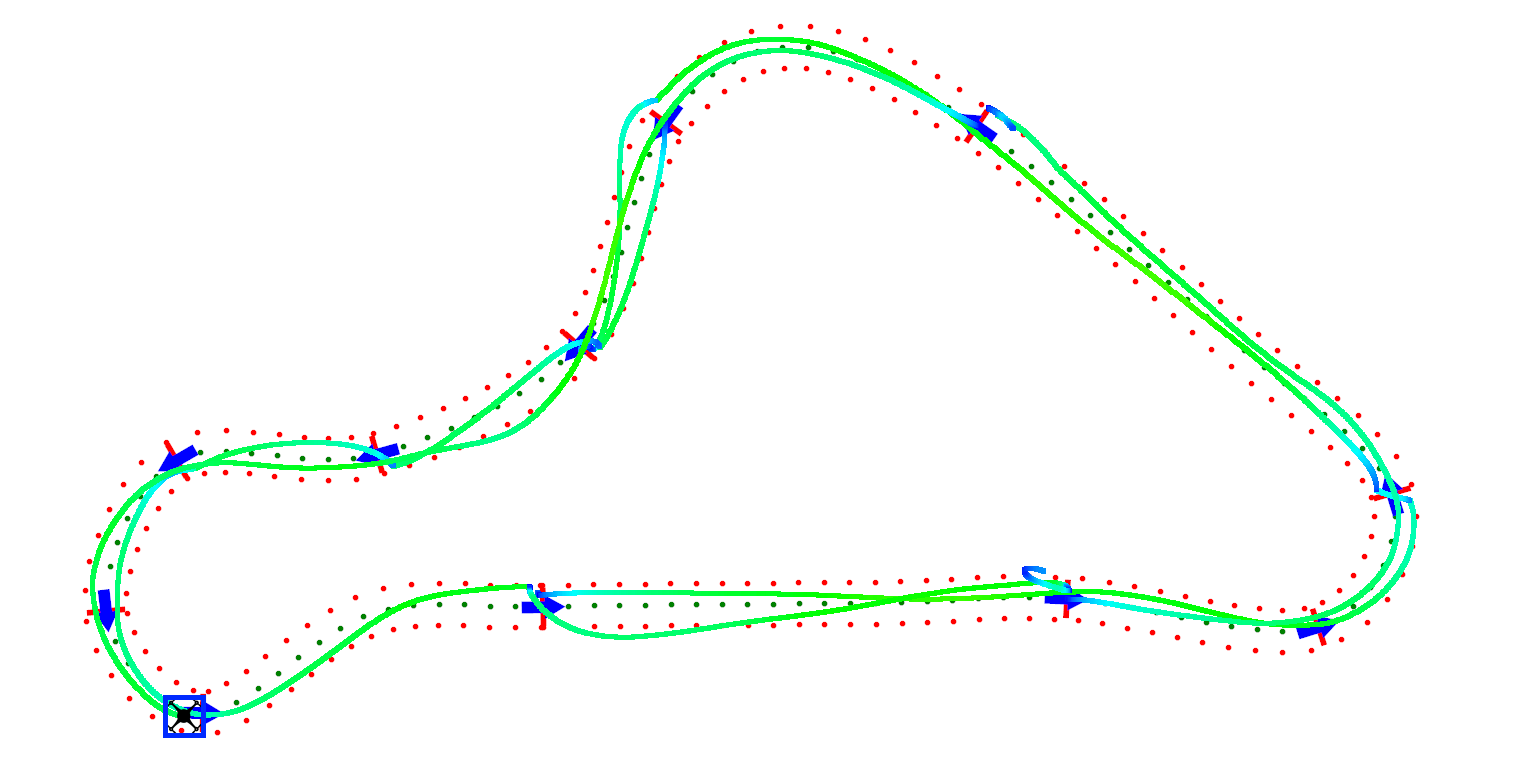} &
		\includegraphics[height=3cm]{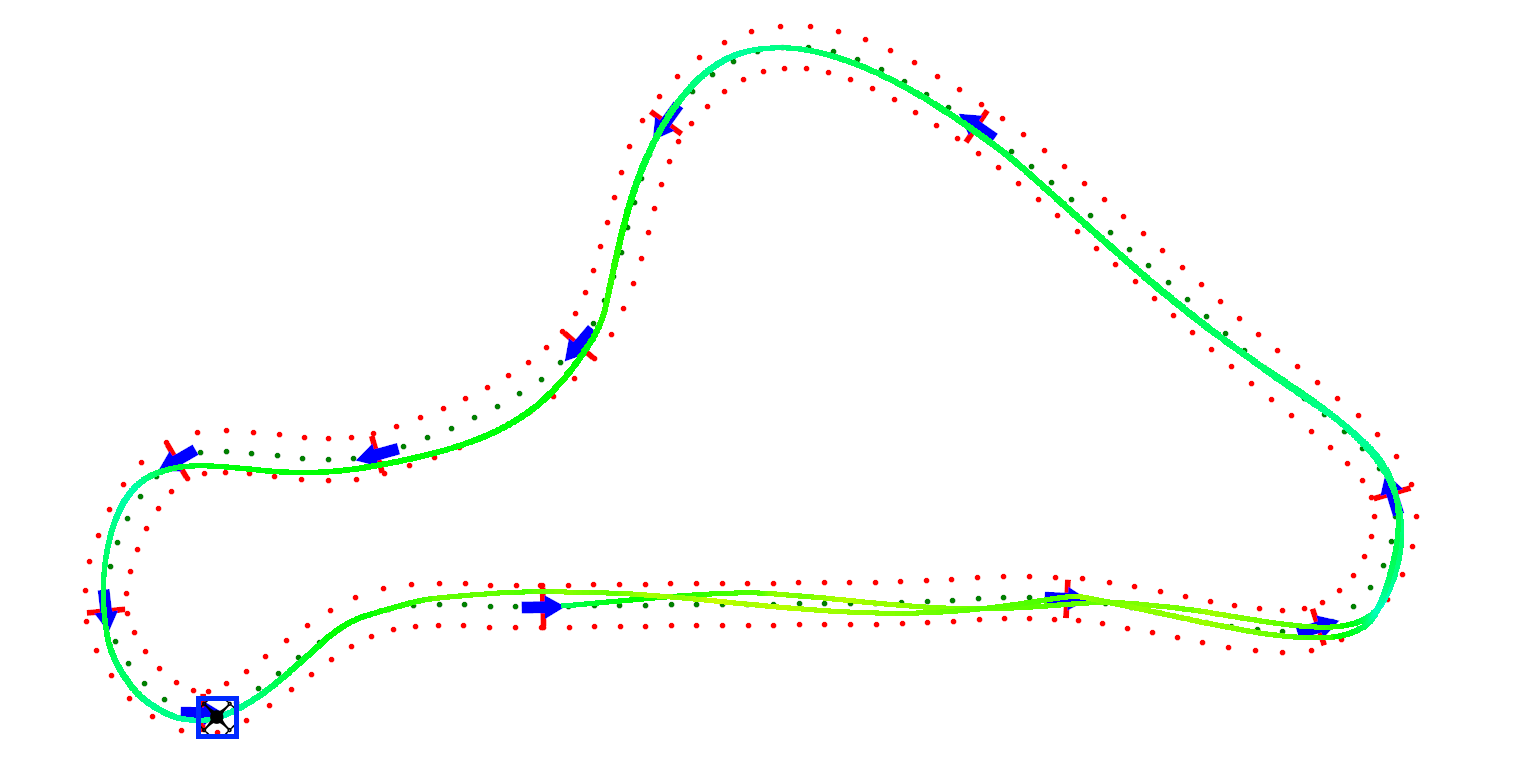} \\
		\small (a) End2End (MAV) & \small (b) End2End (Nvidia) & \small (c) End2End (Ours) \\
		\includegraphics[height=3cm]{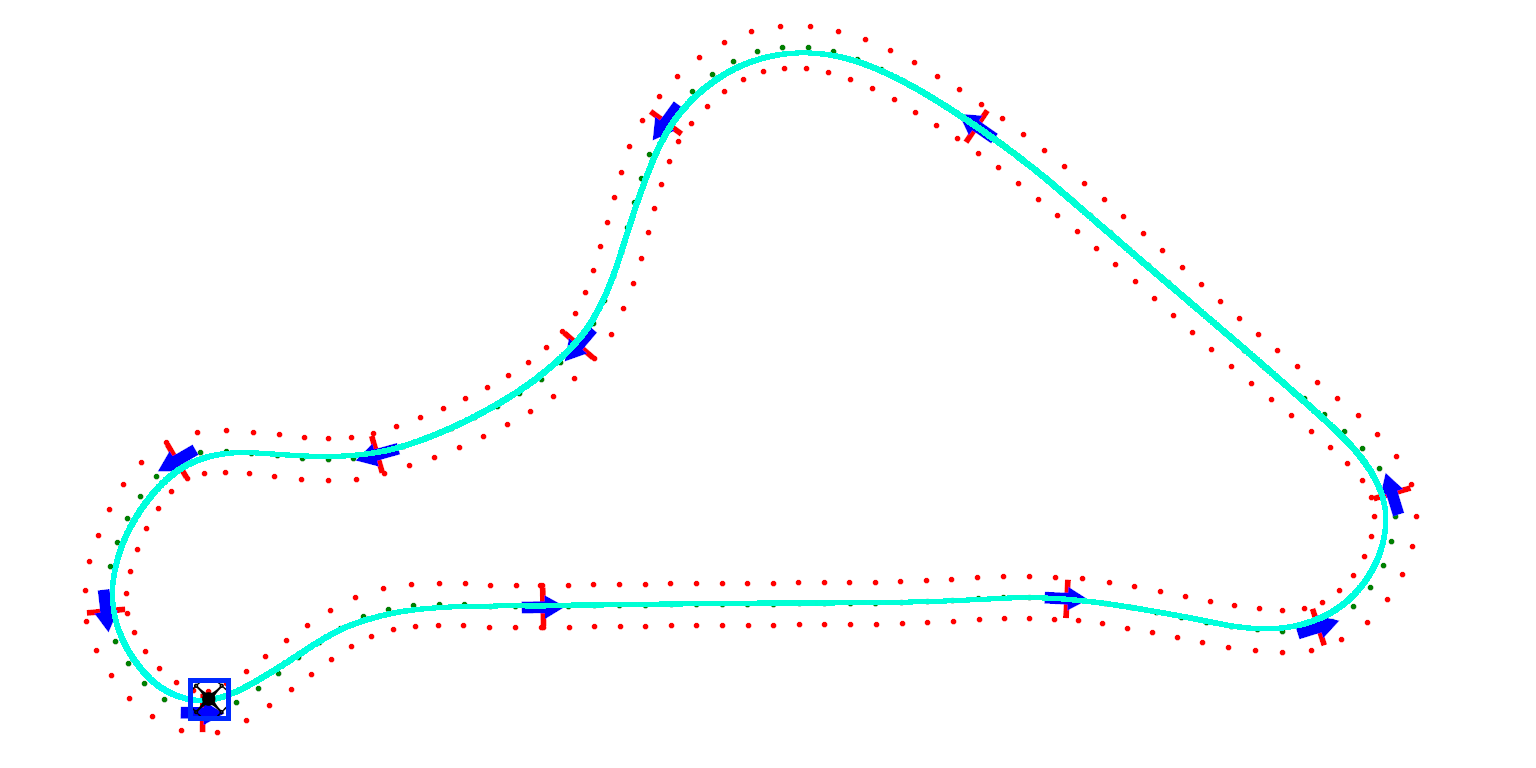} &
		\includegraphics[height=3cm]{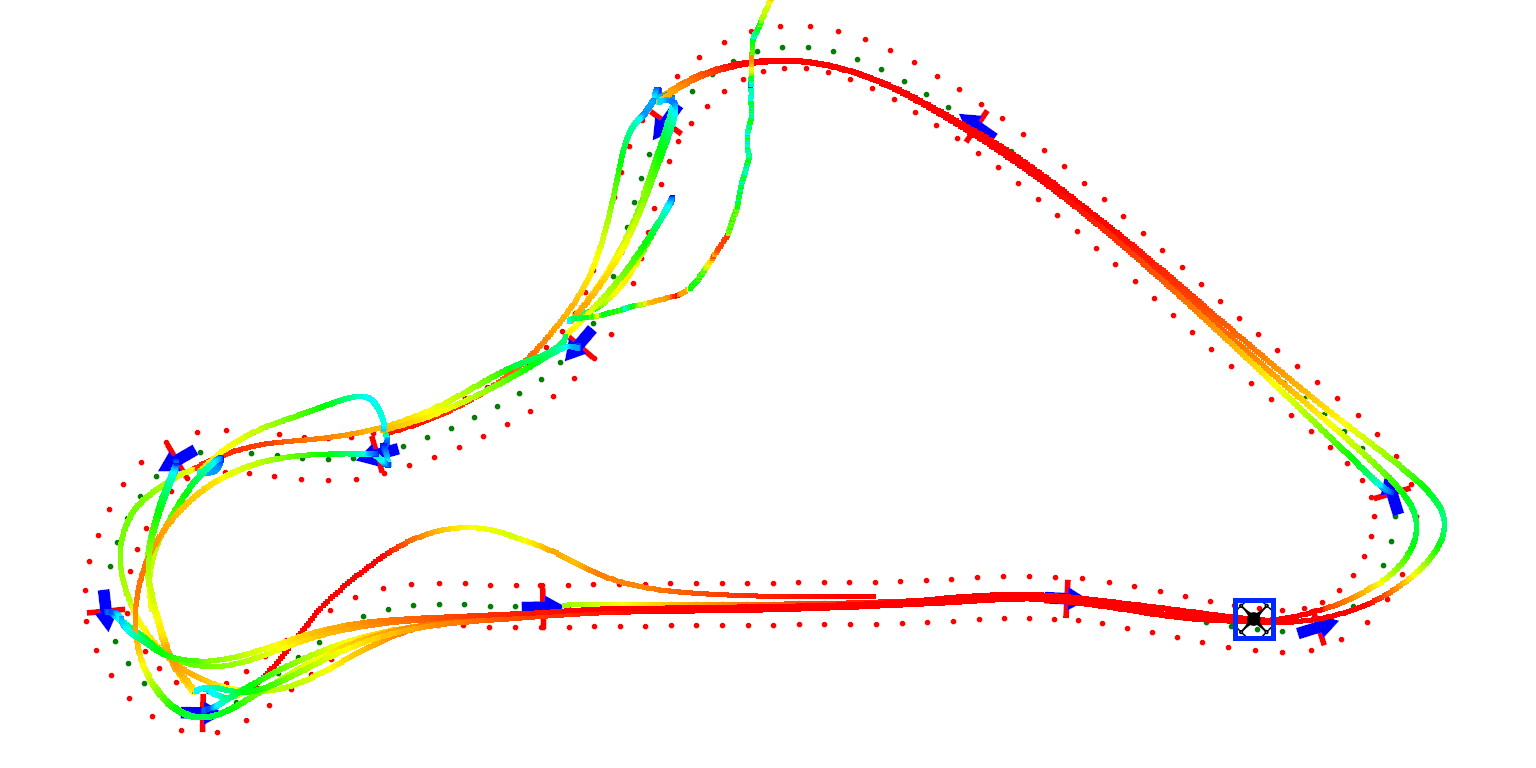} &
		\includegraphics[height=3cm]{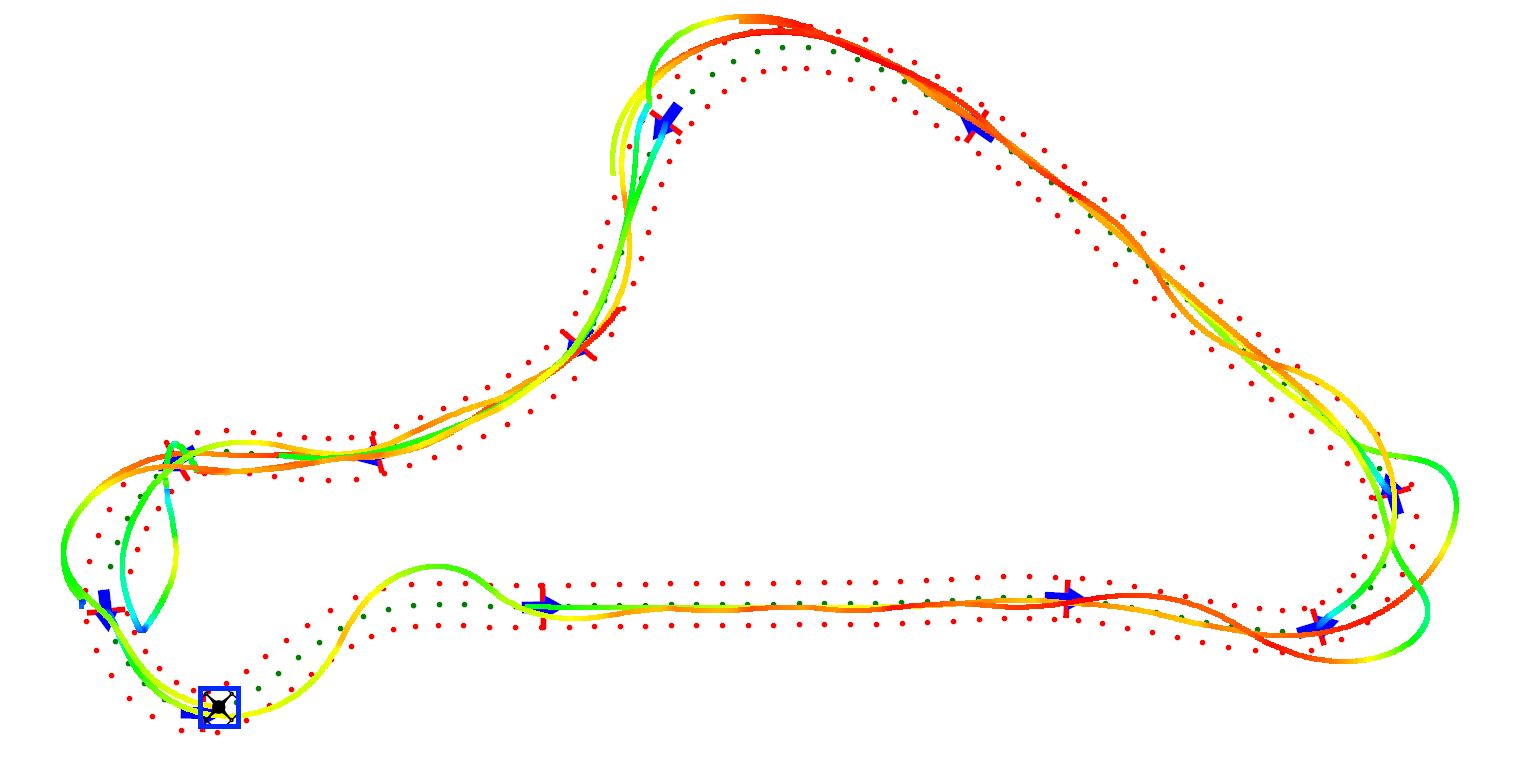} \\
		\small (c) PID1 (Conservative) & \small (d) PID2 (Aggressive) & \small (e) Ours (No Buffer) \\
		\includegraphics[height=3cm]{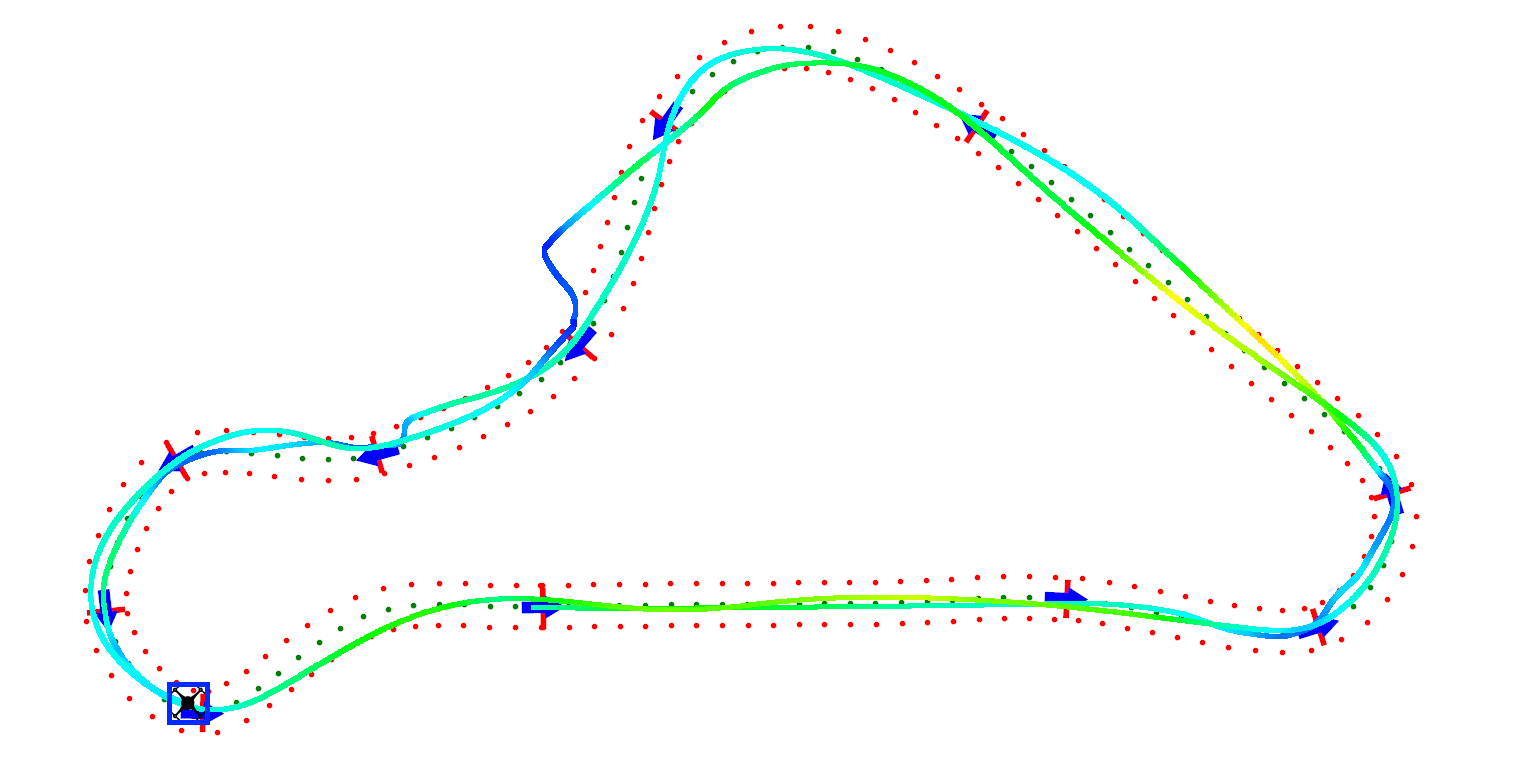} &
		\includegraphics[height=3cm]{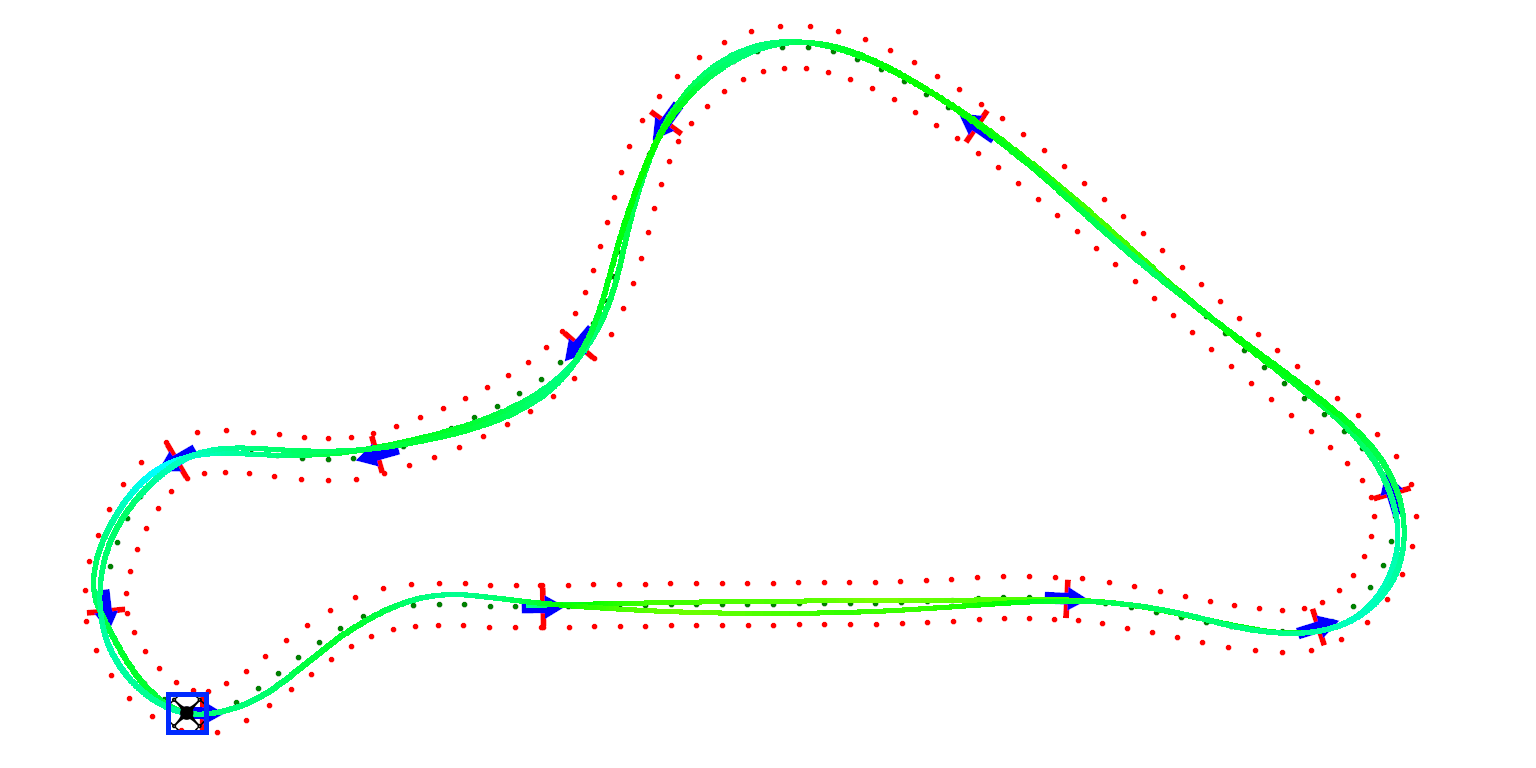} &
		\includegraphics[height=3cm]{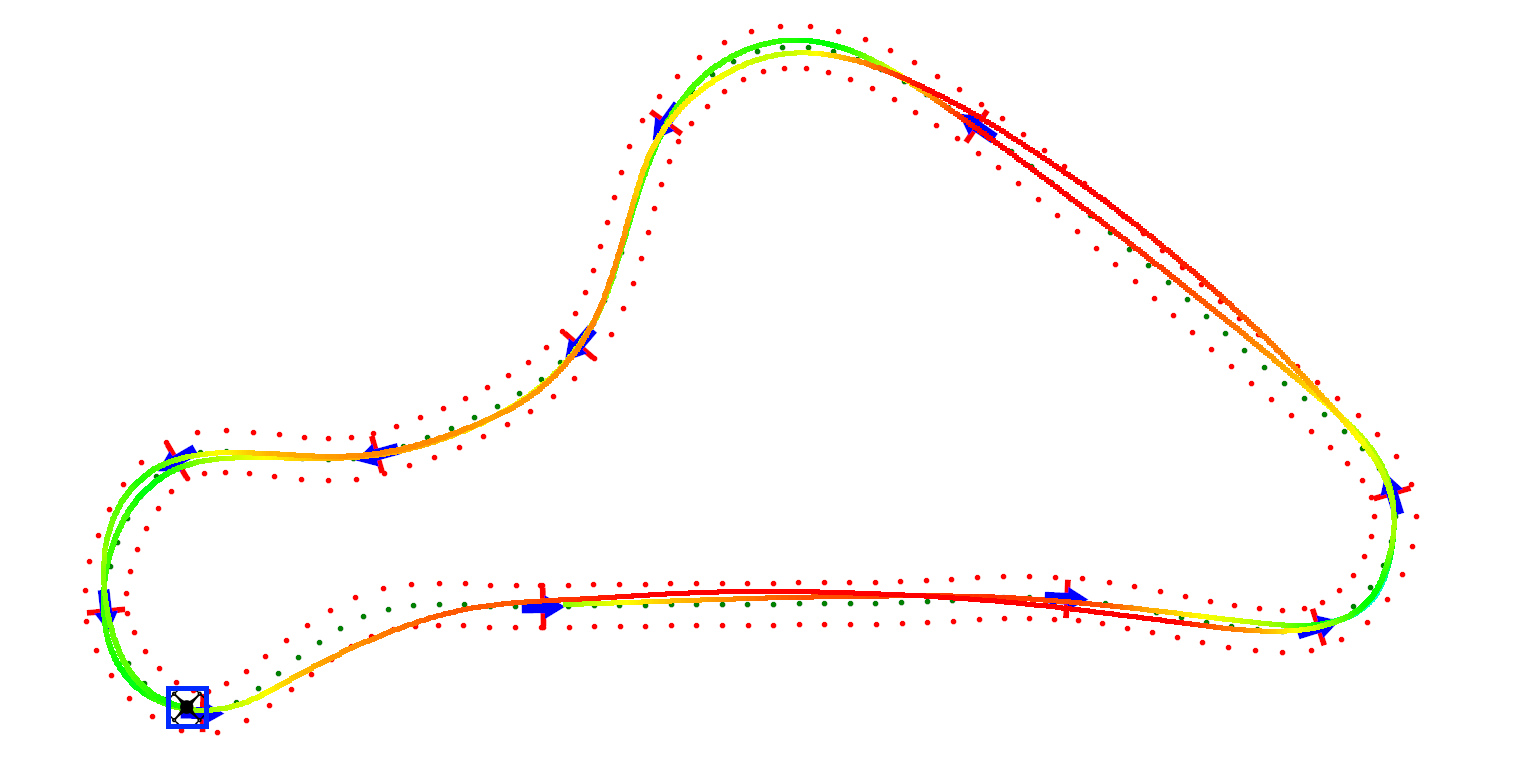} \\
        \small (f) Human (Novice) & \small (g) Human (Intermediate) & \small (h) Human (Professional)\\
		\includegraphics[height=3cm]{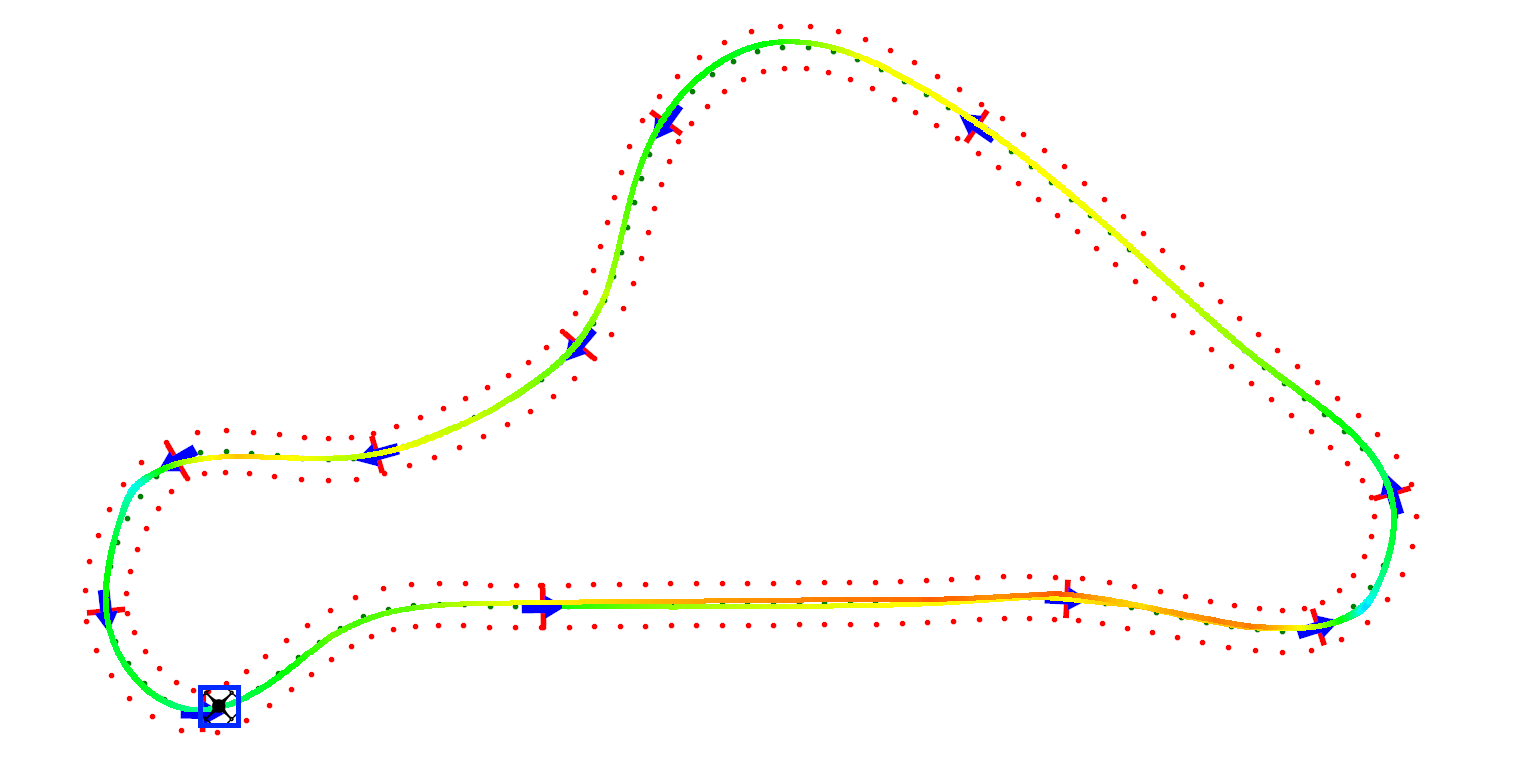} &
		\includegraphics[height=3cm]{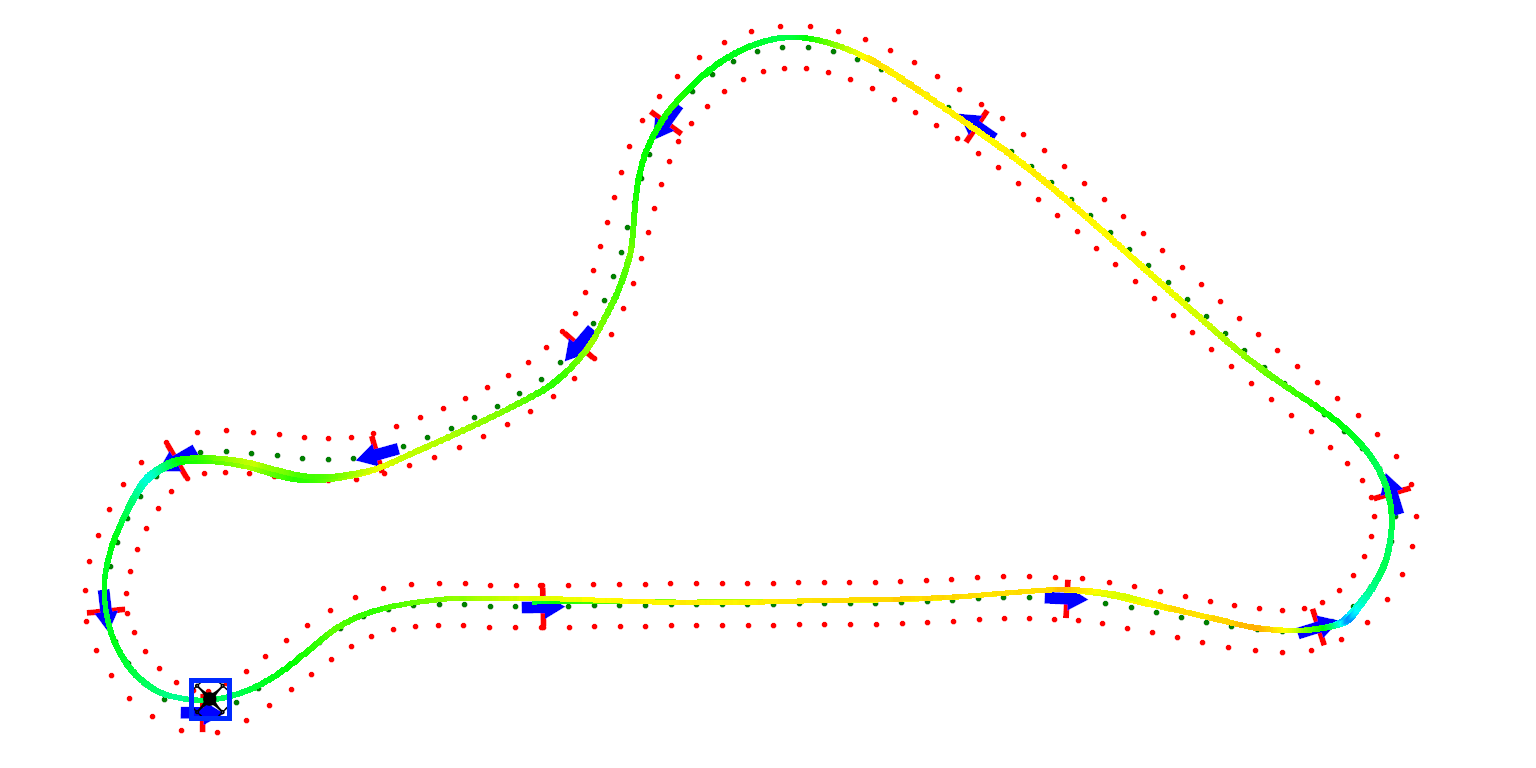} &
		\includegraphics[height=3cm]{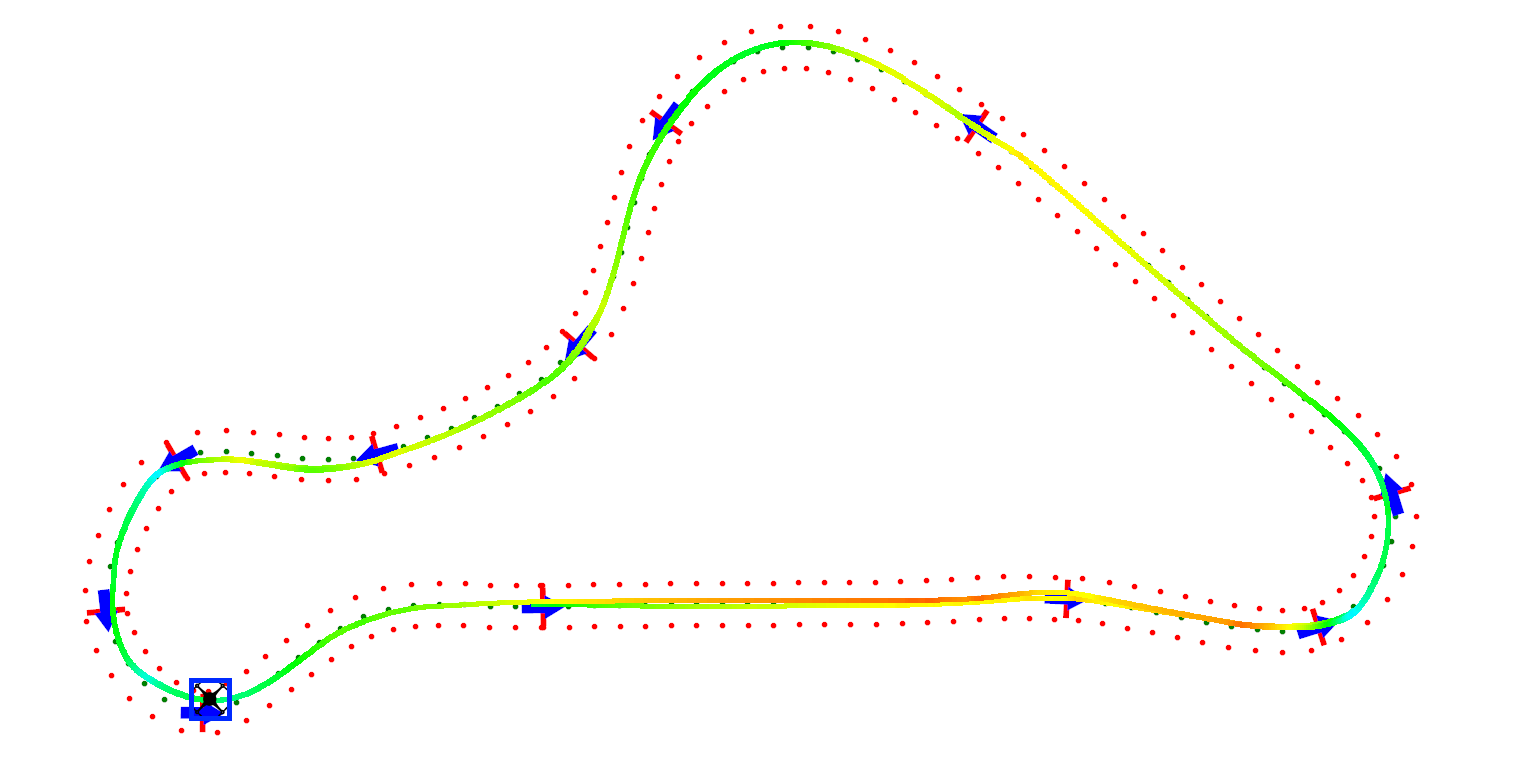} \\
		\small (i) Ours (Reference) & \small (j) Ours (Night) & \small (k)  Ours (Sunrise) \\
		\includegraphics[height=3cm]{sup_figures/track3_ours_grass.png} &
		\includegraphics[height=3cm]{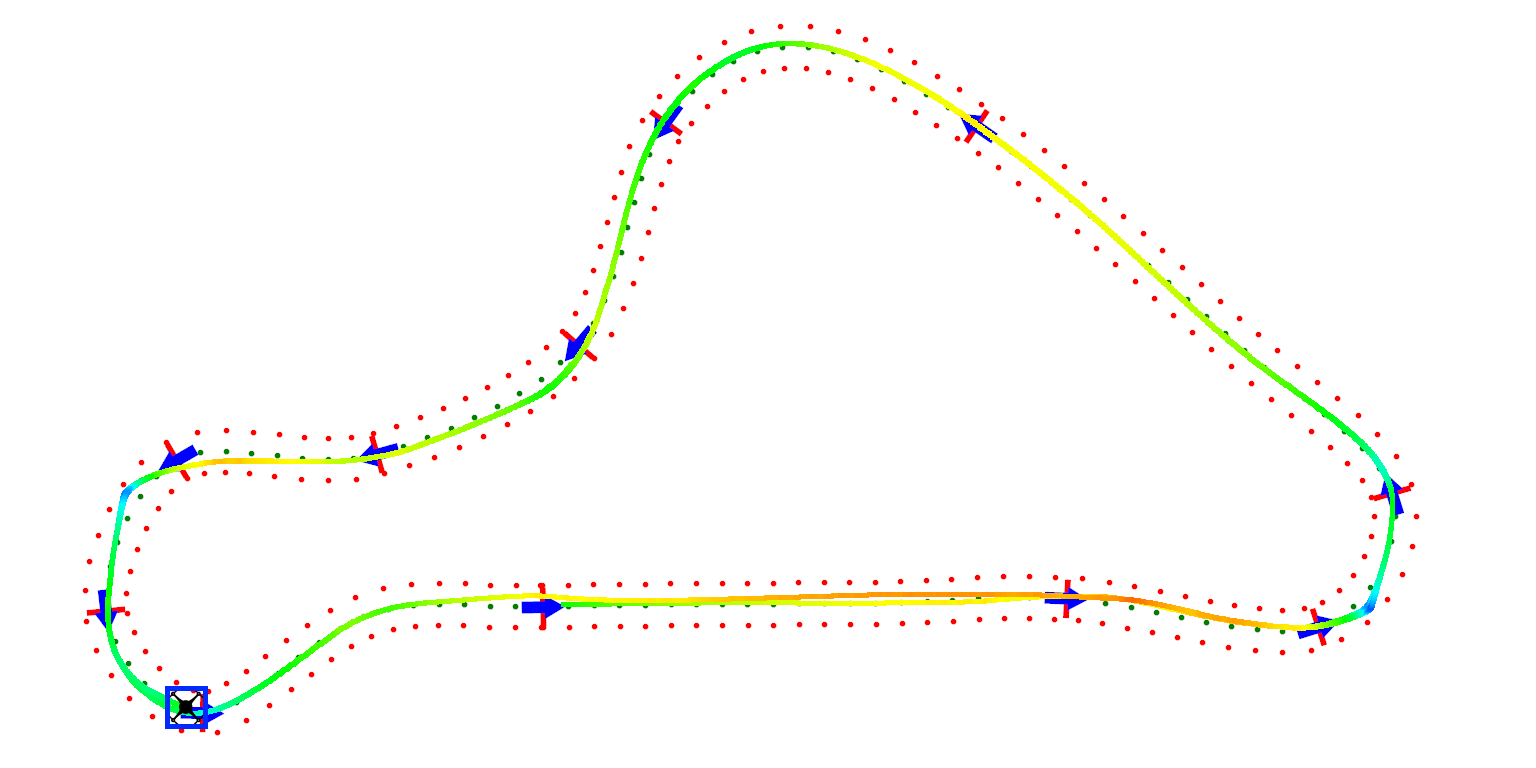} &
		\includegraphics[height=3cm]{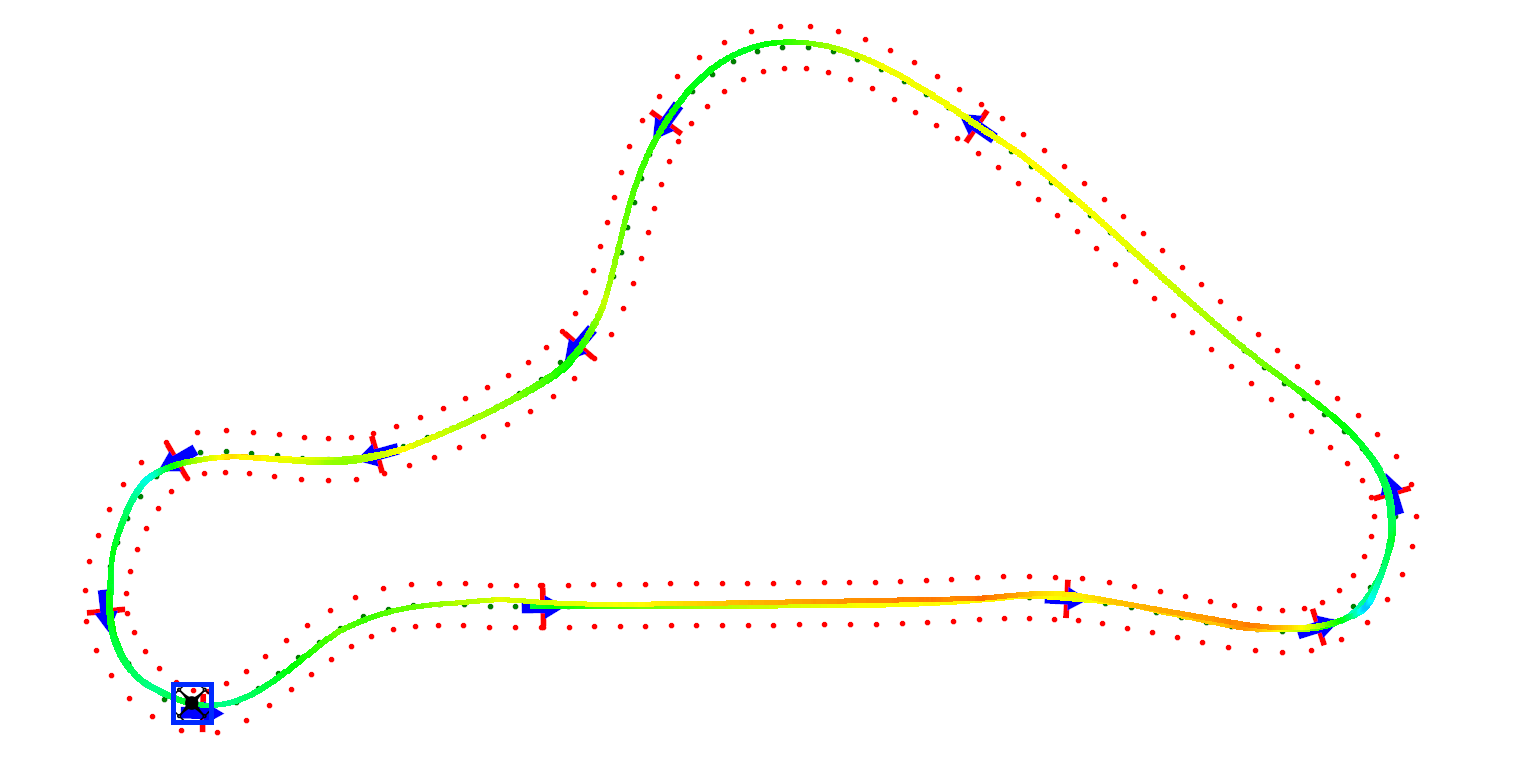} \\
		\small (l) Ours (Reference) & \small (m) Ours (Fog) & \small (o)  Ours (Rain) \\
		\includegraphics[height=3cm]{sup_figures/track3_ours_grass.png} &
		\includegraphics[height=3cm]{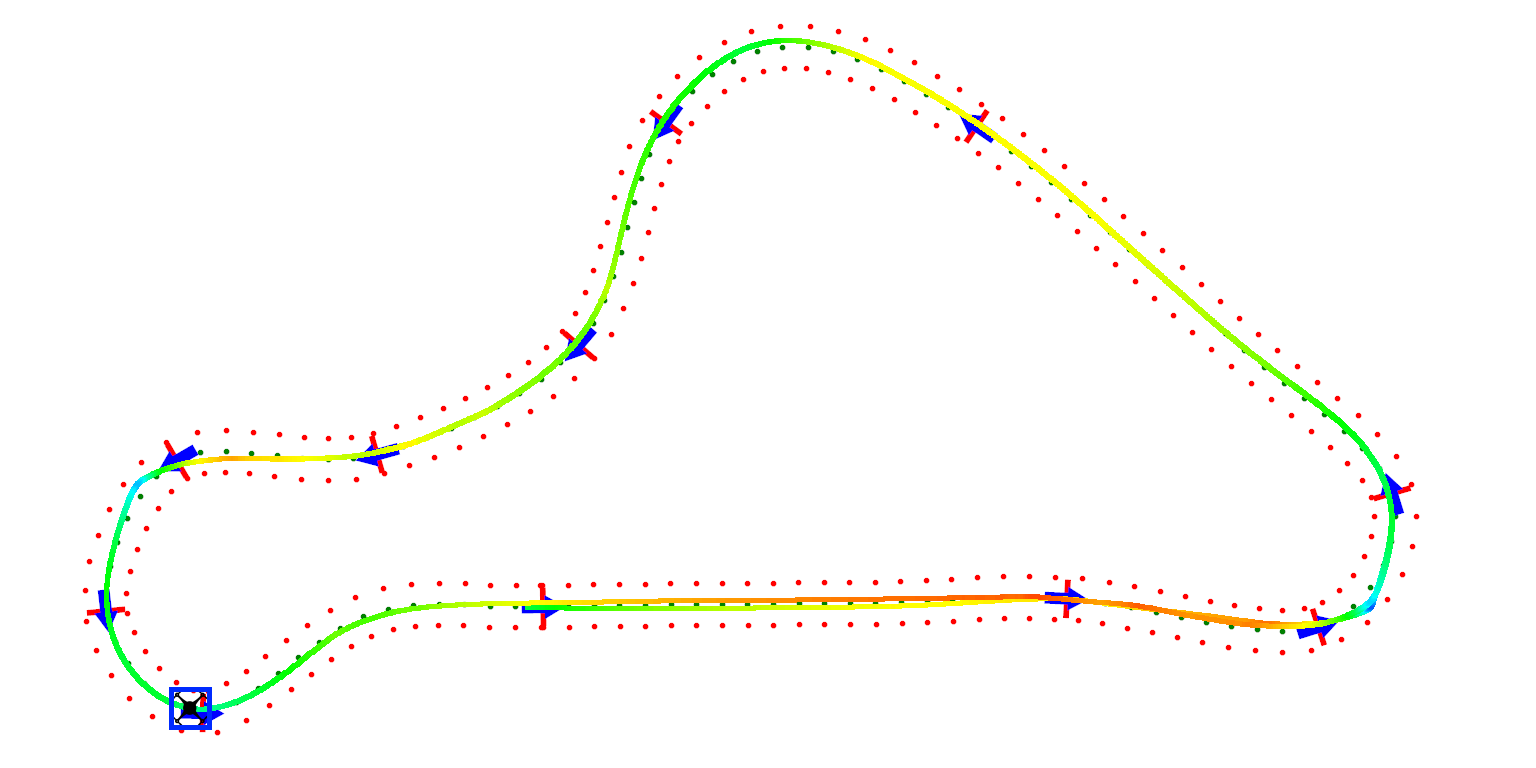} &
		\includegraphics[height=3cm]{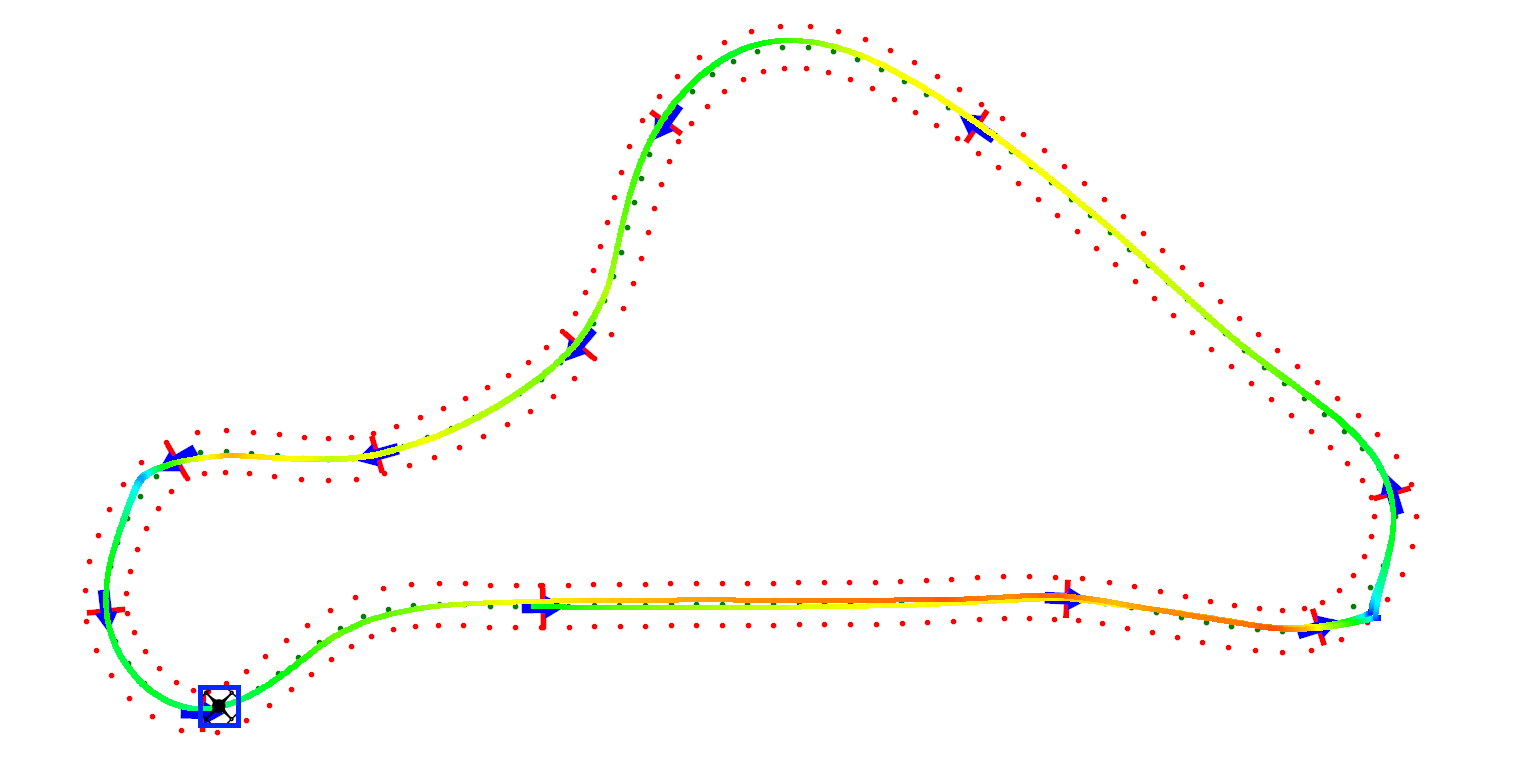} \\
		\small (p) Ours (Grass) & \small (q) Ours (Mud) & \small (r)  Ours (Snow) \\
        \multicolumn{3}{c}{\includegraphics[height=1.2cm]{sup_figures/ColorScaleHorizontal.png}}
\end{tabular}
\captionof{figure}{Qualitative results on track3. The color encodes speed as a heatmap, where blue corresponds to the minimum speed and red to the maximum speed.}
\label{fig:qualitive_results_track3}
\end{figure*}

\begin{figure*}
\centering
\begin{tabular}{@{}c@{\hspace{1mm}}c@{\hspace{1mm}}c@{\hspace{8mm}}c@{}}
		\includegraphics[height=3cm]{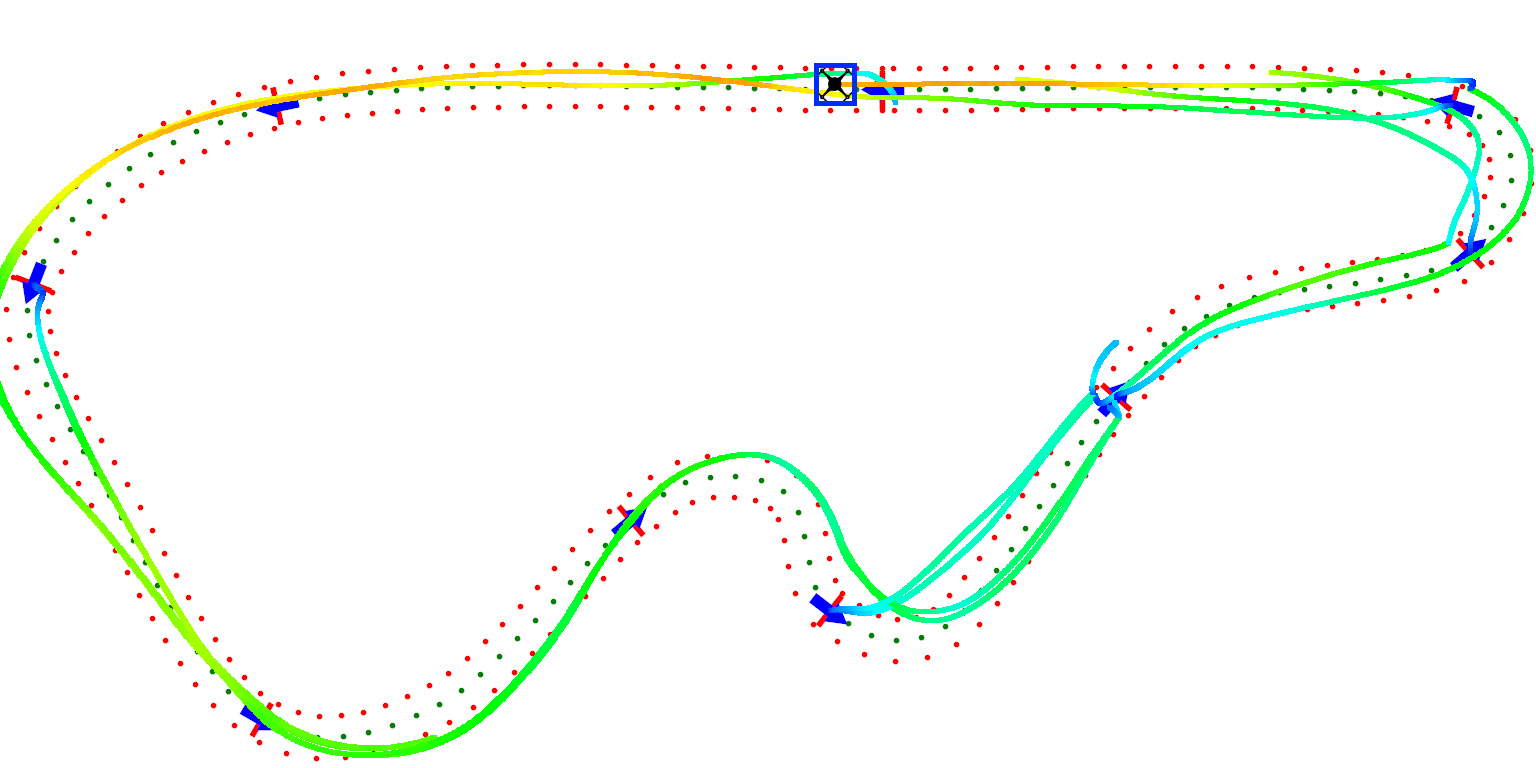} &
		\includegraphics[height=3cm]{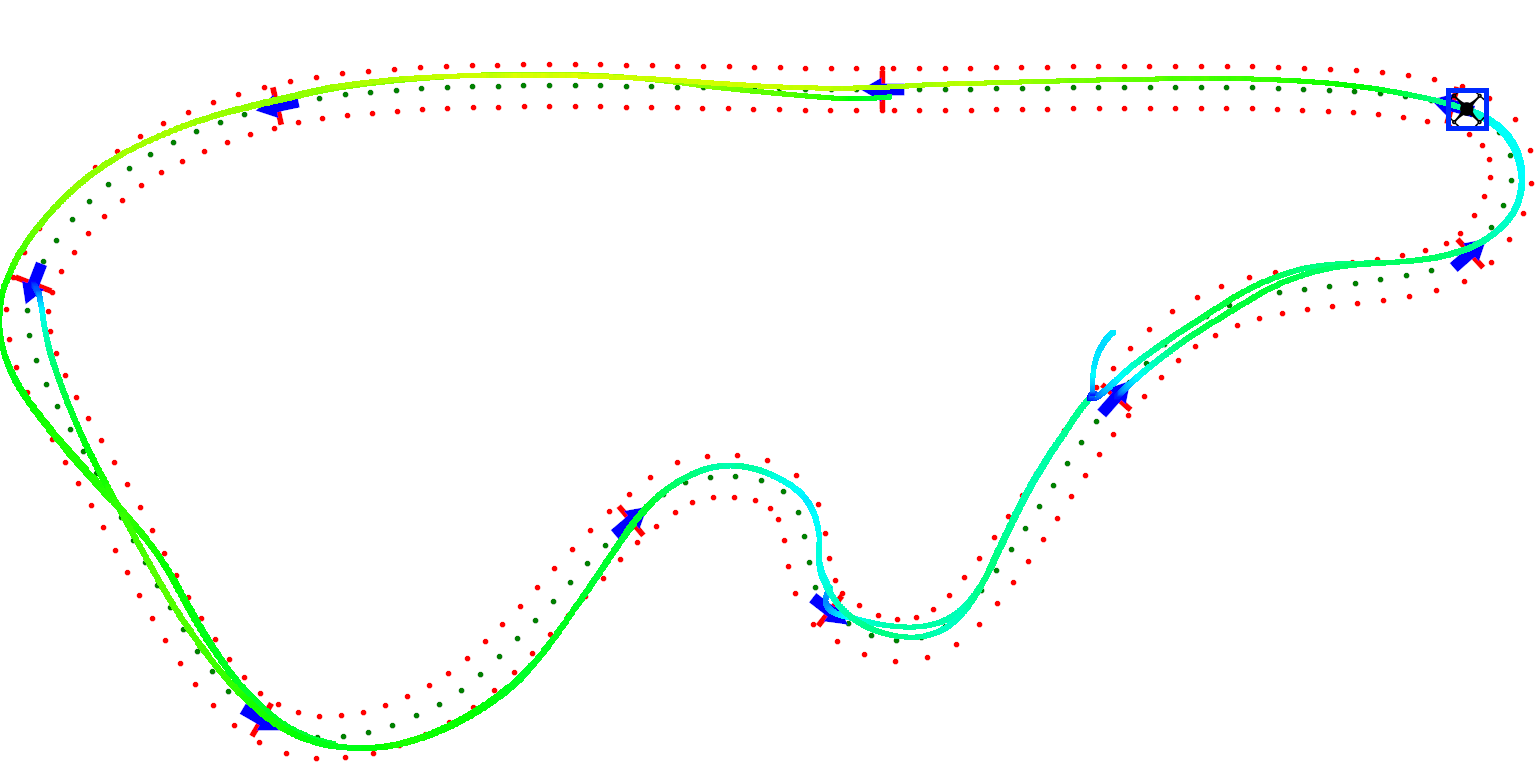} &
		\includegraphics[height=3cm]{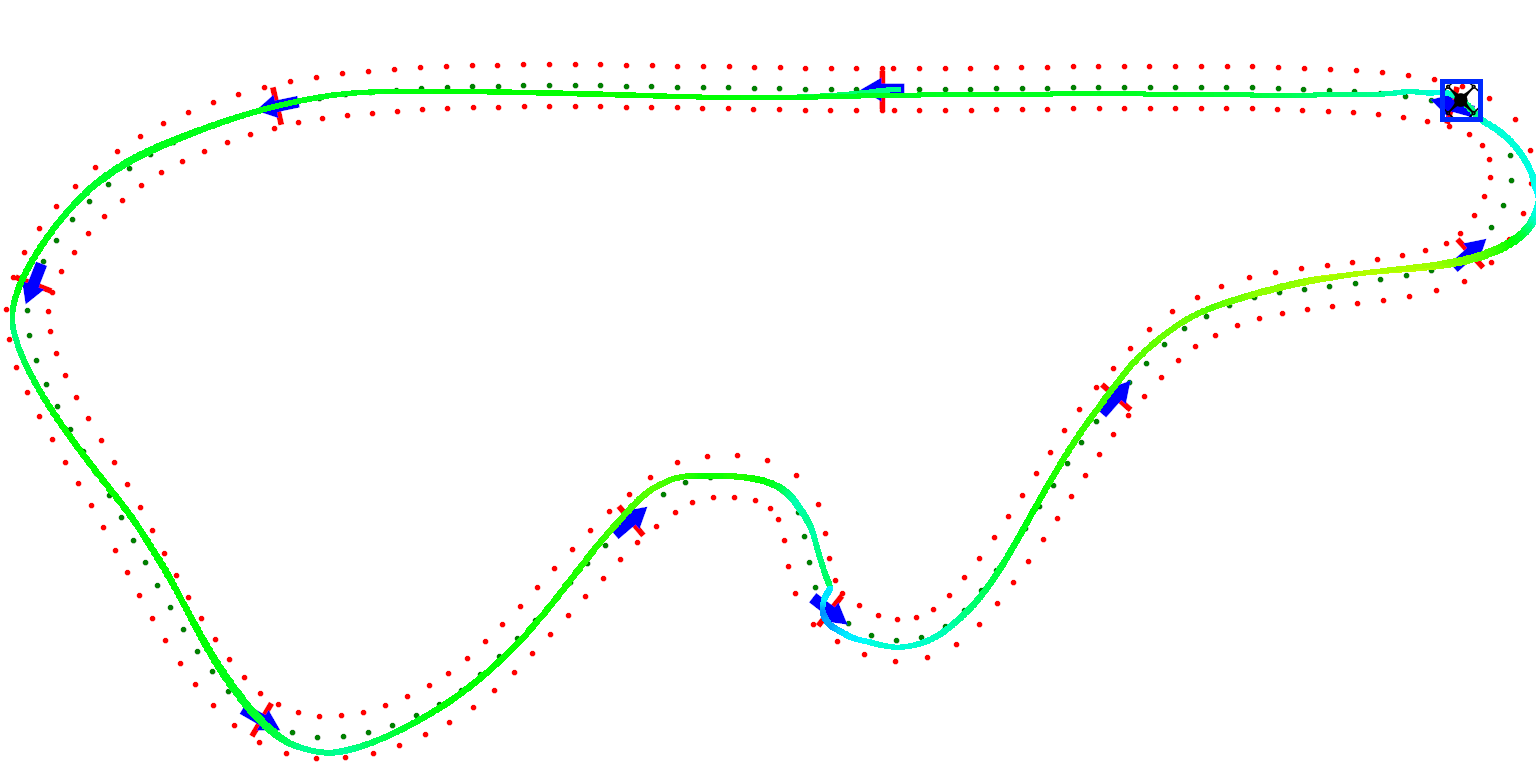} \\
		\small (a) End2End (MAV) & \small (b) End2End (Nvidia) & \small (c) End2End (Ours) \\
		\includegraphics[height=3cm]{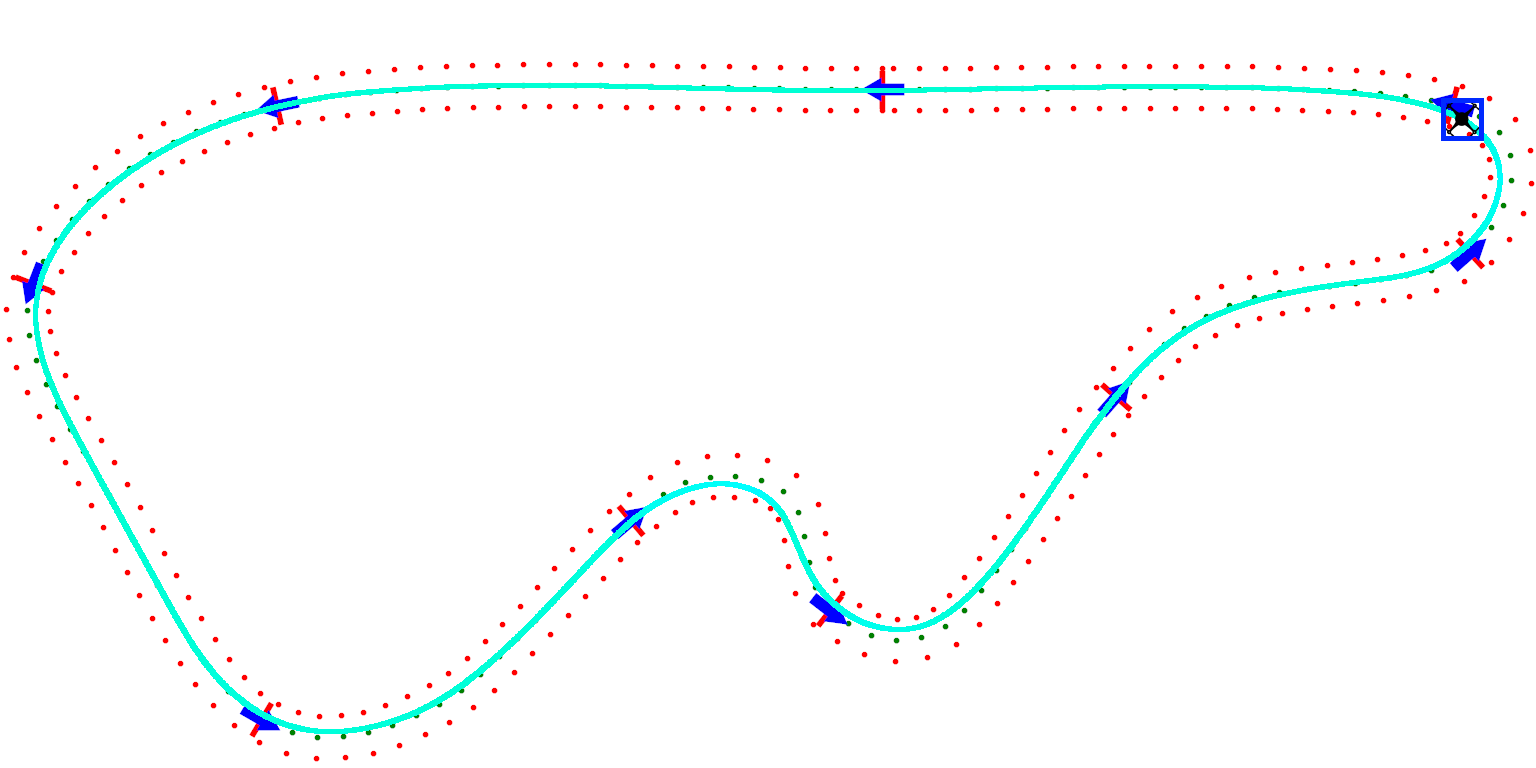} &
		\includegraphics[height=3cm]{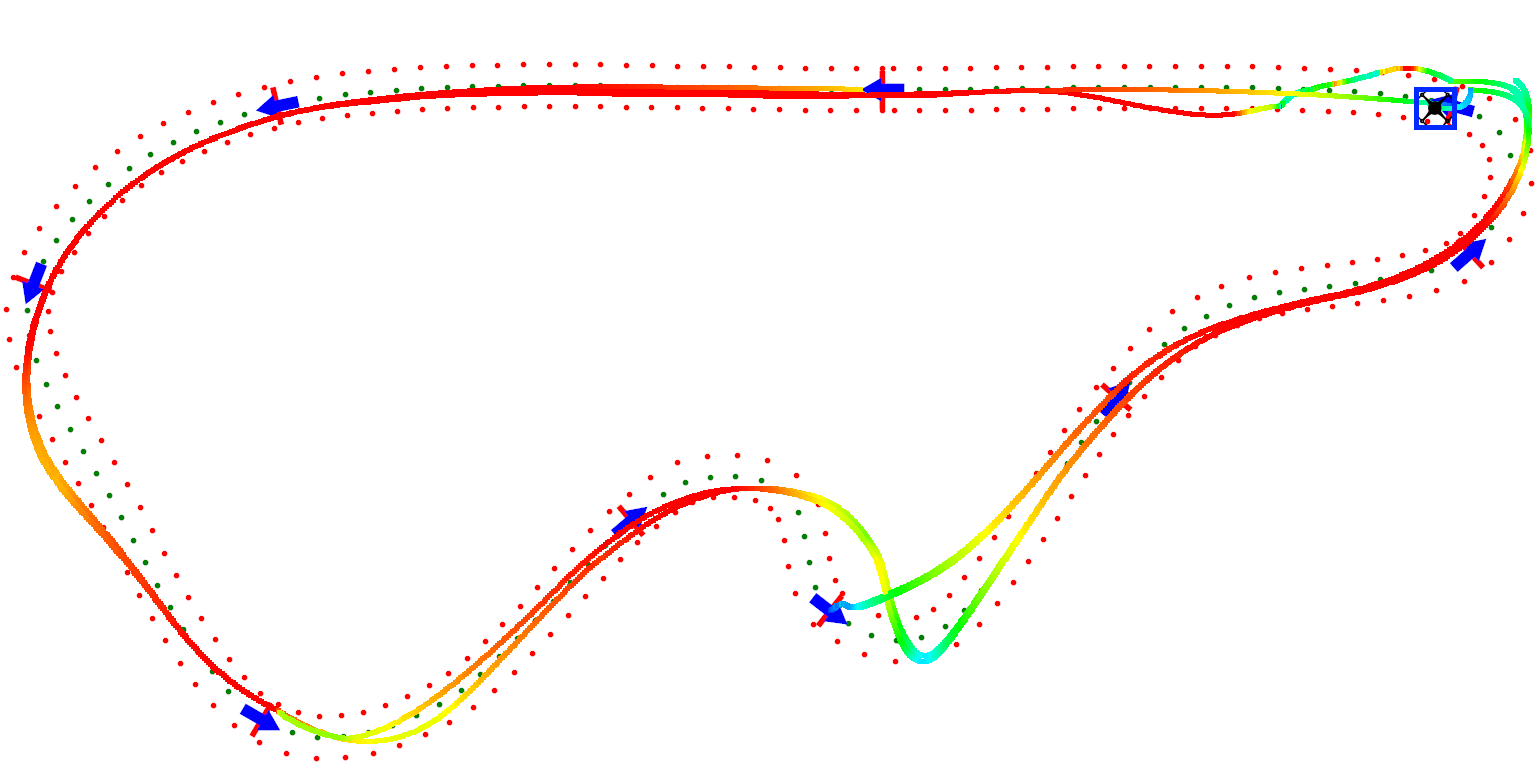} &
		\includegraphics[height=3cm]{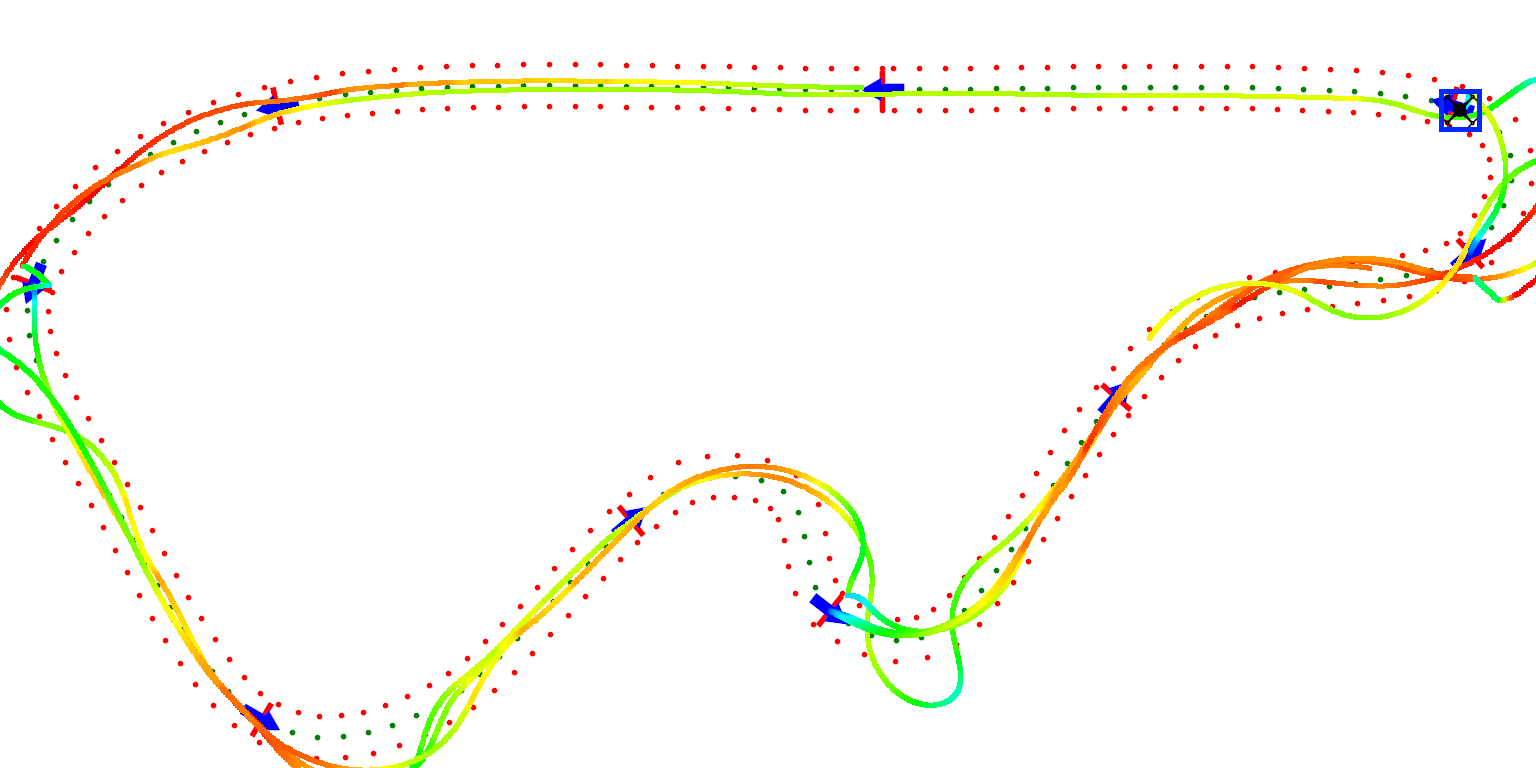} \\
		\small (c) PID1 (Conservative) & \small (d) PID2 (Aggressive) & \small (e) Ours (No Buffer) \\
		\includegraphics[height=3cm]{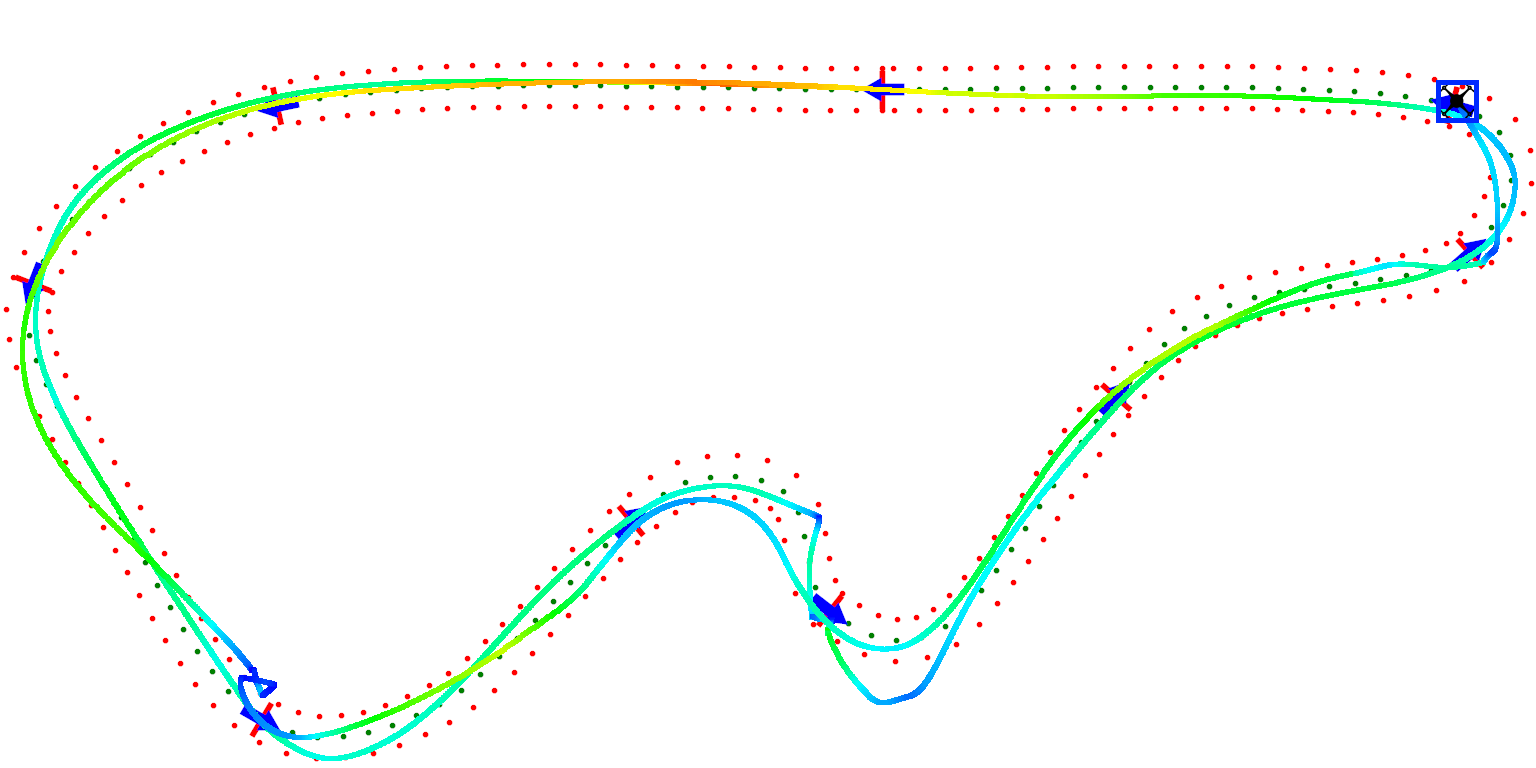} &
		\includegraphics[height=3cm]{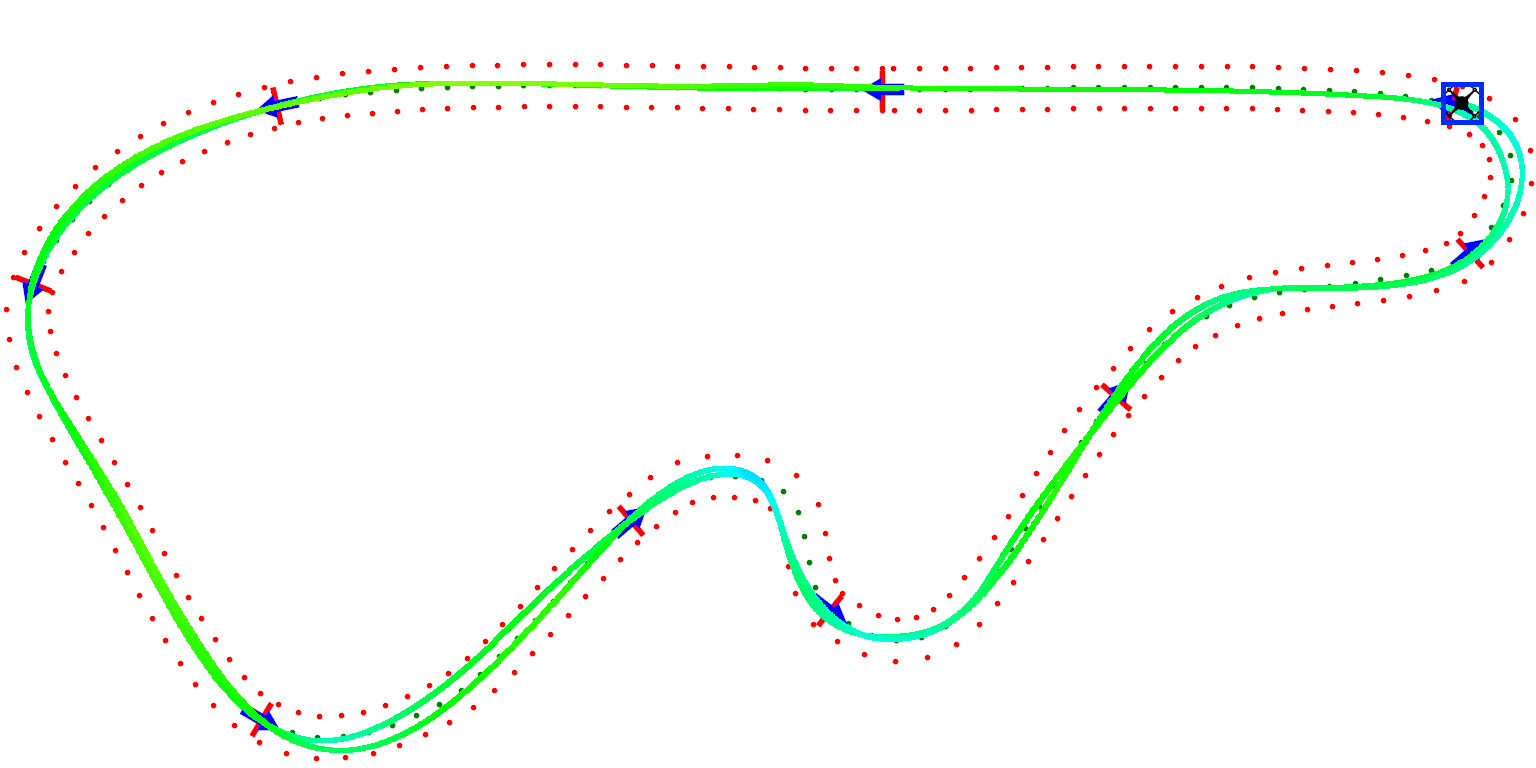} &
		\includegraphics[height=3cm]{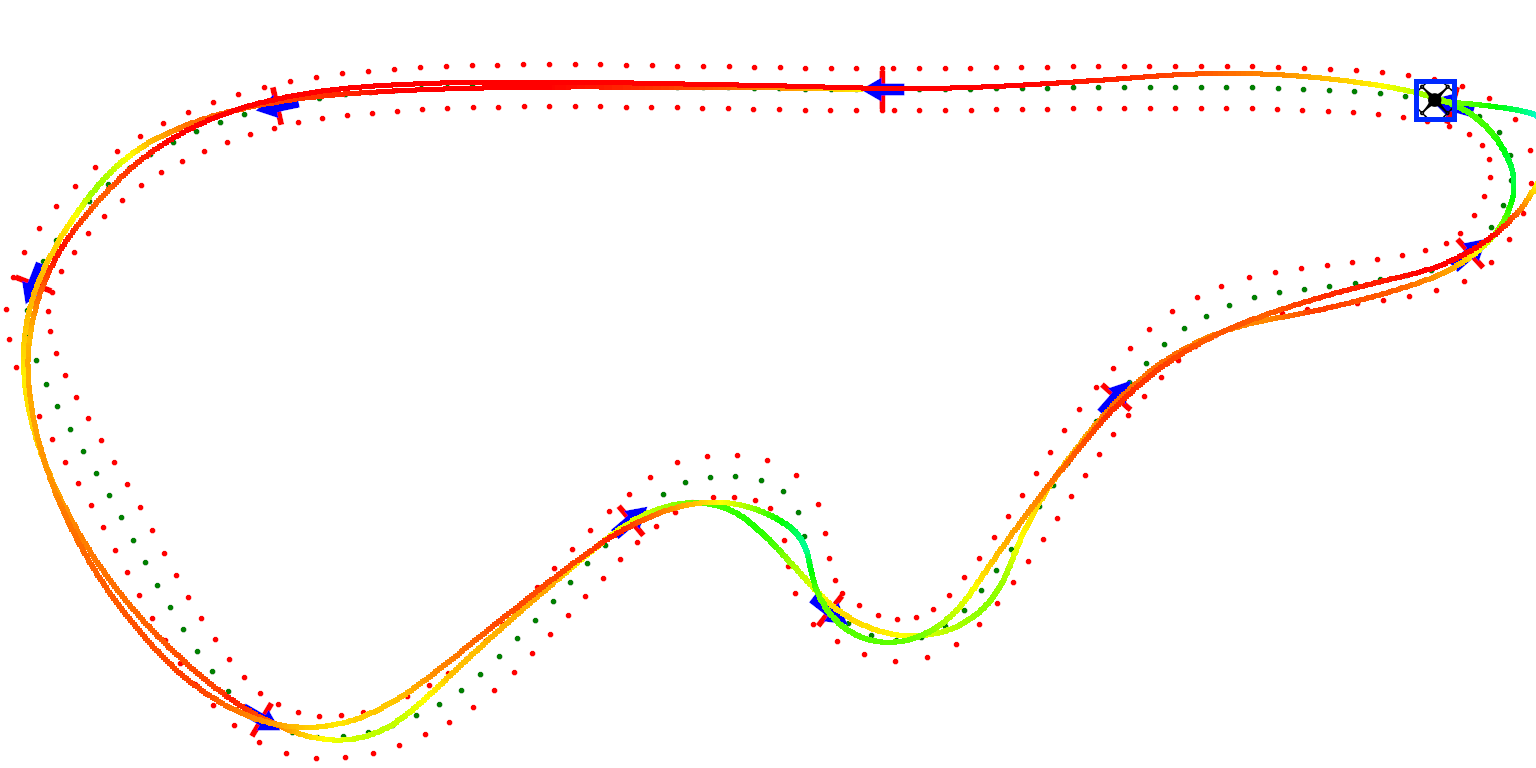} \\
        \small (f) Human (Novice) & \small (g) Human (Intermediate) & \small (h) Human (Professional)\\
		\includegraphics[height=3cm]{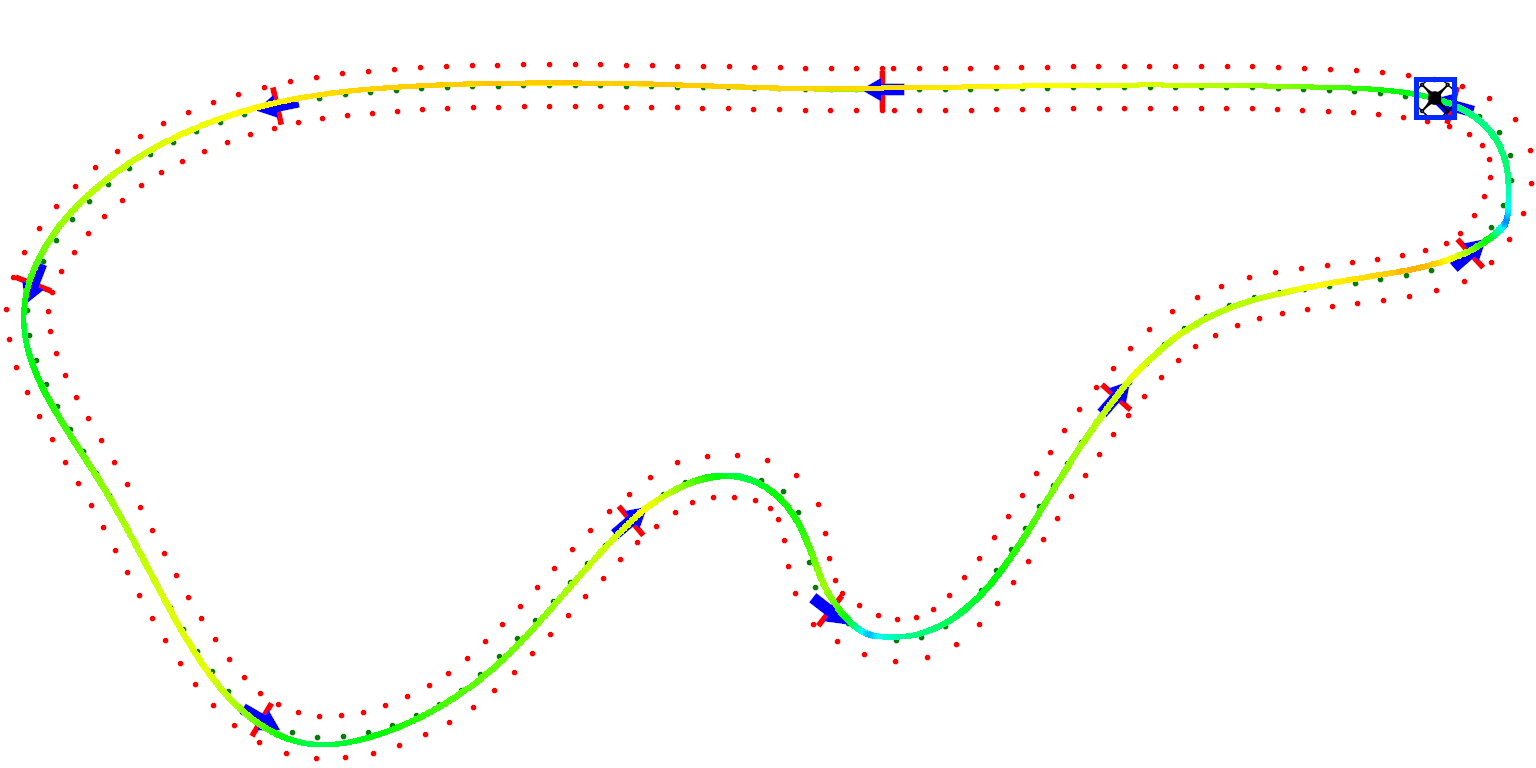} &
		\includegraphics[height=3cm]{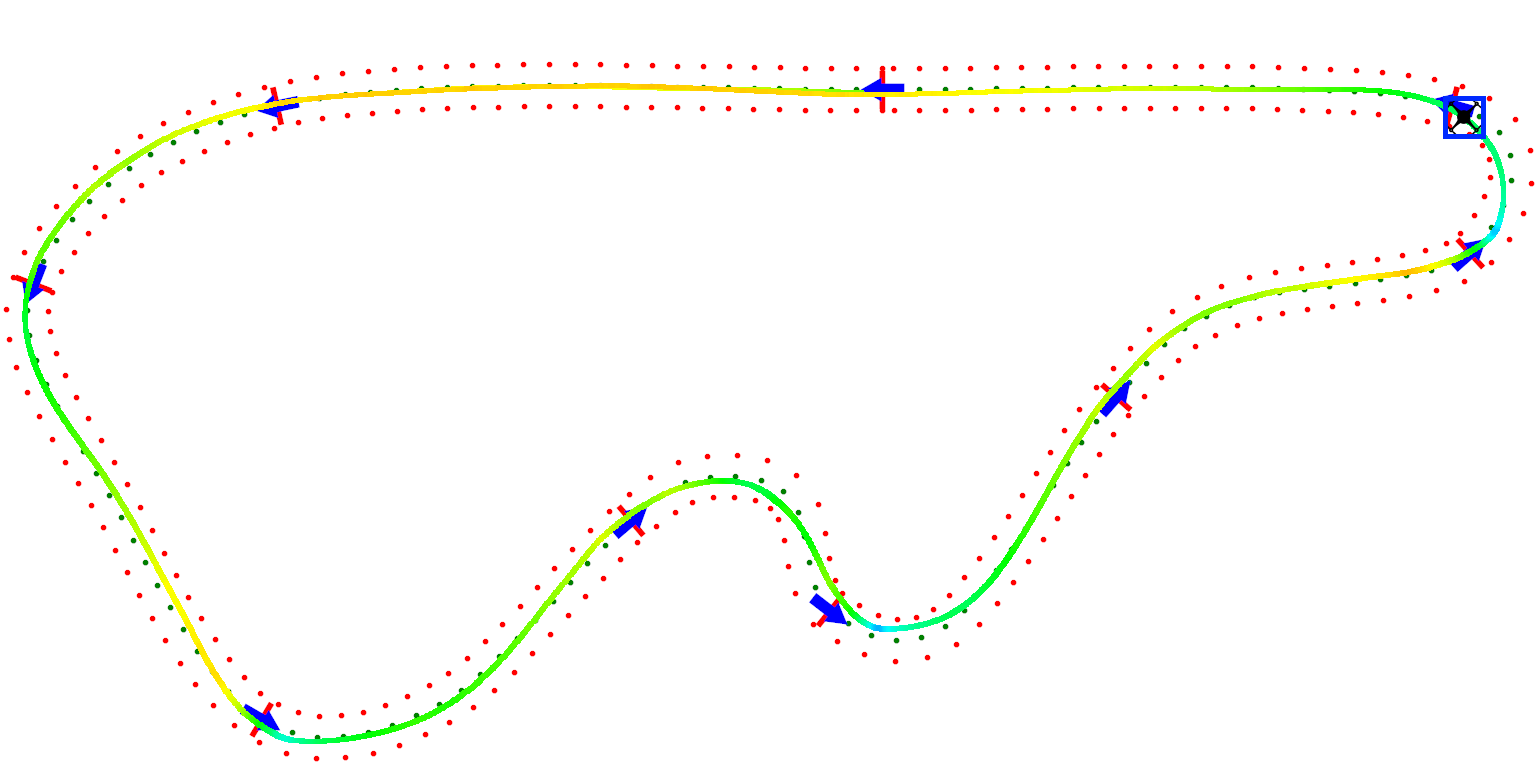} &
		\includegraphics[height=3cm]{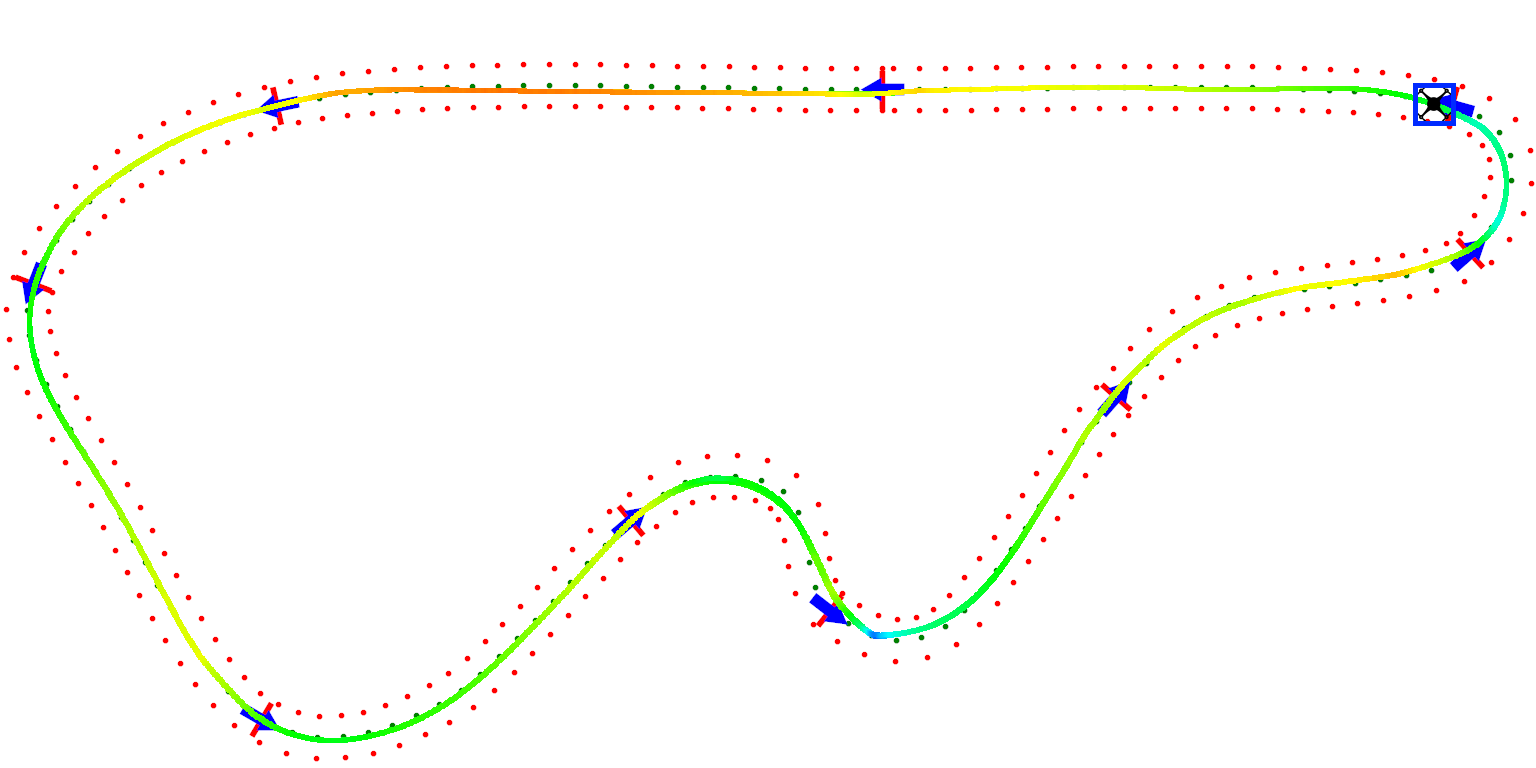} \\
		\small (i) Ours (Reference) & \small (j) Ours (Night) & \small (k)  Ours (Sunrise) \\
		\includegraphics[height=3cm]{sup_figures/track4_ours_grass.png} &
		\includegraphics[height=3cm]{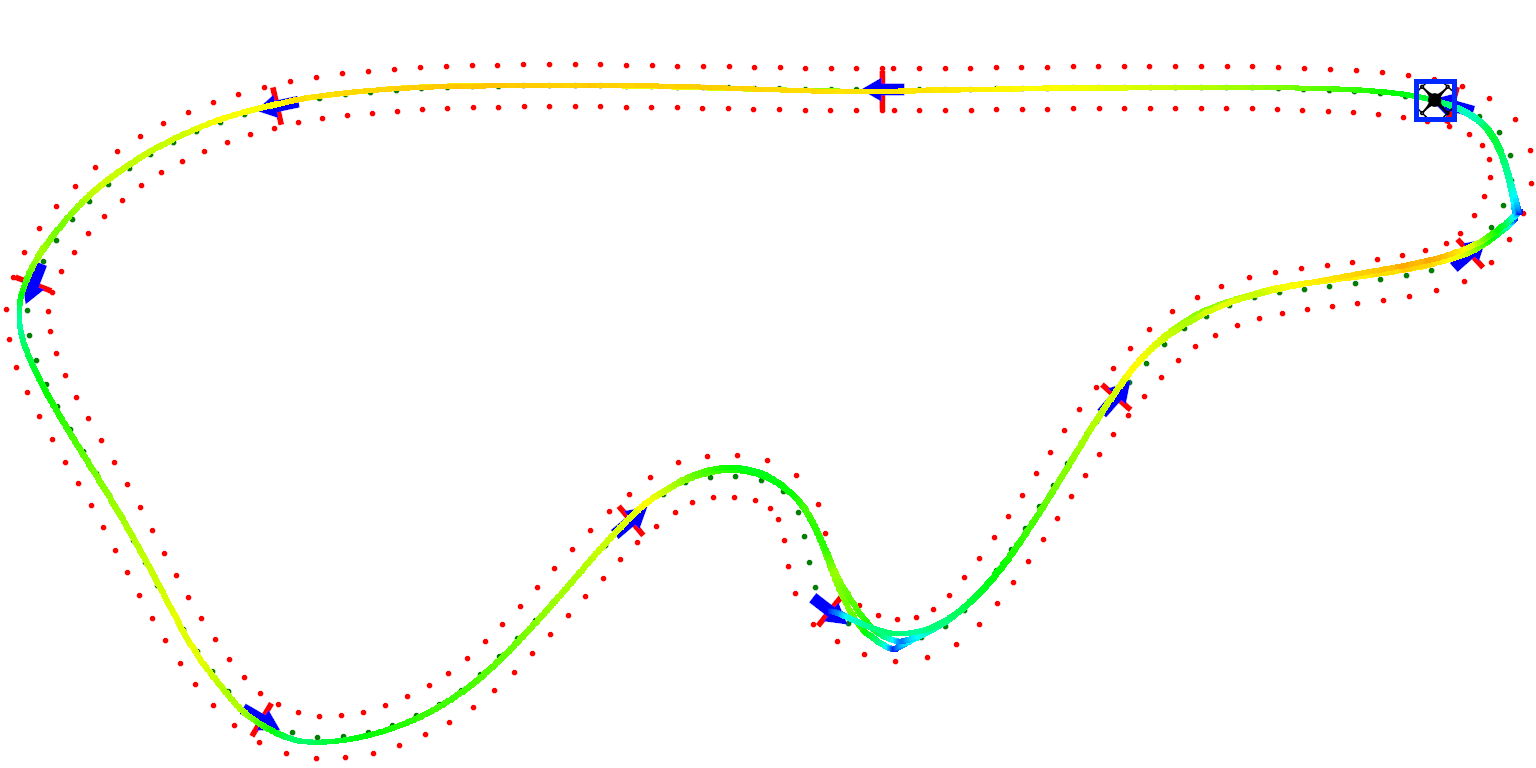} &
		\includegraphics[height=3cm]{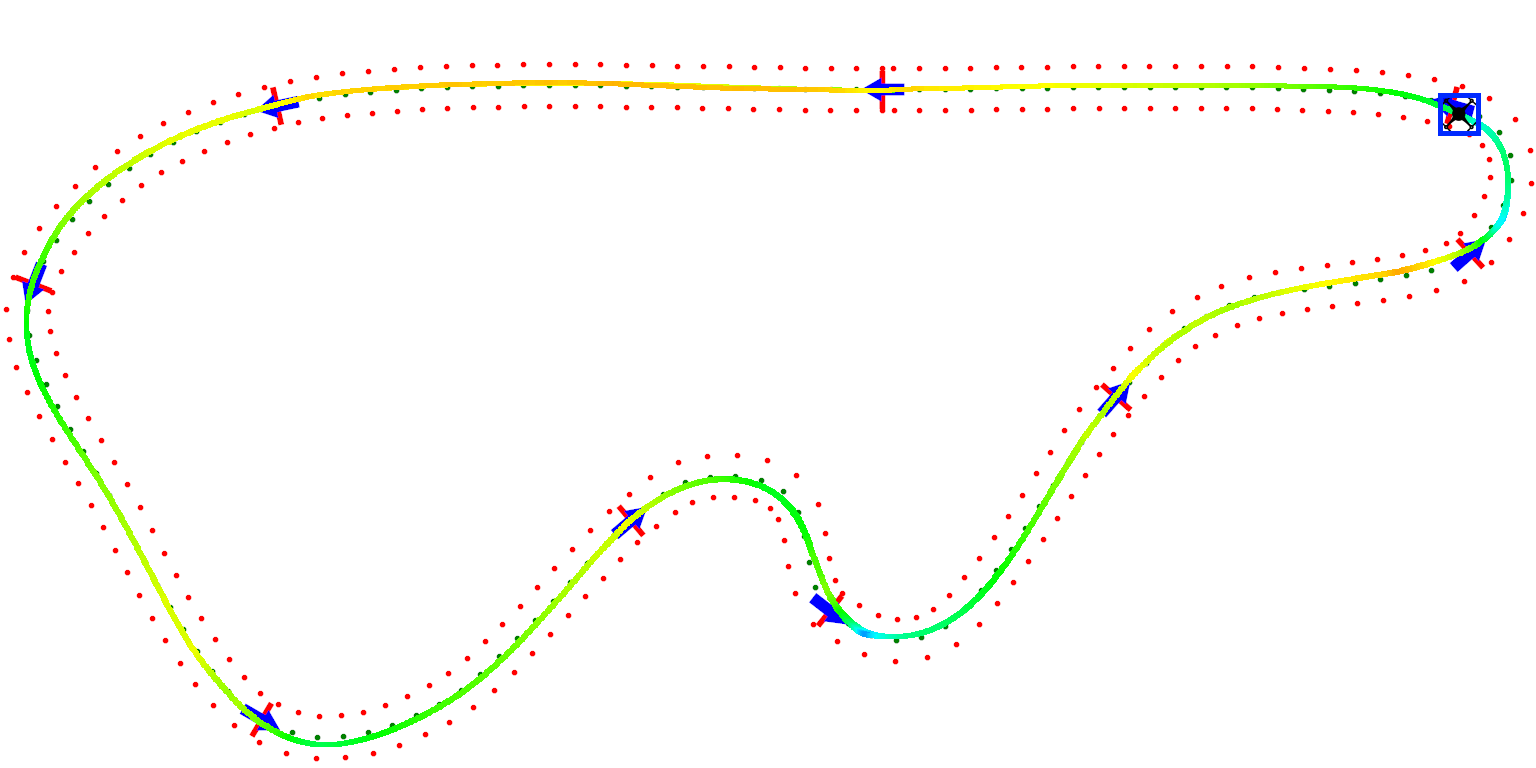} \\
		\small (l) Ours (Reference) & \small (m) Ours (Fog) & \small (o)  Ours (Rain) \\
		\includegraphics[height=3cm]{sup_figures/track4_ours_grass.png} &
		\includegraphics[height=3cm]{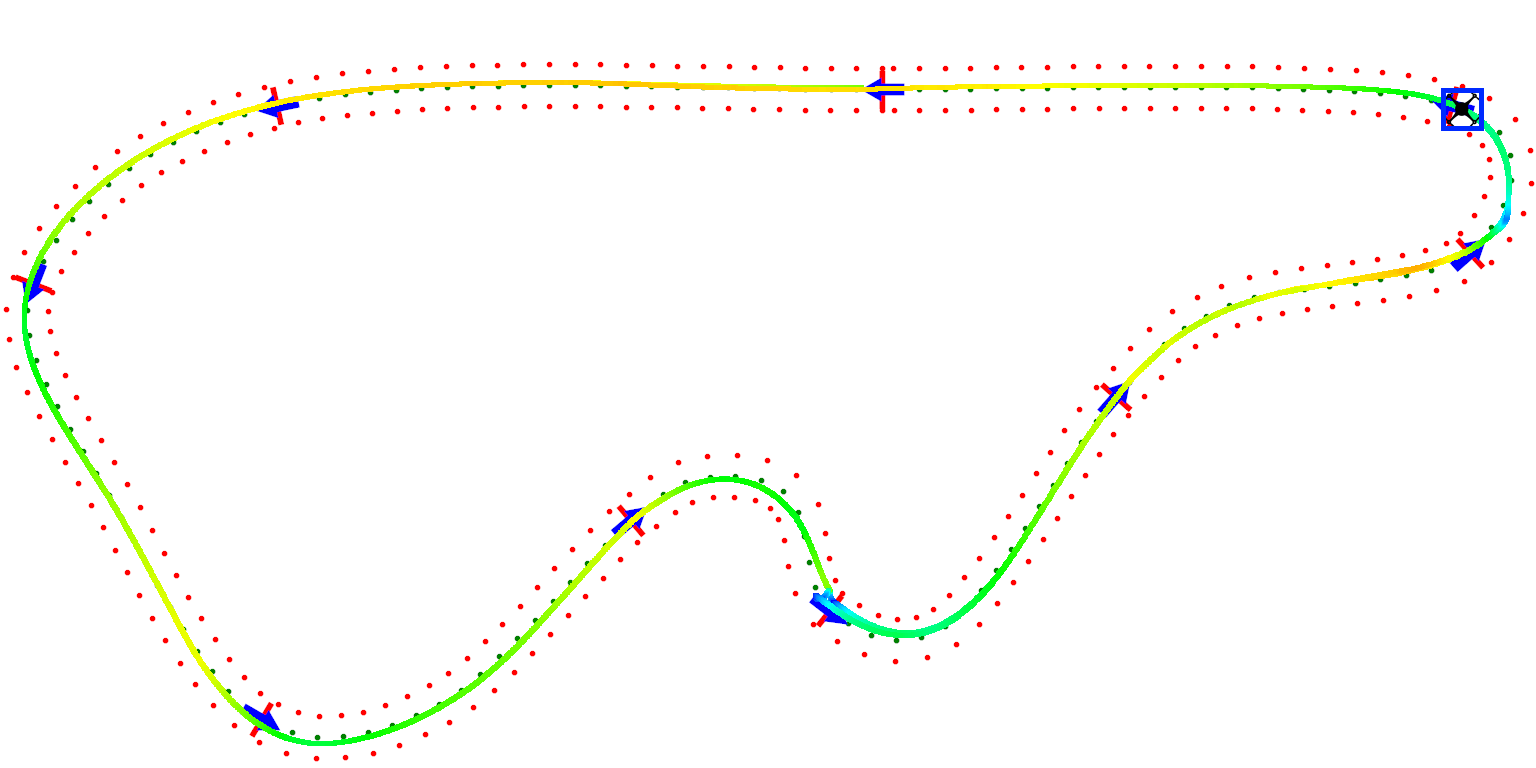} &
		\includegraphics[height=3cm]{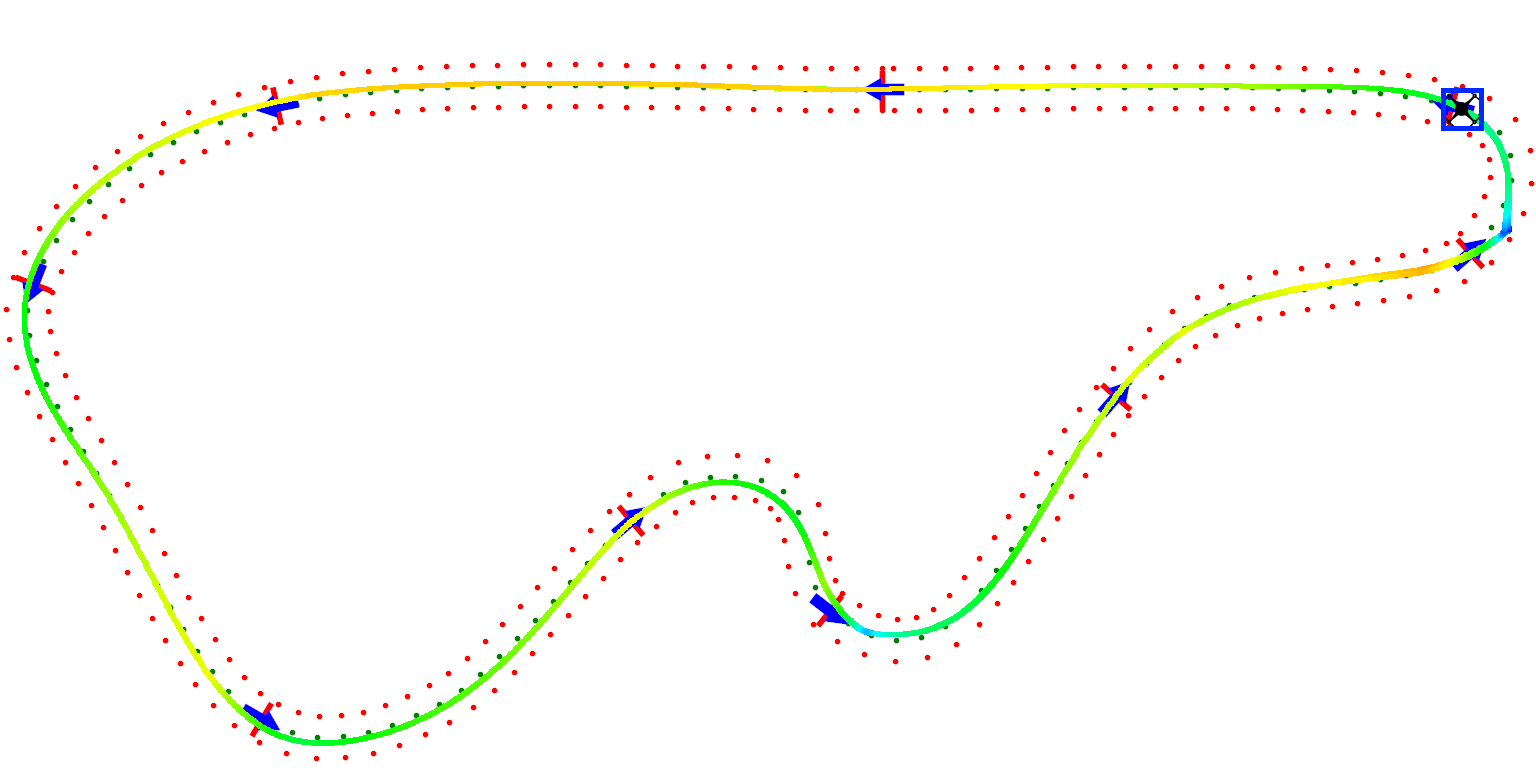} \\
		\small (p) Ours (Grass) & \small (q) Ours (Mud) & \small (r)  Ours (Snow) \\
        \multicolumn{3}{c}{\includegraphics[height=1.2cm]{sup_figures/ColorScaleHorizontal.png}}
\end{tabular}
\captionof{figure}{Qualitative results on track4. The color encodes speed as a heatmap, where blue corresponds to the minimum speed and red to the maximum speed.}
\label{fig:qualitive_results_track4}
\end{figure*}

\begin{figure*}
\centering
\begin{tabular}{@{}c@{\hspace{1mm}}c@{\hspace{1mm}}c@{\hspace{8mm}}c@{}}
		\includegraphics[height=3cm]{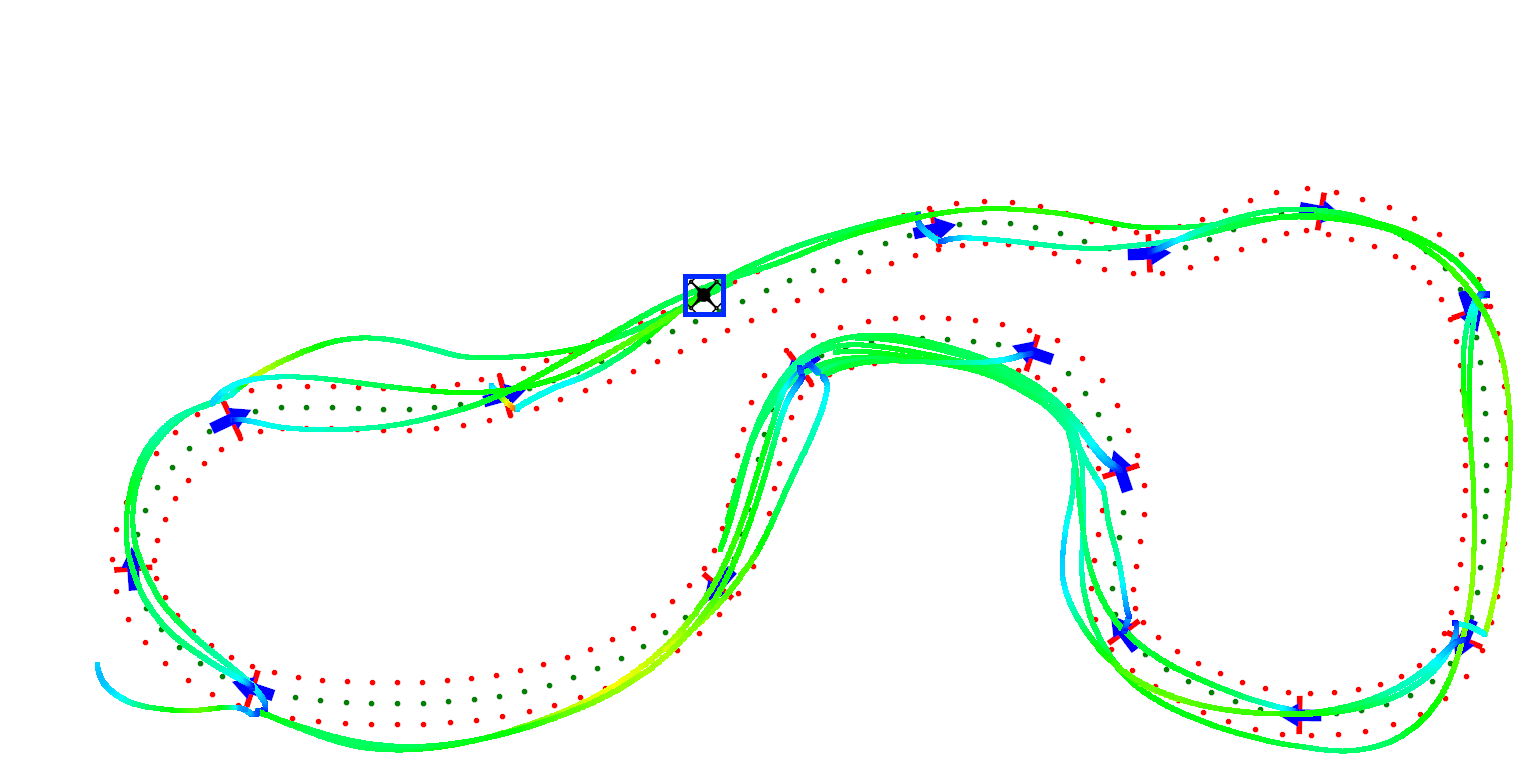} &
		\includegraphics[height=3cm]{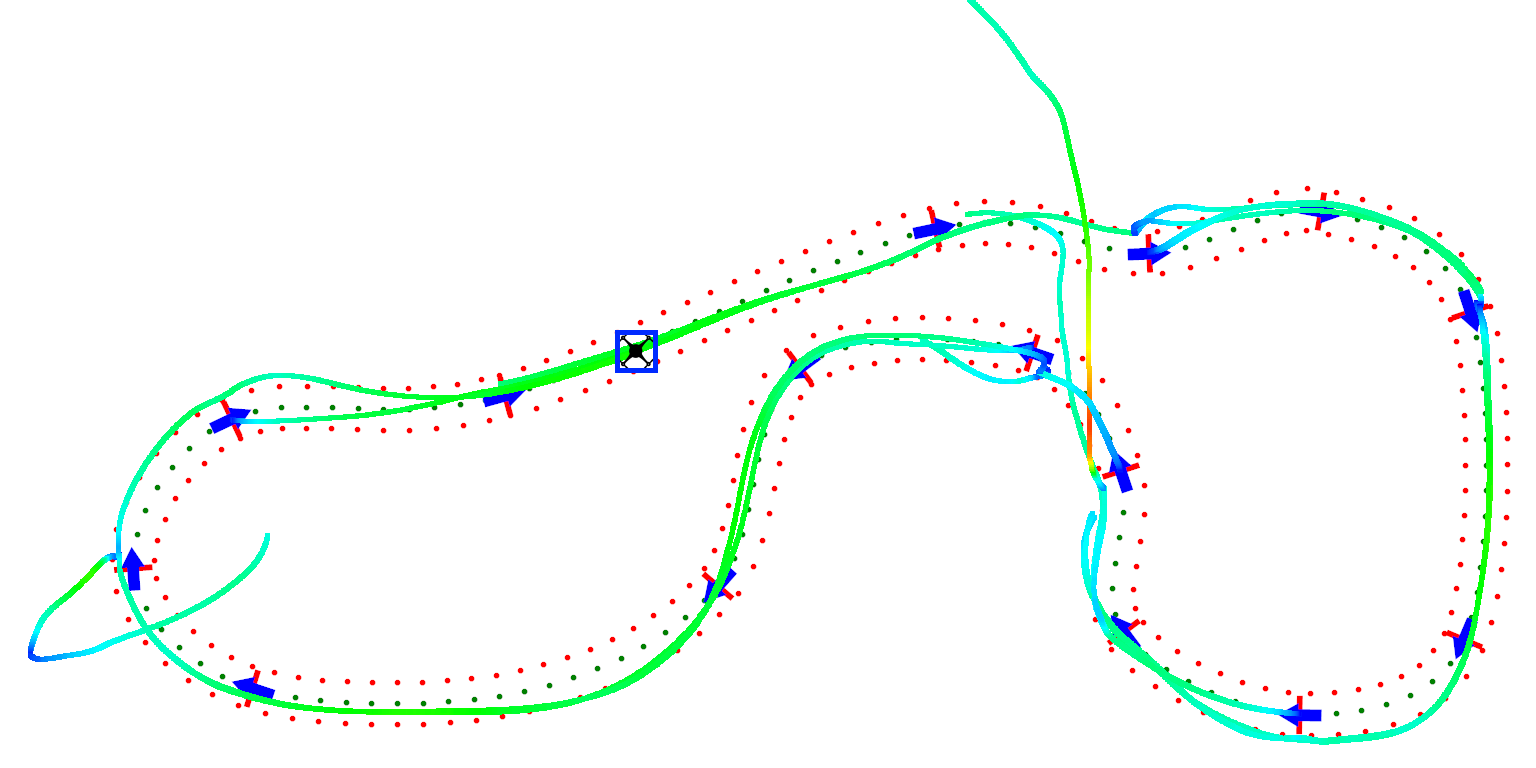} &
		\includegraphics[height=3cm]{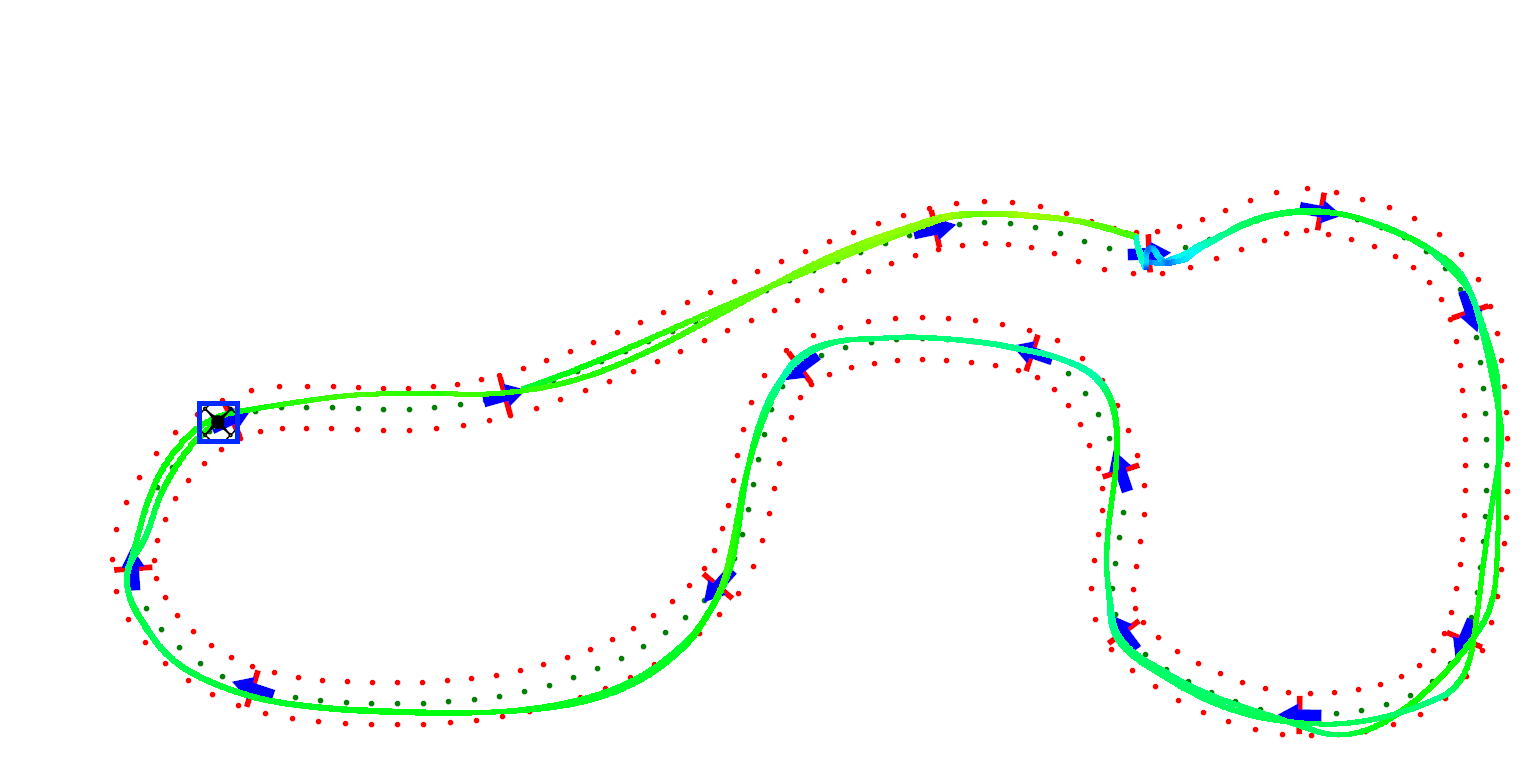} \\
		\small (a) End2End (MAV) & \small (b) End2End (Nvidia) & \small (c) End2End (Ours) \\
		\includegraphics[height=3cm]{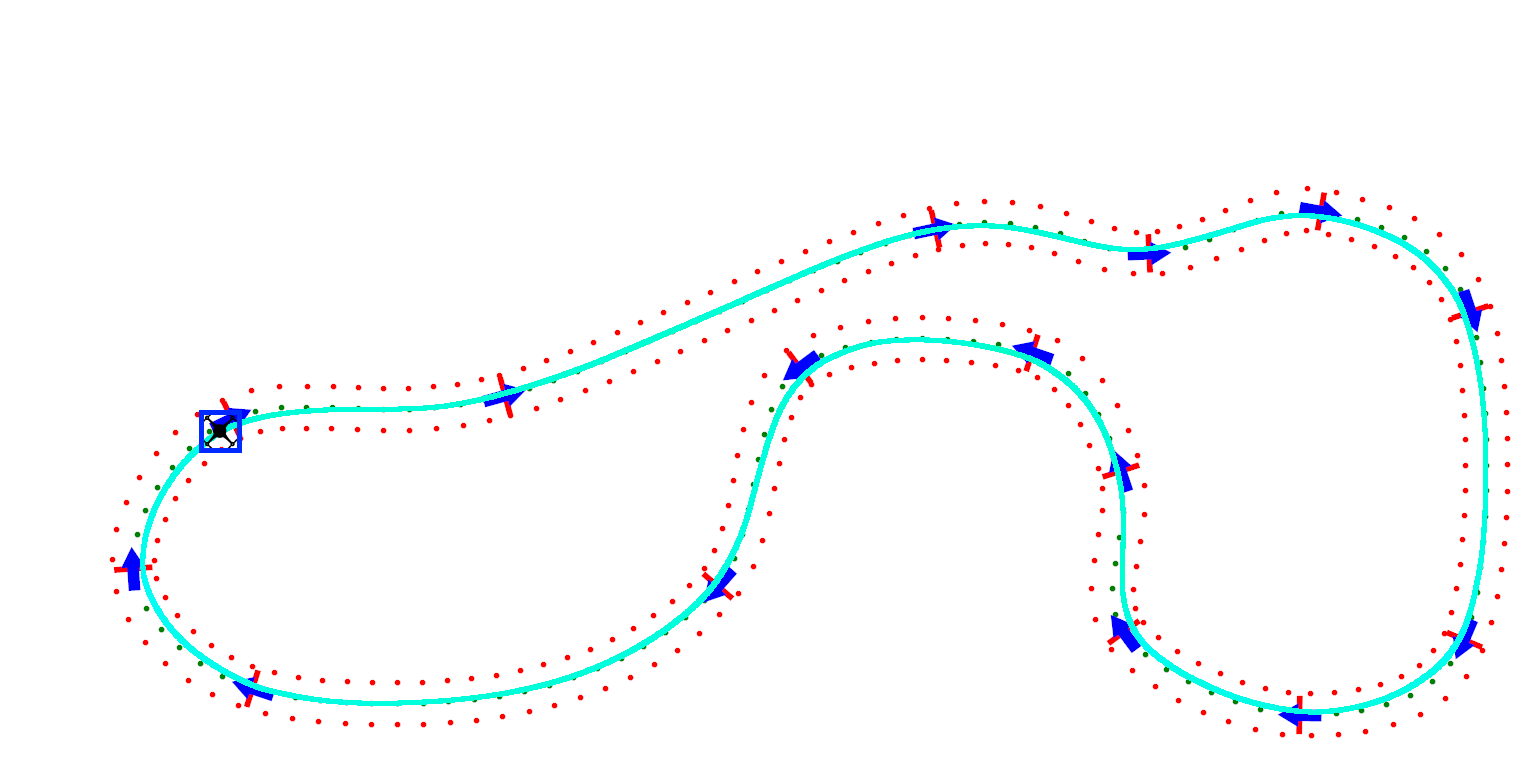} &
		\includegraphics[height=3cm]{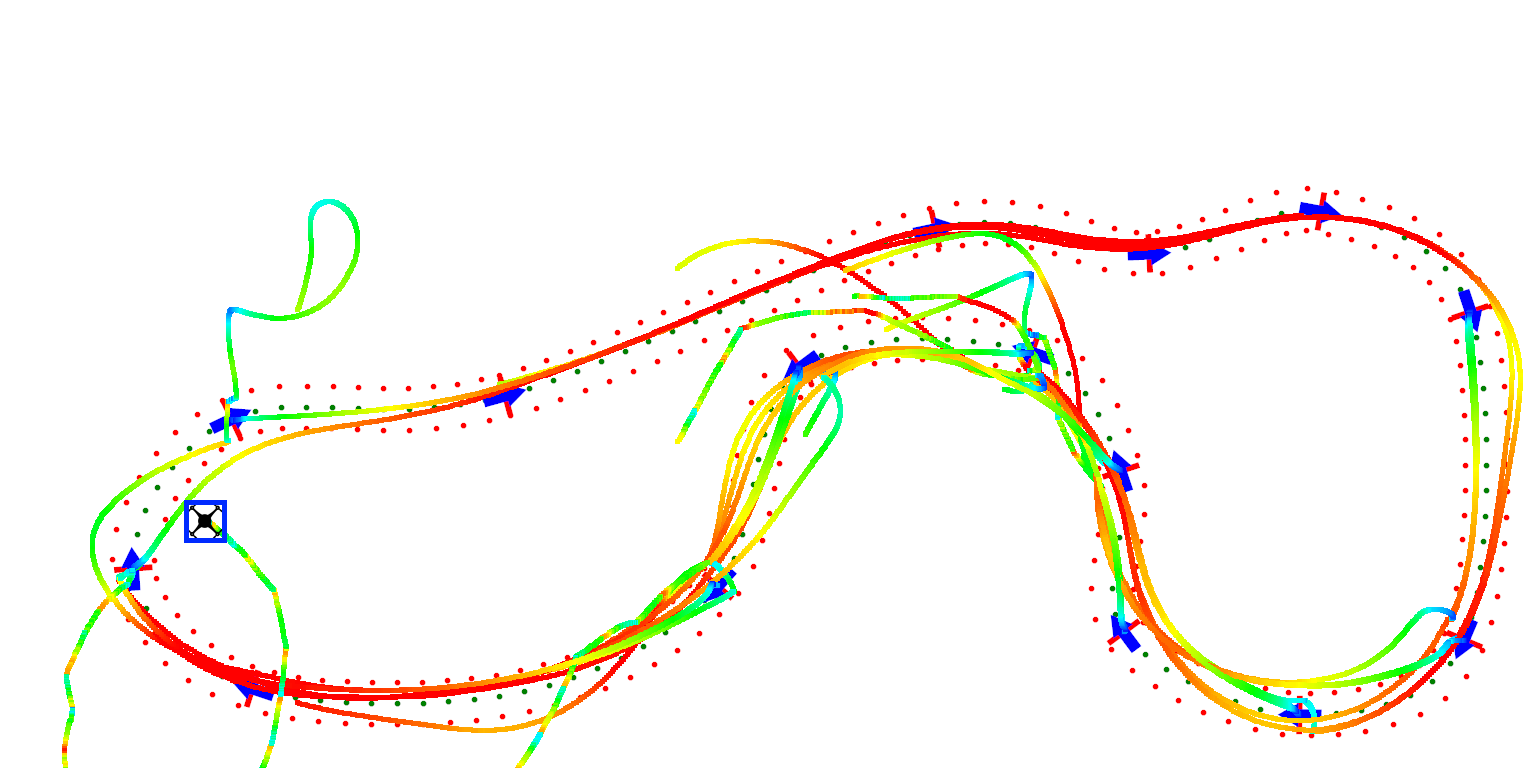} &
		\includegraphics[height=3cm]{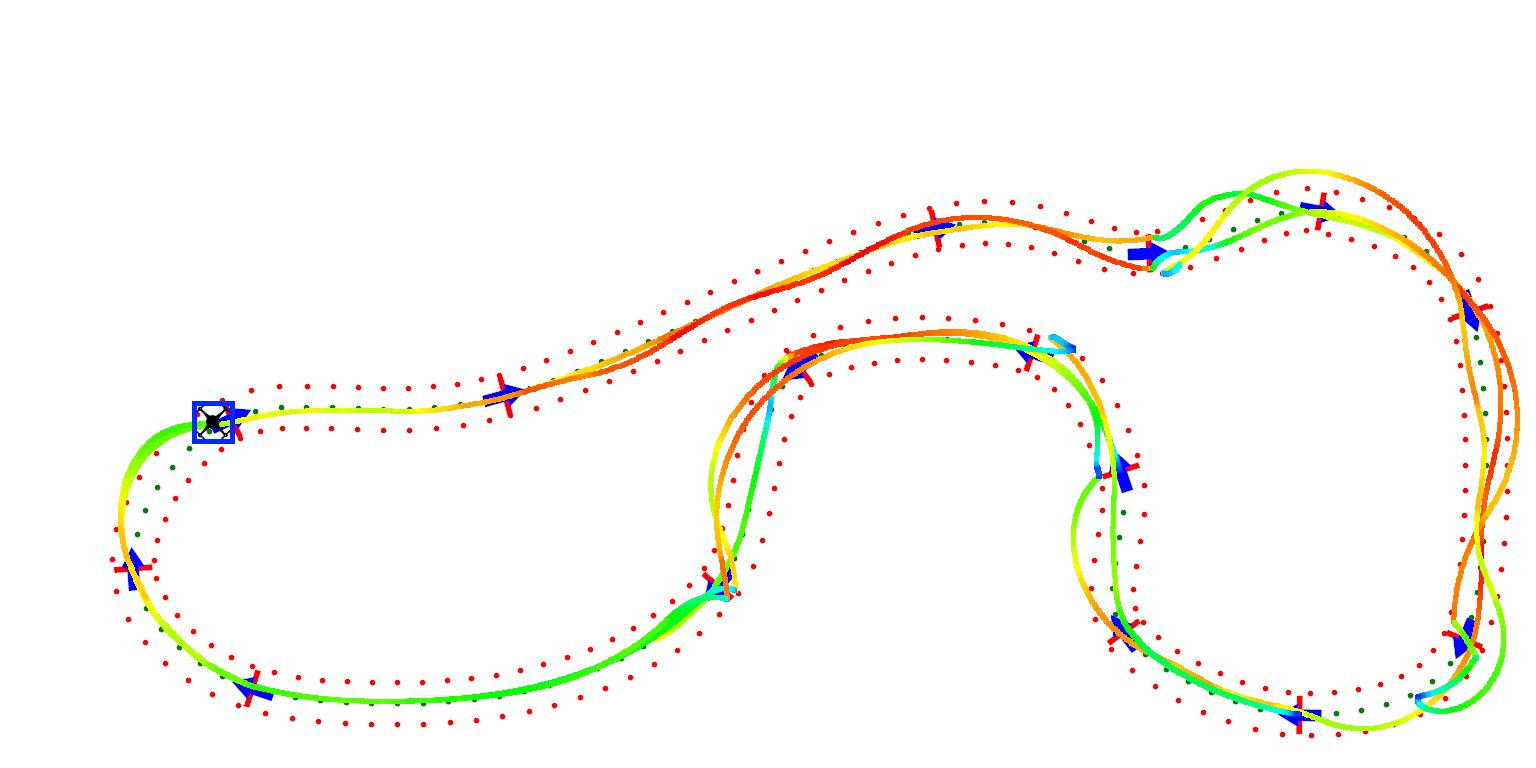} \\
		\small (c) PID1 (Conservative) & \small (d) PID2 (Aggressive) & \small (e) Ours (No Buffer) \\
		\includegraphics[height=3cm]{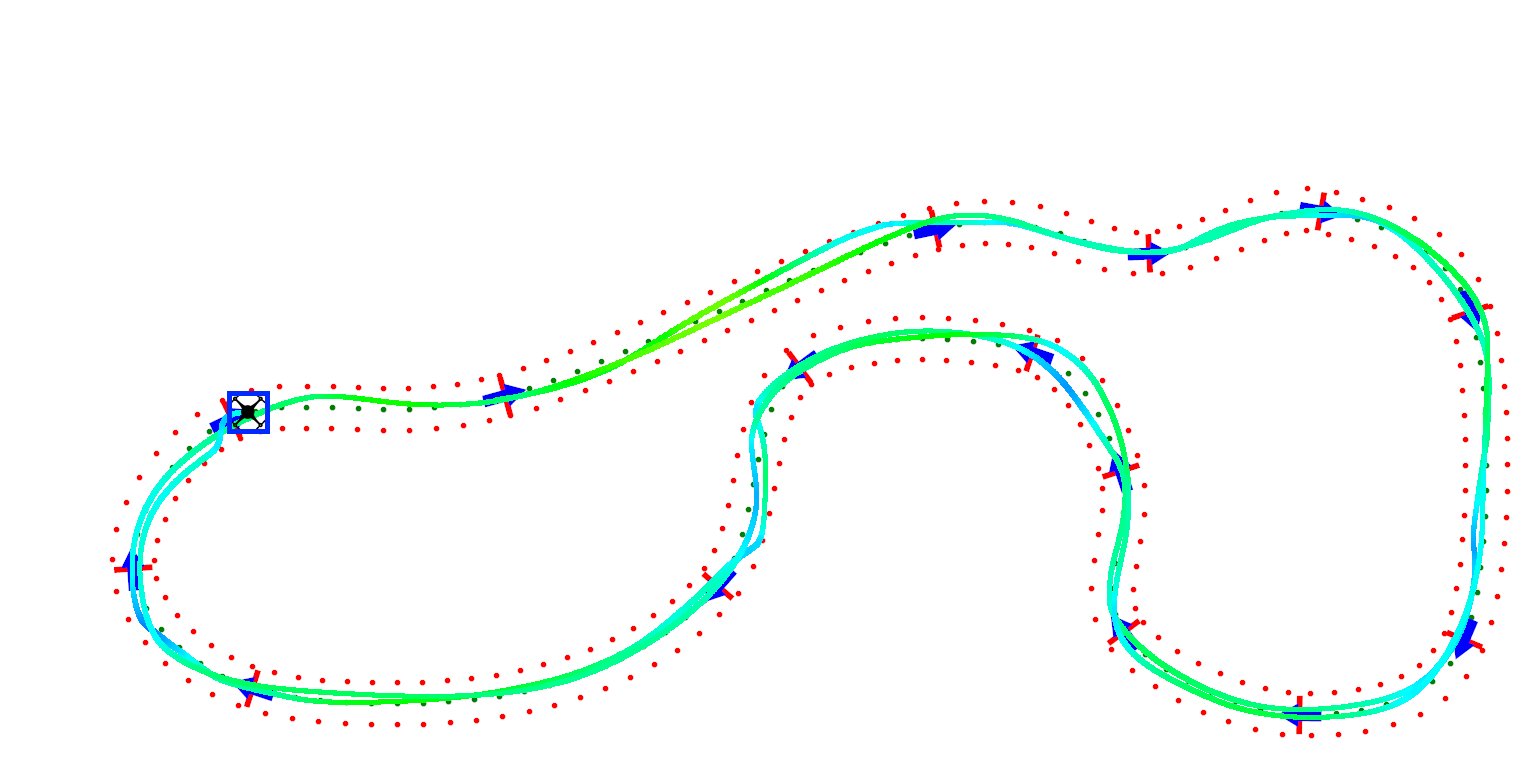} &
		\includegraphics[height=3cm]{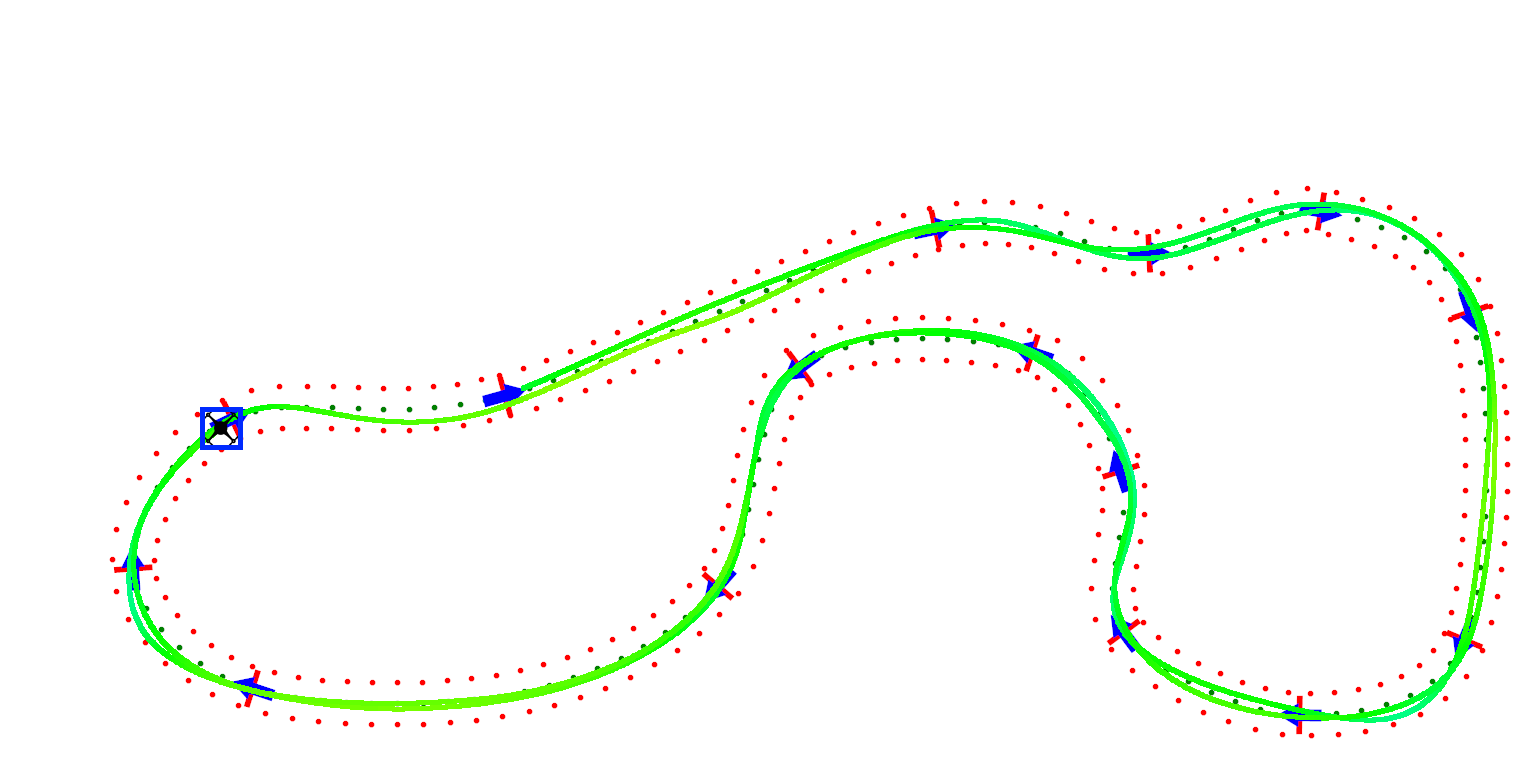} &
		\includegraphics[height=3cm]{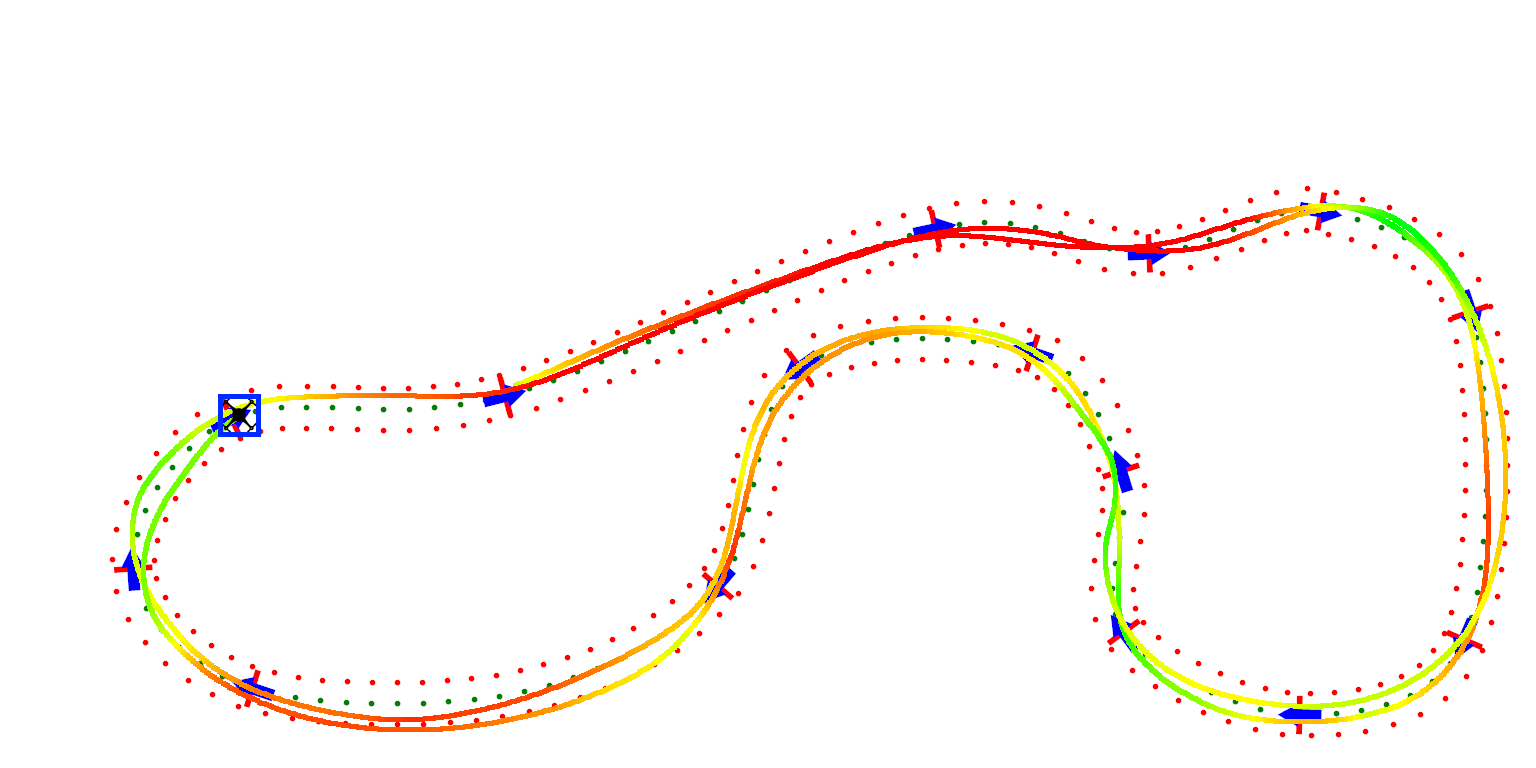} \\
        \small (f) Human (Novice) & \small (g) Human (Intermediate) & \small (h) Human (Professional)\\
		\includegraphics[height=3cm]{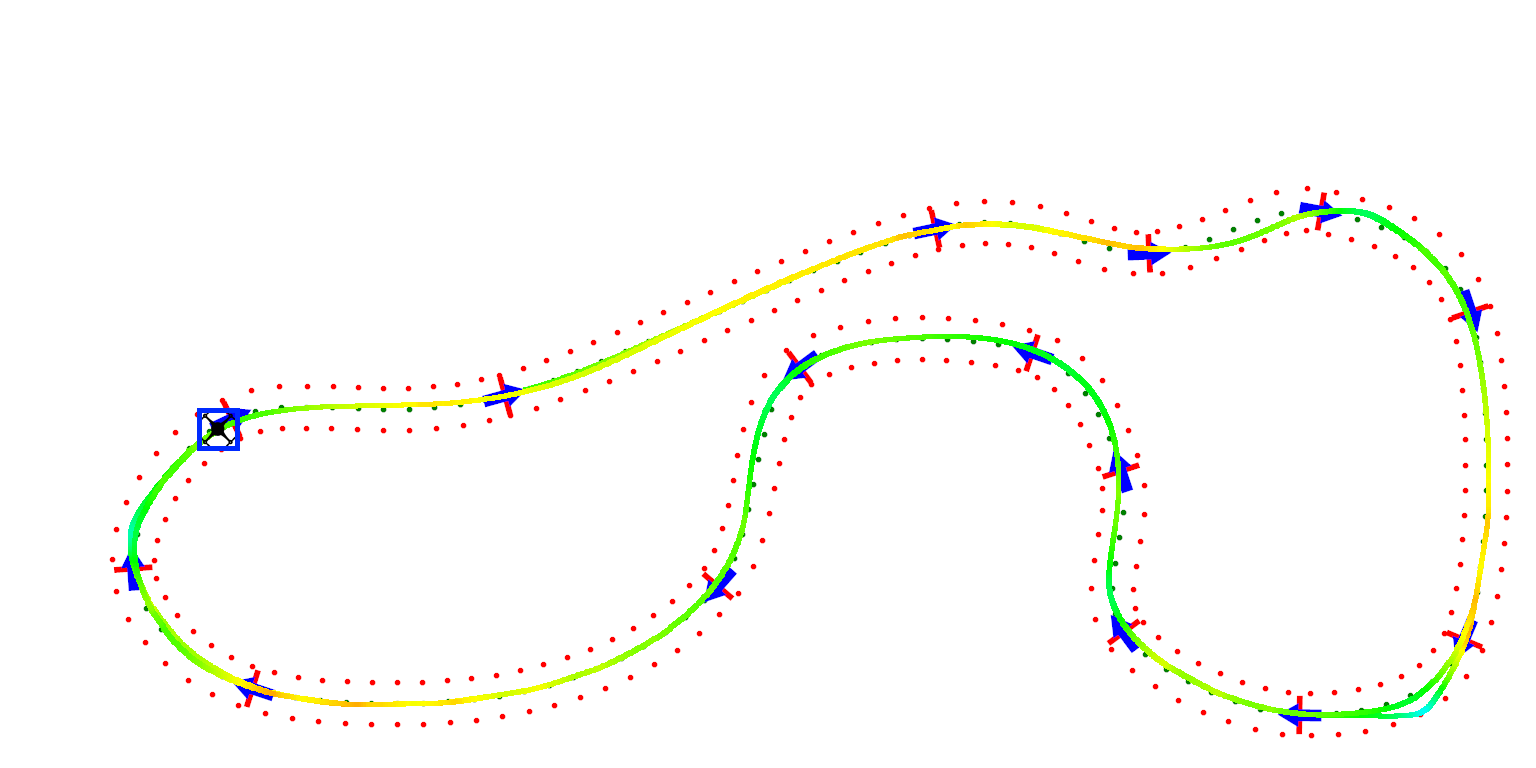} &
		\includegraphics[height=3cm]{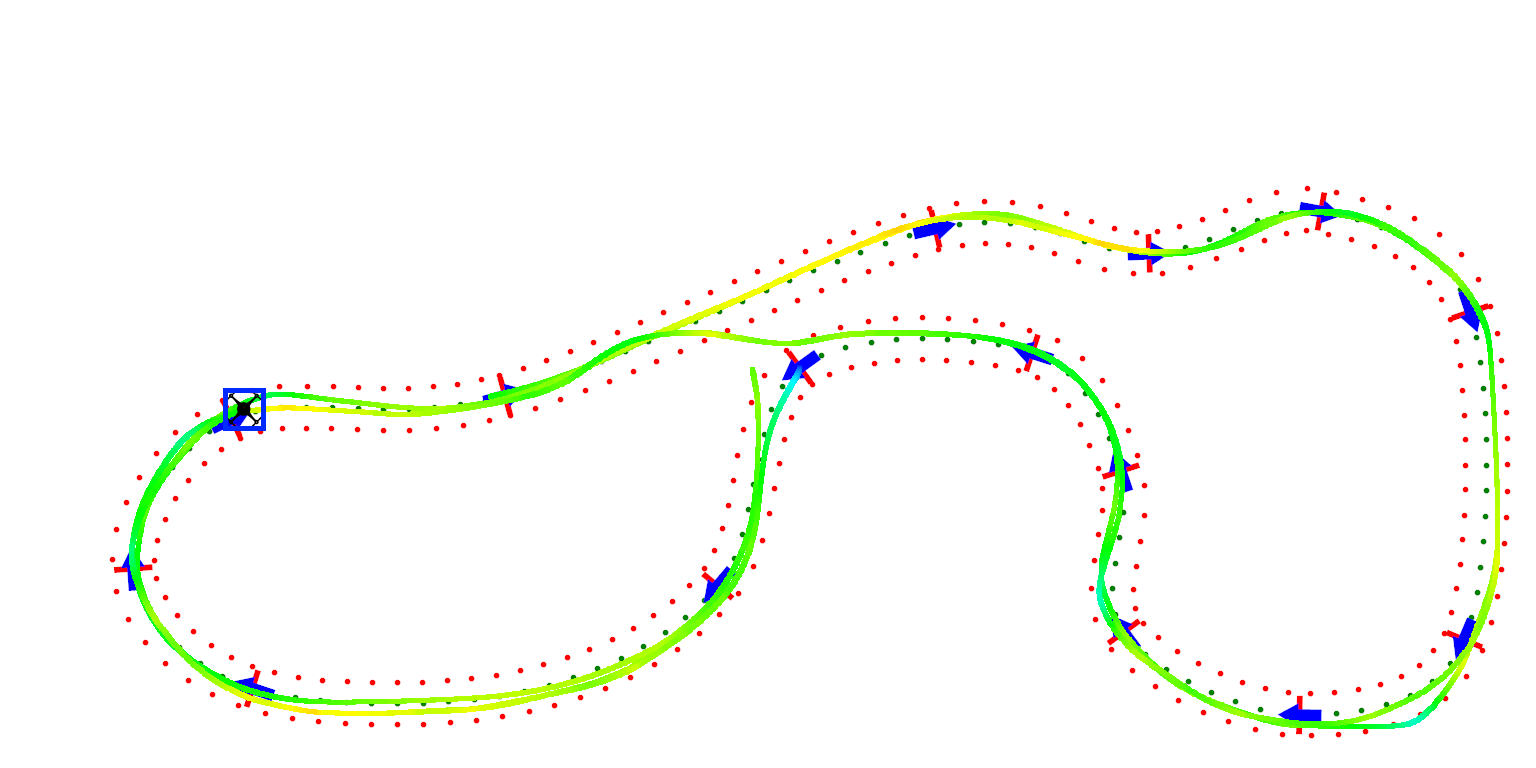} &
		\includegraphics[height=3cm]{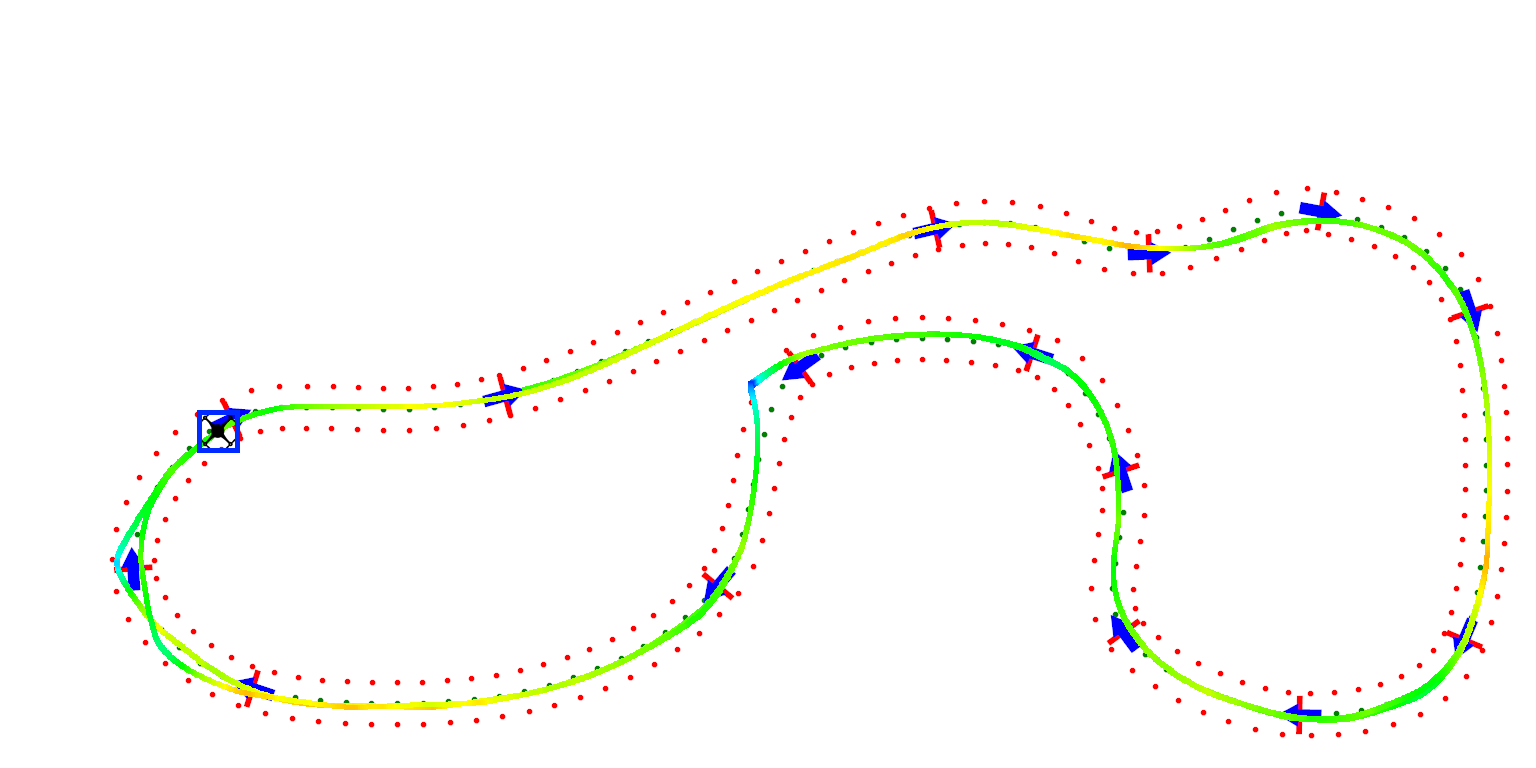} \\
		\small (i) Ours (Reference) & \small (j) Ours (Night) & \small (k)  Ours (Sunrise) \\
		\includegraphics[height=3cm]{sup_figures/track5_ours_grass.png} &
		\includegraphics[height=3cm]{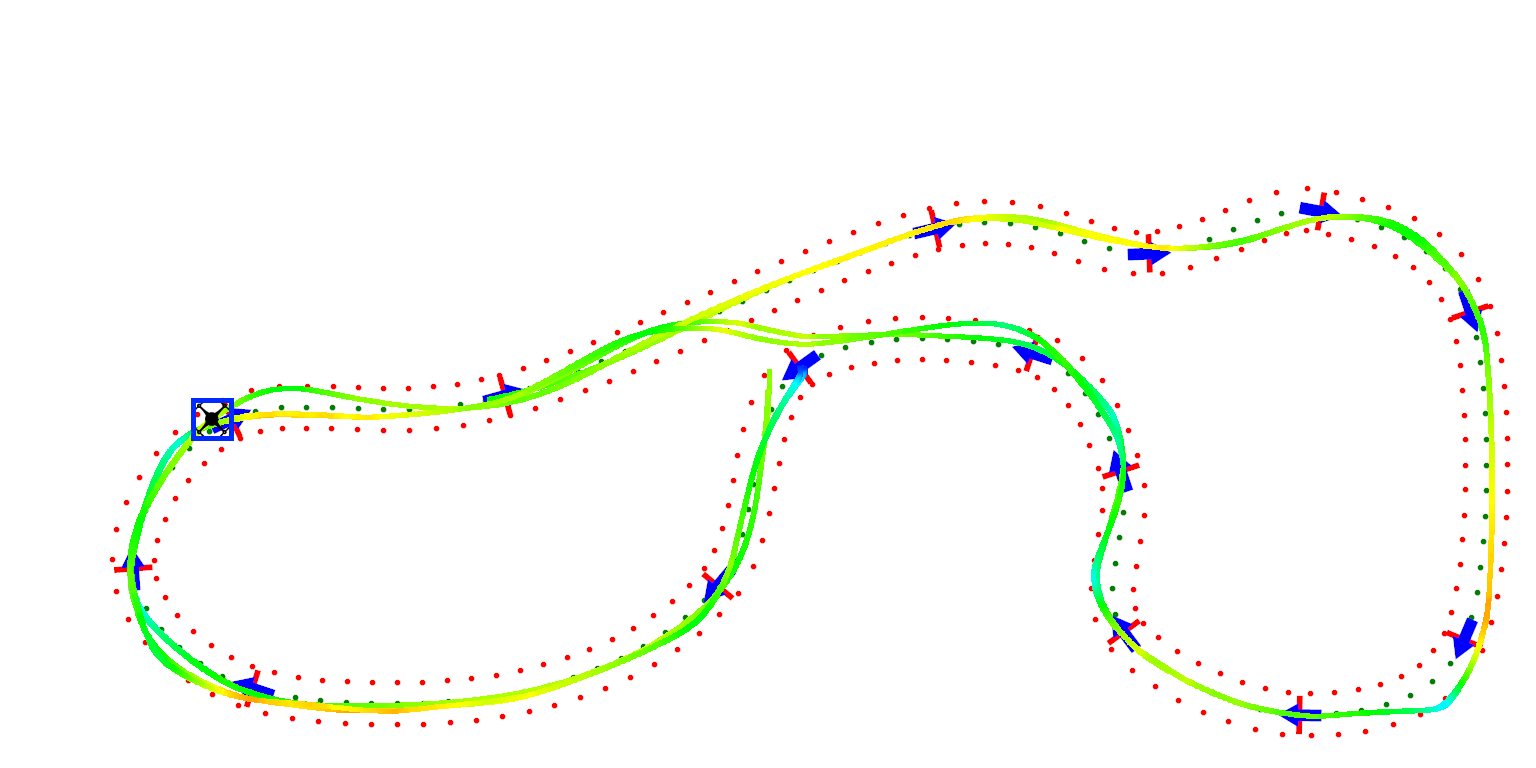} &
		\includegraphics[height=3cm]{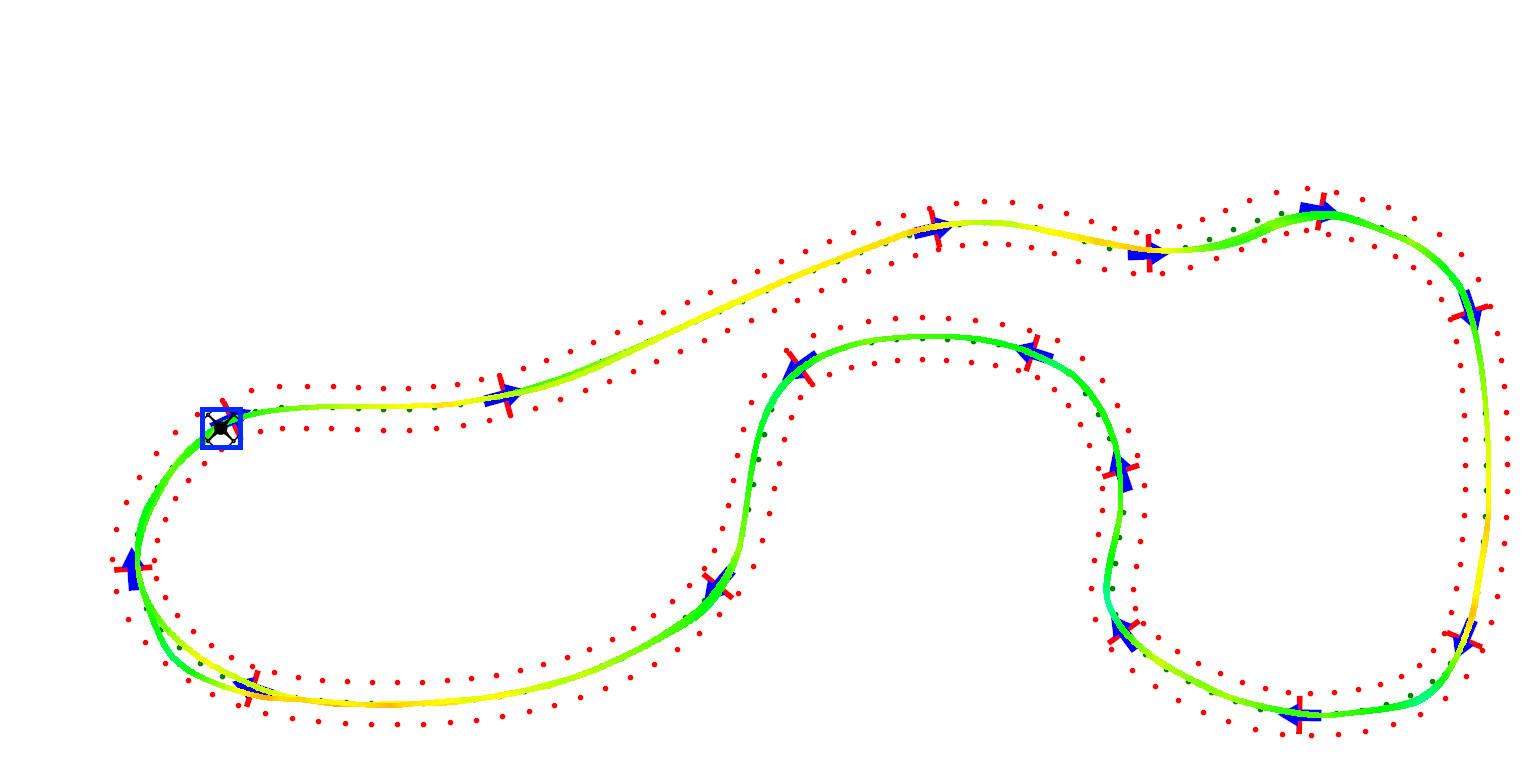} \\
		\small (l) Ours (Reference) & \small (m) Ours (Fog) & \small (o)  Ours (Rain) \\
		\includegraphics[height=3cm]{sup_figures/track5_ours_grass.png} &
		\includegraphics[height=3cm]{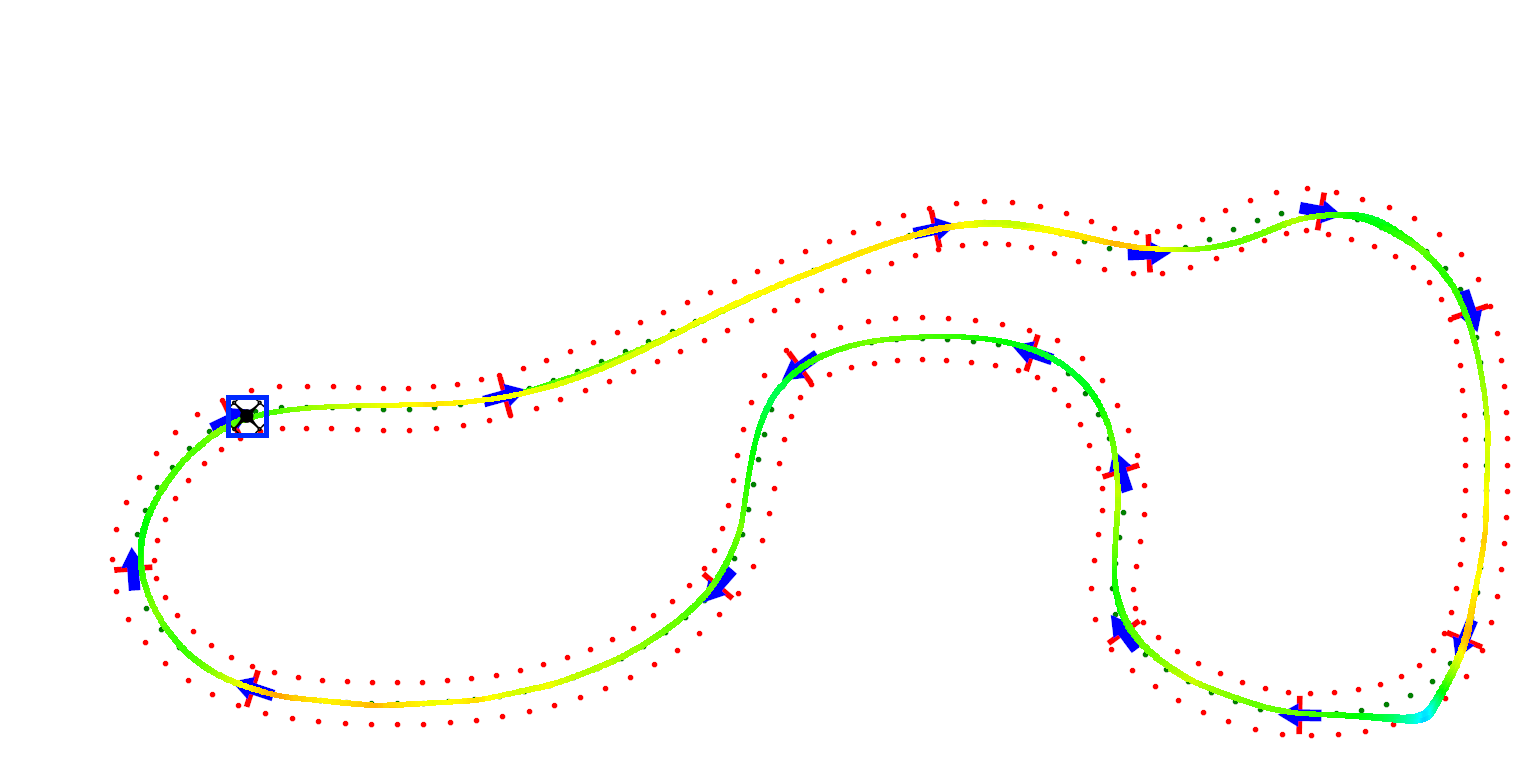} &
		\includegraphics[height=3cm]{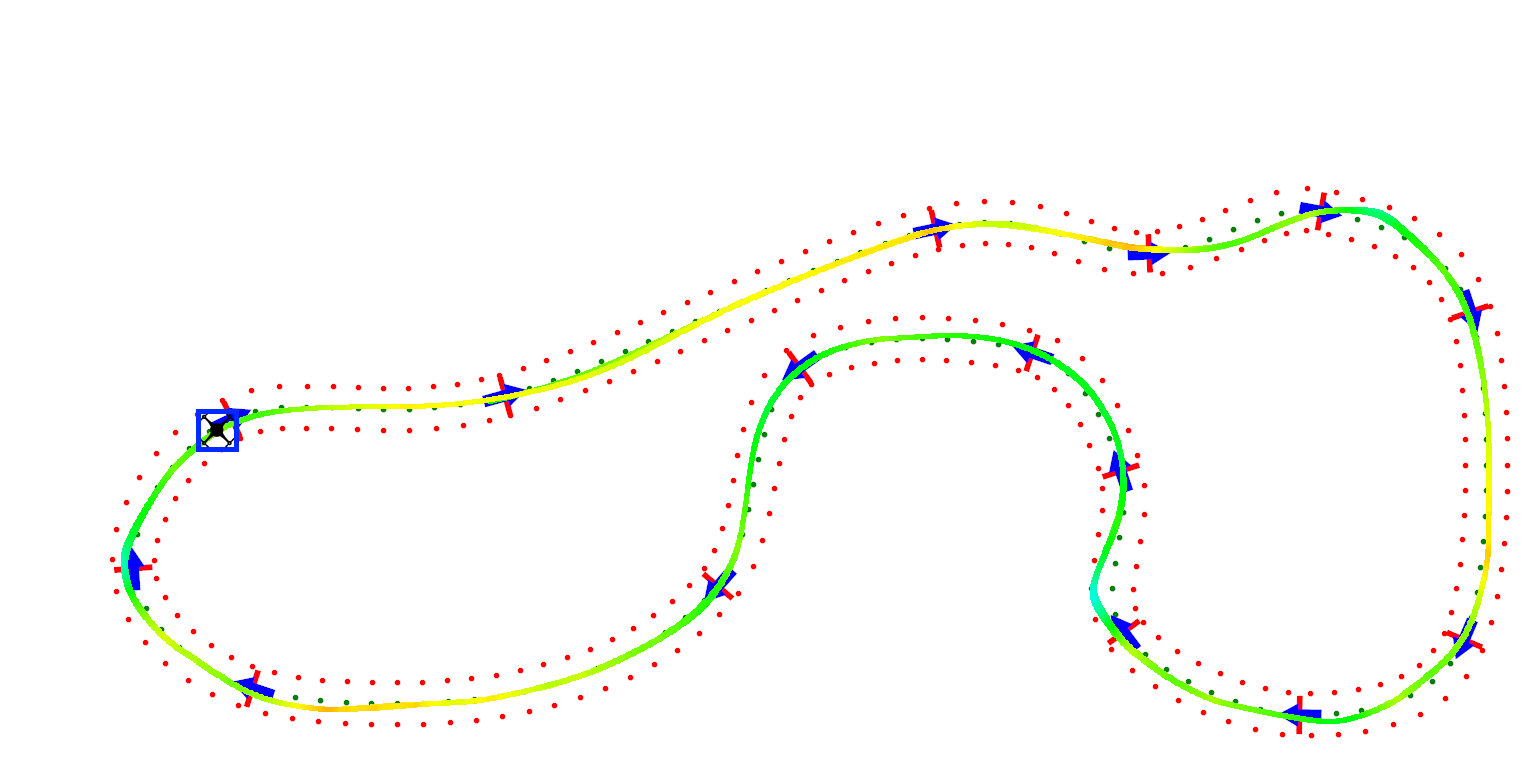} \\
		\small (p) Ours (Grass) & \small (q) Ours (Mud) & \small (r)  Ours (Snow) \\
        \multicolumn{3}{c}{\includegraphics[height=1.2cm]{sup_figures/ColorScaleHorizontal.png}}
\end{tabular}
\captionof{figure}{Qualitative results on track5. The color encodes speed as a heatmap, where blue corresponds to the minimum speed and red to the maximum speed.}
\label{fig:qualitive_results_track5}
\end{figure*}

\begin{figure*}
\centering
\begin{tabular}{@{}c@{\hspace{1mm}}c@{\hspace{1mm}}c@{\hspace{8mm}}c@{}}
		\includegraphics[height=3cm]{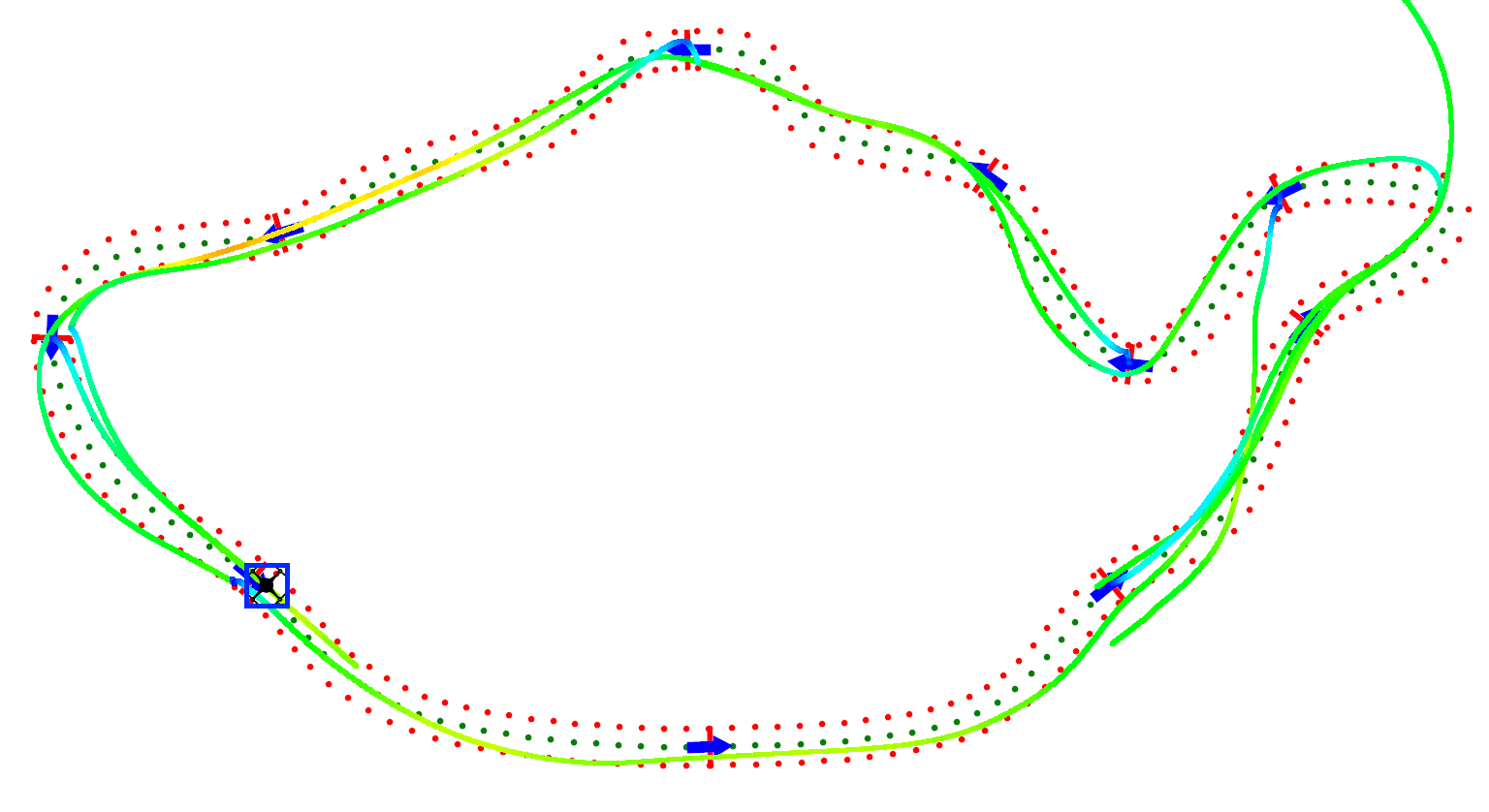} &
		\includegraphics[height=3cm]{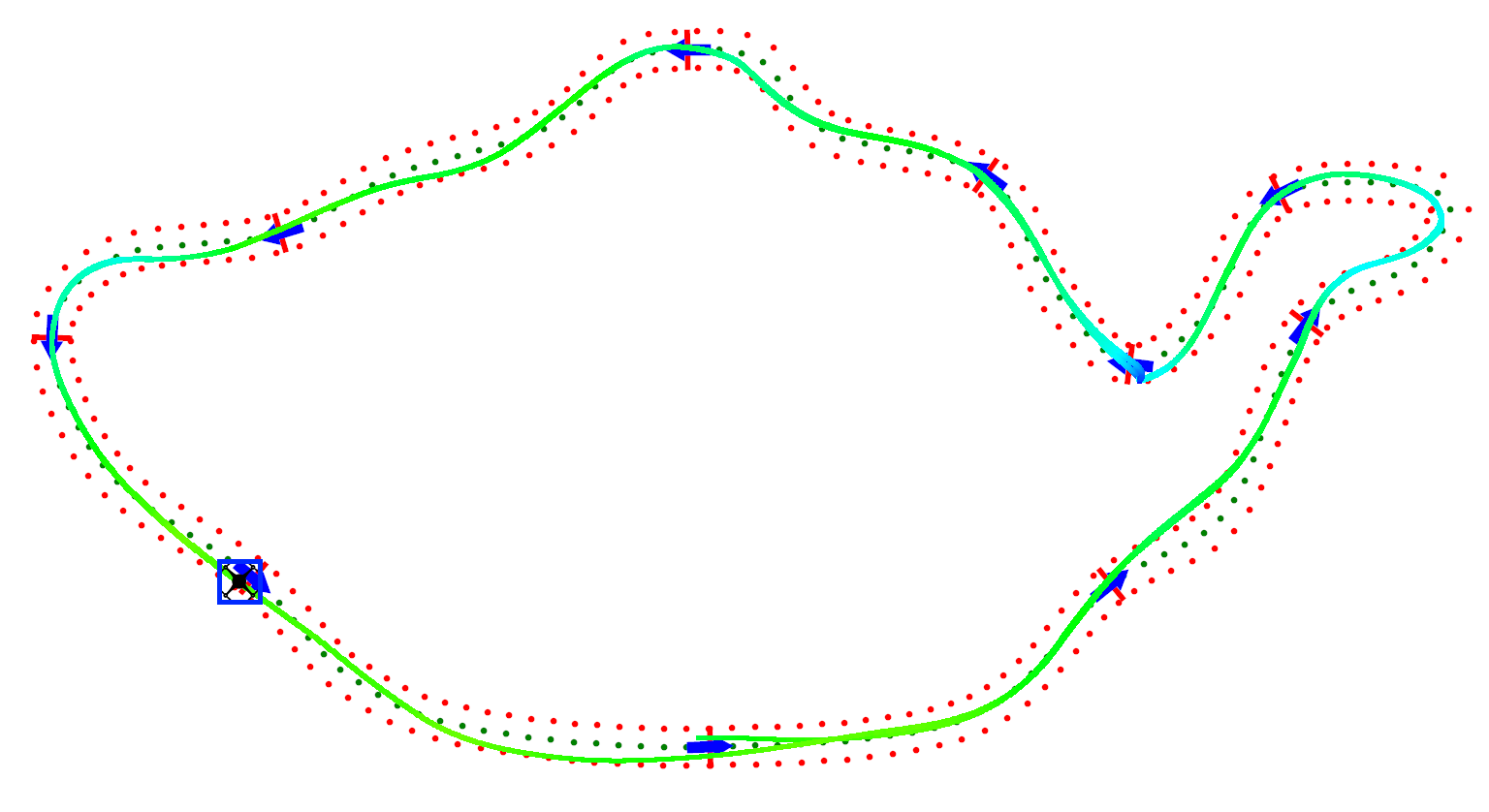} &
		\includegraphics[height=3cm]{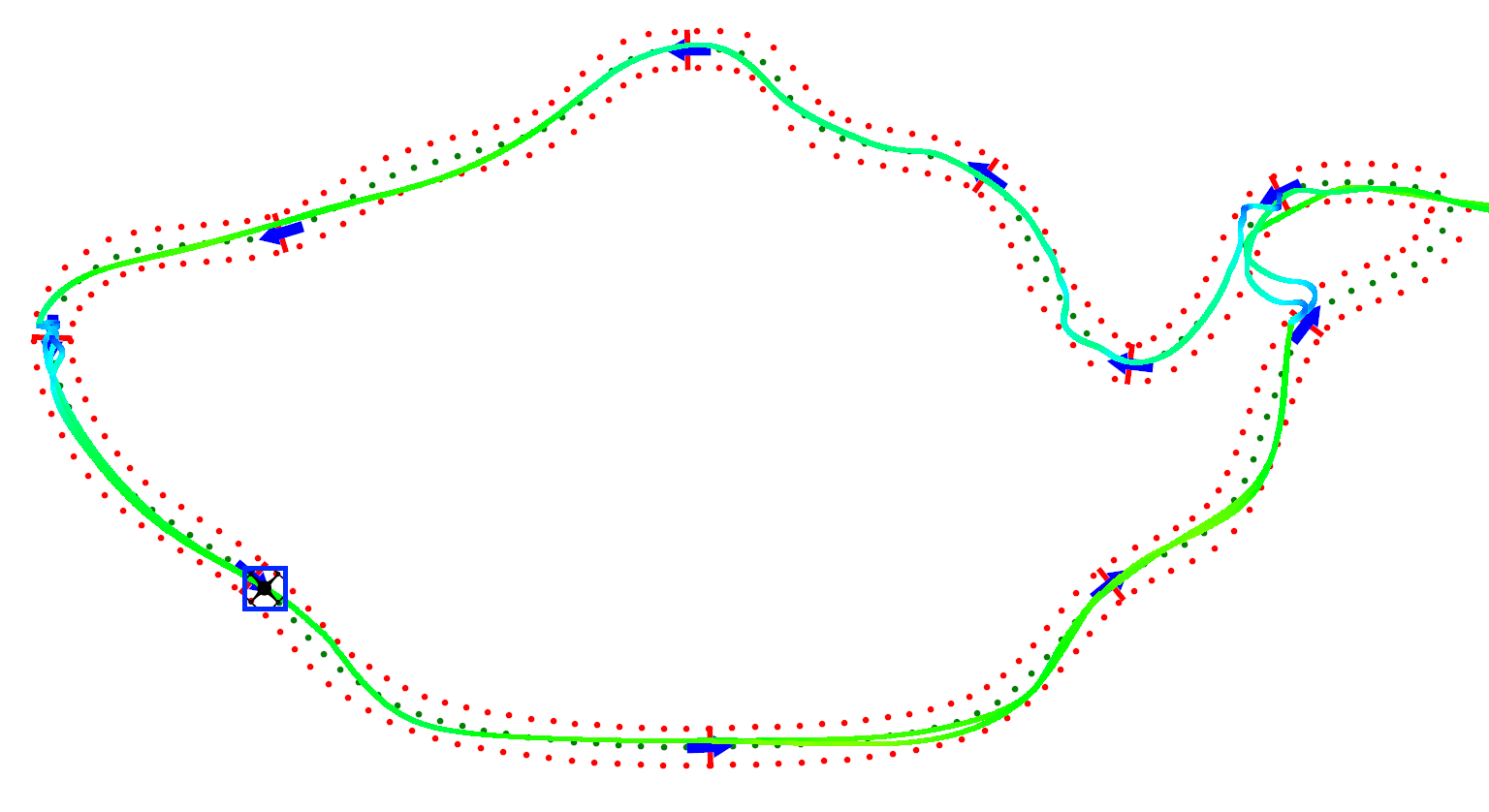} \\
		\small (a) End2End (MAV) & \small (b) End2End (Nvidia) & \small (c) End2End (Ours) \\
		\includegraphics[height=3cm]{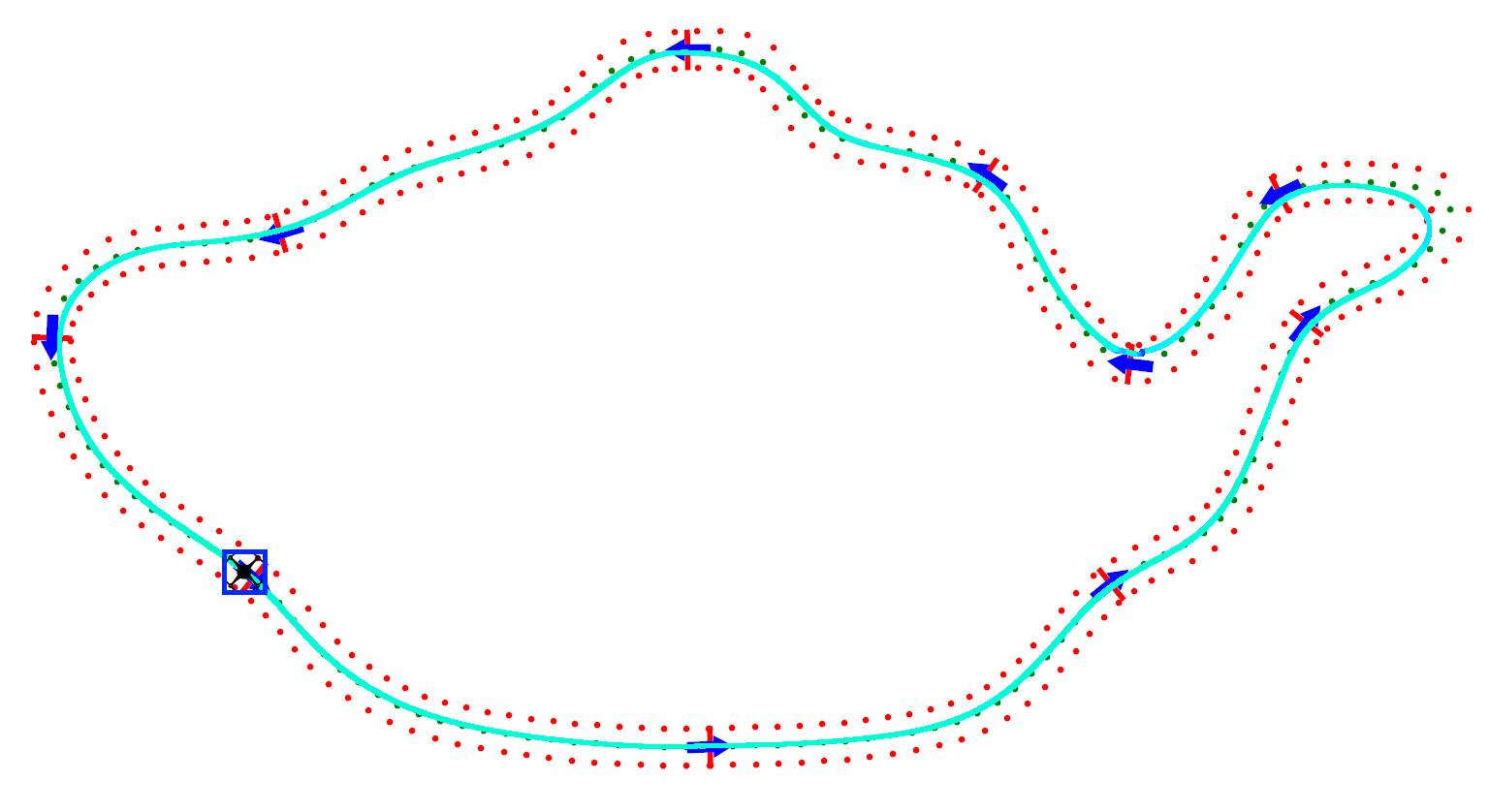} &
		\includegraphics[height=3cm]{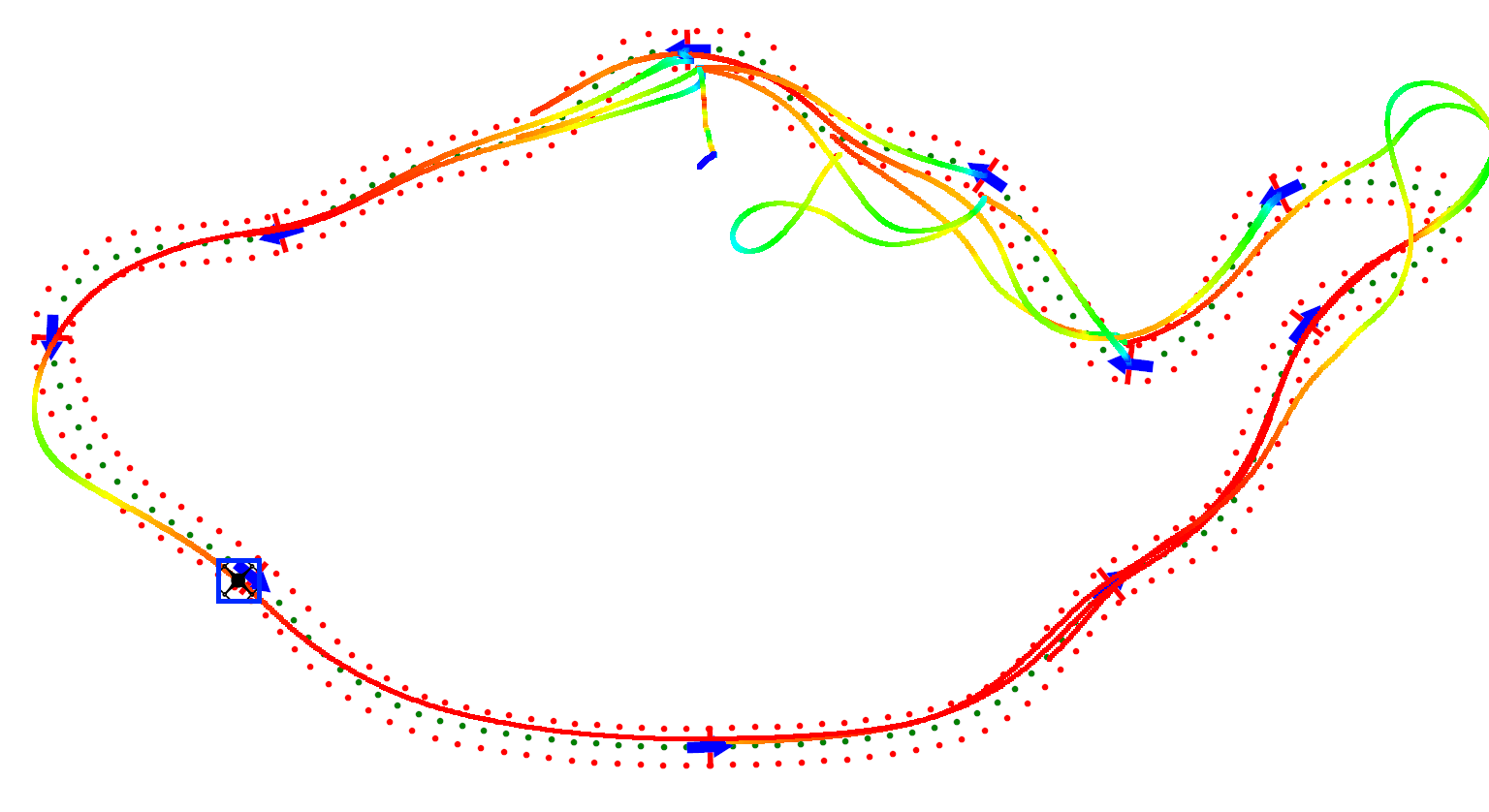} &
		\includegraphics[height=3cm]{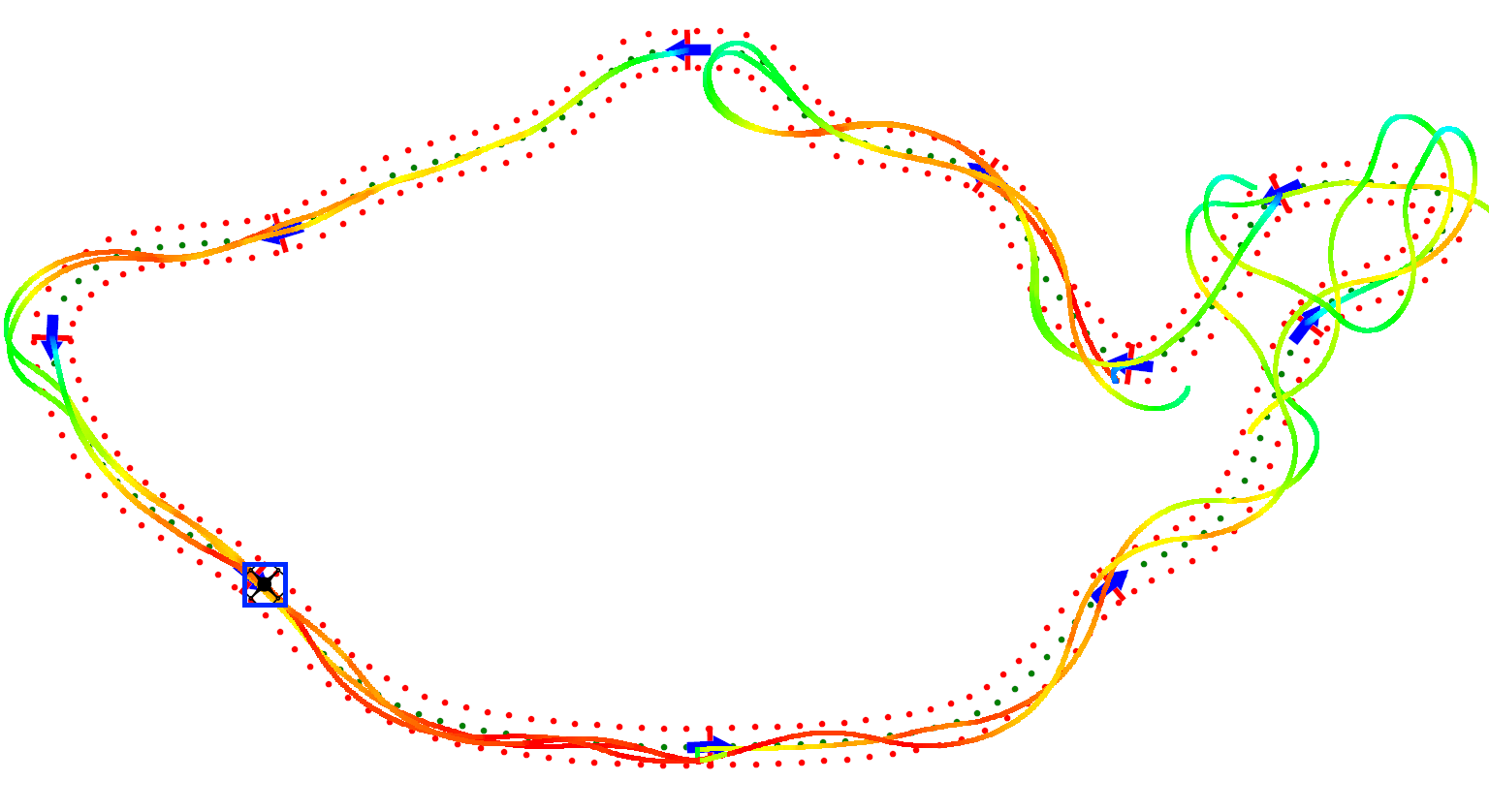} \\
		\small (c) PID1 (Conservative) & \small (d) PID2 (Aggressive) & \small (e) Ours (No Buffer) \\
		\includegraphics[height=3cm]{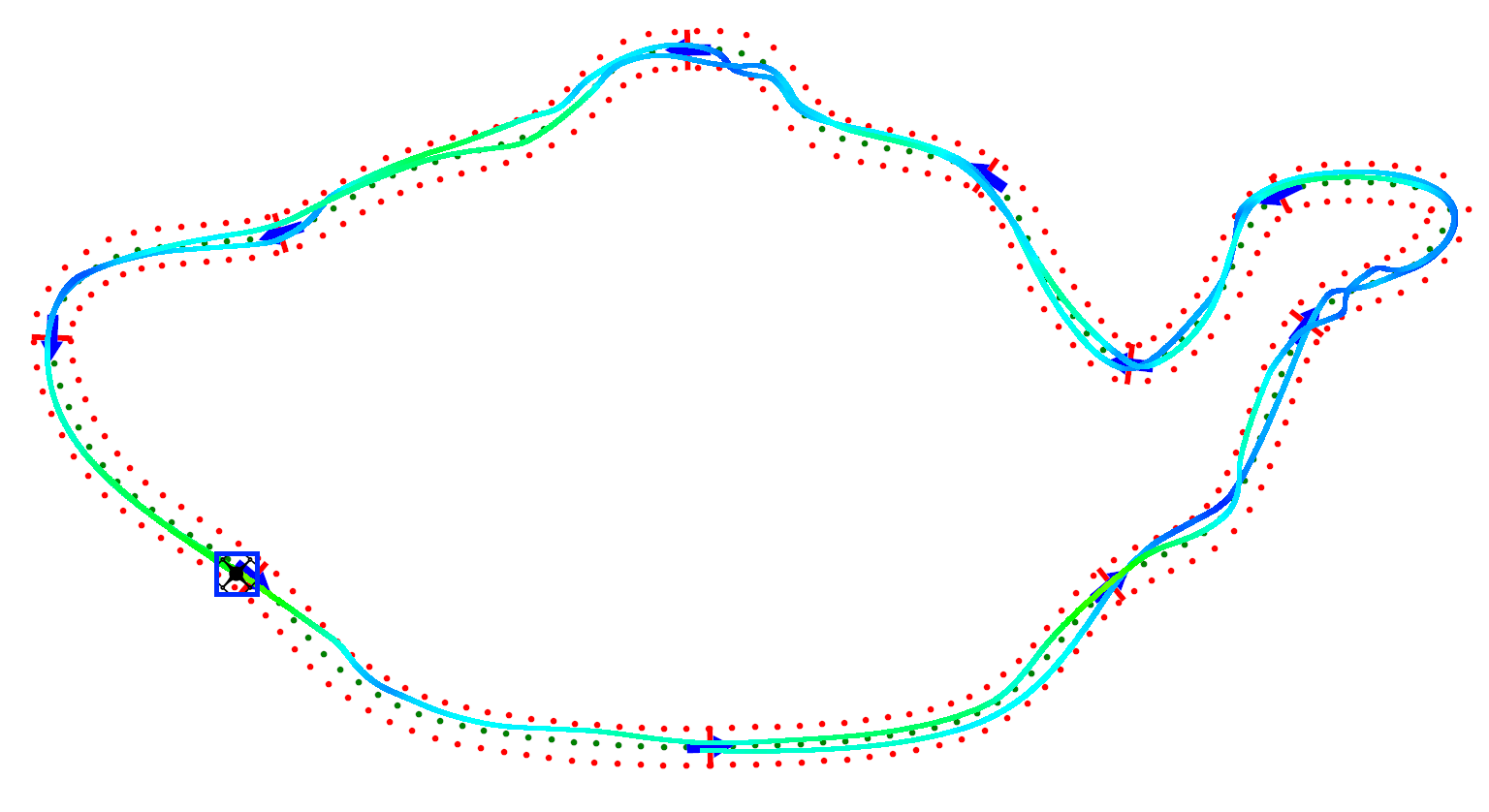} &
		\includegraphics[height=3cm]{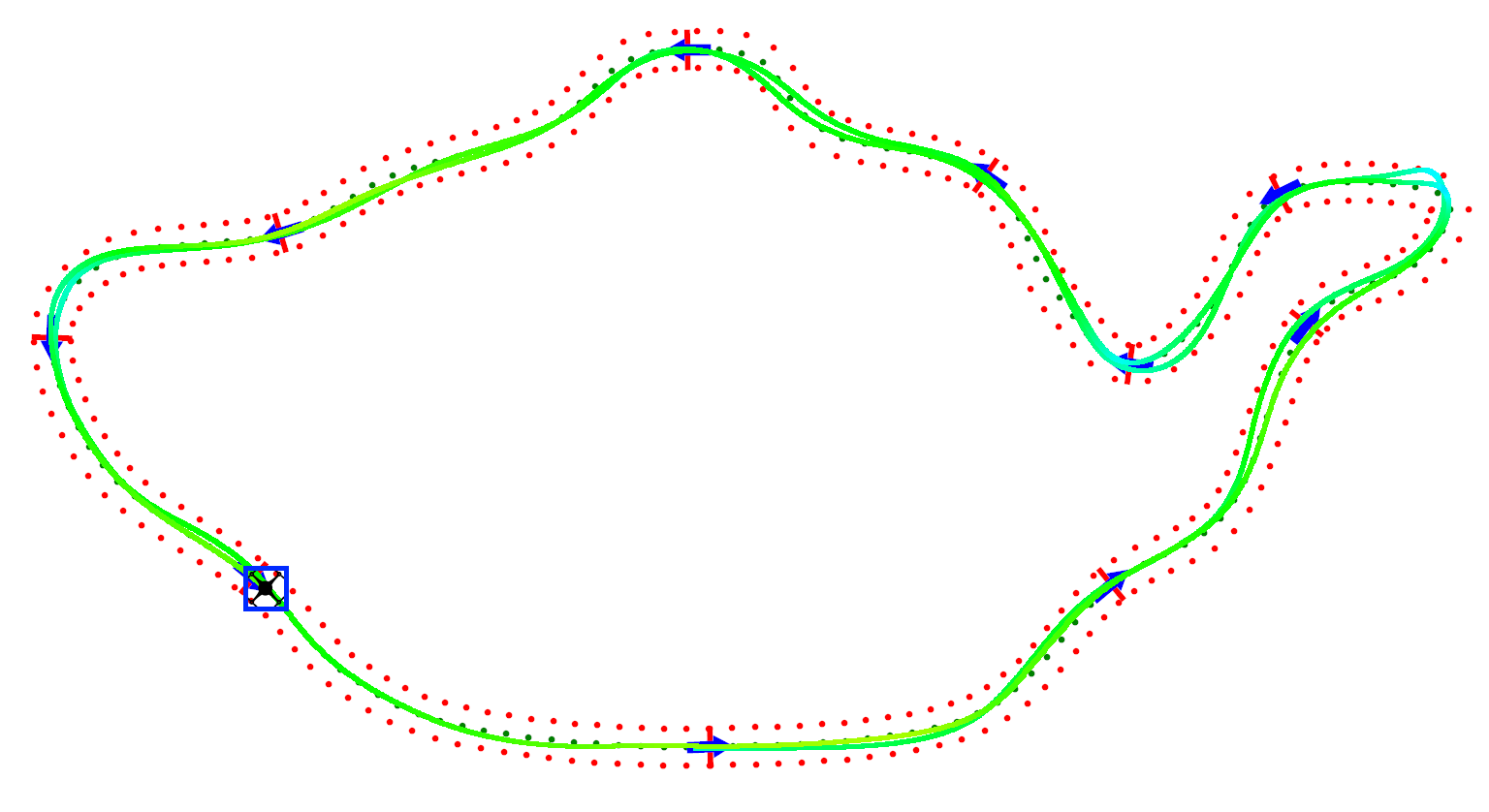} &
		\includegraphics[height=3cm]{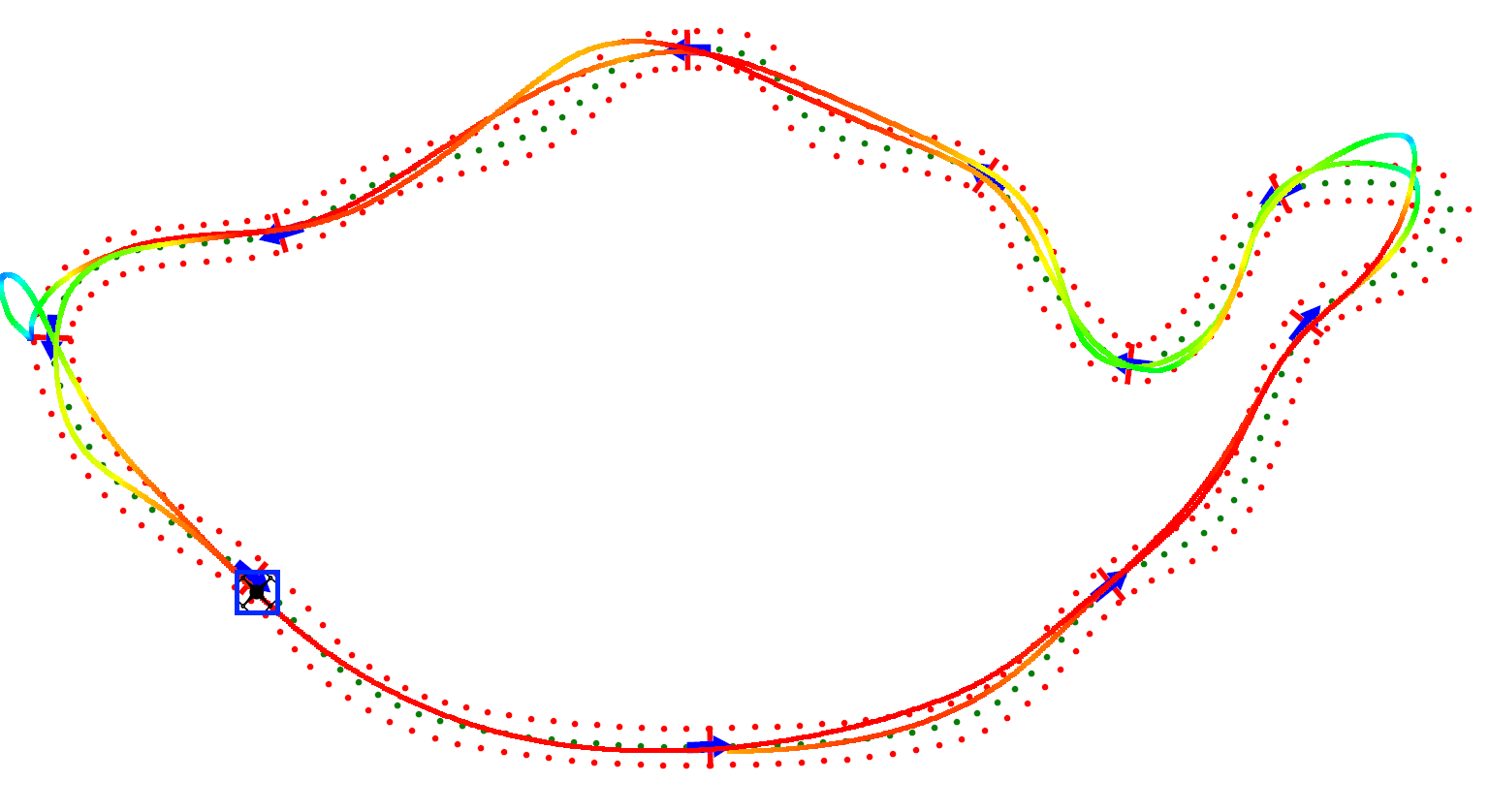} \\
        \small (f) Human (Novice) & \small (g) Human (Intermediate) & \small (h) Human (Professional)\\
		\includegraphics[height=3cm]{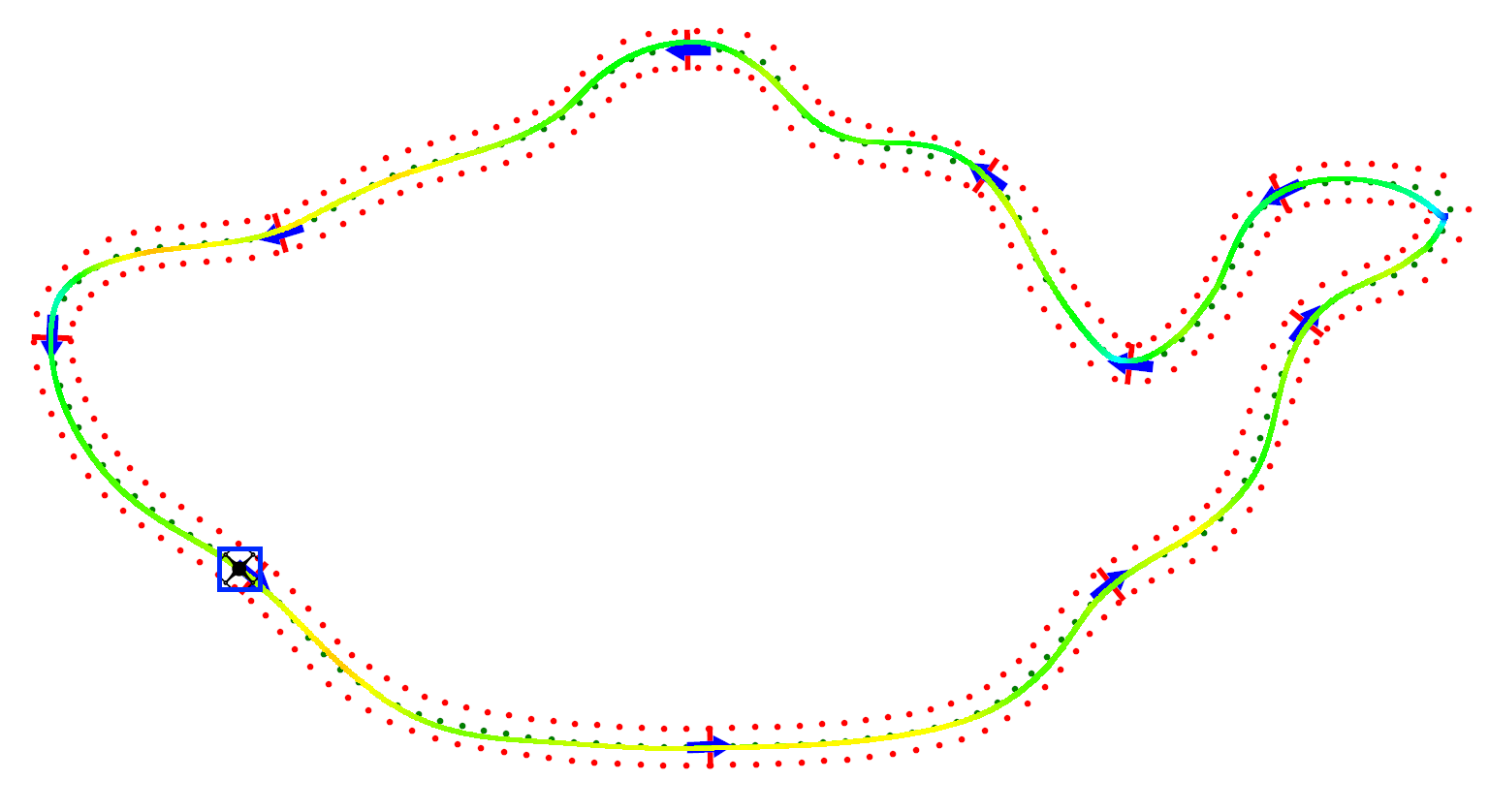} &
		\includegraphics[height=3cm]{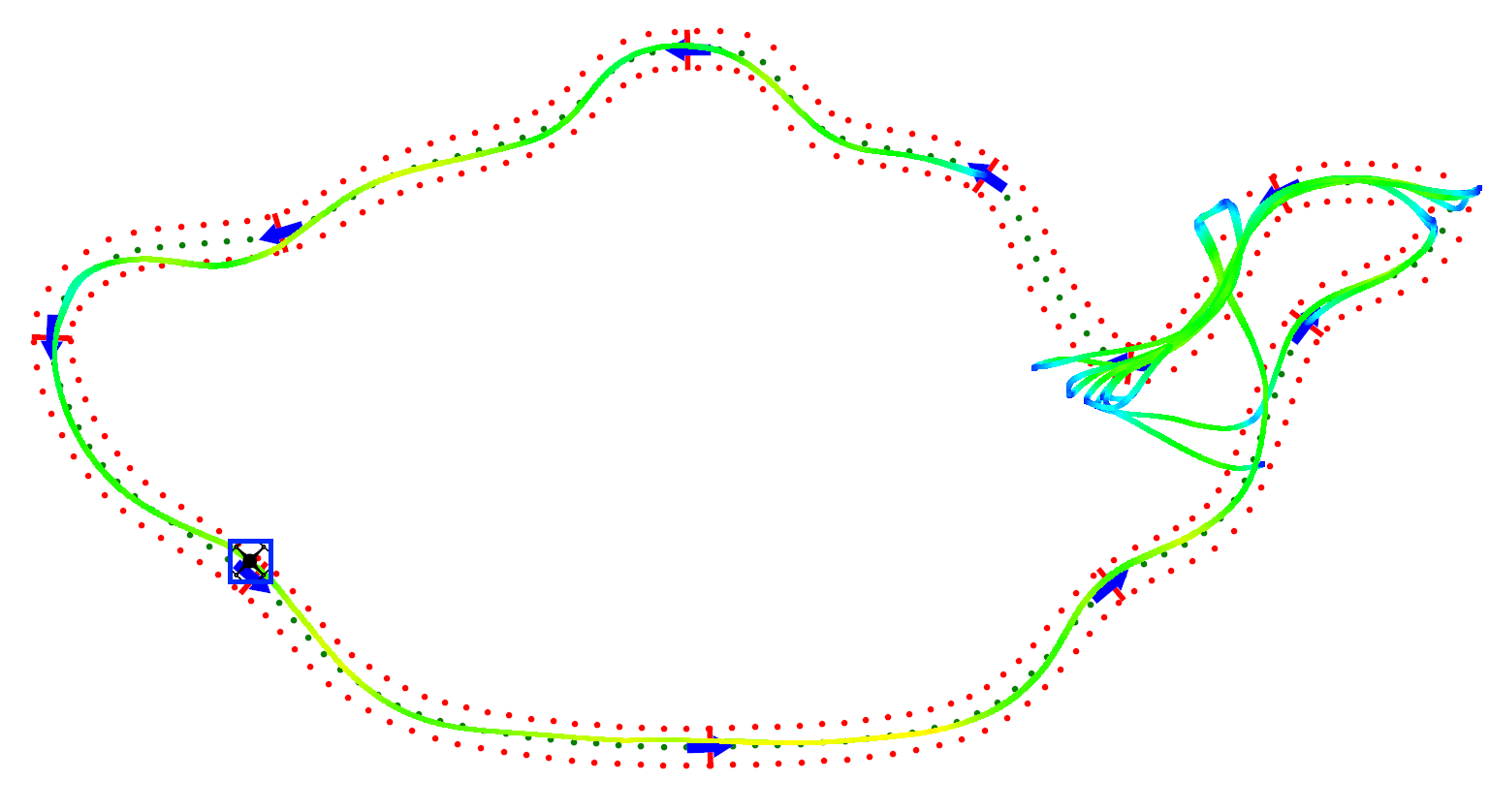} &
		\includegraphics[height=3cm]{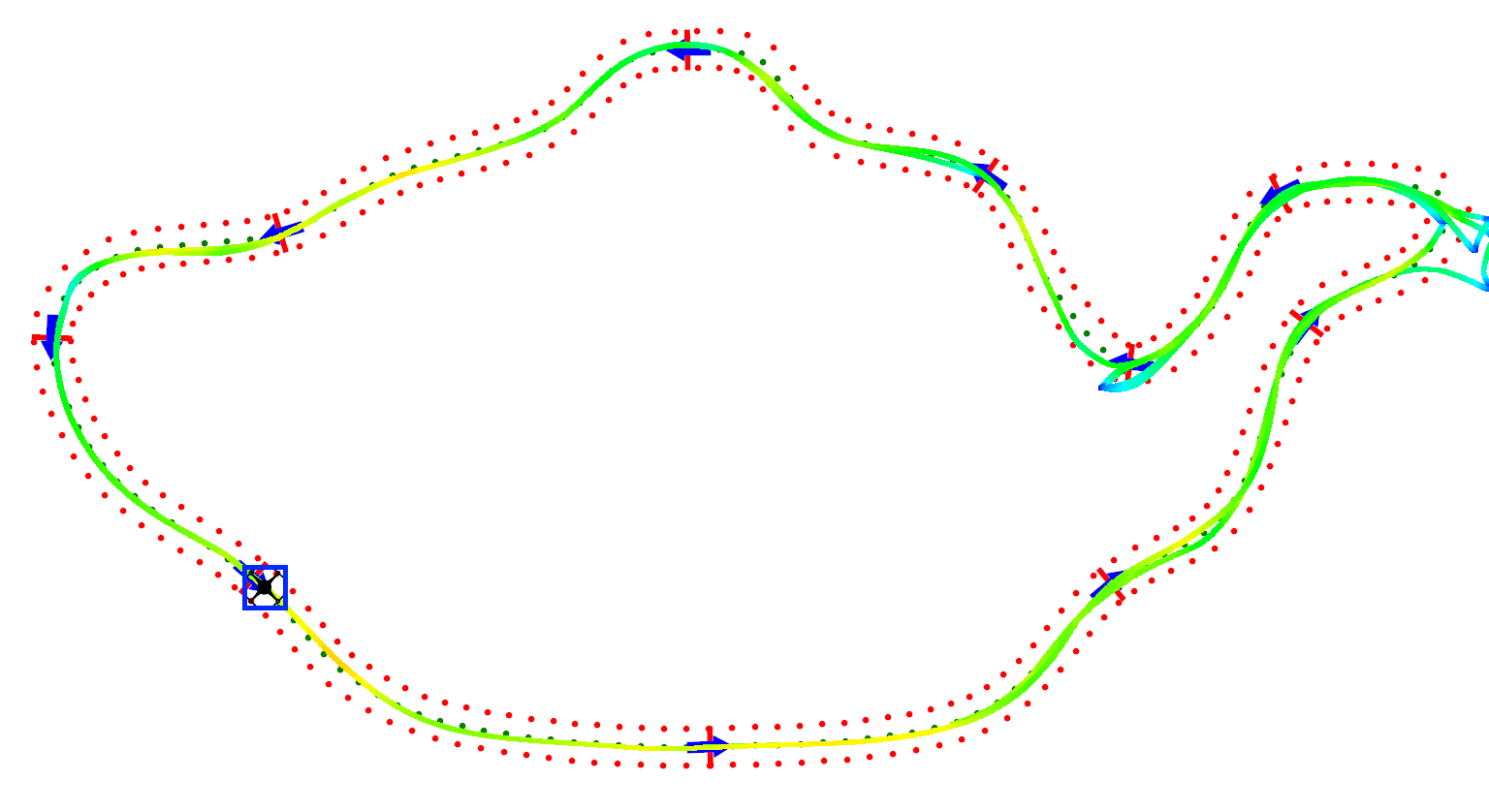} \\
		\small (i) Ours (Reference) & \small (j) Ours (Night) & \small (k)  Ours (Sunrise) \\
		\includegraphics[height=3cm]{sup_figures/track6_ours_grass.png} &
		\includegraphics[height=3cm]{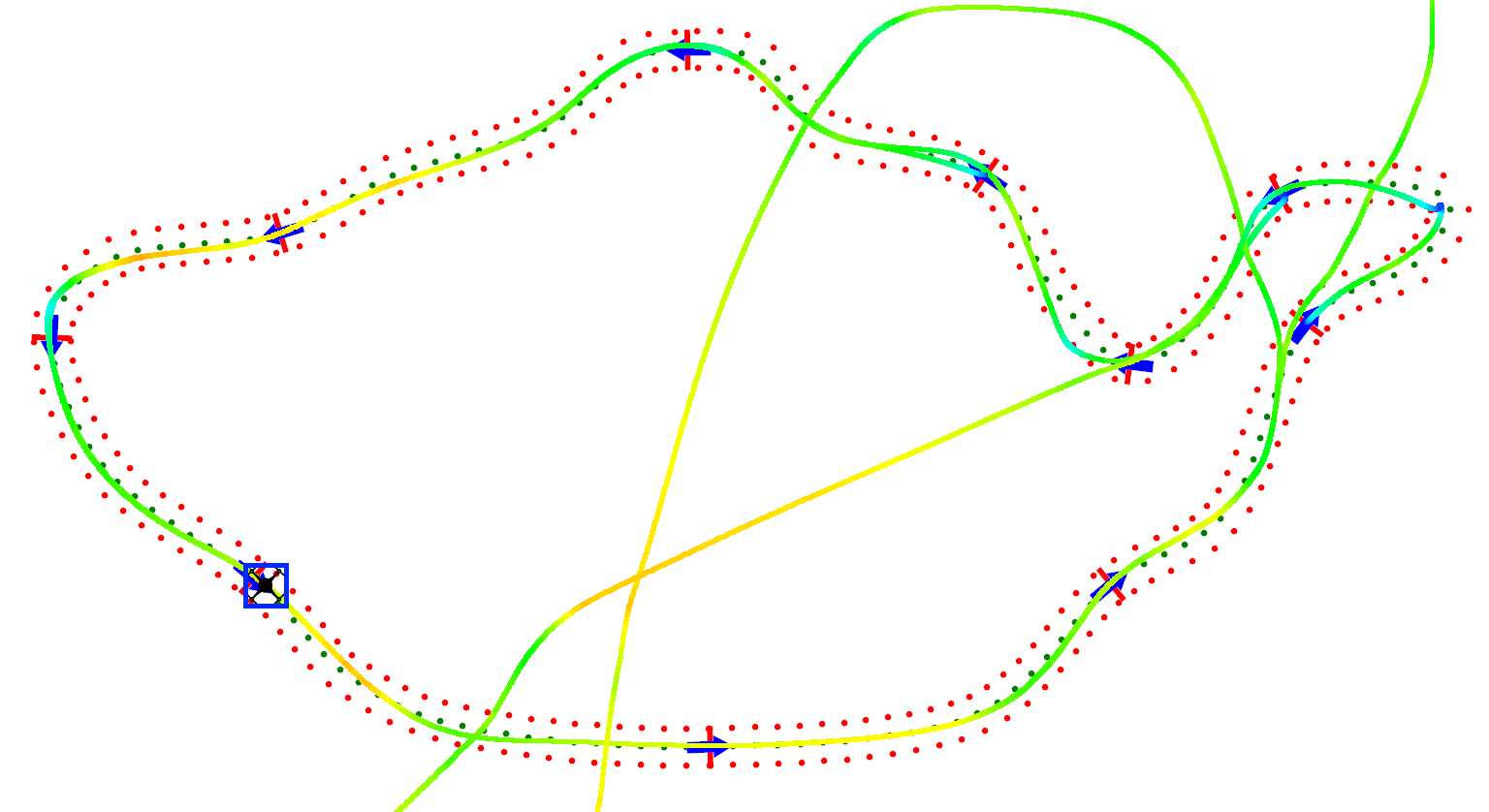} &
		\includegraphics[height=3cm]{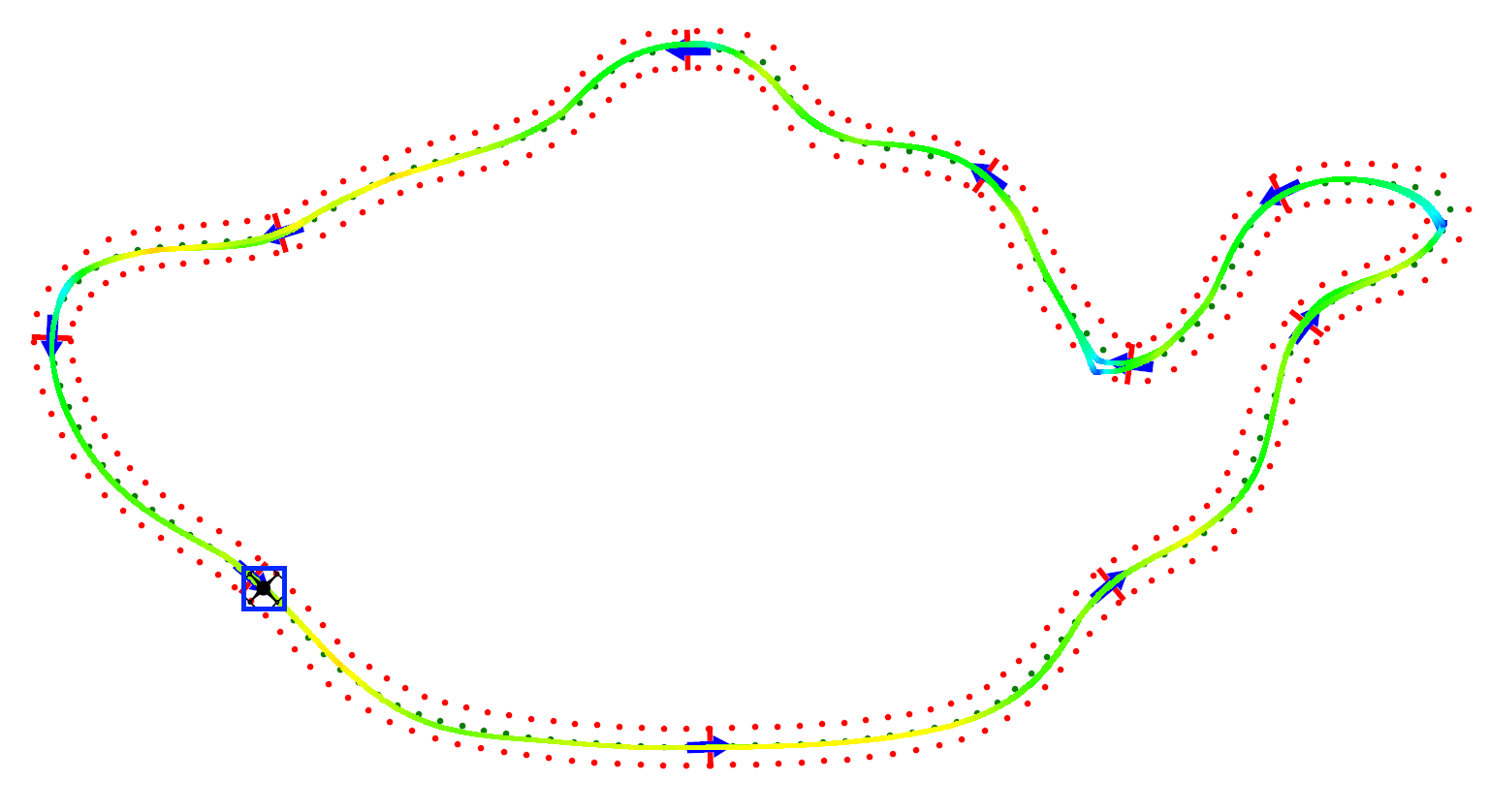} \\
		\small (l) Ours (Reference) & \small (m) Ours (Fog) & \small (o)  Ours (Rain) \\
		\includegraphics[height=3cm]{sup_figures/track6_ours_grass.png} &
		\includegraphics[height=3cm]{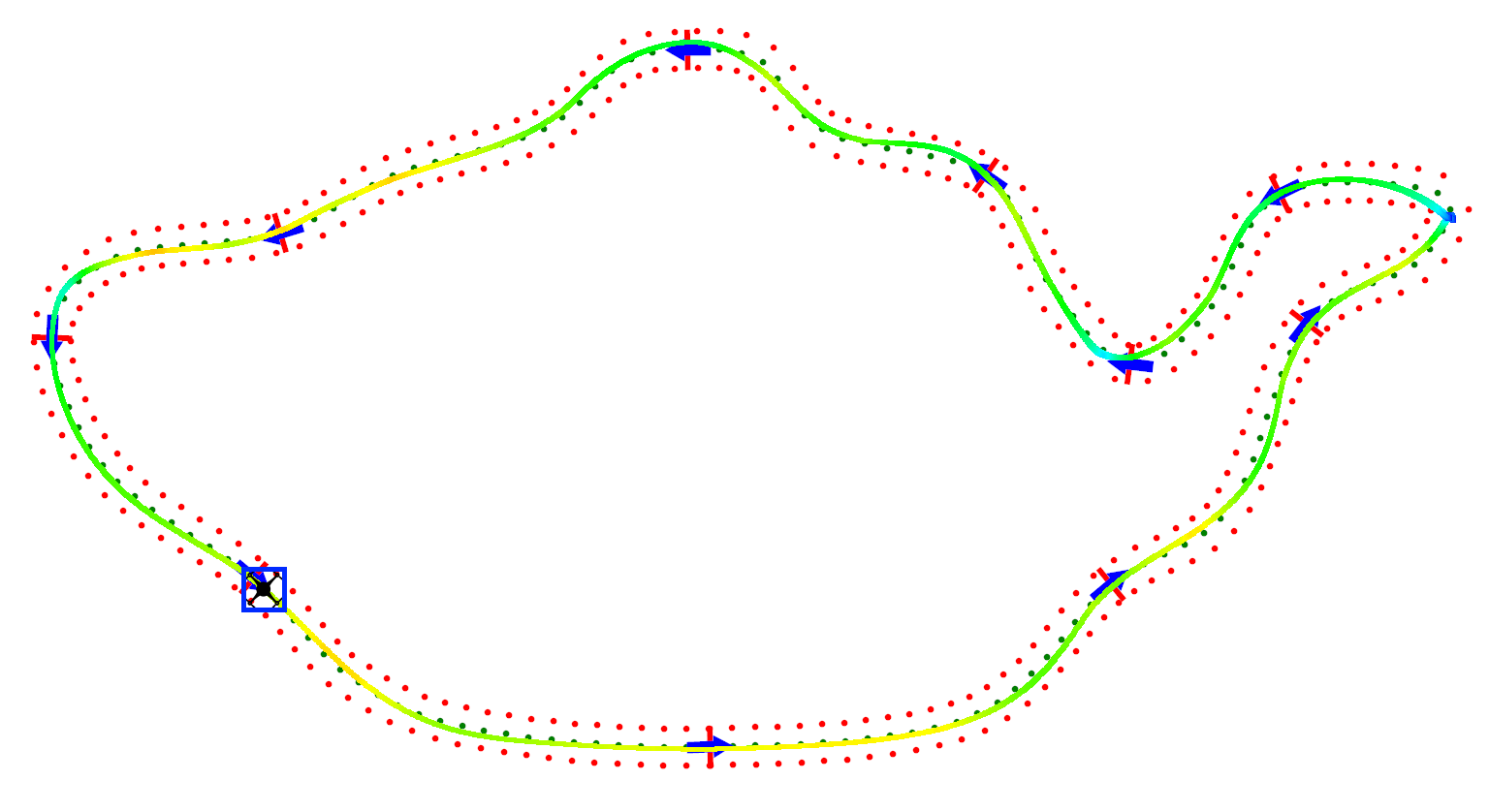} &
		\includegraphics[height=3cm]{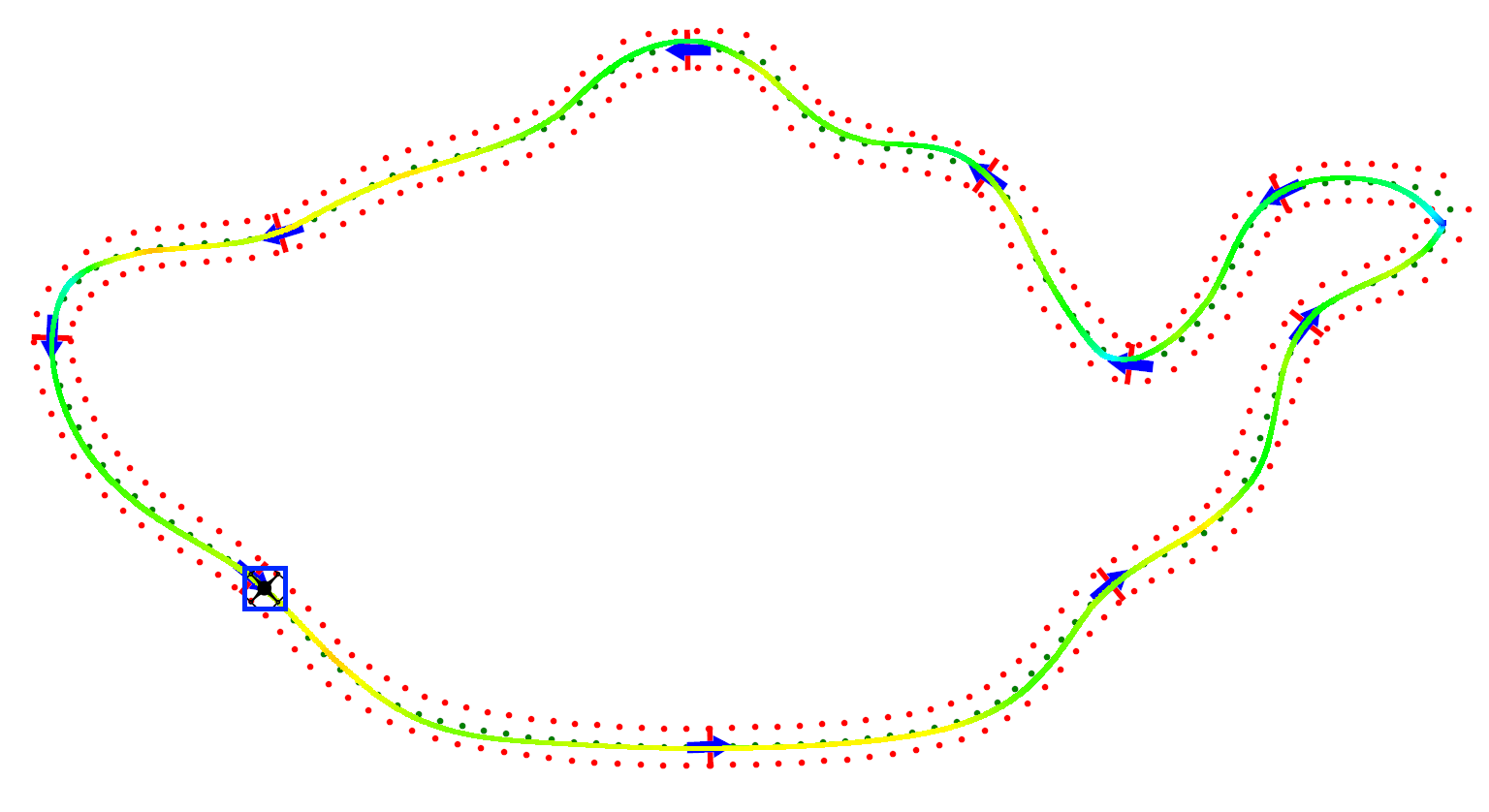} \\
		\small (p) Ours (Grass) & \small (q) Ours (Mud) & \small (r)  Ours (Snow) \\
        \multicolumn{3}{c}{\includegraphics[height=1.2cm]{sup_figures/ColorScaleHorizontal.png}}
\end{tabular}
\captionof{figure}{Qualitative results on track6. The color encodes speed as a heatmap, where blue corresponds to the minimum speed and red to the maximum speed.}
\label{fig:qualitive_results_track6}
\end{figure*}

\begin{figure*}
\centering
\begin{tabular}{@{}c@{\hspace{1mm}}c@{\hspace{1mm}}c@{\hspace{8mm}}c@{}}
		\includegraphics[height=3cm]{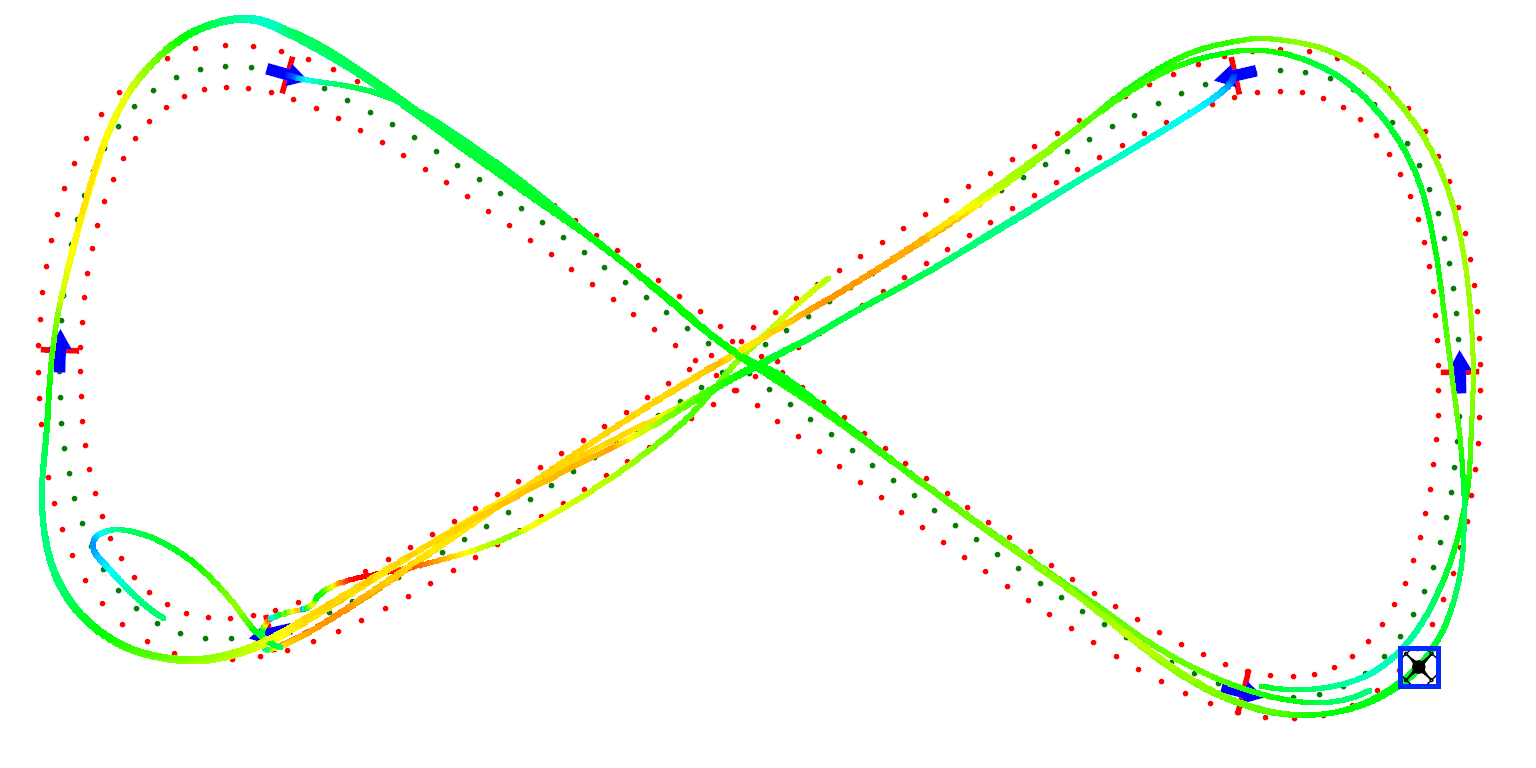} &
		\includegraphics[height=3cm]{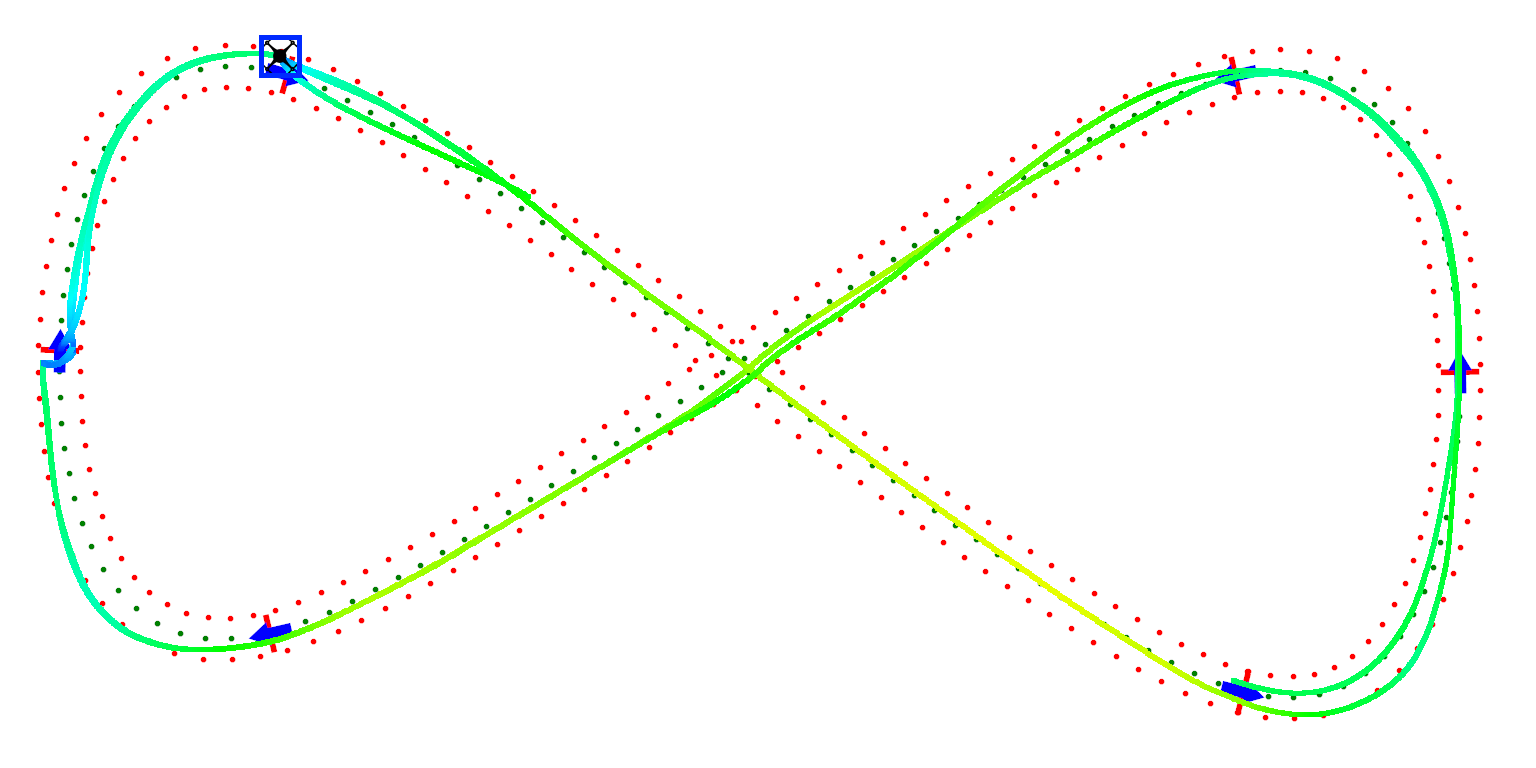} &
		\includegraphics[height=3cm]{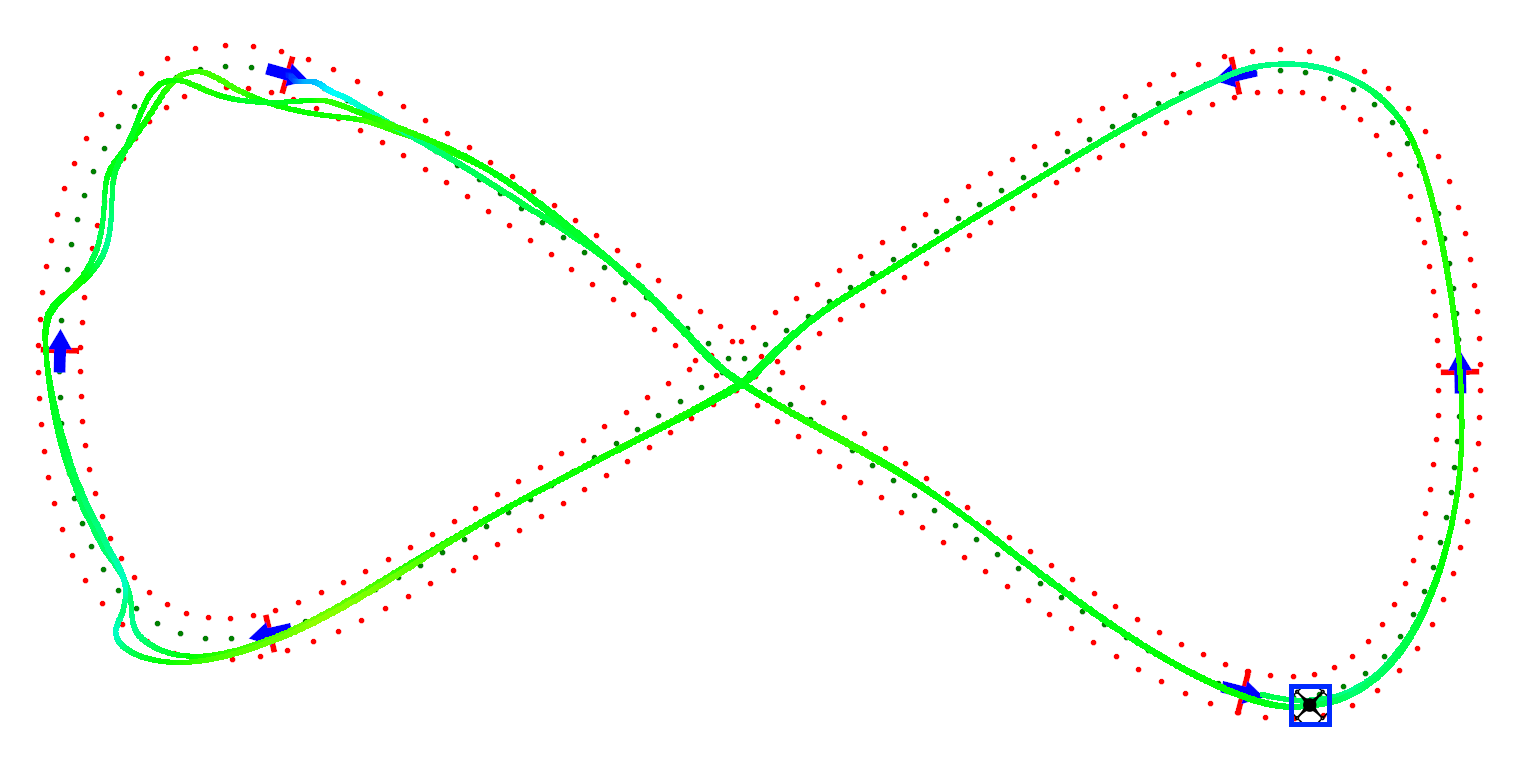} \\
		\small (a) End2End (MAV) & \small (b) End2End (Nvidia) & \small (c) End2End (Ours) \\
		\includegraphics[height=3cm]{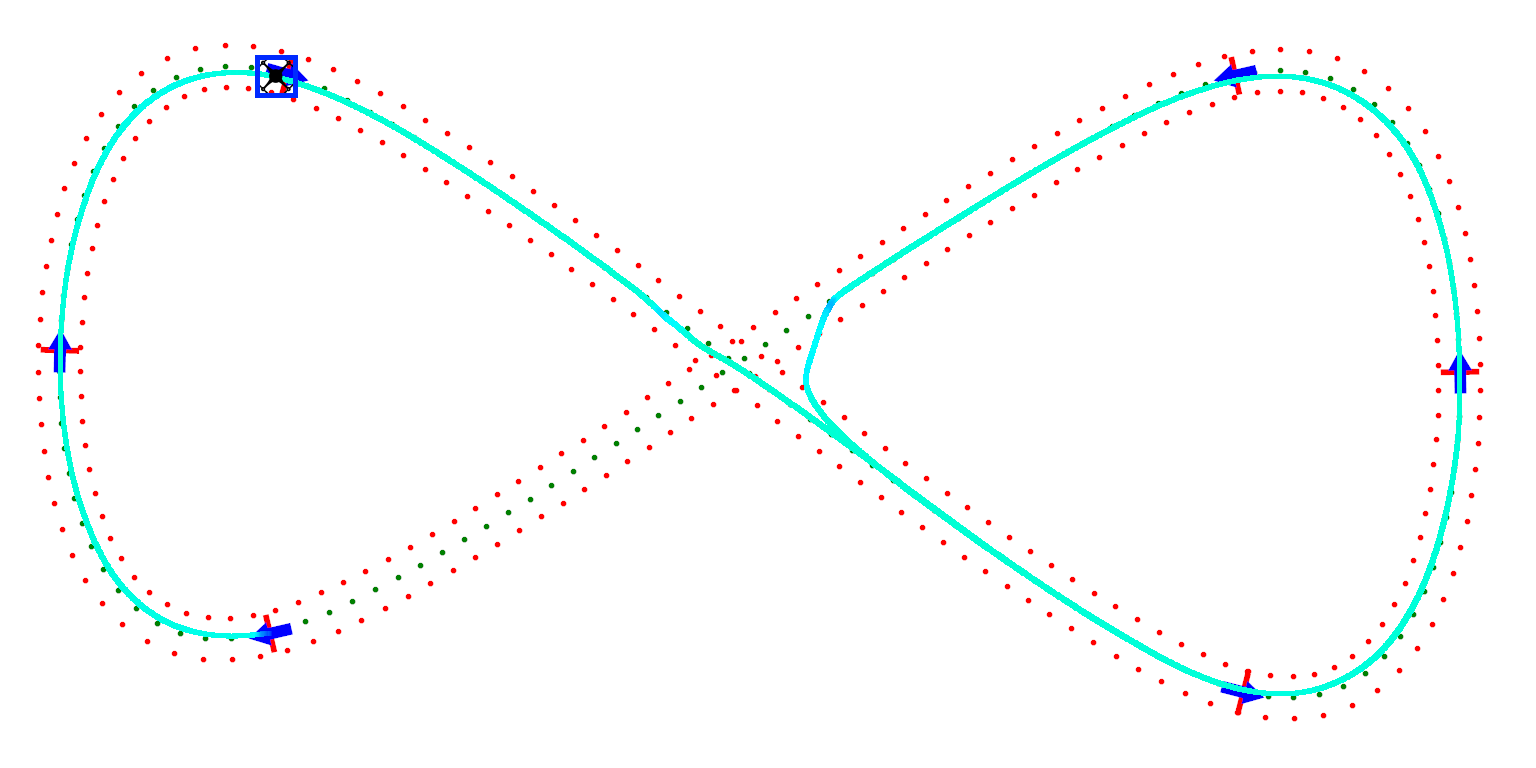} &
		\includegraphics[height=3cm]{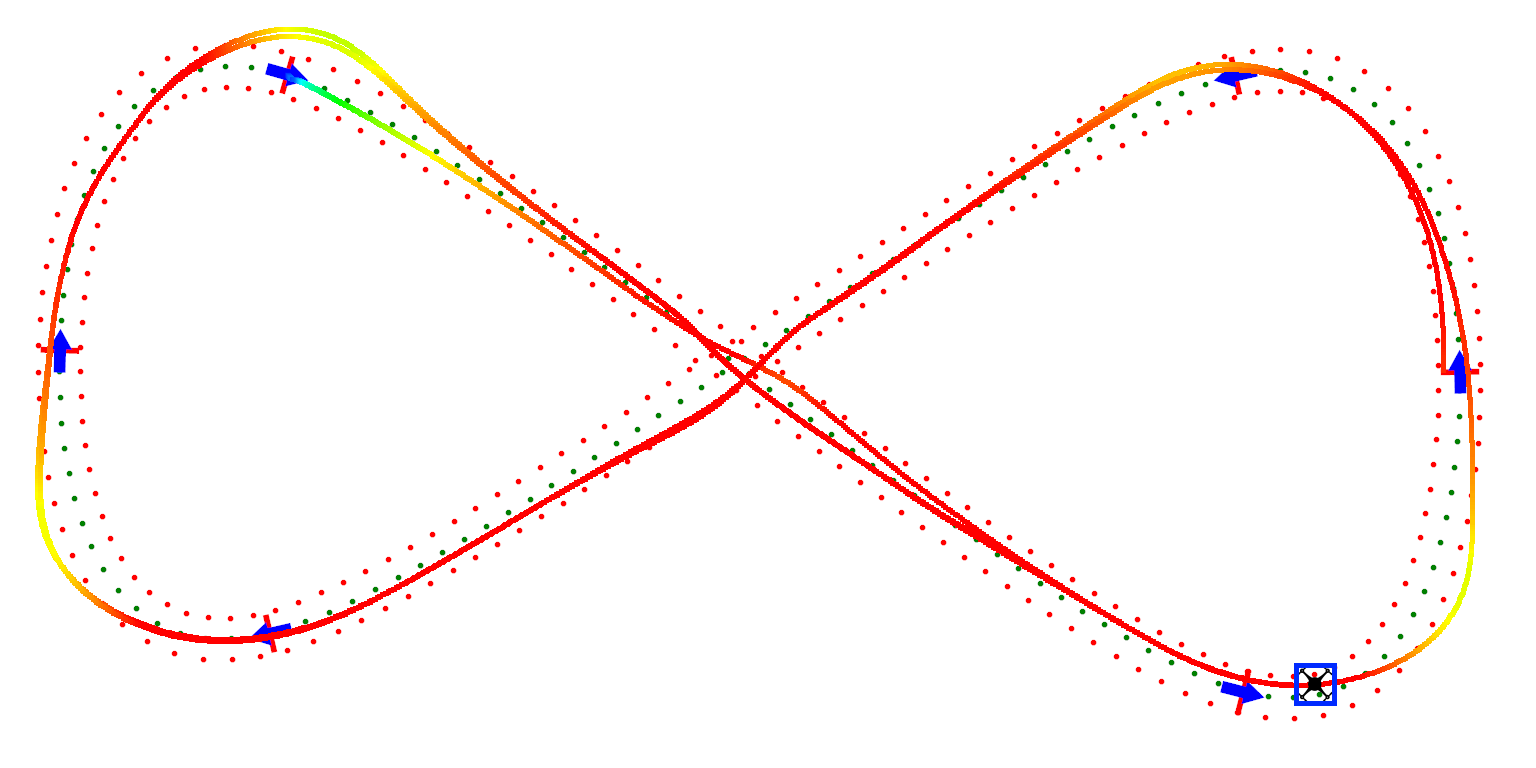} &
		\includegraphics[height=3cm]{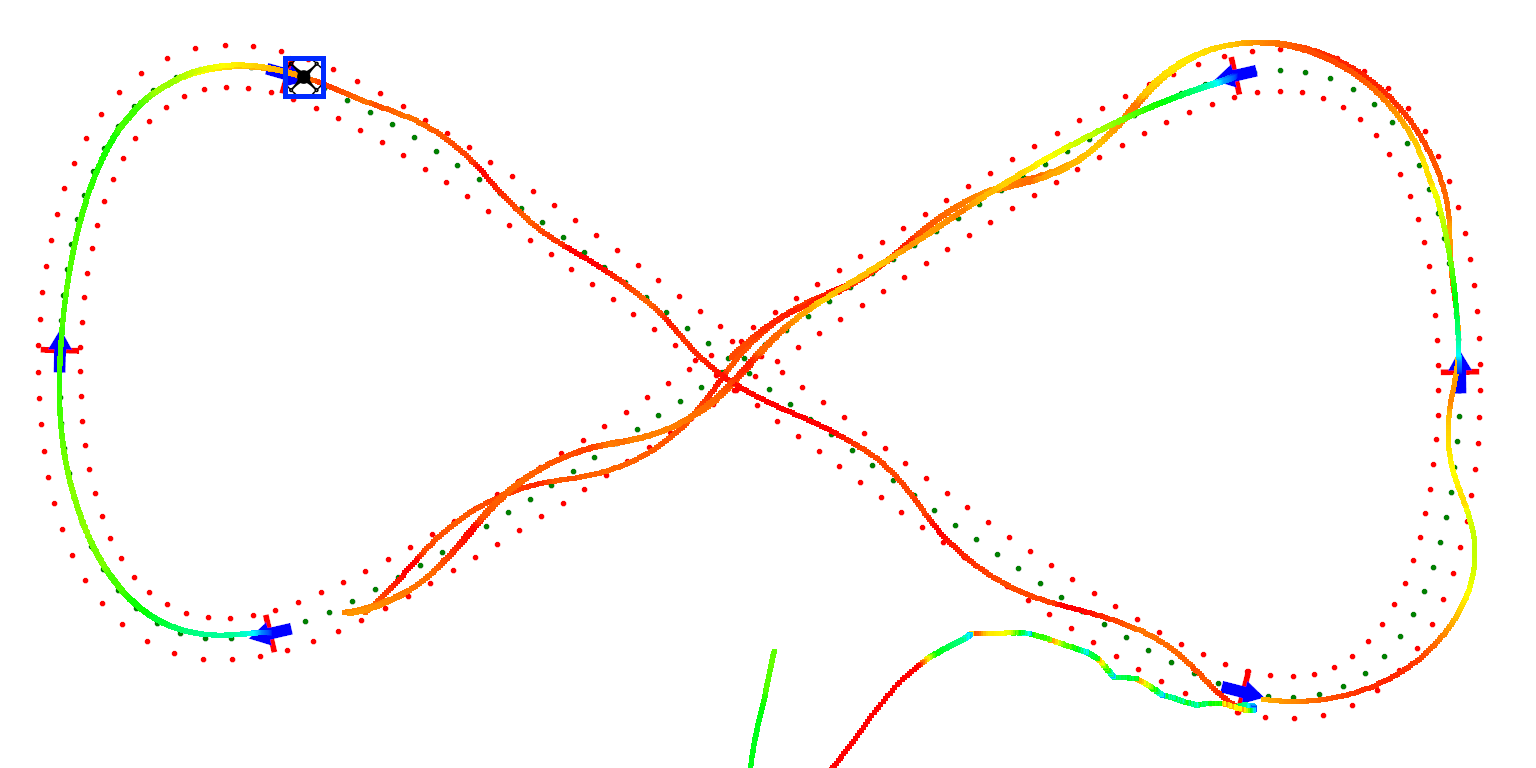} \\
		\small (c) PID1 (Conservative) & \small (d) PID2 (Aggressive) & \small (e) Ours (No Buffer) \\
		\includegraphics[height=3cm]{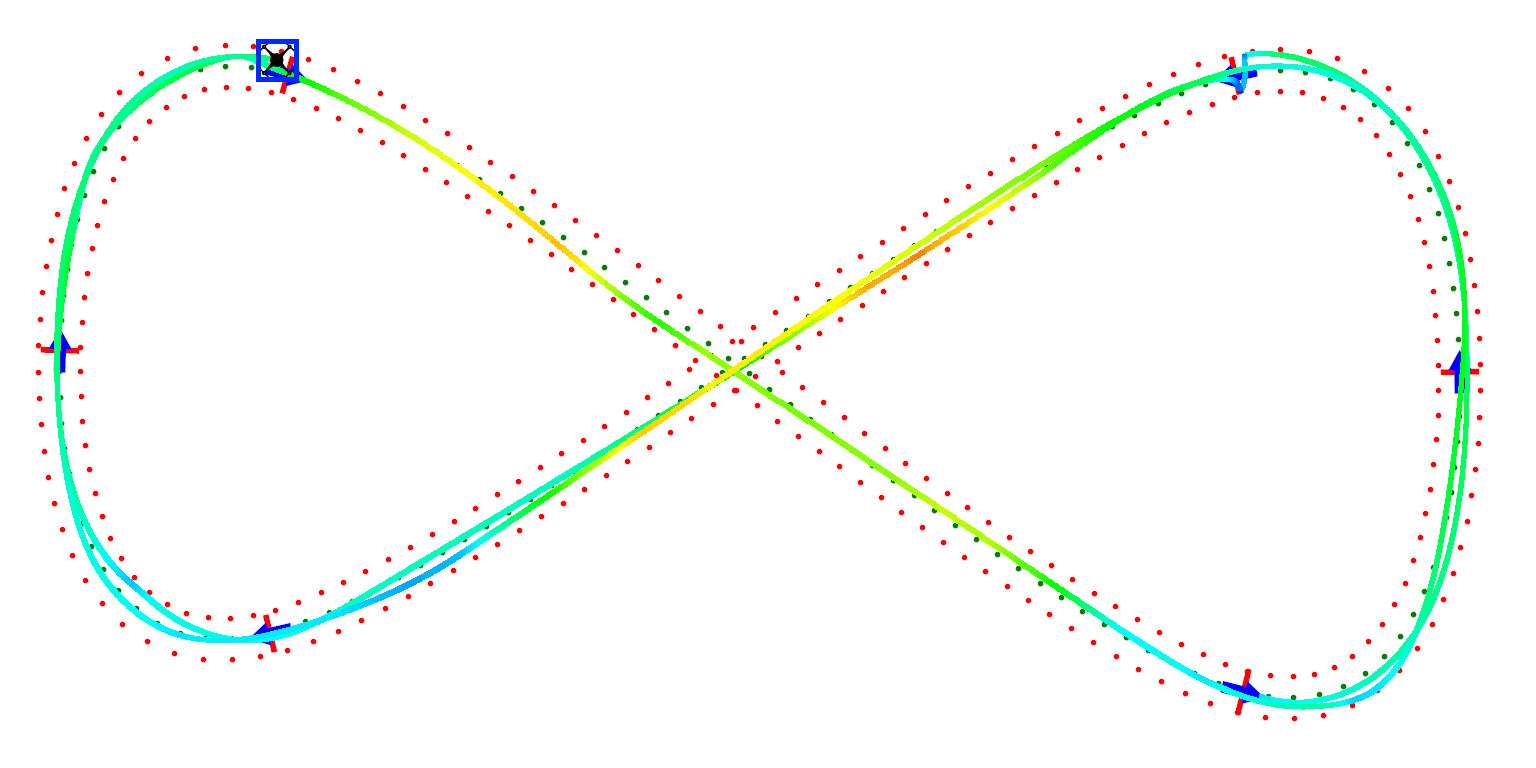} &
		\includegraphics[height=3cm]{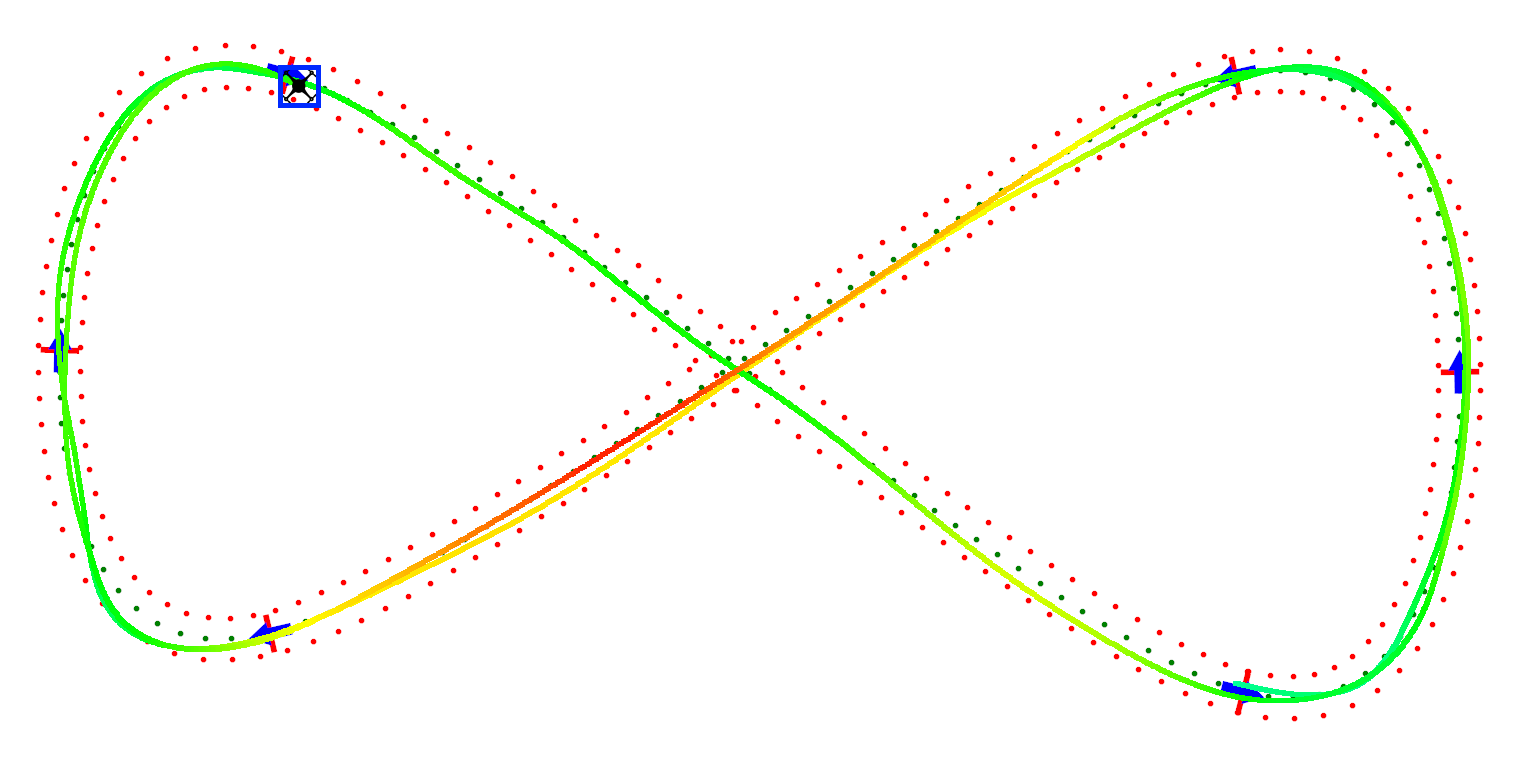} &
		\includegraphics[height=3cm]{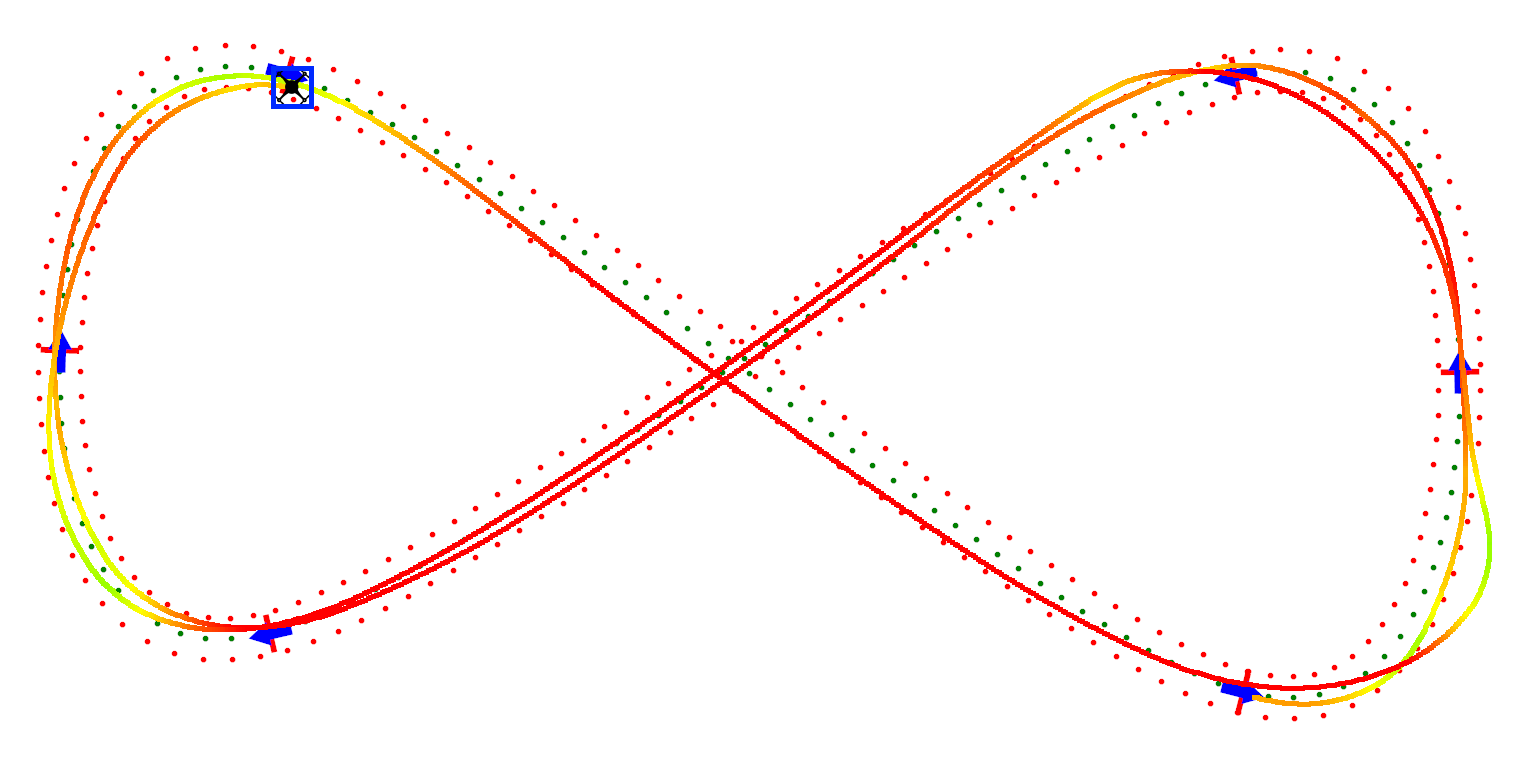} \\
        \small (f) Human (Novice) & \small (g) Human (Intermediate) & \small (h) Human (Professional)\\
		\includegraphics[height=3cm]{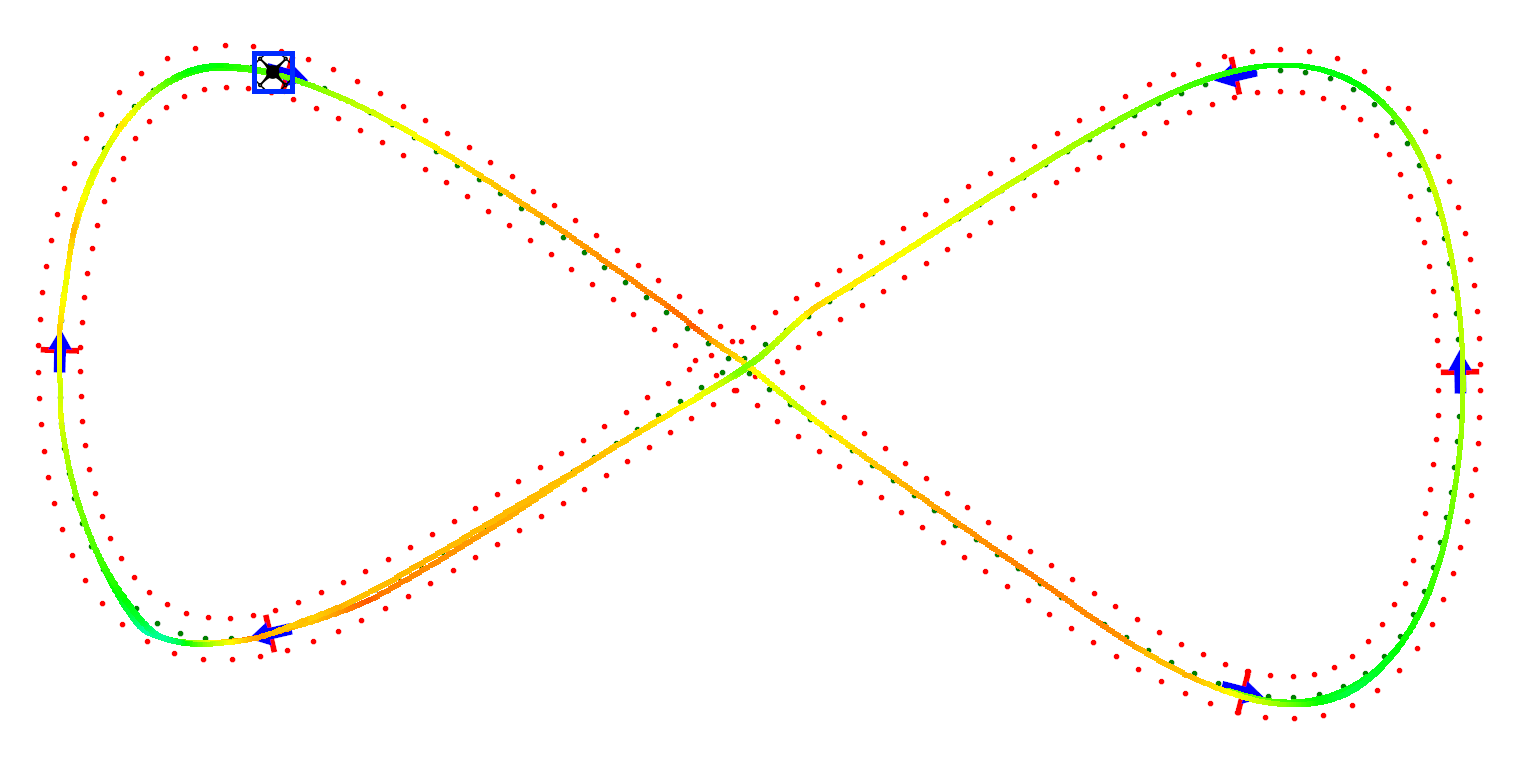} &
		\includegraphics[height=3cm]{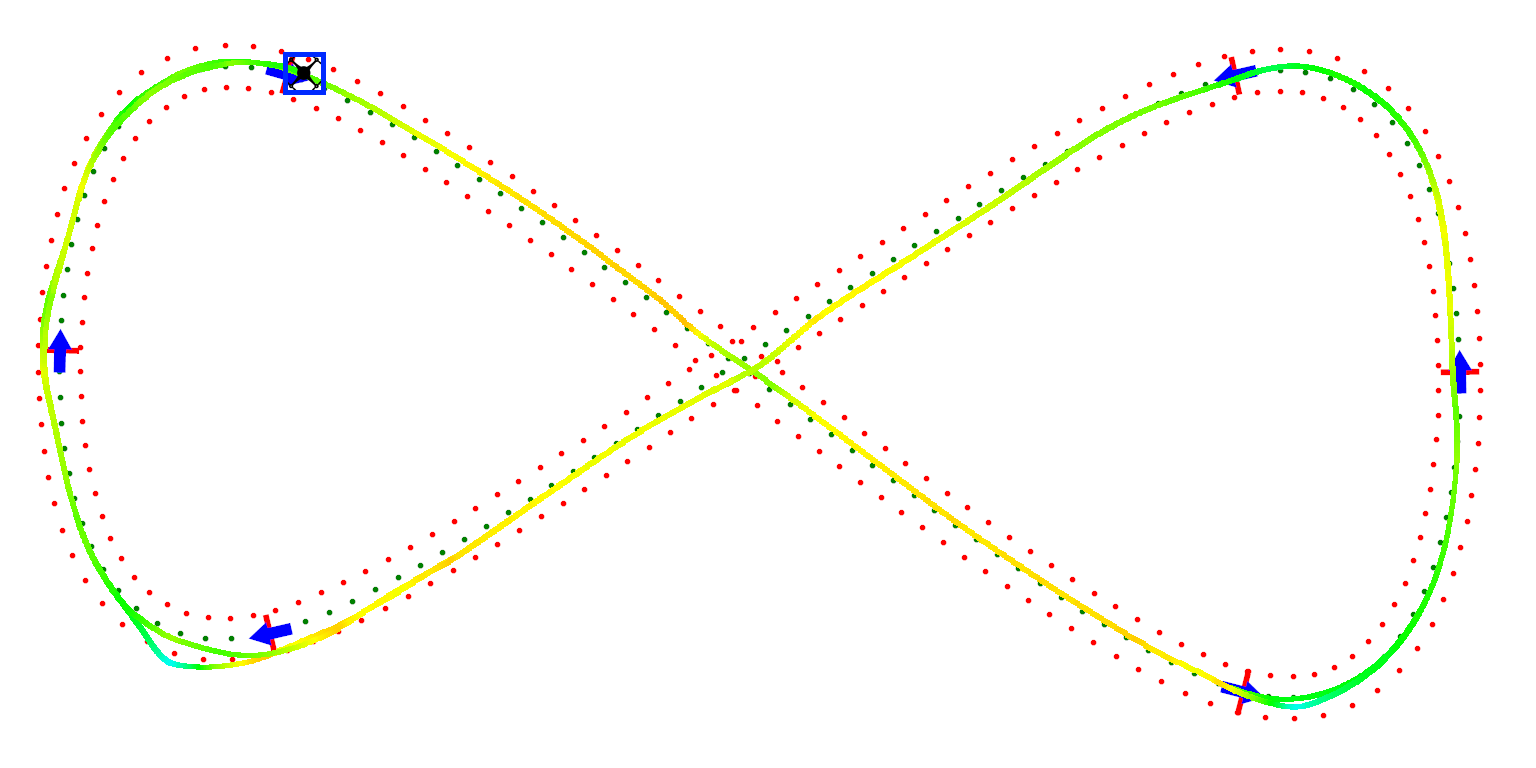} &
		\includegraphics[height=3cm]{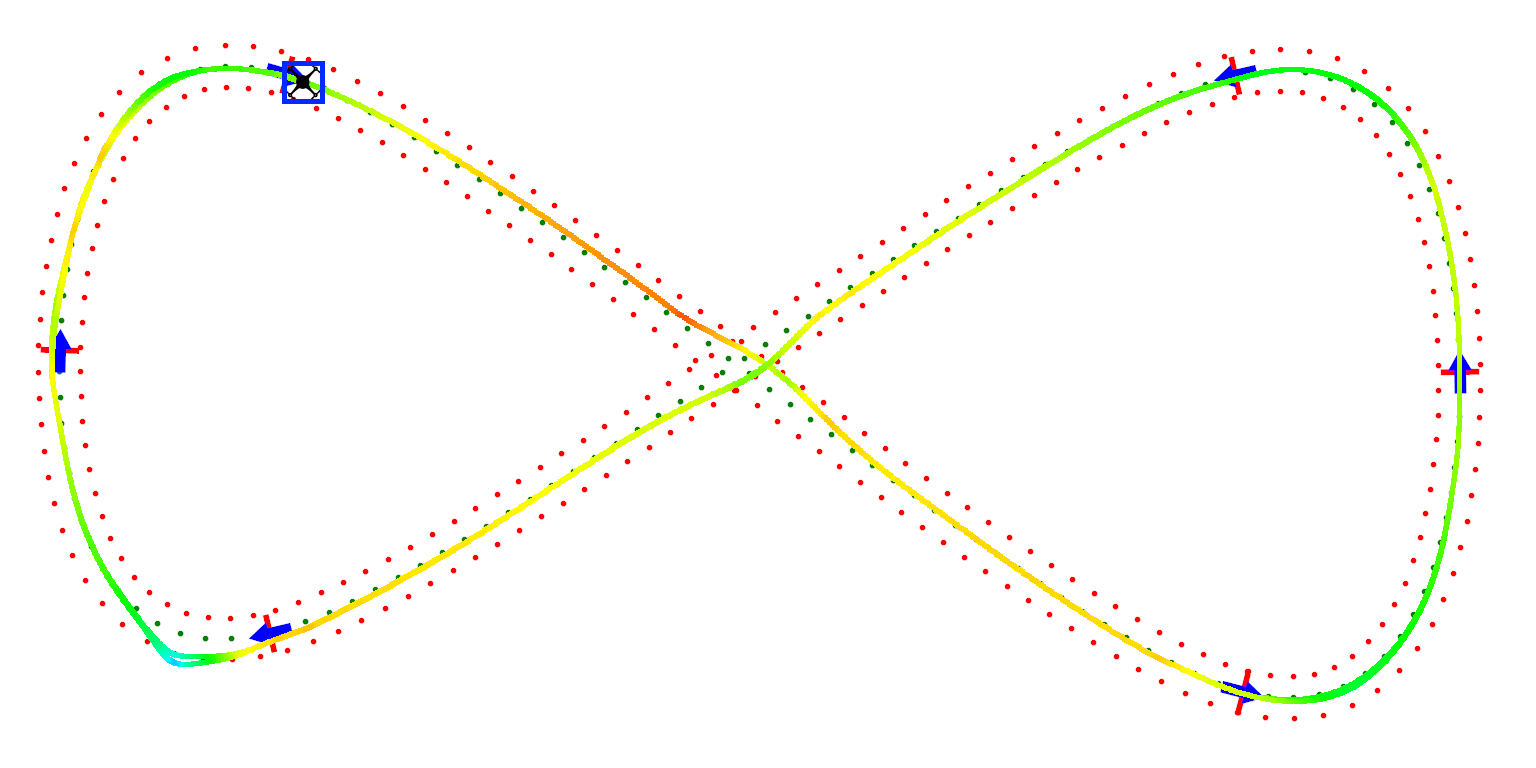} \\
		\small (i) Ours (Reference) & \small (j) Ours (Night) & \small (k)  Ours (Sunrise) \\
		\includegraphics[height=3cm]{sup_figures/track7_ours_grass.png} &
		\includegraphics[height=3cm]{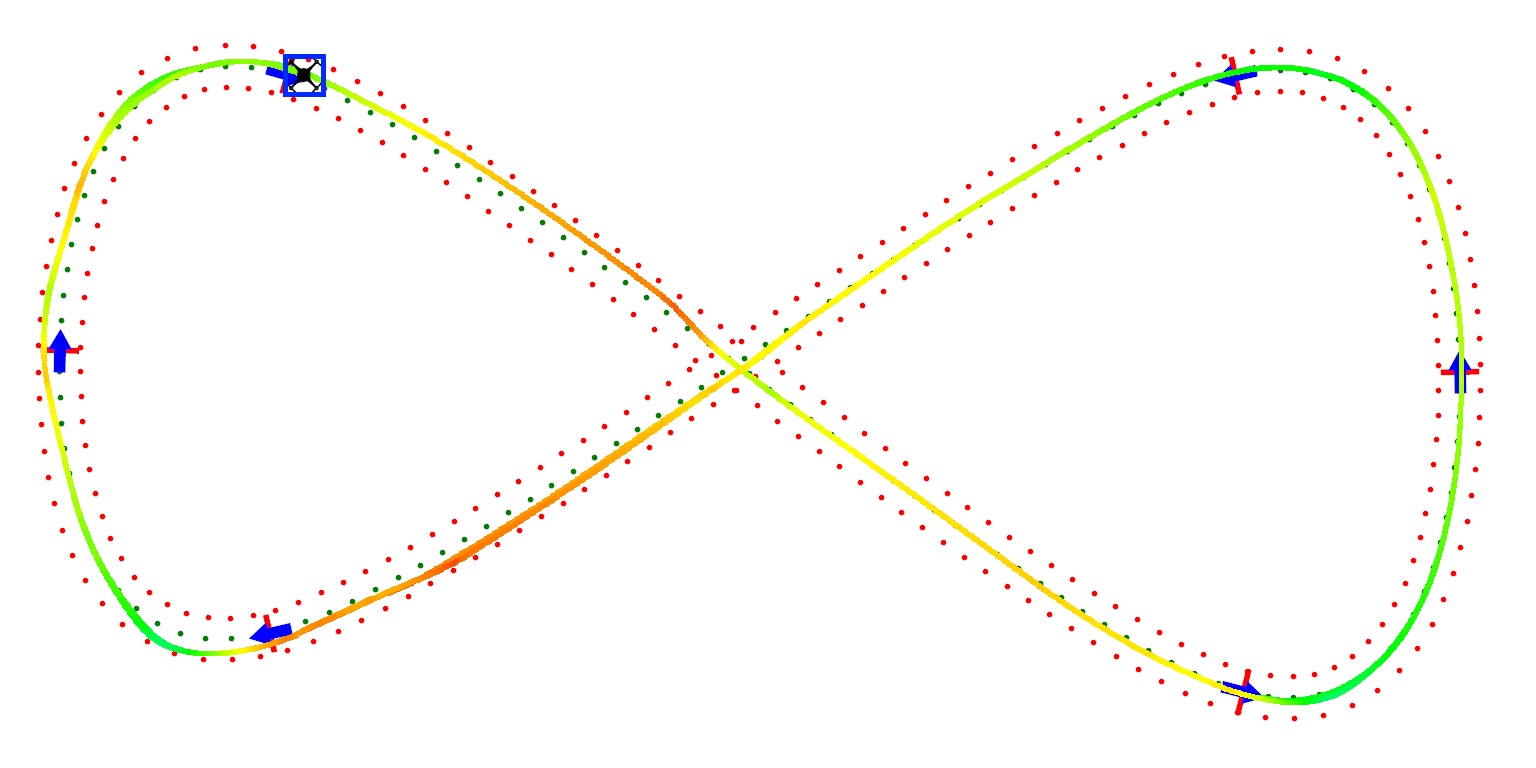} &
		\includegraphics[height=3cm]{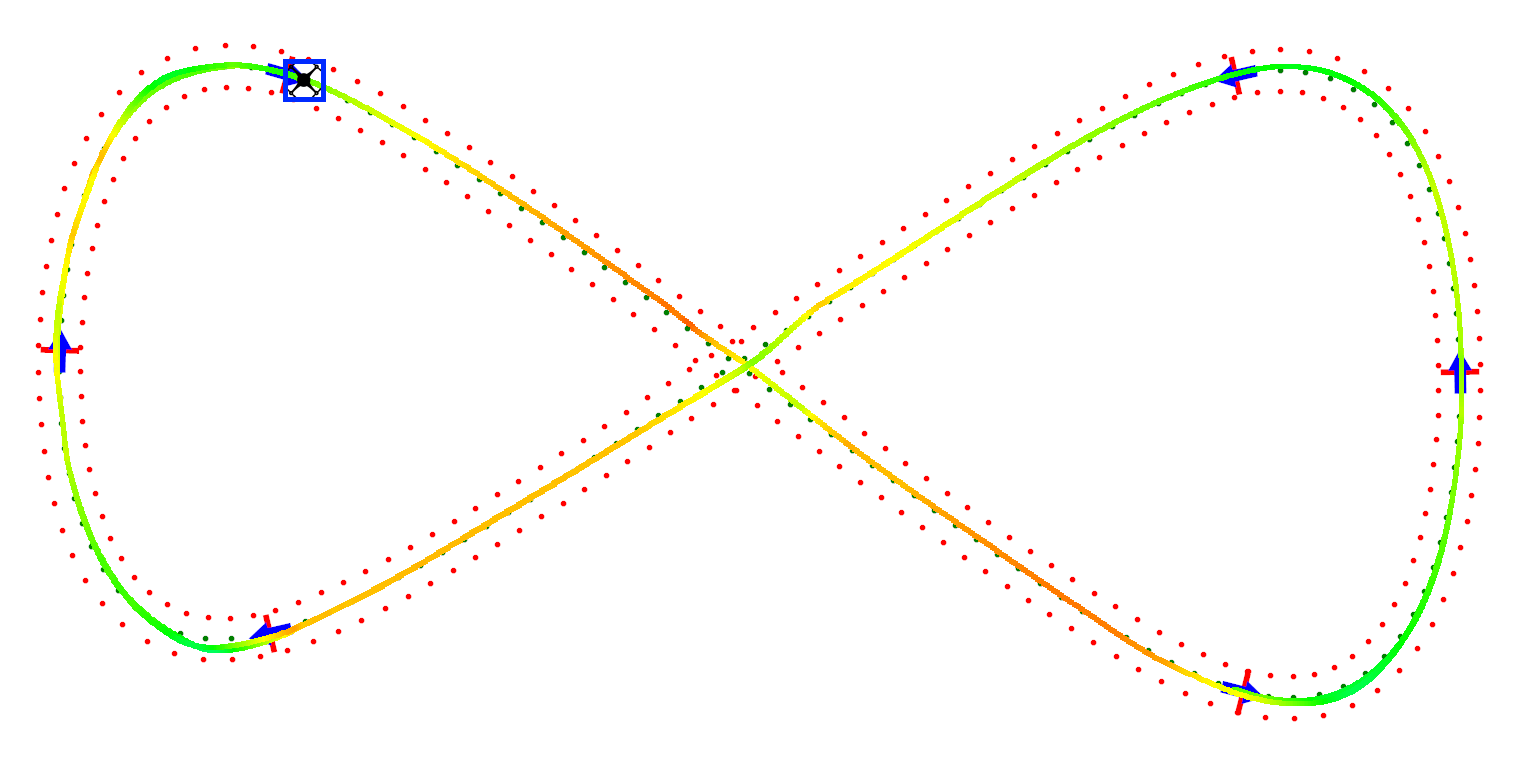} \\
		\small (l) Ours (Reference) & \small (m) Ours (Fog) & \small (o)  Ours (Rain) \\
		\includegraphics[height=3cm]{sup_figures/track7_ours_grass.png} &
		\includegraphics[height=3cm]{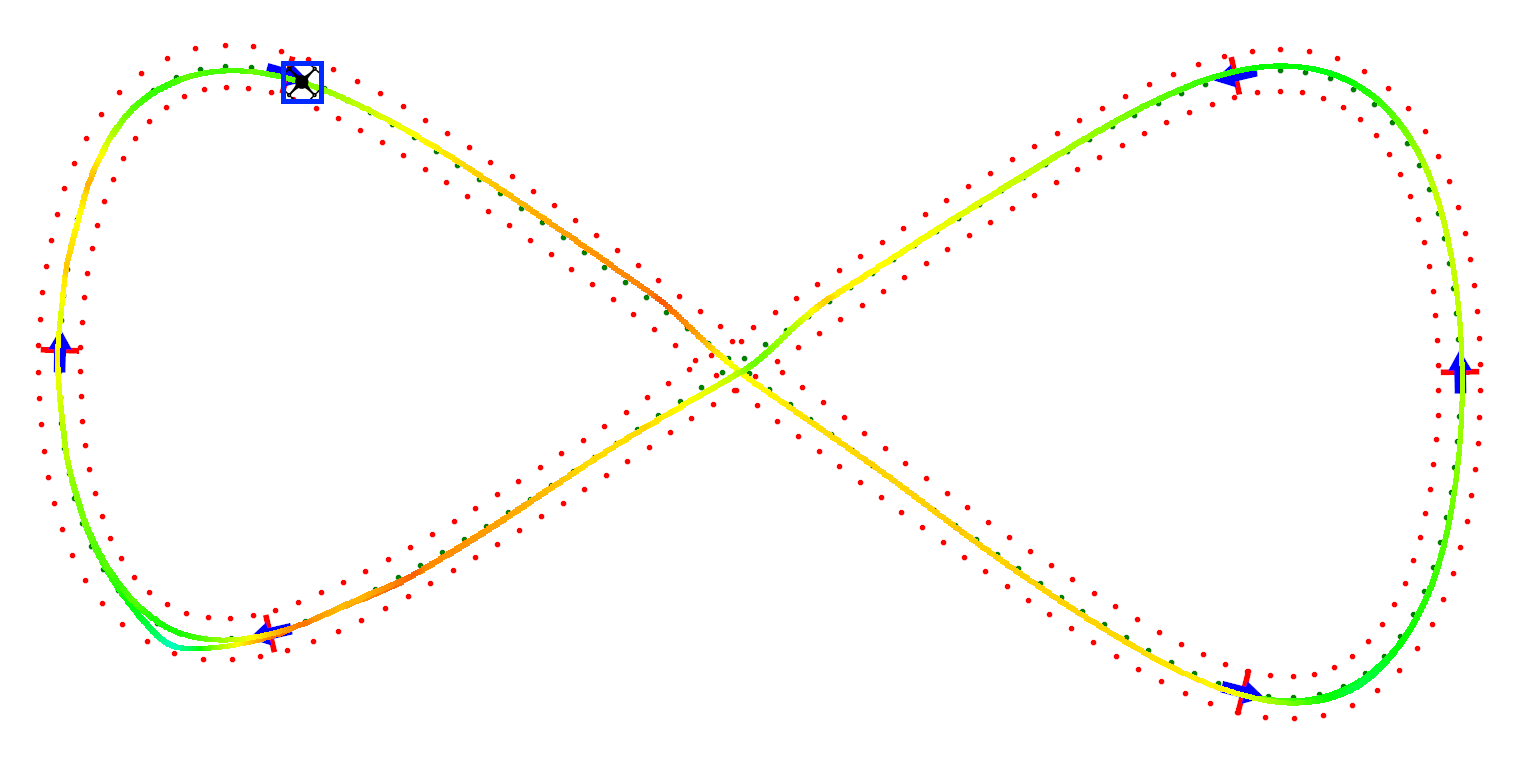} &
		\includegraphics[height=3cm]{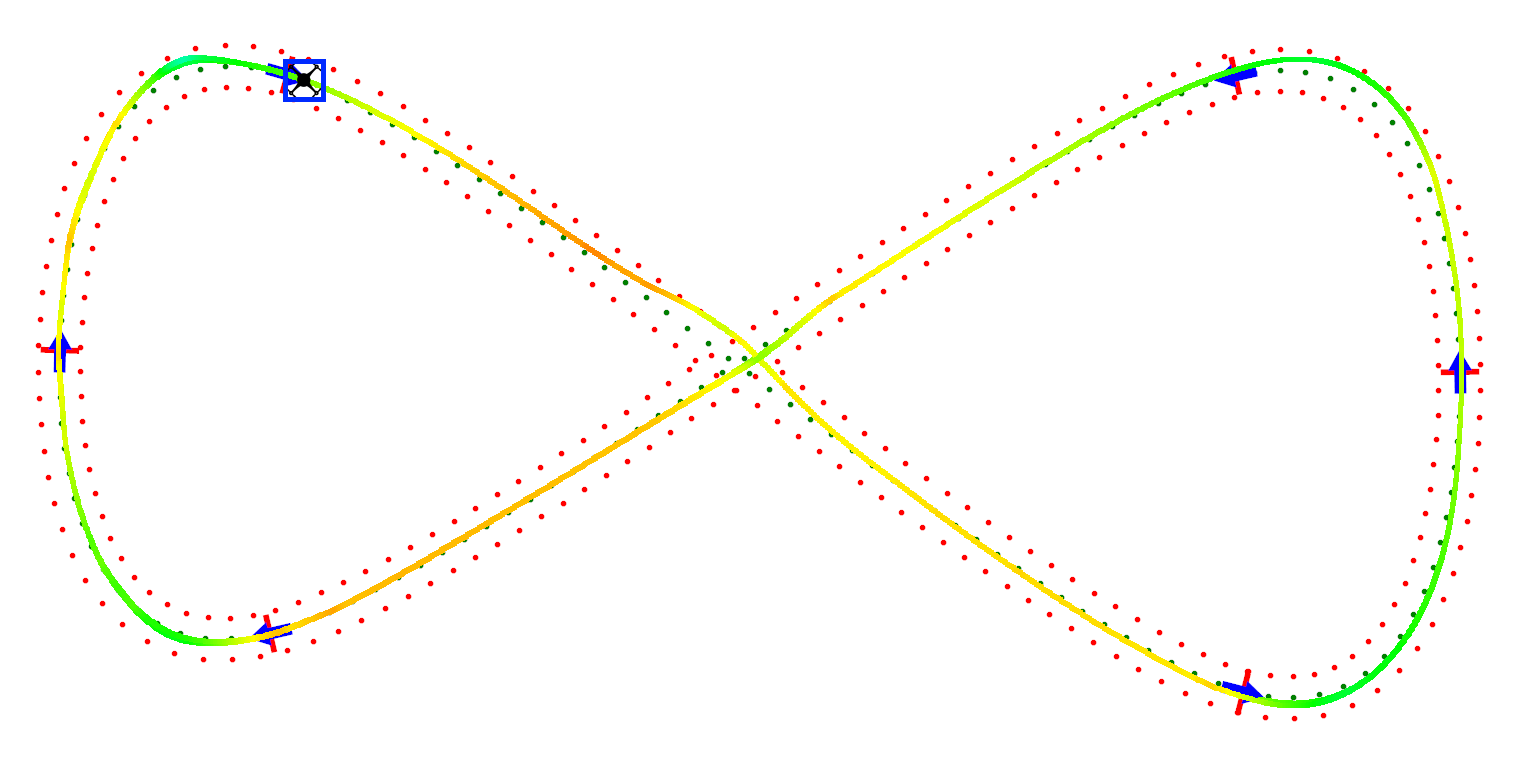} \\
		\small (p) Ours (Grass) & \small (q) Ours (Mud) & \small (r)  Ours (Snow) \\
        \multicolumn{3}{c}{\includegraphics[height=1.2cm]{sup_figures/ColorScaleHorizontal.png}}
\end{tabular}
\captionof{figure}{Qualitative results on track7. The color encodes speed as a heatmap, where blue corresponds to the minimum speed and red to the maximum speed.}
\label{fig:qualitive_results_track7}
\end{figure*}

Figures \ref{fig:qualitive_results_track1},\ref{fig:qualitive_results_track2},\ref{fig:qualitive_results_track3},\ref{fig:qualitive_results_track4},\ref{fig:qualitive_results_track5},\ref{fig:qualitive_results_track6},\ref{fig:qualitive_results_track7} show the measured performance for all trained models and human pilots. 

\begin{table*}[!h]
\centering
\caption{Adaptation with different weather and lighting conditions}
\label{tbl: environment}
\begin{tabular}{lcccccccccccc}
\hline
                & \multicolumn{3}{c}{Ours (Fog)} & \multicolumn{3}{c}{Ours (Rain)} & \multicolumn{3}{c}{Ours (Sunrise)} & \multicolumn{3}{c}{Ours (Night)} \\
                & Score     & Time     & Resets  & Score    & Time     & Resets    & Score      & Time      & Resets    & Score      & Time     & Resets   \\ \hline
Track1          & 12/12     & 57.43    & 0       & 12/12    & 54.97    & 0         & 12/12      & 55.43     & 0         & 12/12      & 58.07    & 0        \\
Track2          & 20/20     & 67.75    & 0       & 20/20    & 67.27    & 0         & 20/20      & 66.50     & 0         & 20/20      & 69.33    & 0        \\
Track3          & 22/22     & 64.32    & 0       & 22/22    & 62.63    & 0         & 22/22      & 62.95     & 0         & 22/22      & 65.17    & 0        \\
Track4          & 17/18     & 93.02    & 1       & 18/18    & 72.42    & 0         & 18/18      & 70.95     & 0         & 18/18      & 72.23    & 0        \\
Track5          & 28/30     & 114.11   & 2       & 30/30    & 73.01    & 0         & 30/30      & 74.10     & 0         & 28/30      & 114.03   & 2        \\
Track6          & 17/20     & 136.71   & 3       & 20/20    & 84.92    & 0         & 19/20      & 118.89    & 1         & 17/20      & 154.74   & 3        \\
Track7          & 12/12     & 65.03    & 0       & 12/12    & 65.58    & 0         & 12/12      & 67.72     & 0         & 12/12      & 68.58    & 0        \\ \hline
\textbf{Avg.} & 95.52\%   & 85.48    & 0.86    & 100\%    & 68.68    & 0         & 99.25\%    & 73.77     & 0.14      & 96.27\%    & 86.01    & 0.71     \\ \hline
\end{tabular}
\end{table*}

\begin{table*}[!h]
\centering
\caption{Adaptation with different textures}
\label{tbl: textures}
\begin{tabular}{lcccccccccccc}
\hline
              & \multicolumn{3}{c}{Ours (Grass)} & \multicolumn{3}{c}{Ours (HD Grass)} & \multicolumn{3}{c}{Ours (Mud)} & \multicolumn{3}{c}{Ours (Snow)} \\
              & Score     & Time     & Resets    & Score      & Time      & Resets     & Score    & Time     & Resets   & Score    & Time     & Resets    \\ \hline
Track1        & 12/12     & 52.20    & 0         & 12/12      & 56.07     & 0          & 12/12    & 53.45    & 0        & 12/12    & 55.13    & 0         \\
Track2        & 20/20     & 64.75    & 0         & 20/20      & 66.98     & 0          & 20/20    & 65.28    & 0        & 20/20    & 64.84    & 0         \\
Track3        & 22/22     & 62.00    & 0         & 22/22      & 64.05     & 0          & 22/22    & 62.03    & 0        & 22/22    & 64.06    & 0         \\
Track4        & 18/18     & 71.93    & 0         & 18/18      & 75.57     & 0          & 18/18    & 73.57    & 0        & 18/18    & 72.18    & 0         \\
Track5        & 30/30     & 71.16    & 0         & 30/30      & 73.78     & 0          & 30/30    & 71.20    & 0        & 30/30    & 73.28    & 0         \\
Track6        & 20/20     & 81.66    & 0         & 20/20      & 85.77     & 0          & 20/20    & 82.21    & 0        & 20/20    & 82.31    & 0         \\
Track7        & 12/12     & 64.86    & 0         & 12/12      & 66.01     & 0          & 12/12    & 65.03    & 0        & 12/12    & 65.99    & 0         \\ \hline
\textbf{Avg.} & 100\%     & 66.94    & 0         & 100\%      & 69.75     & 0          & 100\%    & 67.54    & 0        & 100\%    & 68.26    & 0         \\ \hline
\end{tabular}
\end{table*}

\end{document}